\documentclass[mnsc,nonblindrev]{informs3} 
\usepackage[final]{showkeys}

\usepackage{graphicx}
\usepackage{caption}
\usepackage{subcaption}
\usepackage{booktabs}
\usepackage{multirow}
\OneAndAHalfSpacedXI 

\usepackage{amsmath,amsfonts,amssymb,bm}
\usepackage[english]{babel}
\usepackage{url}
\usepackage{thmtools}
\usepackage{thm-restate}
\usepackage{mathtools}
\usepackage{listings}
\usepackage{colortbl}
\DeclareFixedFont{\ttb}{T1}{txtt}{bx}{n}{9} 
\DeclareFixedFont{\ttm}{T1}{txtt}{m}{n}{9}  
\usepackage{xcolor}
\usepackage{anyfontsize}

\usepackage{parskip} 
\usepackage{natbib}
 \bibpunct[, ]{(}{)}{,}{a}{}{,}%
 \def\bibfont{\normalfont}%
\usepackage[colorlinks=true,bookmarks=false,urlcolor=blue, citecolor=blue,linkcolor=blue,bookmarksopen=false,draft=false, hypertexnames=false]{hyperref}

\usepackage{pgfplots}
\usepackage{pgfplotstable}
\pgfplotsset{compat=1.18}
\usepackage{tikz}
\usetikzlibrary{intersections, pgfplots.fillbetween}
\pgfdeclarelayer{pre main}
\usepackage{bold-extra}
\usetikzlibrary{arrows,shapes}
\usepackage{cleveref}

\declaretheorem[name=Fact]{observation}
\usepackage{xcolor}

\newcommand{\eps}{\epsilon}

\newcommand{\bb}{\mathbb}
\newcommand{\ex}[1]{{\mathbb E} \left[ #1 \right]}
\newcommand{\extwo}[2]{{\mathbb E}_{#1} \left[ #2 \right]}
\newcommand{\expartwo}[2]{{\mathbb E}_{#1} [ #2 ]}

\newcommand{\pr}[1]{{\rm Pr} \left[ #1 \right]}
\newcommand{\prpartwo}[2]{{\rm Pr}_{#1} [ #2 ]}
\newcommand{\prtwo}[2]{{\rm Pr}_{#1} \left[ #2 \right]}
\newcommand{\prpar}[1]{{\rm Pr} [ #1 ]}
\newcommand{\promega}{{\rm Pr}_{\omega}}
\newcommand{\bs}[1]{\boldsymbol{#1}}

\usepackage{color}              

\usepackage{color-edits}
\addauthor{Ali}{red}
\addauthor{Antoine}{orange}

\usepackage{float}

\usepackage{bibunits}

\theoremstyle{plain}

\TheoremsNumberedThrough     
\ECRepeatTheorems
\EquationsNumberedThrough    
\MANUSCRIPTNO{} 

\begin{document}
\begin{bibunit}[ormsv080]

\RUNAUTHOR{Aouad, Desir}

\RUNTITLE{Representing Random Utility Choice Models with Neural Networks}
\TITLE{Representing Random Utility Choice Models \\
with Neural Networks}

\ARTICLEAUTHORS{%
\vspace*{-0.1cm}
\AUTHOR{Ali Aouad}
\AFF{MIT, \EMAIL{maouad@mit.edu}. \URL{}}
\vspace*{-0.1cm}
\AUTHOR{Antoine Désir}
\AFF{INSEAD, \EMAIL{antoine.desir@insead.edu}. \URL{}}
} 
\ABSTRACT{
Motivated by the successes of deep learning, we propose a class of neural network-based discrete choice models, called RUMnets, inspired by the random utility maximization (RUM) framework. This model formulates the agents' random utility function using a sample average approximation. We show that RUMnets sharply approximate the class of RUM discrete choice models: any model derived from random utility maximization has choice probabilities that can be approximated arbitrarily closely by a RUMnet. Reciprocally, any RUMnet is consistent with the RUM principle. Our approach is closely related to ranking-based models and mixtures of multinomial logits proposed in previous literature, in a more general contextual setting. We derive an upper bound on the generalization error of RUMnets fitted on choice data, and provide theoretical insights on their ability to predict choices on new, unseen data depending on critical parameters of the dataset and architecture. The models are estimated by leveraging open-source libraries for training neural networks. We find that RUMnets are competitive against several choice modeling and machine learning methods in terms of predictive accuracy on two real-world datasets. We also conduct synthetic experiments that isolate the effects of each component of the architecture.} 


\maketitle

\vspace*{-0.6cm}

\section{Introduction}

Many businesses offer customers an assortment of products and services to choose from, whether it be a movie to watch, a restaurant to order from, or a product to buy. An important task for these organizations is therefore to predict customers' choices using historical data in order to inform tactical decisions such as assortment, pricing, or matching optimization. The key challenge in these settings is the presence of substitution effects: the demand for a particular product depends on what else is offered. Discrete choice models, which describe a probability measure over the choice alternatives in any given assortment, are often used to represent such substitution behaviors. Research in marketing, economics, and operations research has studied various probabilistic and parametric specifications of discrete choice models.

In recent years, however, research on choice modeling has developed into a new horizon with the incorporation of machine learning (ML) and deep learning (DL) methods. Powered by advances in algorithms, computing power, and the availability of data, such methods are used for a vast array of tasks, such as speech recognition, natural language processing, and image classification. One can hence naturally wonder whether these ML/DL methods could also be useful in the context of demand estimation and choice modeling. The emerging literature in this area focuses on two primary considerations. From an implementation perspective, we may leverage open-source ML/DL libraries (such as automated differentiation tools) to estimate large-scale choice models, bypassing traditional barriers in terms of the number of parameters and volume of data. Along these lines, one approach focuses on (overparametrized) variants of the MNL model, whose implementation is very convenient using deep learning libraries~\citep{wang2020theory,han2020neural}. Other papers, however, have offered a more drastic approach: choice modeling can be formulated as a classification problem with multiple classes. With this view, the estimation of a choice model can be cast into training popular ML/DL classifiers, such as random forests  or deep neural networks; with some oversimplification, we call this a {\em model-free} approach. Under this logic, the probabilistic structure of classical choice models (such as the rationality axiom or Independence of Irrelevant Alternatives property) can be lifted altogether. While training these predictive algorithms, in theory, may require large amounts of data~\citep{feng2022consumer}, this approach was shown to be effective on several real-scale choice datasets~\citep{wong2019reslogit,chen2019use,chen2020decision}.

Our research aims to develop a general, yet structured approach to choice modeling that leverages neural networks for estimation purposes. Our work revisits the \emph{random utility maximization} (RUM) principle, which is the overarching framework for most parametric choice models \citep{mcfadden2000mixed, Train2009}. Assuming that customers are rational, the RUM principle states that customers assign a utility to each product and simply choose the one with the highest utility. The firm cannot observe the customer's utility but only has access to observable attributes of the products and customer. For this reason, the utility is assumed to be stochastic from the firm's point of view. The randomness in the utility allows for capturing unobserved heterogeneity in customers and products. Different structural and distributional assumptions on the utility function lead to various RUM discrete choice models, including the multinomial logit (MNL) model~\citep{Luce59,Mc74}, the latent class MNL~\citep{mcfadden2000mixed}, and the nested logit model~\citep{Williams77}. The validity of a RUM model heavily relies on the utility function and distributional assumption specified by the modeler. 

Considering the RUM principle as a ``minimal'' structure for choice models, we ask the following questions: can the class of {\em all} RUM discrete choice models be efficiently approximated by a compact neural network architecture? Can the resulting neural network be trained from limited choice data? How does this approach compare to model-free methods in terms of prediction accuracy?


\noindent \textbf{Preview of our results.} Our main contribution is to develop a neural network architecture, RUMnet, that provides a sharp approximation of RUM discrete choice models. Contrary to model-free methods based on classification algorithms, our approach retains the RUM principle as an underlying structure of customers' probabilistic choices. The core idea is to approximate the customers' random utility using a {\em sample average approximation} (SAA).  In this context, each ``sample'' is a neural network that corresponds to a realization of the unobserved product or customer attributes.  This approach generalizes the class of ranking-based choice models to feature-based settings, where customers and items have varying attributes. Ranking-based choice models~\citep{farias2013nonparametric} have been popularized in retail operations as a general  approach to model probabilistic choice. Preferences in the population are described by a sparse distribution over  permutations, each representing a certain realization of ordinal preferences, corresponding to a ``sample'' in our setting. To enable estimation with first-order optimization methods, we perturb the utility-based ranking by adding white noise in the final layer of the neural network.  When the white noise is standard Gumbel-distributed, the resulting RUMnets correspond to a structured family of latent-class MNL models with non-linear utility.

We establish that RUMnets tightly capture the class of all RUM discrete choice models.  In particular, their choice probabilities can be approximated arbitrarily closely by those of a RUMnet architecture, and vice versa. This implies that the class of RUMnet discrete choice models is quite general. Analytically, we derive an upper bound on the generalization errors of RUMnets fitted on choice data, i.e., the difference between  their out-of-sample and in-sample log-likelihoods. This bound does not grow with the number of distinct samples of the SAA, and its dependence on the cardinality of the assortments is improved compared to relevant benchmarks. These findings suggest that the estimation of RUMnets may be feasible, even for deep architectures and large assortments. We further show that any RUMnet admits a compact SAA approximation using a relatively small number of samples. A key insight is that the number of samples in the SAA (or the size of the mixture) needed to achieve an accurate model approximation directly depends on the complexity of the utility function.

Empirically, we demonstrate that RUMnets can be  estimated on real-world data by leveraging widely used open-source libraries. RUMnets achieve robust predictive accuracy on two different datasets and are competitive against existing choice modeling methods. In contrast, we find that the performance of model-free methods such as random forests degrades in certain regimes. We hypothesize that, contrary to RUMnets, their predictive performance is negatively affected by the input dimension, which is a function of the number of attributes and unique products in the assortment. This suggests that the RUM structure is beneficial to achieve better generalization. Synthetic experiments in a controlled environment, where the ground truth is known, support these insights and demonstrate the value added by each component of our architecture. Overall, neural network architectures such as RUMnet can be viewed as a compromise between the expressive power of deep learning and the interpretability of utility-based choice models. {\color{black} We also observe that RUMnets (and other similar deep learning architectures) may be more efficient than existing methods to estimate ranking-based choice models in the ``traditional'' offline retail setting where there are no customer attributes and the products' attributes are fixed.}


\noindent \textbf{Directed related literature.} The idea of bridging ML/DL methods with choice modeling has received a great deal of attention in the recent literature. Several papers develop various neural network-based implementations of the MNL choice model~\citep{bentz2000neural,sifringer2020enhancing,wang2020deep}. In particular, \citet{han2020neural} consider a special case of this architecture, called TasteNet, where customers' attributes feed into a neural network that outputs a taste vector, which corresponds to the coefficients of the product attributes within a linear utility function. \cite{wang2021theory} extends the architecture using the residual neural network framework. Along the same lines, \citet{gabel2021product} develops a binary logit choice model and employs neural networks to learn representations of customers' purchase history and other observable attributes.  In these logit-based models, the weights of the neural networks are shared across products, thus allowing the models to be scaled to large assortments~\citep{wang2020theory}.

In contrast with the above approach, a growing number of papers employ generic ML classifiers  and relax the probabilistic structure of parametric choice models altogether. For example, \cite{chen2019use} and~\cite{chen2020decision} train tree ensembles to predict choices, while \cite{jiang2020high} use a graphical Lasso method. The resulting probabilistic models have the ability to capture a variety of choice behaviors including complementarities between products~\citep{jiang2020high}, contextual effects~\citep{rosenfeld2020predicting}, and even irrational behaviors~\citep{berbeglia2018generalized}. Several of these models amount to a class of universal approximators, which generally violates the RUM principle~\citep{mcfadden2000mixed}. Despite their strong predictive power demonstrated in recent literature, these model-free methods are in theory harder to estimate from limited data, and they may not capture generalizable relationships. For example, the importance of modeling the substitution behavior realistically is highlighted by the randomized experiment of \cite{feldman2018customer} in the context of assortment optimization. Reflecting this challenge, our approach combines the RUM principle, as depicted by~\citet{mcfadden2000mixed}, with the expressive power of neural networks. RUMnets are also closely related to the class of rank-based choice models~\citep{rusmevichientong2006nonparametric,farias2013nonparametric}. However, an important limitation of these models is that they cannot easily leverage contextual product and customer attributes. By extending ranking-based models to varying contextual attributes, our work helps advance research on an important ``open problem'' in that line of research.

\section{A Sharp Neural Network Architecture for RUM Choice Models} \label{sec:architecture}

In this section, we introduce our neural network architecture, RUMnet, which closely imitates the RUM principle, meaning that a representative agent chooses over different alternatives by comparing their random utilities. 

\subsection{RUM discrete choice models} \label{subsec:rum}

We introduce the family of RUM discrete choice models using the formalism in~\citet{mcfadden2000mixed}. Each product is associated with a vector of $d_x$ observable features, or attributes, denoted by $\bs{x}$, varying in a compact set ${\cal X}\subseteq {\bb R}^{d_x}$. The features can be of any form---continuous or categorical. Additionally, we let $\bs{\eps}(\bs{x})$ denotes a random vector of unobserved attributes of size $d_{\epsilon}$, corresponding to a random experiment over  $[0,1]^{d_{\eps}}$. This random vector captures unobserved heterogeneity in the products, and corresponds to latent variables in the choice-making process. Here, the notation $\bs{\eps}(\bs{x})$ indicates that the distribution of the alternative's unobserved component might depend on its observed component $\bs{x}$. Similarly, each customer is described by a vector of $d_z$ observable attributes, denoted by $\bs{z}$, varying in a compact set ${\cal Z}\subseteq {\bb R}^{d_z}$. The customer also has a vector of $d_{\nu}$ unobserved (idiosyncratic) attributes $\bs{\nu}(\bs{z})$, which correspond to a random experiment over $[0,1]^{d_{\nu}}$. We further assume the model is in {\em regular canonical form}, meaning that $\bs{\eps}(\bs{x})$ and $\bs{\nu}(\bs{z})$ are mutually independent and uniformly distributed continuous random fields.\footnote{A random function $\eta(\cdot, \cdot): \Omega \times {\cal T}\rightarrow [0,1]^d$  is a uniformly distributed random field over a probabilistic space $\Omega$ and a closed set ${\cal T}$ if   for each $t\in {\cal T}$, $\eta(\cdot, t)$ is a uniformly distributed random vector $[0,1]^{d}$, measurable with respect to $\Omega$, and for a set of $\omega$ occurring with probability $1$, $t\mapsto \eta(\omega, t)$ is a continuous function on ${\cal T}$.} The assumption of independence is standard and assumes that the observable attributes fully explain any probabilistic dependence between offered products and customer characteristics. We call this assumption {\em the independence of unobserved attributes} (IUA); in most plausible scenarios, a principal offers items $\bs{x}$ in the assortment $A$ through a process determined only by observed attributes $\bs{x},\bs{z}$, and thus, the IUA property ensues. The assumption of uniform distribution can be enforced without loss of generality under very mild conditions; see~\citet[Lemma~3]{mcfadden2000mixed}. Importantly, the continuity of the random field realizations can capture a rich correlation structure: for example, as two items $\bs{x}_1$ and $\bs{x}_2$ are closer, their unobserved randomness may be more positively correlated (e.g., this circumvents the so-called red bus and blue bus problem). Moreover, even though $\bs{\eps}(\bs{x})$ and $\bs{\nu}(\bs{z})$ are independent given $\bs{x}$ and $\bs{z}$, their dependence on $\bs{x}$ and $\bs{z}$ may capture relationships between products and individuals. 

Finally, we specify a utility function $U : {\cal X}\times [0,1]^{d_x} \times {\cal Z}\times [0,1]^{d_z} \rightarrow {\bb R}$, which is assumed to be bounded and uniformly continuous in its arguments. For $\bs{x} \in {\cal X}$ and $\bs{z} \in {\cal Z}$, $U(\bs{x}, \bs{\eps}(\bs{x}), \bs{z}, \bs{\nu}(\bs{z}))$ quantifies how a customer with attributes $\bs{z}$  values an offered product with attributes $\bs{x}$. This quantity is  random due to $\bs{\eps}(\bs{x})$ and $\bs{\nu}(\bs{z})$. 

We now describe the probabilistic outcomes for any given {\em choice event} $(\bs{z}, A)$, where a customer with observable attributes $\bs{z} \in {\cal Z}$ is offered a finite assortment of alternatives $A \subseteq {\cal X}$. Throughout the remainder of the paper, we assume $A$ is finite and satisfies $|A| \leq \kappa$ for some integer $\kappa \geq 0$. The RUM principle implies that the customer picks the highest-utility product in the assortment $A$. Specifically, denoting by $\pi(\bs{x}, \bs{z}, A)$  the probability that a customer with observable attributes $\bs{z}$ chooses $\bs{x} \in A$, we have
\begin{eqnarray*}
	\pi(\bs{x}, \bs{z}, A) = \prtwo{\bs{\eps},\bs{\nu}}{ U\left(\bs{x}, \bs{\eps}(\bs{x}), \bs{z}, \bs{\nu}(\bs{z})\right) > U\left(\bs{x}', \bs{\eps}(\bs{x}'), \bs{z}, \bs{\nu}(\bs{z})\right)\ , \forall \bs{x}' \in A\setminus \bs{x}} \ .
\end{eqnarray*}
To ensure that the distribution $\{\pi(\bs{x}, \bs{z}, A)\}_{\bs{x}\in A}$ is well-defined, we assume that ties between the product utilities occur with probability zero. The family of RUM discrete choice models subsumes a large array of models used in practice such as the MNL model, the nested logit model, and their probabilistic mixtures.

\subsection{RUMnet} \label{subsec:rumnet}

In the RUM framework, the utility function is a random field. Our main idea is to develop a neural network architecture that approximates this random field using a sample average approximation (SAA). Each sample expresses the unobserved attributes using a different functional form, which we will then represent as a feed-forward neural network with unknown parameters. More precisely, in the RUM framework presented in Section~\ref{subsec:rum}, there are three functions that we wish to approximate. The first one, $U(\cdot)$, can be straightforwardly replaced by a neural network. The two other functions, $\bs{\epsilon}(\cdot)$ and $\bs{\nu}(\cdot)$, are random and instead of approximating them directly, we propose to use a sample average approximation. More specifically, for some large $K$, we can informally write
\begin{align} \label{eq:sampling}
    \pi(\bs{x}, \bs{z}, A) \ \approx \ \dfrac{1}{K^2} \cdot \sum \limits_{k_1=1}^K \sum \limits_{k_2=1}^K \mathbf{1} \left \{ U\left(\bs{x}, \bs{\eps}_{k_1} (\bs{x}), \bs{z}, \bs{\nu}_{k_2} (\bs{z})\right) > U\left(\bs{x}', \bs{\eps}_{k_1}(\bs{x}'), \bs{z}, \bs{\nu}_{k_2}(\bs{z})\right)\ , \forall \bs{x}' \in A\setminus \bs{x} \right \} ,
\end{align}
where, for each $k_1 \in [K]$, $\bs{\eps}_{k_1}(\cdot)$  denotes an independent sample of the random field $\bs{\eps}(\cdot)$. Analogously, for each $k_2 \in [K]$, $\bs{\nu}_{k_2}(\cdot)$ is an independent sample of the random field $\bs{\nu}(\cdot)$. Since  $\bs{\eps}_{k_1}(\cdot)$ and $\bs{\nu}_{k_2}(\cdot)$ are now deterministic functions, we can try to estimate them using neural networks approximately. Note that we use an equal number of samples $K$ to approximate the random fields $\bs{\eps}(\cdot)$ and $\bs{\nu}(\cdot)$, but in practice, we could use different numbers of samples, $K_1$ and $K_2$,  for the unobserved customer and product attributes. In SAA, the modeler has access to independent samples from the stochastic variable. By contrast, in our setting, we do not have direct access to such samples;  the empirical distribution is obtained by fitting our neural network architecture to the observed choice data.\footnote{Note that a similar interpretation of sample-based approximation was provided for the rank-based choice models~\citep{farias2013nonparametric}. In particular, one can interpret the SAA of RUMnets as an extension of rank-based choice models to contextual choice data. The non-contextual setting corresponds to the special case where $\bs{x}$ is a product indicator and $\bs{z}$ is fixed.} 

In the remainder of this section, we formalize the family of choice models resulting from this combination of neural networks. We subsequently show in Section~\ref{sec:expressive_power} that it approximates any RUM discrete choice model arbitrarily closely.

\noindent \textbf{Feed-forward neural networks.} We use feed-forward neural networks as  building blocks to construct our model {\em architecture}, which refers to the combination of these neural networks. For our purposes, a feed-forward neural network is a function $N(\cdot) : {\bb R}^{d_{\sf input}} \to {\bb R}^{d_{\sf output}}$, where $d_{\sf input}$ is the size of the input and $d_{\sf output}$ the size of the output. We give a more formal account in Appendix~\ref{app:ffnn}. To obtain generalization guarantees, we assume size-based and norm-based capacity restrictions on such neural networks. Specifically, for every scalar $M\geq0$ and integers $\ell,w\geq 0$, let $\Theta^{\ell,w}_M$ be the family of feed-forward neural networks $N(\cdot)$ of depth $\ell$ and width $w$ such that $M$ is an upper bound on the $\ell_1$-norm of incoming weights in each node as well as the $\ell_1$-norm of the final layer's output.  

\noindent \textbf{RUMnet architecture.} Let $d = d_x + d_{\eps} + d_z + d_{\nu}$ be the dimension of the input vector to the utility function $U(\cdot)$ that underlies the customer choices in the RUM framework. Additionally, let $K \geq 0$ be an integer that controls the number of samples we use to approximate the random fields. With this notation at hand, we introduce a RUMnet architecture, which is a family of neural networks, comprising of the following  building blocks:
\begin{enumerate}
\item {\em Utility function:} There is a feed-forward neural network $N_{U}(\cdot) \in \Theta^{\ell,w}_M$ such that  $N_{U}(\cdot)$ is a mapping from ${\bb R}^{d}$ to ${\bb R}$ that serves as an approximation of the function $U(\cdot)$.
\item {\em Unobserved product attributes:} For every $k_1 \in [K]$, there is a feed-forward neural network $N_{\epsilon_{k_1}}(\cdot) \in \Theta^{\ell,w}_M$ such that $N_{\epsilon_{k_1}}(\cdot)$ is a mapping from ${\bb R}^{d_x}$ to ${\bb R}^{d_{\epsilon}}$. Intuitively, each of these neural networks serves as an approximation to one sample of the random field $\bs{\epsilon}(\cdot)$.
\item {\em Unobserved customer attributes:} For every $k_2 \in [K]$, there is a feed-forward neural network $N_{\nu_{k_2}}(\cdot) \in \Theta^{\ell,w}_M$ such  that $N_{\nu_{k_2}}(\cdot)$ is a mapping from ${\bb R}^{d_z}$ to ${\bb R}^{d_{\nu}}$. Intuitively, each of these neural networks serves as an approximation to one sample of the random field $\bs{\nu}(\cdot)$.
\end{enumerate}
Consequently, we denote by ${\cal N}^{\bs{d}}(K,\Theta^{\ell,w}_M)$ the collection of all RUMnet architectures $(N_{U}(\cdot),\{N_{\epsilon_{k_1}}(\cdot)\}_{k_1=1}^{K},\{N_{\nu_{k_2}}(\cdot)\}_{k_2=1}^{K})$, where $\bs{d} = (d_x, d_{\eps},d_z, d_{\nu})$. The parameters of a given architecture are the number of samples of the SAA as well as the depth and width of each building block. The number of samples controls the ``latent'' (random) effects, whereas the size of each neural network determines the ``degree of nonlinearity'' expressed by the utility function. 

\noindent \textbf{RUMnet discrete choice model.} We now specify the discrete choice model induced by any given RUMnet architecture  $N= (N_{U}(\cdot),\{N_{\epsilon_{k_1}}(\cdot)\}_{k_1=1}^{K},\{N_{\nu_{k_2}}(\cdot)\}_{k_2=1}^{K}) \in {\cal N}^{\bs{d}}(K,\Theta^{\ell,w}_M)$. For any choice event $(\bs{z}, A) \in {\cal Z} \times 2^{\cal X}$, we compute the probability $\pi^{{\sf RUMnet}}_N(\bs{x},\bs{z},A)$ that a customer with observable attributes $\bs{z}$ chooses $\bs{x} \in A$ as follows:
\begin{align*}
\pi^{{\sf RUMnet}}_N(\bs{x},\bs{z},A) = &\ \dfrac{1}{K^2} \cdot \sum_{k_1=1}^{K} \sum_{k_2=1}^{K} {\rm Pr}_{\bs{\delta}} \left[ N_{U}\left(\bs{x}, N_{\epsilon_{k_1}}(\bs{x}), \bs{z},N_{\nu_{k_2}}(\bs{z})\right) + \delta_{\bs{x}} > \right.\\
& \qquad \qquad \qquad \qquad \qquad \left. N_{U}\left(\bs{x}', N_{\epsilon_{k_1}}(\bs{x}'), \bs{z},N_{\nu_{k_2}}(\bs{z})\right)  + \delta_{\bs{x}'} \ , \ \forall \bs{x}'\in A\setminus \bs{x}  \right]  \ ,
\end{align*}
where $\bs{\delta} = \{\delta_{\bs{x}}\}_{\bs{x}\in A}$ is a sequence of independent real-valued random variables that follow the same distribution. Comparing the expression of $\pi^{{\sf RUMnet}}_N(\cdot)$ with our sample average approximation~\eqref{eq:sampling}, we have replaced the ``argmax'' indicator with a probability distribution resulting from the white noise~$\bs{\delta}$. Adding noise plays an important role in the training process. In the absence of this property, commonly used loss functions (e.g., log-likelihood) have a gradient equal to zero on a set of measure 1. This is a well-known property for ``combinatorial layers'' in neural networks~\citep{abernethy2016perturbation,berthet2020learning,dalle2022learning}---a small perturbation of the utility parameters cannot reverse the highest-utility alternative unless there are ties. Adding white noise facilitates the training of the neural network architecture.

In theory, we only require that  distribution $\bs{\delta}$ has strictly positive density on ${\bb R}$. However, in remainder of the paper, we focus on the case where the white noise $\bs{\delta}$ follows a standard Gumbel distribution, which amounts to replacing the ``argmax'' indicator with the ``softmax'' operation.\footnote{A similar step is made in the RUM approximation result of~\citet[Proof of Thm.~1]{mcfadden2000mixed}: Gumbel noise is added to smooth the problem and get closed-form expressions for the choice probabilities.}  As shall be discussed in Section~\ref{subsec:discussion}, the benefit of this approach is that the softmax operation is highly optimized in deep learning libraries and the resulting RUMnet model can be interpreted as a discrete mixture of MNLs. Nonetheless, it is important to remark that {\em any alternative} smoothing kernel---i.e., adding white noise with closed-form expressions for the resulting choice probabilities---could be used. For instance, one could use mixtures of exponomial models by adding exponentially distributed white noise~\citep{alptekinouglu2016exponomial,alptekinouglu2021heteroscedastic}, noting that the closed-form expression of the resulting choice probabilities can be implemented via standard neural network operations. More generally, the gradient of the RUMnet architecture admits simple expressions for any distribution in the exponential family~\citep{abernethy2016perturbation}. Variants of the RUMnet implementation could be developed in the future by incorporating such smoothing kernels.

\subsection{Interpretation as a neural network} \label{subsec:neural_net}

For any given RUMnet architecture $N \in {\cal N}^{\bs{d}}(K,\Theta^{\ell,w}_M)$, we can interpret the computation of the probabilities $\{ \pi^{{\sf RUMnet}}_N(\bs{x},\bs{z},A) \}_{\bs{x} \in A}$ associated with a choice event $(\bs{z},A)$ as the output of a highly structured neural network. The input to this neural network consists of a vector formed by concatenating the observed product attributes $\bs{x}_1,\dots,\bs{x}_{|A|}$ together with the observed customer attributes~$\bs{z}$. 
The distribution $\{ \pi^{{\sf RUMnet}}_N(\bs{x},\bs{z},A) \}_{\bs{x} \in A}$ is the output of a computation graph comprising four ``meta-layers'', which sequentially perform the following: 
\begin{itemize}
	\item \textit{Input vector.} Let $\bs{x}_1,\dots,\bs{x}_{|A|}$ be an arbitrary  numbering of the alternatives in the assortment $A$.  The input to our neural network is the vector $(\bigoplus_{i=1}^{|A|} \bs{x}_i) \oplus \bs{z}$ formed by concatenating the observed product attributes $\bs{x}_1,\dots,\bs{x}_{|A|}$ together with the observed customer attributes~$\bs{z}$. 
    \item \textit{Meta-layer 1: Generating the samples of unobserved attributes.} In the first layer, for each $i \in \{1,\dots,|A|\}$ and $k_1 \in [K]$, $\bs{x}_i$ is passed through $N_{\epsilon_{k_1}}(\cdot)$ and, for each $k_2 \in [K]$, $\bs{z}$ is passed through $N_{\nu_{k_2}}(\cdot)$. Here, $N_{\epsilon_{k_1}}(\bs{x}_i)$ can be thought of as the $k_1$-th sample of unobserved product attributes for the alternative $\bs{x}_i$, while  $N_{\nu_{k_2}}(\bs{z})$ is  the ${k_2}$-th sample of unobserved attributes for customer $\bs{z}$. In what follows, we write $\bs{h}^{k_1,k_2}_i = \bs{x}_i \oplus N_{\epsilon_{k_1}}(\bs{x}_i) \oplus \bs{z} \oplus N_{\nu_{k_2}}(\bs{z}) \in \mathbb{R}^d$ for the intermediate variables, i.e., the ouput of the first meta-layer and input to the second layer. 
    \item \textit{Meta-layer 2: Computing the alternative-specific utilities.} For each $i \in \{1,\dots,|A|\}$ and $(k_1,k_2) \in [K^2]$, $\bs{h}^{k_1,k_2}_i$ is passed through $N_U(\cdot)$. The resulting quantity $u^{k_1,k_2}_i = N_U(\bs{h}^{k_1,k_2}_i)$ stands for the utility of the choice alternative $\bs{x}_i$ for the sample $(k_1,k_2)$. The output of this second meta-layer is $\{ \bigoplus_{i=1}^{|A|} u^{k_1,k_2}_i\}_{(k_1,k_2)\in [K^2]}$.
    \item \textit{Meta-layer 3: Converting the utilities into probabilities.} Next, the utilities are converted into probabilities using a softmax layer. More precisely, for each $i \in \{1,\dots,|A|\}$ and $(k_1,k_2) \in [K^2]$, we let $\pi^{k_1,k_2}_i = e^{u^{k_1,k_2}_i} / (\sum_{j=1}^{|A|} e^{u^{k_1,k_2}_j})$.
    \item \textit{Meta-layer 4: Averaging.} For each $i \in \{1,\dots,|A|\}$, the final output of the neural network is  an average over all samples, yielding the choice probabilities $\pi^{{\sf RUMnet}}_N (\bs{x}_i,\bs{z},A) = \frac{1}{K^2}\cdot \sum_{k_1,k_2} \pi^{k_1,k_2}_i$.
\end{itemize}

The design of the first two meta-layers carefully combines several feed-forward neural network building blocks, as illustrated in Figure~\ref{fig:rumnet_layers}.
\begin{figure}[t!]
\centering

\begin{tikzpicture}[scale=0.7, every node/.style={scale=0.7}]
\node at (0,1.5) {\sf Input layer};
\node[draw=black, thick, rectangle, inner sep=5pt, inner ysep=20pt] (x1) at (0,-1) {};
\node at (-0.5,-1) {$\bs{x}_i$};
\node[draw=black, thick, rectangle, inner sep=5pt, inner ysep=20pt] (z1) at (0,-5) {};
\node at (-0.5,-5) {$\bs{z}$};
\node at (3,1.5) {\sf Meta-Layer 1};
\node[draw=black, thick, fill=red!40, rectangle, inner sep=30pt, inner ysep=30pt] (Neps) at (3,-2) {};
\node[draw=black, thick, fill=blue!40, rectangle, inner sep=30pt, inner ysep=30pt] (Nnu) at (3,-6) {};
\node at (3,-2) {$N_{\eps_{k_1}}$};
\node at (3,-6) {$N_{\nu_{k_2}}$};
\node at (6,1.5) {$\bs{h}^{k_1,k_2}_i$};
\node[draw=black, thick, rectangle, inner sep=5pt, inner ysep=20pt] (x2) at (6,0) {};
\node[draw=black, thick, rectangle, inner sep=5pt, inner ysep=20pt] (eps2) at (6,-2) {};
\node[draw=black, thick, rectangle, inner sep=5pt, inner ysep=20pt] (z2) at (6,-4) {};
\node[draw=black, thick, rectangle, inner sep=5pt, inner ysep=20pt] (nu2) at (6,-6) {};
\node at (9,1.5) {\sf Meta-Layer 2};
\node[draw=black, thick, fill=green!40, rectangle, inner sep=40pt, inner ysep=40pt] (Nu) at (9,-3) {};
\node at (9,-3) {$N_{U}$};
\node (u) at (12,-3) {$u^{k_1,k_2}_i$};

\draw[->,thick, -stealth] (Nu) -- (u);
\draw[->,thick, -stealth] ([yshift=-5mm]x1) -| +(-1:1) |- (Neps);
\draw[->,thick, -stealth] ([yshift=-5mm]z1) -| +(1:1) |-  (Nnu);
\draw[->,thick, -stealth] ([yshift=+5mm]x1) -| +(1:1) |- (x2);
\draw[->,thick, -stealth] ([yshift=+5mm]z1) -| +(-1:1) |- (z2);
\draw[->,thick, -stealth] (x2) -- (Nu);
\draw[->,thick, -stealth] (eps2) -- (Nu);
\draw[->,thick, -stealth] (z2) -- (Nu);
\draw[->,thick, -stealth] (nu2) -- (Nu);
\draw[->,thick, -stealth] (Neps) -- (eps2);
\draw[->,thick, -stealth] (Nnu) -- (nu2);

\end{tikzpicture}

\caption{Illustration of the first two meta-layers. Note that we show the computation for one particular product attribute $\bs{x}_i$ and sample $(k_1,k_2)$, but we apply the same transformation for each $\bs{x} \in A$ and $(k_1,k_2) \in K^2$.}
\label{fig:rumnet_layers}
\end{figure} 
The third meta-layer simply consists of a softmax operator. The final meta-layer integrates the choice probabilities over all distinct samples of unobserved attributes. Figure~\ref{fig:rumnet} visualizes the overall architecture, where $\pi^{{\sf RUMnet}}_N(\cdot)$ is determined as the output of a single neural network. 

\begin{figure}[t!]
 \centering

 \begin{tikzpicture}[scale=0.75, every node/.style={scale=0.75}]

\node[draw=red,dashed,fill=red!0, very thick,rectangle, rounded corners, inner sep=20pt, inner ysep=133pt] (ellipse) at (-0.5,-1) {};
\node at (-0.5,4) {\sf Input layer};
\node[rotate=90] at (-1.5,0) {\tiny \sf product features};
\node[draw=black, thick, rectangle, inner sep=5pt, inner ysep=20pt] at (-0.3,2.4) {};
\node at (-0.8,2.5) {$\bs{x}_1$};
\node[rotate=90] at (-0.3,0) {$\dots$ };
\node[draw=black, thick, rectangle, inner sep=5pt, inner ysep=20pt] at (-0.3,-2.4) {};
\node at (-0.8,-2.5) {$\bs{x}_n$};
\node[rotate=90] at (-1.5,-4.6) {\tiny \sf customer features};
\node[draw=black, thick, rectangle, inner sep=5pt, inner ysep=20pt] at (-0.3,-4.6) {};
\node at (-0.8,-4.6) {$\bs{z}$};
\node[draw=blue,dashed,fill=red!0,very thick,rectangle, rounded corners, inner sep=137pt, inner ysep=133pt] (ellipse) at (5.75,-1) {};
\node at (2,4) {\sf Meta-Layer 1};
\node at (4.5,4) {\sf Meta-Layer 2};
\node at (7,4) {\sf Meta-Layer 3};
\node at (9.5,4) {\sf Meta-Layer 4};
\node[draw=black, thick, fill=red!70, rectangle, inner sep=20pt, inner ysep=20pt] at (2.1,2.5) {};
\node[draw=black, thick, fill=red!40, rectangle, inner sep=20pt, inner ysep=20pt] at (2.2,2.4) {};
\node[draw=black, thick, fill=red!10, rectangle, inner sep=20pt, inner ysep=20pt] at (2.3,2.3) {};
\node at (2.3,2.3) {\tiny $N_{\eps_{k_1}}$};
\node[rotate=90] at (2.2,0) {$\dots$ };
\node[draw=black, thick, fill=red!70, rectangle, inner sep=20pt, inner ysep=20pt] at (2.1,-2.3) {};
\node[draw=black, thick, fill=red!40, rectangle, inner sep=20pt, inner ysep=20pt] at (2.2,-2.4) {};
\node[draw=black, thick, fill=red!10, rectangle, inner sep=20pt, inner ysep=20pt] at (2.3,-2.5) {};
\node at (2.3,-2.5) {\tiny $N_{\eps_{k_1}}$};
\node at (1.15,-3.3) {\tiny $K$};
\draw[<->] (1.25,-3.05) -- (1.45,-3.25);

\node[draw=black, thick, fill=blue!70, rectangle, inner sep=20pt, inner ysep=20pt] at (2.1,-4.5) {};
\node[draw=black, thick, fill=blue!40, rectangle, inner sep=20pt, inner ysep=20pt] at (2.2,-4.6) {};
\node[draw=black, thick, fill=blue!10, rectangle, inner sep=20pt, inner ysep=20pt] at (2.3,-4.7) {};
\node at (2.3,-4.7) {\tiny $N_{\nu_{k_2}}$};
\node at (1.15,-5.5) {\tiny $K$};
\draw[<->] (1.25,-5.25) -- (1.45,-5.45);

\node[draw=black, thick, fill=green!40, rectangle, inner sep=20pt, inner ysep=20pt] at (4.5,2.6) {};
\node[draw=black, thick, fill=green!40, rectangle, inner sep=20pt, inner ysep=20pt] at (4.6,2.5) {};
\node[draw=black, thick, fill=green!40, rectangle, inner sep=20pt, inner ysep=20pt] at (4.7,2.4) {};
\node[draw=black, thick, fill=green!40, rectangle, inner sep=20pt, inner ysep=20pt] at (4.8,2.3) {};
\node[draw=black, thick, fill=green!40, rectangle, inner sep=20pt, inner ysep=20pt] at (4.9,2.2) {};
\node[draw=black, thick, fill=green!40, rectangle, inner sep=20pt, inner ysep=20pt] at (5.0,2.1) {};
\node at (5.0,2.1) {\tiny $N_U$};

\node[draw=black, thick, fill=green!40, rectangle, inner sep=20pt, inner ysep=20pt] at (4.5,-2.1) {};
\node[draw=black, thick, fill=green!40, rectangle, inner sep=20pt, inner ysep=20pt] at (4.6,-2.2) {};
\node[draw=black, thick, fill=green!40, rectangle, inner sep=20pt, inner ysep=20pt] at (4.7,-2.3) {};
\node[draw=black, thick, fill=green!40, rectangle, inner sep=20pt, inner ysep=20pt] at (4.8,-2.4) {};
\node[draw=black, thick, fill=green!40, rectangle, inner sep=20pt, inner ysep=20pt] at (4.9,-2.5) {};
\node[draw=black, thick, fill=green!40, rectangle, inner sep=20pt, inner ysep=20pt] at (5.0,-2.6) {};
\node at (5.0,-2.6) {\tiny $N_U$};
\node[rotate=90] at (4.8,0) {$\dots$ };
\node at (3.75,-3.3) {\tiny $K^2$};
\draw[<->] (3.75,-2.95) -- (4.15,-3.35);

\node[draw=black, thick, fill=white, rectangle, inner sep=10pt, inner ysep=90pt] (x1) at (7.0,0.3) {};
\node[draw=black, thick, fill=white, rectangle, inner sep=10pt, inner ysep=90pt] (x1) at (7.1,0.2) {};
\node[draw=black, thick, fill=white, rectangle, inner sep=10pt, inner ysep=90pt] (x1) at (7.2,0.1) {};
\node[draw=black, thick, fill=white, rectangle, inner sep=10pt, inner ysep=90pt] (x1) at (7.3,0) {};
\node[draw=black, thick, fill=white, rectangle, inner sep=10pt, inner ysep=90pt] (x1) at (7.4,-0.1) {};
\node[draw=black, thick, fill=white, rectangle, inner sep=10pt, inner ysep=90pt] (x1) at (7.5,-0.2) {};
\node[rotate = 90] at (7.5,0) {\sf softmax};
\node at (6.45,-3.3) {\tiny $K^2$};
\draw[<->] (6.5,-3) -- (6.9,-3.4);

\node[draw=black, thick, fill=white, rectangle, inner sep=10pt, inner ysep=90pt] (x1) at (9.5,0) {};
\node[rotate = 90] at (9.5,0) {\sf average};

\node[draw=red,dashed,fill=red!0,very thick,rectangle, rounded corners, inner sep=45pt, inner ysep=105pt] (ellipse) at (12.7,0) {};
\node at (12.7,4) {\sf Output layer};
\node[draw=black, thick, circle, inner sep=3pt, inner ysep=3pt] at (11.6,2.4) {};
\node at (13,2.4) {$\pi^{{\sf RUMnet}}_N(\bs{x}_1,\bs{z})$};
\node[draw=black, thick, circle, inner sep=3pt, inner ysep=3pt] at (11.6,-2.4) {};
\node at (13,-2.4) {$\pi^{{\sf RUMnet}}_N(\bs{x}_n,\bs{z})$};
\node[rotate=90] at (12.7,0) {$\dots$ };
\node[rotate=-90] at (14.5,0) {\tiny \sf choice probabilities};
\end{tikzpicture}

\caption{A RUMnet architecture $N=(N_{U}(\cdot),\{N_{\epsilon_{k_1}}(\cdot)\}_{k_1=1}^K,\{N_{\nu_{k_2}}(\cdot)\}_{k_2=1}^K$).}
\label{fig:rumnet}
\end{figure} 

\subsection{Connection with related models} \label{subsec:discussion}
Before we characterize the expressive power of RUMnets, we briefly discuss the connection to logit-based choice models, which are special cases of our model family.

\noindent \textbf{From MNL to RUMnets.} In its simplest form, the MNL model assumes that the utility for a product $\bs{x} \in \mathcal{X}$ is given by 
\begin{align} \tag{{\sf MNL}}
    U(\bs{x}) = \bs{\beta}^T \bs{x} + \epsilon_{\bs{x}}, 
\end{align}
where $\bs{\beta}$ is a vector of parameters and $\{ \epsilon_{\bs{x}} \}_{\bs{x} \in A}$ is an i.i.d. sequence of standard Gumbel shocks. Note that under this linear specification of the utility, the MNL model cannot leverage the customer attributes $\bs{z}$. To overcome this limitation, researchers often manually specify some non-linearity by introducing cross-terms in the utility function. A more recent approach to capture taste heterogeneity across individuals, proposed by~\cite{han2020neural}, is to specify the utility as follows:
\begin{align} \tag{{\sf TasteNet}} 
    U(\bs{x},\bs{z}) &=   \bs{\beta}^T \bs{x} +  N^{{\sf TasteNet}}(\bs{z})^T \bs{x} + \epsilon_{\bs{x}},
\end{align}
where $N^{{\sf TasteNet}}(\cdot) : \mathbb{R}^{d_z} \to \mathbb{R}^{d_x}$ is a feed-forward neural network. Note that TasteNet can express  non-linear transformations $N^{{\sf TasteNet}}(\bs{z})$ of the customer attributes $\bs{z}$. However, the additive term $N^{{\sf TasteNet}}(\cdot)^T\bs{x}$ still assumes a very specific form for how the customer attributes interact with the product attributes. In fact, nothing prevents us from allowing the utility to be a general function of $\bs{x}$ and $\bs{z}$, just as in the RUM framework. Perhaps the most natural approach is to take TasteNet a step further and define DeepMNL as a RUM discrete choice model where the utility is given by:
\begin{align} \tag{{\sf DeepMNL}} 
    U(\bs{x},\bs{z}) &=   N^{{\sf DeepMNL}}(\bs{x},\bs{z}) + \epsilon_{\bs{x}},
\end{align}
where $N^{{\sf DeepMNL}}(\cdot) : \mathbb{R}^{d_x+d_z} \to \mathbb{R}$ is an arbitrary feed-forward neural network. Here, DeepMNL can be viewed as a special case of the RUMnet architecture in which there are no unobserved attributes $(K=0)$; several variants of DeepMNL have been proposed in the literature~\citep{bentz2000neural,sifringer2020enhancing,wang2020deep}. With respect to Figure~\ref{fig:rumnet}, this approach consists in dropping the first meta-layer, and only keeping the second and third meta-layers. Since there is no latent heterogeneity, there is no need for the fourth (averaging) meta-layer either. 

The above sequence of choice models illustrates a gradual increase of complexity and nonlinearity in the deterministic portion of the utility function, which can be conveniently expressed and estimated using neural networks. As noted by \cite{Train2009}, if the researcher could specify $N^{{\sf DeepMNL}}(\bs{x},\bs{z})$ ``{\em sufficiently that the remaining, unobserved portion of utility is essentially white noise}'', then the DeepMNL family would be ideal. However, regardless of how complex the deterministic portion of the utility is, DeepMNL (and the above special cases) are subject to the same restrictions as the MNL choice model. These approaches do not control for latent (unobserved) factors that may influence the utility, which are captured by various probabilistic structures in the previous literature on discrete choice modeling. Contrary to DeepMNL, our RUMnet architecture captures latent heterogeneity by incorporating unobserved product and customer attributes. The empirical (sample-based) distribution for these unobserved attributes endows RUMnets with nearly universal expressive power, as established in the next section.

\noindent \textbf{Connection to ranking-based choice models.} The SAA underlying the design of RUMnets is akin to the SAA that motivated the study of ranking-based choice models~\citep{farias2013nonparametric}. These models, however, are restricted to a setting where there are no customer attributes, i.e., ${\cal Z} = \{\bs{0}\}$, and product attributes are fixed. Concretely, let ${\cal X}^n = \{\bs{x}_i: i\in [n]\}$ be a set of $n$ products. A ranking-based choice model corresponds to a uniform\footnote{A more general formulation assigns probability weights $\{\lambda_i\}_{i\in [S]}$ to each permutation \citep{farias2013nonparametric}. The uniform distribution, however, suffices to approximate  such weighted distributions through the multiplicity of each ranking. The uniform distribution is more natural to represent a sample of customers~\citep{rusmevichientong2006nonparametric}; however, we could extend RUMnets to turn these uniform probabilities into tunable weights.} distribution over $S$ rankings (or permutations) $\sigma^n_1,
\ldots, \sigma^n_S \in {\cal X}^{[n]}$. Feature-based models are useful when either the number of products is large $n\rightarrow +\infty$, the product features vary frequently (e.g., the same item is offered at different prices), or the choice prediction can be personalized. RUMnets can be interpreted as a method to extend ranking-based models in such contexts.

\noindent \textbf{Connection to LC-MNL models.} When $\bs{\delta}$ is standard Gumbel white noise, the resulting RUMnet model $\pi^{{\sf RUMnet}}_N$ is an instance of the latent-class MNL (LC-MNL) model.  However, such mixtures models are often specified with linear utility functions, which have the potential to approximate any random utility choice model~\citep{mcfadden2000mixed}. This well-known result requires specifying an adequate feature map and involves a complex {\em continuous} mixture distribution. By contrast, RUMnets capture non-linear utility functions while using a parsimonious discrete mixture. 

When $\bs{\delta}$ is standard Gumbel-distributed $\bs{\delta}$ and the utility function is linear, the differences between RUMnets and LC-MNLs are more subtle. Since  the neural network architecture mimics a canonical RUM model, it imposes a specific parametrization of the random utilities. Customer segments share a unique utility function, but they differ in the unobserved attributes, which are random inputs to the utility function. In particular, RUMnets leverage the independence of unobserved attributes (IUA) property. One way to understand this is that the LC-MNL model specification does not distinguish between heterogeneity stemming from unobserved product attributes and heterogeneity stemming from unobserved customer attributes. By contrast, the RUMnet model assumes that these two sources of heterogeneity are distinct: they are captured by probabilistically independent latent variables (IUA property). The RUMnet model is designed according to this structural assumption. This difference with LC-MNLs is numerically illustrated in Section~\ref{sec:synthetic}.

To summarize, while both models can theoretically approximate any RUM model, their practical implementations differ. In the case of the LC-MNL model, there is often a gap between its theoretical flexibility, which requires intricate feature engineering and complex mixing distributions, and its typical practical implementation, which imposes structural assumptions for tractability. While RUMnets are also amenable to theoretical analysis, as shown subsequently, our model class can be readily implemented using open-source deep learning libraries.

\section{Expressive Power of RUMnets} \label{sec:expressive_power}

In this section, we show that RUMnets tightly describe the class of all RUM discrete choice models. In particular, we show that any RUM discrete choice model can be approximated arbitrarily closely by a RUMnet. 
\begin{proposition} \label{prop:approx}
For every RUM discrete choice model $\pi(\cdot)$ of dimension $\bs{d}$ and for every $\eta>0$, there exists a RUMnet architecture  $N \in {\cal N}^{\bs{d}}(K,\Theta^{\ell,w}_M)$ such that, for all choice events $(\bs{z},A) \in \mathcal{Z}\times 2^\mathcal{X}$,
$\max_{\bs{x}\in A}\left|\pi(\bs{x},\bs{z},A) - \pi^{{\sf RUMnet}}_N(\bs{x},\bs{z},A)\right|\leq \eta \ .$
\end{proposition}

To prove Proposition~\ref{prop:approx}, we revisit the celebrated result of~\citet[Theorem~1]{mcfadden2000mixed} showing that  mixed MNL models, i.e., {\em continuous mixtures} of multinomial logit models, uniformly approximate the class of RUM discrete choice models. Quoting~\citet{mcfadden2000mixed}, ``one limitation of Theorem 1 [in that paper] is that it provides no practical indication of how to choose parsimonious mixing families, or how many terms are needed to obtain acceptable approximations to $\pi(\bs{x},\bs{z},A)$''. Proposition~\ref{prop:approx} shows the following property: the choice probabilities of RUM discrete choice models are uniformly approximated by finite mixtures of MNL models such as RUMnets on any choice event in the continuous domain. Building on this result, we will establish in Proposition~\ref{prop:representation} that there exist accurate data-dependent approximations according to the ${\sf KL}$-divergence with a relatively small number of samples $K$.

For this purpose, our proof in Appendix~\ref{app:expressive_power} extends the ideas in~\citet{mcfadden2000mixed} by combining a refined covering lemma with concentration bounds. At a high level, the proof consists in showing that when we approximate the random utility function using feed-forward neural networks, the choice probabilities do not change much. Specifically, we analyze the likelihood of a ``preference reversal'' for every pair of alternatives $(\bs{x},\bs{z})$ and $(\bs{x}',\bs{z})$ with $\bs{x} \neq \bs{x}'$ by controlling the variations of the utility function. Key to this analysis is the existence of a finite covering of $\mathcal{X}^2 \times \mathcal{Z}$ on which we can bound the errors incurred by our SAA approximation. Next, we establish that the reciprocal of Proposition~\ref{prop:approx} holds as well.

\begin{observation} \label{prop:belonging}
	For every RUMnet architecture  $N \in {\cal N}^{\bs{d}}(K,\Theta^{\ell,w}_M)$ and for every $\eta > 0$, there exists a RUM discrete choice model $\pi(\cdot)$ of dimension $\bs{d}$ such that, for all choice events $(\bs{z},A) \in \mathcal{Z}\times 2^\mathcal{X}$, 
	$\max_{\bs{x}\in A}\left|\pi^{{\sf RUMnet}}_N(\bs{x},\bs{z},A) - \pi(\bs{x},\bs{z},A) \right| \leq \eta \ .$
\end{observation}
The proof is straightforward: following the interpretation of our neural network architecture given in Section~\ref{subsec:discussion}, RUMnet architectures directly represent the random choices of a utility-maximizing agent, while a small random perturbation of the RUMnet ensures that the regularity conditions of~\citet{mcfadden2000mixed} are met.\footnote{Note that RUMnets deviate from the canonical form of RUM choice models as we do not enforce that the unobserved attributes are uniformly distributed.}

\section{Dealing with the Curse of Dimensionality} \label{sec:curse_dimensionality}

Given the expressive power of RUMnets demonstrated by Proposition~\ref{prop:approx}, one important concern is the risk of overfitting, which could affect the model's ability to generalize to new, previously unseen data. Hence, in this section, we provide theoretical guarantees on the out-of-sample error of the fitted RUMnets. More specifically, we provide an upper bound on the {\em generalization error}, which measures the degree of overfitting, i.e., it characterizes how the (out-of-sample) expected risk is related to the minimum (in-sample) empirical risk attained by our hypothesis class. We exploit this generalization error bound to establish a compact representation property, showing that any RUMnet architecture can be accurately represented using a relatively ``small'' number of samples.

\subsection{Estimation framework} \label{subsec:estimation}

We formulate the estimation of the neural network of Section~\ref{subsec:neural_net} in the standard Empirical Risk Minimization (ERM) framework; e.g., see~\citet[Chap.~2]{shalev2014understanding}. We assume that our data set is given by a sample of $T$ i.i.d. observations $S =\{(\bs{y}_1,\bs{z}_1,A_1),\ldots,(\bs{y}_T,\bs{z}_T,A_T)\}$, where $(\bs{z}_t,A_t)$ is the $t$-th choice event and $\bs{y}_t \in A_t$ is the product picked by the corresponding customer. To ease the exposition, we further assume that the assortments have a uniform cardinality $|A_t| = \kappa$ for all $t\in [T]$. Notation-wise, ${\cal D}$ stands for the marginal distribution of each observation in the sample, i.e., $S \sim {\cal D}^T$. By a slight abuse of notation, we sometimes use $(\bs{z}_t, A_t) \sim {\cal D}$ (instead of $(\bs{y}_t,\bs{z}_t, A_t) \sim {\cal D}$) to indicate that the choice event is generated according to ${\cal D}$. We do not impose the so-known {\em realizability assumption}, meaning that our modeling approach can be misspecified, i.e., we do not assume that ${\cal D}$ is described by some RUM discrete choice model.

We proceed by formulating our estimation criterion. As an input to our estimation procedure, we specify the parameters $\bs{d},K,w,\ell,M\geq 0$ so that we restrict our estimation procedure to the hypothesis class ${\cal N}^{\bs{d}}(K,\Theta^{\ell,w}_M)$ of RUMnet architectures. To fully specify this class of neural networks, the activation functions are chosen as ReLUs (Rectified Linear Units), which are in popular use. Additionally, we assume that all customer and product attributes are pre-normalized to lie in the range $[-1,1]$.   Our estimator is based on the ERM principle with respect to the negative log-likelihood loss function. Specifically, given a sample $S$ and a discrete choice model $\pi(\cdot)$, we let $L_S(\pi)$ be the empirical negative log-likelihood, namely
\begin{eqnarray} \label{eq:logloss}
L_S\left(\pi\right) = - \dfrac{1}{T} \cdot \sum_{t=1}^T  \log\left(\pi\left(\bs{y}_t, \bs{z}_t,A_t\right)\right)\ .
\end{eqnarray}
With this definition, our approach is to choose the RUMnet architecture that minimizes the above empirical loss over all RUMnet architectures. Formally, $   N^{\sf ERM}_S = \argmin_{N \in {\cal N}^{\bs{d}}(K,\Theta^{\ell,w}_M)}	L_S(\pi_N)$.  Note that the fitted RUMnet architecture $N^{\sf ERM}_S$ is a function of the observed (random) sample $S$. Additionally, let $\pi^{\sf ERM}_S(\cdot)$ be the associated discrete choice model, i.e., $\pi^{\sf ERM}_S (\cdot)= \pi^{{\sf RUMnet}}_{N^{\sf ERM}_S} (\cdot)$. It is worth highlighting that the ERM principle is equally applicable to other loss functions such as mean-squared error or accuracy. Nonetheless, our analysis will focus on log-likelihood-based estimation, in accordance with the fitting procedure used  for other classes of choice models. From a practical perspective, estimating a RUMnet discrete choice model using the ERM principle reduces to training a neural network for multi-label classification. The log-likelihood loss function is often referred to as the cross-entropy loss in the ML practice. Thus, the ERM rule can be implemented using graph optimization and automatic differentiation tools such as Keras~\citep{chollet2015keras}.

\subsection{Learning error guarantees and compact representation}

We define the true error as the expected out-of-sample loss, where the expectation is taken over the unknown distribution $\mathcal{D}$. Specifically, for any discrete choice model $\pi(\cdot)$, let $L^{\sf true}_{\mathcal{D}} (\pi) = \expartwo{S' \sim{\cal D^T}}{L_{S'}(\pi)}$.  Note that this is ideally what we want to minimize. However, since we cannot directly measure out-of-sample performance, as the distribution is unknown, we minimize the empirical error defined in Equation~\eqref{eq:logloss}. The next claim quantifies the gap between these errors.

\begin{proposition} \label{prop:gen-error}
With probability at least $(1- \delta)$, the sampled training set $S$ satisfies
\[
L^{\sf true}_{\mathcal{D}} \left( \pi^{\sf ERM}_S \right)  \leq L_S \left( \pi^{\sf ERM}_S \right) + c_1\cdot \sqrt{\frac{\kappa^3 \log(d)}{T}}  \cdot e^{2 M} (2M)^{\ell}    + (8M+4\log \kappa) \cdot  \sqrt{\frac{2 \cdot \ln(4/\delta)}{T}}  \ .
\]
\end{proposition}

The proof is presented in Appendix~\ref{app:prop:gen-error}, and it follows from standard notions of Rademacher calculus applied to the RUMnet architecture. As expected for neural networks with per-unit $\ell_1$-bounds,  the error bound of Proposition~\ref{prop:gen-error} indicates an exponential dependence on depth $\ell$ of the neural network. Moreover,  we expect a sample complexity of $\Omega(1/p_{\min}) = \Omega(\kappa e^{2M})$ to estimate choice probabilities that can be as small as $p_{\min} = \frac{1}{\kappa}\cdot e^{-2M}$ (see Claim~\ref{clm:pmin} in Appendix~\ref{app:prop:gen-error}). Yet, our bound on the generalization error reveals two interesting properties. First, the bound does not depend on the number of samples $K$. In other words, the  generalization error bound of Proposition~\ref{prop:gen-error} does not degrade with the added complexity from increasing the sample size $K$, contrary to other dimensions of the RUMnet architecture. Second, it is worth noticing a subquadratic dependence $O(\kappa^{\frac{3}{2}} T^{-\frac{1}{2}})$ on the number of choice alternatives $\kappa$, which is better than an existing bound of $\Omega(\kappa^{\ell/2}T^{-\frac{1}{2}})$ by~\citet{wang2020theory}. For another comparison point, choice modeling can be viewed as a structured form of classification problem. In this context, classical Rademacher complexity bounds on margin-based learning for multi-label classification have a quadratic scaling $O(\kappa^2T^{-\frac{1}{2}})$~\citep{koltchinskii2000rademacher,cortes2013multi,mohri2018foundations}, unless further restrictions are imposed on the hypothesis class or a more complex analysis is used.

Building on Proposition~\ref{prop:gen-error}, we next explore the number of samples $K$ required in the RUMnet architecture to accurately describe any RUMnet choice model for a fixed data distribution. Our next claim shows that, with a small $O(\eps)$ loss in accuracy in terms of expected KL-divergence, the number of samples $K$ can be chosen in the order of $O(\frac{1}{\eps^2}{\rm poly}( \log \frac{1}{\eps}, \ell, \kappa, e^M))$. This finding suggests that the latent heterogeneity of the model should be commensurate to the complexity of the feed-forward neural network building blocks of our architecture. 

\begin{proposition} \label{prop:representation}
For every neural network architecture $N \in {\cal N}^{\bs{d}}(K,\Theta^{\ell,w}_M)$, let $N' \in {\cal N}^{\bs{d}}(K',\Theta^{\ell,w}_M)$ be the random neural network architecture obtained from $N$ by taking $K' = \lceil \frac{1}{2\eps^2} \cdot \log \delta \cdot (\kappa e^{2M})^2 \log(\lceil\max\{\frac{c_1}{\eps^2}\kappa^3 \log(d) e^{4M} (2M)^{2\ell}, 128(8M+\log \kappa)^2\ln\frac{4}{\delta}\}\rceil) \rceil$  samples of the unobserved attributes $\{N_{\epsilon_{k_1}}(\cdot)\}_{k_1=1}^{K}$ and $\{N_{\nu_{k_2}}(\cdot)\}_{k_2=1}^{K}$ uniformly at random with replacement. With probability at least $1-\delta$, we have
\begin{align*}
\extwo{(\bs{z},A)\sim {\cal D}}{{\sf KL} \left( \left. \pi^{{\sf RUMnet}}_{N}(\cdot,\bs{z},A)\right| \pi^{{\sf RUMnet}}_{N'}(\cdot,\bs{z},A) \right)} \leq  3\eps \ .
\end{align*}
\end{proposition}
This claim can be viewed as the counterpart of Theorem~4 in the paper by~\citet{chierichetti2018discrete} for feature-dependent choice models in the continuous domain, rather than the discrete domain. The proof appears in Appendix~\ref{app:prop:representation} and combines Proposition~\ref{prop:gen-error} with  concentration bounds.

\section{Numerical Estimation on Synthetic Data} \label{sec:synthetic}

In this section, we present numerical experiments on synthetic data. The objective is to gain insights into the expressive capabilities of RUMnets by successively varying the ground truth model that generates the synthetic choice data. In particular, we show that the proposed model and estimation method can infer non-linearity in the utility function as well as the presence of customer heterogeneity.  The setup is purposely as simple as possible, and, in particular, it does not include customer attributes.

\subsection{Experiment setup} 

We begin by describing three generative processes to construct synthetic datasets. For each setting, we consider a sequence of $T=10,000$ customers, each being presented with an assortment $A_t$ of $\kappa = 10$ products chosen uniformly at random from a universe of 50 products. Each product is endowed with a vector of attributes $\bs{x} = (x_1,x_2,\bs{\delta})$ where $x_1$ and $x_2$ are picked uniformly over the interval $[0,1]$ and $\bs{\delta} \in \mathbb{R}^{50}$ is an indicator vector allowing us to introduce fixed effects for each product. More precisely, $\delta_i = 1$ if $i$ is the index of the chosen product and $0$ otherwise. In each setting, we assume that customers choose according to a random utility model. However, the utility specification differs in each setting. We use this ground truth model to generate a choice event $\bs{y}_t \in A_t$, which corresponds to the product picked by the corresponding customer. Using these generated  observations $S= \{ (\bs{y}_1,A_1), \dots, (\bs{y}_T,A_T)\}$, we use the framework described in Section \ref{subsec:estimation} to fit various RUMnet models. In particular, we experiment with $(\ell,w) \in \{(0,0), (1,3), (2,5) \}$, where recall that $\ell$ denotes the depth of the network and $w$ its width. We denote these different neural network architectures \textsf{linear}, \textsf{non-linear} and \textsf{non-linear$+$} respectively. For our RUMnet architecture, we also test out different number of samples $K \in \{0, 2,5\}$ controlling for the latent heterogeneity of customers and products. Note that $K=0$ corresponds to the DeepMNL model which is a model that does not capture any latent heterogeneity (see Section~\ref{subsec:discussion}). Finally, we also estimated LC-MNL models with $K \in \{0, 2,5\}$ latent classes as an extra benchmark. Remark that both \textsf{linear} DeepMNL and LC-MNL with $K=0$ reduce to the standard MNL model. For each setting, we use 20\% of the data as a validation set for early stopping (see Appendix~\ref{app:implementation} for additional details on the implementation).  Unless specified otherwise, we also generate a sequence of $1,000$ customers that we use as a testing set. We report the log-likelihood loss on the test set averaged over ten different instances, except in Setting 4 where we repeat the experiment on 50 samples.

\noindent \textbf{Setting 1: MNL model.}  The ground truth model is simply an MNL model. In particular, for every vector of product attributes $\bs{x}$, the utility is given by
\begin{align*}
    U^1(\bs{x}) &= \bs{\beta}^T \bs{x}  +  \epsilon,
\end{align*}
where the entries of $\bs{\beta}$ are picked uniformly at random over the interval $[-1,1]$ and $\epsilon$ is a standard Gumbel shock, which is sampled independently across product and customers.

\noindent \textbf{Setting 2: nonlinear utility.} Here, we assume that the ground truth model is described by a non-linear utility function. In particular, for every vector of product attributes $\bs{x}$, the utility is given by
\begin{align*}
    U^2(\bs{x}) &= \beta_1 \cdot x_1 + \beta_2 \cdot x_2 + \beta_3 \cdot x_1^2 + \beta_4 \cdot x_1 \cdot x_2 + \beta_5 \cdot  x_2^2 + \bs{\gamma}^T \bs{\delta} + \epsilon,
\end{align*}
where each entry of $\bs{\beta}$ and $\bs{\gamma}$ is picked uniformly over the interval $[-1,1]$ and $\epsilon$ is a standard Gumbel shock, which is sampled independently across products and customers. Additionally, recall that  $\delta_i = 1$ if $i$ is the index of the chosen product and $0$ otherwise. To accentuate the non-linearity effects, for this setting only, we sample  $x_1$ and $x_2$ uniformly at random over the interval $[0,10]$.

\noindent \textbf{Setting 3: latent heterogeneity.} Our last ground truth model exhibits customer heterogeneity. Specifically, for every vector of product attributes $\bs{x}$, the utility is given by
\begin{align*}
    U^3(\bs{x}) &= b \cdot \bs{\beta}^T \bs{x} + (1-b) \cdot \bs{\gamma}^T \bs{x} + \epsilon,
\end{align*}
where the coordinates of $\bs{\beta}$ and $\bs{\gamma}$ are picked uniformly over the interval $[-50,50]$, $\epsilon$ is a standard Gumbel shock, which is sampled independently across product and customers, and $b$ is a Bernoulli random variable with probability of success $\prpar{b=1}=0.3$. Note that this model is a latent class MNL model with two customer segments. The randomness in $b$ can be equivalently interpreted as unobserved customer heterogeneity in how much the product features are valued. We increase the scale of the parameters $\bs{\beta}$ and $\bs{\gamma}$ to accentuate the violation of the IIA property. 

\noindent \textbf{Setting 4: independence of unobserved attributes.} When the utility function is linear,  RUMnets and latent-class MNL models have very similar architectures. One difference, however, is that RUMnets leverage the probabilistic independence between the unobserved attributes $\eps(\bs{x})$ and $\nu(\bs{z})$, as mentioned in Section~\ref{subsec:discussion}. Here, we test the effect of this structural difference by considering a generative setting with a linear utility function. In particular, our objective is to compare the sample efficiency of RUMnets and LC-MNLs, when both methods are well-specified.

Specifically, for each vector of product attributes $\bs{x},\bs{z}$, the utility is given by
\begin{equation} \label{U4:eq2}
\begin{aligned}
    U^4(\bs{x},\bs{z}) &= b_z \cdot \bs{\beta}^T \bs{x} + (1-b_z) \cdot \bs{\gamma}^T \bs{x} +b_x \cdot \bs{z}^T\Omega_1\bs{x} + (1-b_x) \cdot  \bs{z}^T\Omega_2\bs{x}  + \epsilon \\
    & = b_z \cdot \bs{\beta}^T \bs{x} + (1-b_z) \cdot \bs{\gamma}^T \bs{x} +b_x \cdot Tr(\Omega_1 \cdot \bs{x}\cdot \bs{z}^T) + (1-b_x) \cdot  Tr(\Omega_2\cdot\bs{x}\cdot\bs{z}^T)  + \epsilon  \ ,
\end{aligned}
\end{equation}
Here, $\bs{\beta}, \bs{\gamma},\bs{x}$ are generated as before, and $\bs{z} = (z_1,z_2)$ are observed customer attributes with coordinates $z_i$ drawn uniformly over the interval $[0,1]$. The matrices $\Omega_1,\Omega_2$ have all their entries drawn uniformly over the interval $[-50,50]$. Like in Setting 3, the Bernoulli random variable $b_z \in \{0,1\} $ corresponds to unobserved customer attributes that affect how much the product features are valued. Additionally, the Bernoulli random variable $b_x \in \{0,1\}$ captures unobserved product attributes: depending on the realization of $b_x$, we apply different product feature mappings $x\mapsto \Omega_1 x$ or $x\mapsto \Omega_2 x$. By hypothesis, $b_x,b_z$ are independent.

This model can be represented as linear-utility LC-MNL model with a uniform mixture over 4 customer classes, corresponding to each distinct realization of the vector $(b_z,b_x)$. Equivalently, this model can be specified as a specialized RUMnet architecture with $K=2$ samples.  Both specifications are linear, as follows from Equation~\eqref{U4:eq2} when we provide the the vector $\bs{x}$ as well as the matrix $\bs{x}\cdot\bs{z}^T$ as inputs to the utility function (rather than just $\bs{x},\bs{z}$). We vary the number of samples in the training set $T\in \{100,500,1000,5000,10000\}$ and compare the fitted models' out-of-sample error. We repeat each experiment for fixed values of $T$ on 50 randomly generated synthetic instances.

\subsection{Results}

\begin{figure}[t]
\centering
\begin{subfigure}{.45\textwidth}
  \centering
  \begin{tikzpicture}[scale=0.8]
  \begin{axis}[xlabel={$K$}, ylabel={Log-likelihood loss}, axis y line=left, axis x line=bottom, xmin=0, xmax=0.4, ymin=2.1, ymax=2.3, legend pos=north east,
        xtick=\empty,
	    xtick pos=bottom,
	    extra x ticks={0.1,0.2,0.3},
		extra x tick style={grid=none},
	    extra x tick labels={{0},{2},{5}}] 
        \addplot[smooth,blue,very thick] coordinates {(0.1, 2.156691861152649) (0.2, 2.15685396194458 ) (0.3, 2.15668203830719)};
         \addlegendentry{RUMnet \textsf{linear}}
        \addplot[smooth,blue,very thick, dashed] coordinates {(0.1, 2.156691861152649) (0.2, 2.15685396194458 ) (0.3, 2.15668203830719)};
        \addlegendentry{RUMnet \textsf{non-linear}}
        \addplot[smooth,blue,very thick, dotted] coordinates {(0.1, 2.156555414199829) (0.2, 2.157793879508972 ) (0.3, 2.1583486795425415)};
        \addlegendentry{RUMnet \textsf{non-linear$+$}}
        \addplot[smooth,purple,very thick] coordinates {(0.1, 2.156691861152649) (0.2, 2.1567511320114137 ) (0.3, 2.156590390205383)};
        \addlegendentry{LC-MNL}
        
        \addplot[smooth,orange,very thick] coordinates {(0.1, 2.153219033698573) (0.2, 2.153219033698573 ) (0.3, 2.153219033698573)};
        \addlegendentry{Ground truth}
       
        \end{axis}
 \end{tikzpicture}
  \caption{Setting 1: MNL model}
  \label{fig:synthetic_1}
\end{subfigure}
\begin{subfigure}{.45\textwidth}
  \centering
 \begin{tikzpicture}[scale=0.8]
     \begin{axis}[xlabel={$K$}, ylabel={Log-likelihood loss}, axis y line=left, axis x line=bottom, xmin=0, xmax=0.4, ymin=0.15, ymax=0.45, xtick=\empty,
	    xtick pos=bottom,
	    extra x ticks={0.1,0.2,0.3},
		extra x tick style={grid=none},
	    extra x tick labels={{0},{2},{5}},
            ytick={0.1,0.2,0.3,0.4,0.5},
            yticklabels={0.1,0.2,0.3,0.4,0.5},] 
        \addplot[blue,very thick] coordinates {(0.1, 0.40911174938082695) (0.2, 0.40014880821108817 ) (0.3, 0.3980639398097992)};
        \addplot[blue,very thick, dashed] coordinates {(0.1, 0.32557238265872) (0.2, 0.23920998796820642 ) (0.3, 0.24567788392305373)};
        \addplot[blue,very thick, dotted] coordinates {(0.1, 0.22605194076895713) (0.2, 0.22970900759100915 ) (0.3, 0.2373579502105713)};
        \addplot[purple,very thick] coordinates {(0.1, 0.40911174938082695) (0.2, 0.39988917112350464 ) (0.3, 0.39921413734555244)};
  \addplot[orange,very thick] coordinates {(0.1, 0.21475911063409053) (0.2, 0.21475911063409053 ) (0.3, 0.21475911063409053)};
        \end{axis}
 \end{tikzpicture}
  \caption{Setting 2: Non-linearity}
  \label{fig:synthetic_2}
\end{subfigure}%
\begin{subfigure}{.45\textwidth}
  \centering
 \begin{tikzpicture}[scale=0.8]
    \begin{axis}[xlabel={$K$}, ylabel={Log-likelihood loss}, axis y line=left, axis x line=bottom, xmin=0, xmax=0.4, ymin=0.5, ymax=1.2,legend pos=north east,
        xtick=\empty,
	    xtick pos=bottom,
	    extra x ticks={0.1,0.2,0.3},
		extra x tick style={grid=none},
	    extra x tick labels={{0},{2},{5}}] 
        \addplot[blue,very thick] coordinates {(0.1, 1.1670116424560546) (0.2, 0.7360277771949768 ) (0.3, 0.6860689282417297)};
         \addplot[blue,very thick, dashed] coordinates {(0.1, 1.1241455793380737) (0.2, 0.7378921627998352 ) (0.3, 0.6909700334072113)};
       \addplot[blue,very thick, dotted] coordinates {(0.1, 1.1064848065376283) (0.2, 0.7377565383911133 ) (0.3, 0.6961209416389466)};
        
        \addplot[ purple,very thick] coordinates {(0.1, 1.1670116424560546) (0.2, 0.7417652249336243 ) (0.3, 0.7358592748641968)};
        
        \addplot[orange,very thick] coordinates {(0.1, 0.6562055471637595) (0.2, 0.6562055471637595 ) (0.3, 0.6562055471637595)};
        
        \end{axis}
 \end{tikzpicture}
  \caption{Setting 3: Heterogeneity}
  \label{fig:synthetic_3}
\end{subfigure}
\caption{Average out-of-sample log-likelihood loss for different settings and models.}
\label{fig:synthetic}
\end{figure}

\begin{figure}[t]
\centering

\begin{subfigure}{.5\textwidth}
  \centering
 \begin{tikzpicture}[scale=0.8]
    \begin{axis}[xlabel={Training sample size}, ylabel={Log-likelihood loss}, axis y line=left, axis x line=bottom, xmin=5.5, xmax=14, ymin=0.9, ymax=1.9, legend pos=north east, legend style={at={(1,1.2)}, anchor=north , font=\small},
        xmode=log,
        xmin=100, xmax=10000]
     \addlegendentry{MNL}
        \addplot[smooth,blue,very thick] coordinates 
{(100, 1.679126)
(500, 1.559159)
(1000, 1.533763)
(5000.5, 1.528092)
(10000, 1.494944)
};
\addlegendentry{RUMnet (K=2)}
        \addplot[smooth,blue,very thick,dashed] coordinates 
{(100, 1.741240)
(500, 1.449057)
(1000, 1.271586)
(5000, 1.104755)
(10000, 1.052423)
};
\addlegendentry{LC-MNL (K=2)}
        \addplot[smooth,purple,very thick] coordinates 
{(100, 1.732867)
(500, 1.493742)
(1000, 1.366671)
(5000, 1.271969)
(10000, 1.231649)
};
\addlegendentry{LC-MNL (K=4)}
        \addplot[smooth,purple,very thick,dashed] coordinates 
{(100, 1.730340)
(500,  1.464199)
(1000, 1.313536)
(5000, 1.112478)
(10000, 1.048818)
};
\addlegendentry{LC-MNL (K=8)}
        \addplot[smooth,purple,very thick,dotted] coordinates 
{(100, 1.727490)
(500, 1.445289)
(1000, 1.308969)
(5000, 1.119124)
(10000, 1.052102)
};
\addlegendentry{Ground truth}
        \addplot[smooth,orange,very thick] coordinates 
{(100, 0.997805)
(500, 0.997805)
(1000, 0.997805)
(5000, 0.997805)
(10000, 0.997805)
};
        \end{axis}
 \end{tikzpicture}
  \caption{Log-likelihoods loss (NLL) as a function of $T$.}
  \label{fig:synthetic_4}
\end{subfigure}
\begin{subfigure}{.45\textwidth}
  \begin{tikzpicture}[scale=0.8]
    \begin{axis}[
        xlabel={Training sample size},
        ylabel={Difference in negative log-likelihood},
        xmode=log,
        xmin=100, xmax=10000,
        ymin=-0.07, ymax=0.02,
        width=8cm,
        height=8cm,
        legend pos=north east,
        axis lines=left, 
        scaled y ticks=false,
        draw = none,
        ytick=\empty,
	    extra y ticks={-0.08,-0.06,-0.04,-0.02,0,0.02,0.04},
		extra y tick style={grid=none},
	    extra y tick labels={-0.08,-0.06,-0.04,-0.02,0,0.02,0.04}
    ]
    \addplot [
        thick,
        blue,
        mark=none,
    ] coordinates {
        (100, 0.010900163650512695) (500, -0.015142292976379394) (1000, -0.04194948434829712) (5000, -0.007722471952438355) (10000, 0.003604609966278076)
    };
    \addplot [
        name path=A,
        draw=none,
    ] coordinates {
        (100, 0.007494980540068163) (500, -0.025696226056709195) (1000, -0.053631274859001064) (5000, -0.017418113495328583) (10000, -0.009248187267668006)
    };
    \addplot [
        name path=B,
        draw=none,
    ] coordinates {
        (100, 0.014305346760957228) (500, -0.004588359896049592) (1000, -0.03026769383759318) (5000, 0.0019731695904518743) (10000, 0.016457407200224157)
    };
    \addplot [blue, opacity=0.2] fill between [of = A and B];
    \end{axis}
\end{tikzpicture}
  \label{fig:gap}
  \caption{NLL gap: LC-MNL($K=4$) versus RUMnets}
\end{subfigure}
\caption{Setting 4: Average out-of-sample log-likelihood loss and performance gap of RUMnets and LC-MNLs as a function of the training sample size $T$.}
\label{fig:synthetic_5}
\end{figure}

In Setting~1, the dataset is generated using an MNL model. In this case, the utility function is linear and does not exhibit any heterogeneity. As a result, as illustrated in Figure~\ref{fig:synthetic_1}, the average log-likelihood loss is insensitive to the complexity of the feed-forward neural network $(\ell,w)$ as well as the number of samples $K$. Despite being highly over-parametrized, RUMnets almost recover the ground truth's performance; this  validates the efficiency of our early-stopping regularization method. In Setting~2, we use a non-linear utility function to generate the synthetic dataset. As can be seen in Figure~\ref{fig:synthetic_2}, RUMnet is able to capture this non-linearity by increasing the complexity of the the feed-forward neural network. In particular, as $(\ell,w)$ ``increases'', i.e., going from \textsf{linear} to \textsf{non-linear} to \textsf{non-linear$+$}, the out-of-sample log-likelihood of RUMnet improves and it approaches that of the ground truth model, which is represented by the orange line. In this case, note that increasing $K$ does not add much value. This is consistent with the fact that our ground truth model does not exhibit any form of customer heterogeneity. Unlike Setting~2, recall that Setting~3 has two customer classes each choosing according to a distinct MNL model. In this case, as illustrated in Figure~\ref{fig:synthetic_3}, the performance of RUMnet improves with $K$ but it is mostly insensitive to the  complexity $(\ell,w)$ of the feed-forward neural networks. 

To compare RUMnets and LC-MNLs, Setting 4 exhibits both customer and product heterogeneity. Both models are well-specified. Thus, as expected, their out-of-sample log-likelihoods are very close to each other and converge to the ground truth level for a sufficiently large sample $T\approx 10^4$, as illustrated in Figure~\ref{fig:synthetic_5}. On the other extreme, for very small samples $T\approx 10^2$,  both the RUMnets' and LC-MNLs' performances ``shrink'' to that of the standard MNL model. However, in the intermediary regime $T\approx 10^3$, we observe some statistically significant differences between RUMnets and LC-MNLs. Specifically, the gap in average log-likelihood loss is in the order of 4\% for $T=10^3$, indicating that it is marginally more efficient to estimate RUMnets than LC-MNLs on relatively small data sets. Based on the architecture design, this is due to the IUA property and the weight sharing it induces across customer segments in the RUMnet architecture. The sample efficiency of LC-MNLs does not improve much by over-specifying $K=8$  or under-specifying $K=2$.

To summarize, our synthetic experiments validate each component of the RUMnet architecture. 

\subsection{Additional synthetic experiments} 

\paragraph{Identifiability.} It is well-known that model identifiability breaks even for simple LC-MNL models; see, e.g., \cite{ammar2014s}.  Nevertheless, since the RUMnet architecture directly models the utility function, we can numerically study the extent to which the fitted RUMnets recover the ground truth utility. We conduct such experiments in Appendix~\ref{app:uniqueness} where we observe that there is a strong correlation between the learned utility and the ground truth utility. 

\paragraph{Ranking-based models.}  Section~\ref{subsec:discussion} showed the connection between RUMnets and ranking-based models. In this setting, the observable product attributes $\bs{x}$ are a one-hot encoding of a finite universe of products. There are no observable customer attributes $\bs{z}$ (or equivalently, these are constant).  RUMnets provide a new methodology to estimate and approximate ranking-based models. We explore this in Appendix~\ref{app:rank} on a synthetic dataset. We observe that RUMnets tend to be more efficient than existing implementations of ranking-based models, in terms of predictive accuracy and running times. However, the edge comes from the estimation method. This suggests that ML/DL libraries offer new efficient computational methods to estimate existing choice models.

\section{Numerical Estimation on Real Data} \label{sec:numerical}

We test the predictive accuracy of RUMnets against a comprehensive set of benchmarks on two real-world datasets, which reflect complementary choice prediction settings.

\subsection{Benchmarks} \label{subsec:benchmark}

We implement various choice models and classification algorithms proposed in previous literature. 

\noindent \textbf{MNL and neural network-based extensions.} The first set of benchmarks comprises the MNL model and the two neural network-based extensions, TasteNet and DeepMNL, which we presented in Section~\ref{subsec:discussion}. While the MNL model specifies a linear utility function, the subsequent models capture nonlinear effects of increasing complexity, and can be viewed as state-of-the-art benchmarks. From an implementation perspective, we formulate each of these models as a neural network with a softmax prediction layer. 

\noindent \textbf{Model-free approaches.} In addition, we test two model-free benchmarks that treat the choice prediction task as a multi-label classification problem and forego any additional structure on the probabilistic choices: 
\begin{enumerate}
\item {\em Vanilla Neural Network (VNN):}  The input to this feed-forward neural network is the concatenation of all product and customer attributes. The output is a vector of length $|A|$ that represents the utility of each choice alternative, which is then passed through a softmax layer to compute the corresponding choice probabilities.

\item {\em Random Forest (RF):} We train a variant of the random forests for choice prediction which was very recently introduced by~\citet{chen2019use} and~\citet{chen2020decision}. Importantly, previous work focus on a different observational setting, where the assortment composition varies but each product's attributes are essentially fixed. Due to contextual variations in product and customer attributes, we implement a featurized version of this methodology, which was suggested by~\citet{chen2019use}. The input is the concatenation of product features in the assortment and customer features; we include a binary variable to indicate which products are offered in the assortment (see Appendix~\ref{app:expedia}). We note that the optimization-based estimation methods developed by~\citet{chen2020decision} are not easily applicable in this setting.

\end{enumerate}

\subsection{Implementation specifications}

Except for the RF model, we implement all other benchmarks using Keras. Appendix~\ref{app:implementation} details our approach for estimating the neural networks-based models.  
We use label smoothing as a norm-based regularization method on the neural network's outputs and specify Exponential Linear Unit (ELU) activation functions~\citep{clevert2015fast}.  For each model, we conduct an extensive parameter search in the training process. We use the validation set to tune the hyper-parameters on each data split.  We test different complexities for the building-block neural networks. In particular, for each neural network-based model, we vary the parameters $(\ell,w) \in \{ (3,10),(5,20), (10,30) \}$, where recall that $\ell$ denotes the depth of the network and $w$ its width.\footnote{We omit certain instantiations of $(\ell,w)$ if the corresponding running times are prohibitive, or if the number of parameters becomes excessive.} For our RUMnet architecture, we also vary number of samples $K \in \{5,10\}$ controlling for the latent heterogeneity of customers and products. We report the performance of the DeepMNL model of Section~\ref{subsec:discussion}, which is essentially a RUMnet model that does not capture any latent heterogeneity, i.e., $K=1$. For RFs, we jointly vary the number of trees in the forest in $\{50,100,200,400\}$ and the maximum tree depth in $\{5,10,20\}$; we then pick the forest with the best loss on the validation set. 

We use 10-fold cross-validation and report the averaged log-likelihood loss and classification over the ten splits. More precisely, we use a 80/10/10 split for the train/validation/test sets respectively. For each model, we use the validation set to determine the best value of the hyperparameters, such as the depth and width of the network for RUMnet, and we then evaluate the corresponding models on the test set.  Importantly, the test set is never seen during training, which enables us to assess the out-of-sample performance of the models.  In addition to the log-likelihood loss defined in Equation~\eqref{eq:logloss}, we also report the {\em accuracy}, defined as the percentage of correct predictions for the chosen alternative. In terms of running time, RUMnets are more computationally intensive than  other methods, as we  discuss in Appendix~\ref{app:implementation}.

\subsection{Results} \label{subsec:data-swiss}

The first dataset (Swissmetro data) consists of responses to a survey in Switzerland to assess the potential demand for a new mode of public transportation, called Swissmetro. The alternatives include Swissmetro, Train or Car ($\kappa =3$), and the number of observations in the data is in the order of 10K. The second one (Expedia data) is a larger dataset. This dataset counts around 400K distinct search queries, 36 hotel features, and 56 customer and search features.  This setting mirrors large-scale transaction data with assortments formed by many distinct alternatives ($\kappa = 39$). More details on each dataset are presented in Appendices~\ref{app:swissmetro} and~\ref{app:expedia}. 

\begin{table}\centering
\caption{Out-of-sample predictive performance of the fitted choice models averaged over ten folds. Standard errors are reported in parentheses.} \label{tab:numerics_summary}
\setlength\extrarowheight{2pt} 
\resizebox{12cm}{!}{
\begin{tabular}{lcccc}
&  \multicolumn{2}{c}{Swissmetro} & \multicolumn{2}{c}{Expedia}    \\
 \cmidrule[1pt](lr){2-3} \cmidrule[1pt](lr){4-5}
  Model  & Log-likelihood loss & Accuracy & Log-likelihood loss & Accuracy   \\
  \cmidrule[1pt](lr){1-5} 
  MNL & \begin{tabular}{@{}c@{}} 0.842 \vspace{-1ex} \\ \tiny ($0.5\cdot 10^{-2}$)\end{tabular}  &  \begin{tabular}{@{}c@{}} 0.623 \vspace{-1ex} \\ \tiny ($3.6\cdot 10^{-3}$)\end{tabular}  & \begin{tabular}{@{}c@{}} 2.482 \vspace{-1ex} \\ \tiny ($1.9\cdot 10^{-3}$)\end{tabular}  & \begin{tabular}{@{}c@{}} 0.306 \vspace{-1ex} \\ \tiny ($7.1\cdot 10^{-4}$)\end{tabular}  \\
  Latent class MNL & \begin{tabular}{@{}c@{}}  0.764 \vspace{-1ex} \\ \tiny ($0.4\cdot 10^{-2}$)\end{tabular}  &  \begin{tabular}{@{}c@{}} 0.652 \vspace{-1ex} \\ \tiny ($2.1\cdot 10^{-3}$)\end{tabular}  & \begin{tabular}{@{}c@{}} 2.434 \vspace{-1ex} \\ \tiny ($1.9\cdot 10^{-3}$)\end{tabular}  & \begin{tabular}{@{}c@{}} 0.317 \vspace{-1ex} \\ \tiny ($6.3\cdot 10^{-4}$)\end{tabular}  \\
  TasteNet & \begin{tabular}{@{}c@{}} 0.568 \vspace{-1ex} \\ \tiny ($1.2\cdot 10^{-2}$)\end{tabular} &  \begin{tabular}{@{}c@{}} 0.782 \vspace{-1ex} \\ \tiny ($3.7\cdot 10^{-3}$)\end{tabular} & \begin{tabular}{@{}c@{}} 2.092 \vspace{-1ex} \\ \tiny ($1.8\cdot 10^{-3}$)\end{tabular}  & \begin{tabular}{@{}c@{}} 0.413 \vspace{-1ex} \\ \tiny ($1.0\cdot 10^{-3}$)\end{tabular}  \\
  {DeepMNL} & \begin{tabular}{@{}c@{}} 0.592 \vspace{-1ex} \\ \tiny ($1.1\cdot 10^{-2}$)\end{tabular}  &  \begin{tabular}{@{}c@{}} 0.771 \vspace{-1ex} \\ \tiny ($3.2\cdot 10^{-3}$)\end{tabular} & \begin{tabular}{@{}c@{}} 2.069 \vspace{-1ex} \\ \tiny ($3.6\cdot 10^{-3}$)\end{tabular}  & \begin{tabular}{@{}c@{}} 0.412 \vspace{-1ex} \\ \tiny ($1.5\cdot 10^{-3}$)\end{tabular}  \\
   \cmidrule (lr){1-5} 
  RUMnet & \begin{tabular}{@{}c@{}} 0.561 \vspace{-1ex} \\ \tiny ($1.5\cdot 10^{-2}$)\end{tabular}  & \begin{tabular}{@{}c@{}} \cellcolor{green!75} 0.793\vspace{-1ex} \\ \tiny ($4.3\cdot 10^{-3}$)\end{tabular}  & \begin{tabular}{@{}c@{}}\cellcolor{green!75} 2.018\vspace{-1ex} \\ \tiny ($2.1\cdot 10^{-3}$)\end{tabular}  & \begin{tabular}{@{}c@{}} \cellcolor{green!75} 0.425 \vspace{-1ex} \\ \tiny ($6.2\cdot 10^{-4}$)\end{tabular}  \\ 
 \cmidrule(lr){1-5} 
  Vanilla Neural Network &  \begin{tabular}{@{}c@{}} 0.612 \vspace{-1ex} \\ \tiny ($1.0\cdot 10^{-2}$)\end{tabular} & \begin{tabular}{@{}c@{}} 0.749 \vspace{-1ex} \\ \tiny ($3.3\cdot 10^{-3}$)\end{tabular} & \begin{tabular}{@{}c@{}} 2.407 \vspace{-1ex} \\ \tiny ($2.6\cdot 10^{-3}$)\end{tabular}  & \begin{tabular}{@{}c@{}} 0.328 \vspace{-1ex} \\ \tiny ($6.8\cdot 10^{-4}$)\end{tabular}  \\
  Random Forest &  \begin{tabular}{@{}c@{}} \cellcolor{green!75} 0.523 \vspace{-1ex} \\ \tiny ($0.7\cdot 10^{-2}$)\end{tabular} & \begin{tabular}{@{}c@{}} 0.776  \vspace{-1ex} \\ \tiny ($4.7\cdot 10^{-3}$)\end{tabular} & \begin{tabular}{@{}c@{}} 2.712 \vspace{-1ex} \\ \tiny ($2.0\cdot 10^{-3}$)\end{tabular}  & \begin{tabular}{@{}c@{}} 0.309 \vspace{-1ex} \\ \tiny ($8.1\cdot 10^{-4}$)\end{tabular}  \\
 \cmidrule[1pt](lr){1-5} 
\end{tabular}
}
\end{table} 

The predictive performance of the models is reported in Table~\ref{tab:numerics_summary}. RUMnet emerges as the most predictive method out of the benchmarks in terms of accuracy. All the tested models significantly outperform the basic MNL benchmark in terms of  log-likelihood loss and prediction accuracy. The gaps in the log-likelihoods exceed $20\%$ on both datasets, thereby indicating that a linear utility-based model is too restrictive to predict choices in these settings accurately. Now, comparing TasteNets and DeepMNLs to RUMnets, we can attribute the gain in predictive performance to the incorporation of  latent heterogeneity. Table~\ref{tab:numerics_detailed} in Appendix~\ref{app:full_table} provides a more detailed perspective on the effects of the architecture design. Considering TasteNets and DeepMNLs, we see that increasing the complexity of the neural networks from  $(\ell,w) = (3,10)$ to $(\ell,w) = (5,20)$ improves predictive performance; this phenomenon  can be imputed to a more complex nonlinear utility function. However, there are no marginal gains from fitting even larger neural networks $(\ell,w) = (10,30)$.  These observations suggest that it is impossible to ``make up'' for the lack of latent heterogeneity using a more complex utility function. This aligns with our controlled synthetic experiments in Section~\ref{sec:synthetic}, where we varied the parametric and probabilistic specification of the ground truth utility. The differences in predictive performance between RUMnets and DeepMNL are statistically significant on both datasets.\footnote{In terms of log-likelihood loss, the paired t-tests with respect to the 10 splits yield p-values of $5.4 \cdot 10^{-2}$ on Swissmetro data and $9.33 \cdot 10^{-10}$ on Expedia data.}

The model-free methods, RF and VNN, achieve strong predictive performance on the Swissmetro data. In fact, RFs achieve lower negative log-likelihoods than RUMnets in that setting. In stark contrast, both methods perform poorly on the Expedia dataset.  We note that RFs suffer from a high level of overfitting, as indicated by the gap in performance on our training and test data (see Table~\ref{tab:numerics_detailed}, Appendix~\ref{app:full_table}).\footnote{Note that the hyper-parameters of RFs are tuned using the validation set: our estimation method enables selecting smaller tree depths or fewer trees in the forest. Yet, these choices do not result in better out-of-sample performance.} This noteworthy phenomenon might be related to the large number of distinct alternatives and the assortment size of Expedia data. Indeed, for such model-free methods, an explosion of the number of parameters seems unavoidable as the input dimension increases (see Table~\ref{tab:params}, Appendix~\ref{app:implementation}). This potential explanation is supported by controlled synthetic experiments, which we present in the next section.  Another possible cause is that it is difficult to encode assortments of varying sizes for  Random Forests (see Appendix~\ref{app:expedia}).

\subsection{Model-free methods on high-dimensional synthetic data} \label{sec:rfs}

Although the predictive performance of the random forest (RF) approach is comparable to that of RUMnet on the Swissmetro dataset, this method has a poor performance on the Expedia dataset with a high level of overfitting (see Section~\ref{subsec:data-swiss}). To explain this phenomenon, we hypothesize that the performance of RFs degrades as the dimension of the underlying prediction problem increases, i.e.,  the number of product and customer attributes and/or the number of distinct choice alternatives which together determine the dimension of the RF inputs. In this section, we conduct synthetic experiments that support this hypothesis. 

In particular, we re-use Setting~1 of our synthetic experiments, described in Section~\ref{sec:synthetic}. Recall that in this setting, we generate observations using an MNL ground truth model. Each customer is presented with $\kappa$ products chosen uniformly randomly from a universe of $P$ products. We fix $\kappa = 5$ and experiment with $P \in  \{5,10,25,50,100\}$. Each product is endowed with a vector of attributes $\bs{x} = (x_1,x_2,\bs{\delta})$ where $x_1$ and $x_2$ are picked uniformly over the interval $[0,1]$ and $\bs{\delta} \in \mathbb{R}^{P}$ is an indicator vector allowing us to introduce fixed effects for each product. Here, note that the input size grows with the number of products in the universe $P$ allowing to test how the performance of the random forest scales with $P$. Moreover, for the random forest method, we vary the number of trees in the forest in $\{200,300,400\}$ and the maximum tree depth in $\{5,10,20\}$; we then pick the forest with the best loss on the validation set.

\begin{figure}[t]
  \centering
 \begin{tikzpicture}[scale=0.8]
    \begin{axis}[xlabel={Number of products in the universe}, ylabel={Log-likelihood loss}, axis y line=left, axis x line=bottom, xmin=0, xmax=100, legend style={at={(0.5,0.5)},anchor=west},yticklabel style={
        /pgf/number format/fixed,
        /pgf/number format/precision=3
},
scaled y ticks=false]
                \addplot[smooth,blue,very thick] coordinates {(5,1.48310403823852)(10,1.50701283216476) (25,1.5071481347084) (50,1.50239645242691) (100,1.49113008975982)};
         \addlegendentry{RUMnet}
         \addplot[smooth,blue,very thick, dashed] coordinates {(5,1.51138885664381)(10,1.55728749981111) (25,1.5722936013494) (50,1.58314795095763) (100,1.58935394243988)};
        \addlegendentry{Random forest}
        \addplot[smooth,orange,very thick, dashed] coordinates {(5,1.609437912) (10,1.609437912) (25,1.609437912) (50,1.609437912) (100,1.609437912)};
        \addlegendentry{Random guess}
        \addplot[smooth, orange, very thick] coordinates {(5,1.48292136165527) (10,1.50532667460634) (25,1.50430741525726) (50,1.49871819833564) (100,1.48303601143115) };
        \addlegendentry{Ground Truth}
        \end{axis}
 \end{tikzpicture}
\caption{Log-likelihood loss on the test set as a function of the number of products. RUMnet has parameters $\boldsymbol{(\ell,w)=(2,5)}$ and $\boldsymbol{K=5}$. Note that the y-axis log-likelihood losses  vary from 1.48 to 1.61.}
\label{fig:rf}
\end{figure}

Figure~\ref{fig:rf} shows the average log-likelihood loss on the test set averaged over ten splits. We also report on this figure the average log-likelihood loss of the RUMnet model with parameters $(\ell,w)=(2,5)$ and $K=5$, as well as the average log-likelihood loss of a naive model, termed \emph{RandomGuess}, that prescribes a uniform choice probability distribution over the $\kappa$ offered products. We observe that as $P$ increases, the performance of RFs diverges from that of the ground truth model and nearly matches that of RandomGuess. This shows that when the input dimension is large, the random forest approach is unable to learn generalizable patterns from the data. On the other hand, the predictive performance of the RUMnet model is quite close to the ground truth model. Although this is expected because the ground truth model is in the RUM family, the performance of RUMnet is insensitive to the value of the number of products $P$, as $P$ increases. This highlights that our method scales well with respect to the number of attributes and distinct products in the universe in contrast with model-free methods.

\subsection{Visualization} \label{app:viz_rf}

Neural networks are often viewed as ``black-box'' tools that are hard to interpret. Hence, it is important to explore what type of substitution patterns are captured by the fitted RUMnets and contrast them with other methods. As a first step, we visualize how the RUMnet predictions vary as a function of the price, which is arguably one of the most important operational dimension, using the Swissmetro dataset. In order to visualize how RUMnets predict the customers' substitution patterns, we use the $K$-means method to determine a balanced partition of the customer features and subsequently define the centroid of each cluster as a distinct customer type. Figure~\ref{fig:types} plots the predicted choice probabilities of each customer types as a function of the Swissmetro cost. Our clustering reveals three distinct choice behaviors: Customer~1 is mostly price insensitive and chooses the Swissmetro with high probability even when increasing its cost. Both Customers~2 and 3 are price sensitive and choose the Swissmetro option when its cost is low enough. Interestingly, they differ in what they substitute to when the cost of the Swissmetro is too high: Customer~2 chooses the Train option whereas Customer~3 chooses the Car option.

For all customer types, the choice probability of the Swissmetro decreases with its cost. This shows that RUMnets capture realistic substitution patterns with respect to price.  Note that in practice, these relationships may or may not be observed due to a variety of issues, including the presence of endogeneity and overfitting. In fact, in the case of random forests, the predicted choice probability for Swissmetro does not always decrease as its price increases. Figure~\ref{fig:monotonicityRF} in Appendix~\ref{app:add_viz} illustrates this phenomenon. This finding shows that model-free methods may capture unrealistic dependencies on product attributes, in spite of their strong predictive accuracy.  

\begin{figure}[t!]
\centering
\begin{subfigure}{.3\textwidth}
  \centering
  \begin{tikzpicture}[scale=0.6]
\newcommand{\probaTrain}{
( 0.00000 , 0.01161 )
( 0.00303 , 0.01171 )
( 0.00606 , 0.01180 )
( 0.00909 , 0.01190 )
( 0.01212 , 0.01200 )
( 0.01515 , 0.01210 )
( 0.01818 , 0.01220 )
( 0.02121 , 0.01230 )
( 0.02424 , 0.01241 )
( 0.02727 , 0.01252 )
( 0.03030 , 0.01263 )
( 0.03333 , 0.01274 )
( 0.03636 , 0.01285 )
( 0.03939 , 0.01297 )
( 0.04242 , 0.01309 )
( 0.04545 , 0.01321 )
( 0.04848 , 0.01333 )
( 0.05152 , 0.01345 )
( 0.05455 , 0.01357 )
( 0.05758 , 0.01370 )
( 0.06061 , 0.01382 )
( 0.06364 , 0.01395 )
( 0.06667 , 0.01408 )
( 0.06970 , 0.01421 )
( 0.07273 , 0.01435 )
( 0.07576 , 0.01448 )
( 0.07879 , 0.01462 )
( 0.08182 , 0.01476 )
( 0.08485 , 0.01489 )
( 0.08788 , 0.01504 )
( 0.09091 , 0.01518 )
( 0.09394 , 0.01532 )
( 0.09697 , 0.01547 )
( 0.10000 , 0.01561 )
( 0.10303 , 0.01576 )
( 0.10606 , 0.01591 )
( 0.10909 , 0.01606 )
( 0.11212 , 0.01621 )
( 0.11515 , 0.01637 )
( 0.11818 , 0.01652 )
( 0.12121 , 0.01668 )
( 0.12424 , 0.01684 )
( 0.12727 , 0.01700 )
( 0.13030 , 0.01716 )
( 0.13333 , 0.01733 )
( 0.13636 , 0.01749 )
( 0.13939 , 0.01766 )
( 0.14242 , 0.01783 )
( 0.14545 , 0.01800 )
( 0.14848 , 0.01817 )
( 0.15152 , 0.01835 )
( 0.15455 , 0.01853 )
( 0.15758 , 0.01871 )
( 0.16061 , 0.01889 )
( 0.16364 , 0.01907 )
( 0.16667 , 0.01925 )
( 0.16970 , 0.01944 )
( 0.17273 , 0.01963 )
( 0.17576 , 0.01982 )
( 0.17879 , 0.02002 )
( 0.18182 , 0.02021 )
( 0.18485 , 0.02041 )
( 0.18788 , 0.02061 )
( 0.19091 , 0.02082 )
( 0.19394 , 0.02102 )
( 0.19697 , 0.02123 )
( 0.20000 , 0.02145 )
( 0.20303 , 0.02166 )
( 0.20606 , 0.02188 )
( 0.20909 , 0.02210 )
( 0.21212 , 0.02232 )
( 0.21515 , 0.02255 )
( 0.21818 , 0.02278 )
( 0.22121 , 0.02301 )
( 0.22424 , 0.02325 )
( 0.22727 , 0.02349 )
( 0.23030 , 0.02373 )
( 0.23333 , 0.02398 )
( 0.23636 , 0.02423 )
( 0.23939 , 0.02449 )
( 0.24242 , 0.02475 )
( 0.24545 , 0.02501 )
( 0.24848 , 0.02528 )
( 0.25152 , 0.02555 )
( 0.25455 , 0.02582 )
( 0.25758 , 0.02610 )
( 0.26061 , 0.02639 )
( 0.26364 , 0.02668 )
( 0.26667 , 0.02697 )
( 0.26970 , 0.02727 )
( 0.27273 , 0.02757 )
( 0.27576 , 0.02788 )
( 0.27879 , 0.02819 )
( 0.28182 , 0.02851 )
( 0.28485 , 0.02884 )
( 0.28788 , 0.02917 )
( 0.29091 , 0.02951 )
( 0.29394 , 0.02985 )
( 0.29697 , 0.03020 )
( 0.30000 , 0.03055 )
}
\newcommand{\probaSM}{
( 0.00000 , 0.97868 )
( 0.00303 , 0.97858 )
( 0.00606 , 0.97849 )
( 0.00909 , 0.97839 )
( 0.01212 , 0.97829 )
( 0.01515 , 0.97819 )
( 0.01818 , 0.97809 )
( 0.02121 , 0.97799 )
( 0.02424 , 0.97788 )
( 0.02727 , 0.97777 )
( 0.03030 , 0.97766 )
( 0.03333 , 0.97755 )
( 0.03636 , 0.97744 )
( 0.03939 , 0.97732 )
( 0.04242 , 0.97720 )
( 0.04545 , 0.97709 )
( 0.04848 , 0.97697 )
( 0.05152 , 0.97684 )
( 0.05455 , 0.97672 )
( 0.05758 , 0.97659 )
( 0.06061 , 0.97647 )
( 0.06364 , 0.97634 )
( 0.06667 , 0.97621 )
( 0.06970 , 0.97608 )
( 0.07273 , 0.97594 )
( 0.07576 , 0.97581 )
( 0.07879 , 0.97567 )
( 0.08182 , 0.97554 )
( 0.08485 , 0.97540 )
( 0.08788 , 0.97526 )
( 0.09091 , 0.97511 )
( 0.09394 , 0.97497 )
( 0.09697 , 0.97482 )
( 0.10000 , 0.97468 )
( 0.10303 , 0.97453 )
( 0.10606 , 0.97438 )
( 0.10909 , 0.97423 )
( 0.11212 , 0.97408 )
( 0.11515 , 0.97392 )
( 0.11818 , 0.97377 )
( 0.12121 , 0.97361 )
( 0.12424 , 0.97345 )
( 0.12727 , 0.97329 )
( 0.13030 , 0.97313 )
( 0.13333 , 0.97296 )
( 0.13636 , 0.97280 )
( 0.13939 , 0.97263 )
( 0.14242 , 0.97246 )
( 0.14545 , 0.97229 )
( 0.14848 , 0.97212 )
( 0.15152 , 0.97194 )
( 0.15455 , 0.97176 )
( 0.15758 , 0.97159 )
( 0.16061 , 0.97140 )
( 0.16364 , 0.97122 )
( 0.16667 , 0.97104 )
( 0.16970 , 0.97085 )
( 0.17273 , 0.97066 )
( 0.17576 , 0.97047 )
( 0.17879 , 0.97027 )
( 0.18182 , 0.97008 )
( 0.18485 , 0.96988 )
( 0.18788 , 0.96968 )
( 0.19091 , 0.96947 )
( 0.19394 , 0.96927 )
( 0.19697 , 0.96906 )
( 0.20000 , 0.96885 )
( 0.20303 , 0.96863 )
( 0.20606 , 0.96841 )
( 0.20909 , 0.96819 )
( 0.21212 , 0.96797 )
( 0.21515 , 0.96774 )
( 0.21818 , 0.96751 )
( 0.22121 , 0.96728 )
( 0.22424 , 0.96704 )
( 0.22727 , 0.96680 )
( 0.23030 , 0.96656 )
( 0.23333 , 0.96631 )
( 0.23636 , 0.96606 )
( 0.23939 , 0.96580 )
( 0.24242 , 0.96555 )
( 0.24545 , 0.96528 )
( 0.24848 , 0.96502 )
( 0.25152 , 0.96474 )
( 0.25455 , 0.96447 )
( 0.25758 , 0.96419 )
( 0.26061 , 0.96390 )
( 0.26364 , 0.96362 )
( 0.26667 , 0.96332 )
( 0.26970 , 0.96302 )
( 0.27273 , 0.96272 )
( 0.27576 , 0.96241 )
( 0.27879 , 0.96210 )
( 0.28182 , 0.96178 )
( 0.28485 , 0.96145 )
( 0.28788 , 0.96112 )
( 0.29091 , 0.96078 )
( 0.29394 , 0.96044 )
( 0.29697 , 0.96009 )
( 0.30000 , 0.95974 )
}
\newcommand{\probaCar}{
( 0.00000 , 0.00971 )
( 0.00303 , 0.00971 )
( 0.00606 , 0.00971 )
( 0.00909 , 0.00971 )
( 0.01212 , 0.00971 )
( 0.01515 , 0.00971 )
( 0.01818 , 0.00971 )
( 0.02121 , 0.00971 )
( 0.02424 , 0.00971 )
( 0.02727 , 0.00971 )
( 0.03030 , 0.00971 )
( 0.03333 , 0.00971 )
( 0.03636 , 0.00971 )
( 0.03939 , 0.00971 )
( 0.04242 , 0.00971 )
( 0.04545 , 0.00971 )
( 0.04848 , 0.00971 )
( 0.05152 , 0.00971 )
( 0.05455 , 0.00971 )
( 0.05758 , 0.00971 )
( 0.06061 , 0.00971 )
( 0.06364 , 0.00971 )
( 0.06667 , 0.00971 )
( 0.06970 , 0.00971 )
( 0.07273 , 0.00971 )
( 0.07576 , 0.00971 )
( 0.07879 , 0.00971 )
( 0.08182 , 0.00971 )
( 0.08485 , 0.00971 )
( 0.08788 , 0.00971 )
( 0.09091 , 0.00971 )
( 0.09394 , 0.00971 )
( 0.09697 , 0.00971 )
( 0.10000 , 0.00971 )
( 0.10303 , 0.00971 )
( 0.10606 , 0.00971 )
( 0.10909 , 0.00971 )
( 0.11212 , 0.00971 )
( 0.11515 , 0.00971 )
( 0.11818 , 0.00971 )
( 0.12121 , 0.00971 )
( 0.12424 , 0.00971 )
( 0.12727 , 0.00971 )
( 0.13030 , 0.00971 )
( 0.13333 , 0.00971 )
( 0.13636 , 0.00971 )
( 0.13939 , 0.00971 )
( 0.14242 , 0.00971 )
( 0.14545 , 0.00971 )
( 0.14848 , 0.00971 )
( 0.15152 , 0.00971 )
( 0.15455 , 0.00971 )
( 0.15758 , 0.00971 )
( 0.16061 , 0.00971 )
( 0.16364 , 0.00971 )
( 0.16667 , 0.00971 )
( 0.16970 , 0.00971 )
( 0.17273 , 0.00971 )
( 0.17576 , 0.00971 )
( 0.17879 , 0.00971 )
( 0.18182 , 0.00971 )
( 0.18485 , 0.00971 )
( 0.18788 , 0.00971 )
( 0.19091 , 0.00971 )
( 0.19394 , 0.00971 )
( 0.19697 , 0.00971 )
( 0.20000 , 0.00971 )
( 0.20303 , 0.00971 )
( 0.20606 , 0.00971 )
( 0.20909 , 0.00971 )
( 0.21212 , 0.00971 )
( 0.21515 , 0.00971 )
( 0.21818 , 0.00971 )
( 0.22121 , 0.00971 )
( 0.22424 , 0.00971 )
( 0.22727 , 0.00971 )
( 0.23030 , 0.00971 )
( 0.23333 , 0.00971 )
( 0.23636 , 0.00971 )
( 0.23939 , 0.00971 )
( 0.24242 , 0.00971 )
( 0.24545 , 0.00971 )
( 0.24848 , 0.00971 )
( 0.25152 , 0.00971 )
( 0.25455 , 0.00971 )
( 0.25758 , 0.00971 )
( 0.26061 , 0.00971 )
( 0.26364 , 0.00971 )
( 0.26667 , 0.00971 )
( 0.26970 , 0.00971 )
( 0.27273 , 0.00971 )
( 0.27576 , 0.00971 )
( 0.27879 , 0.00971 )
( 0.28182 , 0.00971 )
( 0.28485 , 0.00971 )
( 0.28788 , 0.00971 )
( 0.29091 , 0.00971 )
( 0.29394 , 0.00971 )
( 0.29697 , 0.00971 )
( 0.30000 , 0.00971 )
}

     \begin{axis}[xlabel={Cost of Swissmetro}, ylabel={Choice probability}, axis y line=left, axis x line=bottom, xmin=0, xmax=0.3, ymin=0, ymax=1,legend pos=north east,
        xtick=\empty,
	    xtick pos=bottom,
	    extra x ticks={0.1,0.2,0.3},
		extra x tick style={grid=none},
	    extra x tick labels={0.1,0.2,0.3},] 
        \addplot[smooth,blue,very thick] coordinates {\probaTrain};
       \addplot[smooth,orange,very thick] coordinates {\probaSM};
        \addplot[smooth,black!40!green,very thick] coordinates {\probaCar};
        \end{axis}
 \end{tikzpicture}
  
  \caption{Customer type 1}
\end{subfigure}
\begin{subfigure}{.3\textwidth}
  \centering
 \begin{tikzpicture}[scale=0.6]
\newcommand{\probaTrain}{
( 0.00000 , 0.01207 )
( 0.00303 , 0.01290 )
( 0.00606 , 0.01400 )
( 0.00909 , 0.01547 )
( 0.01212 , 0.01740 )
( 0.01515 , 0.01993 )
( 0.01818 , 0.02324 )
( 0.02121 , 0.02746 )
( 0.02424 , 0.03277 )
( 0.02727 , 0.03938 )
( 0.03030 , 0.04753 )
( 0.03333 , 0.05746 )
( 0.03636 , 0.06941 )
( 0.03939 , 0.08363 )
( 0.04242 , 0.10033 )
( 0.04545 , 0.11969 )
( 0.04848 , 0.14183 )
( 0.05152 , 0.16678 )
( 0.05455 , 0.19451 )
( 0.05758 , 0.22486 )
( 0.06061 , 0.25758 )
( 0.06364 , 0.29232 )
( 0.06667 , 0.32866 )
( 0.06970 , 0.36609 )
( 0.07273 , 0.40410 )
( 0.07576 , 0.44218 )
( 0.07879 , 0.47981 )
( 0.08182 , 0.51655 )
( 0.08485 , 0.55203 )
( 0.08788 , 0.58592 )
( 0.09091 , 0.61802 )
( 0.09394 , 0.64816 )
( 0.09697 , 0.67625 )
( 0.10000 , 0.70218 )
( 0.10303 , 0.72593 )
( 0.10606 , 0.74760 )
( 0.10909 , 0.76731 )
( 0.11212 , 0.78521 )
( 0.11515 , 0.80143 )
( 0.11818 , 0.81612 )
( 0.12121 , 0.82941 )
( 0.12424 , 0.84144 )
( 0.12727 , 0.85232 )
( 0.13030 , 0.86218 )
( 0.13333 , 0.87110 )
( 0.13636 , 0.87919 )
( 0.13939 , 0.88654 )
( 0.14242 , 0.89320 )
( 0.14545 , 0.89927 )
( 0.14848 , 0.90479 )
( 0.15152 , 0.90983 )
( 0.15455 , 0.91443 )
( 0.15758 , 0.91863 )
( 0.16061 , 0.92248 )
( 0.16364 , 0.92601 )
( 0.16667 , 0.92925 )
( 0.16970 , 0.93223 )
( 0.17273 , 0.93498 )
( 0.17576 , 0.93751 )
( 0.17879 , 0.93985 )
( 0.18182 , 0.94201 )
( 0.18485 , 0.94401 )
( 0.18788 , 0.94587 )
( 0.19091 , 0.94760 )
( 0.19394 , 0.94920 )
( 0.19697 , 0.95069 )
( 0.20000 , 0.95208 )
( 0.20303 , 0.95338 )
( 0.20606 , 0.95460 )
( 0.20909 , 0.95573 )
( 0.21212 , 0.95680 )
( 0.21515 , 0.95780 )
( 0.21818 , 0.95873 )
( 0.22121 , 0.95961 )
( 0.22424 , 0.96044 )
( 0.22727 , 0.96122 )
( 0.23030 , 0.96196 )
( 0.23333 , 0.96266 )
( 0.23636 , 0.96331 )
( 0.23939 , 0.96393 )
( 0.24242 , 0.96452 )
( 0.24545 , 0.96508 )
( 0.24848 , 0.96561 )
( 0.25152 , 0.96611 )
( 0.25455 , 0.96659 )
( 0.25758 , 0.96704 )
( 0.26061 , 0.96747 )
( 0.26364 , 0.96788 )
( 0.26667 , 0.96827 )
( 0.26970 , 0.96865 )
( 0.27273 , 0.96900 )
( 0.27576 , 0.96934 )
( 0.27879 , 0.96967 )
( 0.28182 , 0.96998 )
( 0.28485 , 0.97028 )
( 0.28788 , 0.97056 )
( 0.29091 , 0.97084 )
( 0.29394 , 0.97110 )
( 0.29697 , 0.97135 )
( 0.30000 , 0.97159 )
}
\newcommand{\probaSM}{
( 0.00000 , 0.97822 )
( 0.00303 , 0.97739 )
( 0.00606 , 0.97628 )
( 0.00909 , 0.97482 )
( 0.01212 , 0.97289 )
( 0.01515 , 0.97035 )
( 0.01818 , 0.96704 )
( 0.02121 , 0.96282 )
( 0.02424 , 0.95751 )
( 0.02727 , 0.95089 )
( 0.03030 , 0.94273 )
( 0.03333 , 0.93279 )
( 0.03636 , 0.92083 )
( 0.03939 , 0.90660 )
( 0.04242 , 0.88989 )
( 0.04545 , 0.87051 )
( 0.04848 , 0.84836 )
( 0.05152 , 0.82338 )
( 0.05455 , 0.79563 )
( 0.05758 , 0.76526 )
( 0.06061 , 0.73251 )
( 0.06364 , 0.69774 )
( 0.06667 , 0.66138 )
( 0.06970 , 0.62391 )
( 0.07273 , 0.58587 )
( 0.07576 , 0.54777 )
( 0.07879 , 0.51011 )
( 0.08182 , 0.47333 )
( 0.08485 , 0.43783 )
( 0.08788 , 0.40390 )
( 0.09091 , 0.37178 )
( 0.09394 , 0.34162 )
( 0.09697 , 0.31350 )
( 0.10000 , 0.28755 )
( 0.10303 , 0.26378 )
( 0.10606 , 0.24210 )
( 0.10909 , 0.22237 )
( 0.11212 , 0.20446 )
( 0.11515 , 0.18823 )
( 0.11818 , 0.17352 )
( 0.12121 , 0.16022 )
( 0.12424 , 0.14818 )
( 0.12727 , 0.13729 )
( 0.13030 , 0.12743 )
( 0.13333 , 0.11850 )
( 0.13636 , 0.11040 )
( 0.13939 , 0.10305 )
( 0.14242 , 0.09638 )
( 0.14545 , 0.09031 )
( 0.14848 , 0.08478 )
( 0.15152 , 0.07974 )
( 0.15455 , 0.07514 )
( 0.15758 , 0.07093 )
( 0.16061 , 0.06708 )
( 0.16364 , 0.06354 )
( 0.16667 , 0.06030 )
( 0.16970 , 0.05732 )
( 0.17273 , 0.05457 )
( 0.17576 , 0.05203 )
( 0.17879 , 0.04969 )
( 0.18182 , 0.04753 )
( 0.18485 , 0.04553 )
( 0.18788 , 0.04367 )
( 0.19091 , 0.04194 )
( 0.19394 , 0.04034 )
( 0.19697 , 0.03884 )
( 0.20000 , 0.03745 )
( 0.20303 , 0.03615 )
( 0.20606 , 0.03493 )
( 0.20909 , 0.03380 )
( 0.21212 , 0.03273 )
( 0.21515 , 0.03173 )
( 0.21818 , 0.03080 )
( 0.22121 , 0.02991 )
( 0.22424 , 0.02908 )
( 0.22727 , 0.02830 )
( 0.23030 , 0.02757 )
( 0.23333 , 0.02687 )
( 0.23636 , 0.02621 )
( 0.23939 , 0.02559 )
( 0.24242 , 0.02500 )
( 0.24545 , 0.02444 )
( 0.24848 , 0.02392 )
( 0.25152 , 0.02341 )
( 0.25455 , 0.02294 )
( 0.25758 , 0.02248 )
( 0.26061 , 0.02205 )
( 0.26364 , 0.02164 )
( 0.26667 , 0.02125 )
( 0.26970 , 0.02088 )
( 0.27273 , 0.02052 )
( 0.27576 , 0.02018 )
( 0.27879 , 0.01985 )
( 0.28182 , 0.01954 )
( 0.28485 , 0.01924 )
( 0.28788 , 0.01896 )
( 0.29091 , 0.01868 )
( 0.29394 , 0.01842 )
( 0.29697 , 0.01817 )
( 0.30000 , 0.01792 )
}
\newcommand{\probaCar}{
( 0.00000 , 0.00971 )
( 0.00303 , 0.00971 )
( 0.00606 , 0.00971 )
( 0.00909 , 0.00971 )
( 0.01212 , 0.00971 )
( 0.01515 , 0.00972 )
( 0.01818 , 0.00972 )
( 0.02121 , 0.00972 )
( 0.02424 , 0.00973 )
( 0.02727 , 0.00973 )
( 0.03030 , 0.00974 )
( 0.03333 , 0.00975 )
( 0.03636 , 0.00976 )
( 0.03939 , 0.00977 )
( 0.04242 , 0.00978 )
( 0.04545 , 0.00980 )
( 0.04848 , 0.00981 )
( 0.05152 , 0.00983 )
( 0.05455 , 0.00986 )
( 0.05758 , 0.00988 )
( 0.06061 , 0.00991 )
( 0.06364 , 0.00994 )
( 0.06667 , 0.00997 )
( 0.06970 , 0.01000 )
( 0.07273 , 0.01003 )
( 0.07576 , 0.01006 )
( 0.07879 , 0.01009 )
( 0.08182 , 0.01012 )
( 0.08485 , 0.01014 )
( 0.08788 , 0.01017 )
( 0.09091 , 0.01020 )
( 0.09394 , 0.01022 )
( 0.09697 , 0.01024 )
( 0.10000 , 0.01027 )
( 0.10303 , 0.01028 )
( 0.10606 , 0.01030 )
( 0.10909 , 0.01032 )
( 0.11212 , 0.01033 )
( 0.11515 , 0.01035 )
( 0.11818 , 0.01036 )
( 0.12121 , 0.01037 )
( 0.12424 , 0.01038 )
( 0.12727 , 0.01039 )
( 0.13030 , 0.01039 )
( 0.13333 , 0.01040 )
( 0.13636 , 0.01041 )
( 0.13939 , 0.01041 )
( 0.14242 , 0.01042 )
( 0.14545 , 0.01042 )
( 0.14848 , 0.01043 )
( 0.15152 , 0.01043 )
( 0.15455 , 0.01044 )
( 0.15758 , 0.01044 )
( 0.16061 , 0.01044 )
( 0.16364 , 0.01045 )
( 0.16667 , 0.01045 )
( 0.16970 , 0.01045 )
( 0.17273 , 0.01045 )
( 0.17576 , 0.01045 )
( 0.17879 , 0.01046 )
( 0.18182 , 0.01046 )
( 0.18485 , 0.01046 )
( 0.18788 , 0.01046 )
( 0.19091 , 0.01046 )
( 0.19394 , 0.01046 )
( 0.19697 , 0.01046 )
( 0.20000 , 0.01047 )
( 0.20303 , 0.01047 )
( 0.20606 , 0.01047 )
( 0.20909 , 0.01047 )
( 0.21212 , 0.01047 )
( 0.21515 , 0.01047 )
( 0.21818 , 0.01047 )
( 0.22121 , 0.01047 )
( 0.22424 , 0.01047 )
( 0.22727 , 0.01047 )
( 0.23030 , 0.01047 )
( 0.23333 , 0.01047 )
( 0.23636 , 0.01048 )
( 0.23939 , 0.01048 )
( 0.24242 , 0.01048 )
( 0.24545 , 0.01048 )
( 0.24848 , 0.01048 )
( 0.25152 , 0.01048 )
( 0.25455 , 0.01048 )
( 0.25758 , 0.01048 )
( 0.26061 , 0.01048 )
( 0.26364 , 0.01048 )
( 0.26667 , 0.01048 )
( 0.26970 , 0.01048 )
( 0.27273 , 0.01048 )
( 0.27576 , 0.01048 )
( 0.27879 , 0.01048 )
( 0.28182 , 0.01048 )
( 0.28485 , 0.01048 )
( 0.28788 , 0.01048 )
( 0.29091 , 0.01048 )
( 0.29394 , 0.01048 )
( 0.29697 , 0.01048 )
( 0.30000 , 0.01048 )
}

     \begin{axis}[xlabel={Cost of Swissmetro}, ylabel={Choice probability}, axis y line=left, axis x line=bottom, xmin=0, xmax=0.3, ymin=0, ymax=1,legend style={at={(1,0.65)}},
        xtick=\empty,
	    xtick pos=bottom,
	    extra x ticks={0.1,0.2,0.3},
		extra x tick style={grid=none},
	    extra x tick labels={0.1,0.2,0.3},] 
        \addplot[smooth,blue,very thick] coordinates {\probaTrain};
         \addlegendentry{Train}
       \addplot[smooth,orange,very thick] coordinates {\probaSM};
        \addlegendentry{SwissMetro}
        \addplot[smooth,black!40!green,very thick] coordinates {\probaCar};
        \addlegendentry{Car}
        \end{axis}
 \end{tikzpicture}
  \caption{Customer type 2}
\end{subfigure}%
\begin{subfigure}{.3\textwidth}
  \centering
 \begin{tikzpicture}[scale=0.6]
    \newcommand{\probaTrain}{
( 0.00000 , 0.00976 )
( 0.00303 , 0.00978 )
( 0.00606 , 0.00981 )
( 0.00909 , 0.00986 )
( 0.01212 , 0.00993 )
( 0.01515 , 0.01002 )
( 0.01818 , 0.01013 )
( 0.02121 , 0.01027 )
( 0.02424 , 0.01043 )
( 0.02727 , 0.01060 )
( 0.03030 , 0.01077 )
( 0.03333 , 0.01094 )
( 0.03636 , 0.01109 )
( 0.03939 , 0.01121 )
( 0.04242 , 0.01132 )
( 0.04545 , 0.01140 )
( 0.04848 , 0.01146 )
( 0.05152 , 0.01151 )
( 0.05455 , 0.01155 )
( 0.05758 , 0.01158 )
( 0.06061 , 0.01161 )
( 0.06364 , 0.01163 )
( 0.06667 , 0.01164 )
( 0.06970 , 0.01165 )
( 0.07273 , 0.01166 )
( 0.07576 , 0.01167 )
( 0.07879 , 0.01167 )
( 0.08182 , 0.01168 )
( 0.08485 , 0.01168 )
( 0.08788 , 0.01169 )
( 0.09091 , 0.01169 )
( 0.09394 , 0.01169 )
( 0.09697 , 0.01169 )
( 0.10000 , 0.01169 )
( 0.10303 , 0.01169 )
( 0.10606 , 0.01170 )
( 0.10909 , 0.01170 )
( 0.11212 , 0.01170 )
( 0.11515 , 0.01170 )
( 0.11818 , 0.01170 )
( 0.12121 , 0.01170 )
( 0.12424 , 0.01170 )
( 0.12727 , 0.01170 )
( 0.13030 , 0.01170 )
( 0.13333 , 0.01170 )
( 0.13636 , 0.01170 )
( 0.13939 , 0.01170 )
( 0.14242 , 0.01170 )
( 0.14545 , 0.01170 )
( 0.14848 , 0.01170 )
( 0.15152 , 0.01170 )
( 0.15455 , 0.01170 )
( 0.15758 , 0.01170 )
( 0.16061 , 0.01170 )
( 0.16364 , 0.01170 )
( 0.16667 , 0.01170 )
( 0.16970 , 0.01170 )
( 0.17273 , 0.01170 )
( 0.17576 , 0.01170 )
( 0.17879 , 0.01170 )
( 0.18182 , 0.01170 )
( 0.18485 , 0.01170 )
( 0.18788 , 0.01170 )
( 0.19091 , 0.01170 )
( 0.19394 , 0.01170 )
( 0.19697 , 0.01170 )
( 0.20000 , 0.01170 )
( 0.20303 , 0.01170 )
( 0.20606 , 0.01170 )
( 0.20909 , 0.01170 )
( 0.21212 , 0.01170 )
( 0.21515 , 0.01170 )
( 0.21818 , 0.01170 )
( 0.22121 , 0.01170 )
( 0.22424 , 0.01170 )
( 0.22727 , 0.01170 )
( 0.23030 , 0.01170 )
( 0.23333 , 0.01170 )
( 0.23636 , 0.01170 )
( 0.23939 , 0.01170 )
( 0.24242 , 0.01170 )
( 0.24545 , 0.01170 )
( 0.24848 , 0.01170 )
( 0.25152 , 0.01170 )
( 0.25455 , 0.01170 )
( 0.25758 , 0.01170 )
( 0.26061 , 0.01170 )
( 0.26364 , 0.01170 )
( 0.26667 , 0.01170 )
( 0.26970 , 0.01170 )
( 0.27273 , 0.01170 )
( 0.27576 , 0.01170 )
( 0.27879 , 0.01170 )
( 0.28182 , 0.01170 )
( 0.28485 , 0.01170 )
( 0.28788 , 0.01170 )
( 0.29091 , 0.01170 )
( 0.29394 , 0.01170 )
( 0.29697 , 0.01170 )
( 0.30000 , 0.01170 )
}
\newcommand{\probaSM}{
( 0.00000 , 0.95778 )
( 0.00303 , 0.94679 )
( 0.00606 , 0.93046 )
( 0.00909 , 0.90670 )
( 0.01212 , 0.87393 )
( 0.01515 , 0.83045 )
( 0.01818 , 0.77513 )
( 0.02121 , 0.70808 )
( 0.02424 , 0.63114 )
( 0.02727 , 0.54795 )
( 0.03030 , 0.46331 )
( 0.03333 , 0.38219 )
( 0.03636 , 0.30930 )
( 0.03939 , 0.24788 )
( 0.04242 , 0.19780 )
( 0.04545 , 0.15786 )
( 0.04848 , 0.12645 )
( 0.05152 , 0.10195 )
( 0.05455 , 0.08291 )
( 0.05758 , 0.06810 )
( 0.06061 , 0.05657 )
( 0.06364 , 0.04756 )
( 0.06667 , 0.04048 )
( 0.06970 , 0.03489 )
( 0.07273 , 0.03045 )
( 0.07576 , 0.02690 )
( 0.07879 , 0.02405 )
( 0.08182 , 0.02174 )
( 0.08485 , 0.01986 )
( 0.08788 , 0.01833 )
( 0.09091 , 0.01706 )
( 0.09394 , 0.01602 )
( 0.09697 , 0.01515 )
( 0.10000 , 0.01442 )
( 0.10303 , 0.01381 )
( 0.10606 , 0.01329 )
( 0.10909 , 0.01286 )
( 0.11212 , 0.01248 )
( 0.11515 , 0.01217 )
( 0.11818 , 0.01189 )
( 0.12121 , 0.01165 )
( 0.12424 , 0.01145 )
( 0.12727 , 0.01127 )
( 0.13030 , 0.01112 )
( 0.13333 , 0.01098 )
( 0.13636 , 0.01086 )
( 0.13939 , 0.01076 )
( 0.14242 , 0.01066 )
( 0.14545 , 0.01058 )
( 0.14848 , 0.01051 )
( 0.15152 , 0.01044 )
( 0.15455 , 0.01039 )
( 0.15758 , 0.01033 )
( 0.16061 , 0.01029 )
( 0.16364 , 0.01024 )
( 0.16667 , 0.01021 )
( 0.16970 , 0.01017 )
( 0.17273 , 0.01014 )
( 0.17576 , 0.01011 )
( 0.17879 , 0.01009 )
( 0.18182 , 0.01006 )
( 0.18485 , 0.01004 )
( 0.18788 , 0.01002 )
( 0.19091 , 0.01000 )
( 0.19394 , 0.00999 )
( 0.19697 , 0.00997 )
( 0.20000 , 0.00996 )
( 0.20303 , 0.00995 )
( 0.20606 , 0.00993 )
( 0.20909 , 0.00992 )
( 0.21212 , 0.00991 )
( 0.21515 , 0.00990 )
( 0.21818 , 0.00989 )
( 0.22121 , 0.00989 )
( 0.22424 , 0.00988 )
( 0.22727 , 0.00987 )
( 0.23030 , 0.00986 )
( 0.23333 , 0.00986 )
( 0.23636 , 0.00985 )
( 0.23939 , 0.00985 )
( 0.24242 , 0.00984 )
( 0.24545 , 0.00984 )
( 0.24848 , 0.00983 )
( 0.25152 , 0.00983 )
( 0.25455 , 0.00982 )
( 0.25758 , 0.00982 )
( 0.26061 , 0.00982 )
( 0.26364 , 0.00981 )
( 0.26667 , 0.00981 )
( 0.26970 , 0.00981 )
( 0.27273 , 0.00980 )
( 0.27576 , 0.00980 )
( 0.27879 , 0.00980 )
( 0.28182 , 0.00979 )
( 0.28485 , 0.00979 )
( 0.28788 , 0.00979 )
( 0.29091 , 0.00979 )
( 0.29394 , 0.00979 )
( 0.29697 , 0.00978 )
( 0.30000 , 0.00978 )
}
\newcommand{\probaCar}{
( 0.00000 , 0.03246 )
( 0.00303 , 0.04343 )
( 0.00606 , 0.05973 )
( 0.00909 , 0.08344 )
( 0.01212 , 0.11614 )
( 0.01515 , 0.15953 )
( 0.01818 , 0.21474 )
( 0.02121 , 0.28165 )
( 0.02424 , 0.35843 )
( 0.02727 , 0.44146 )
( 0.03030 , 0.52591 )
( 0.03333 , 0.60687 )
( 0.03636 , 0.67961 )
( 0.03939 , 0.74090 )
( 0.04242 , 0.79088 )
( 0.04545 , 0.83074 )
( 0.04848 , 0.86208 )
( 0.05152 , 0.88653 )
( 0.05455 , 0.90554 )
( 0.05758 , 0.92031 )
( 0.06061 , 0.93182 )
( 0.06364 , 0.94082 )
( 0.06667 , 0.94788 )
( 0.06970 , 0.95346 )
( 0.07273 , 0.95789 )
( 0.07576 , 0.96143 )
( 0.07879 , 0.96428 )
( 0.08182 , 0.96658 )
( 0.08485 , 0.96846 )
( 0.08788 , 0.96999 )
( 0.09091 , 0.97125 )
( 0.09394 , 0.97229 )
( 0.09697 , 0.97316 )
( 0.10000 , 0.97389 )
( 0.10303 , 0.97449 )
( 0.10606 , 0.97501 )
( 0.10909 , 0.97545 )
( 0.11212 , 0.97582 )
( 0.11515 , 0.97614 )
( 0.11818 , 0.97641 )
( 0.12121 , 0.97665 )
( 0.12424 , 0.97685 )
( 0.12727 , 0.97703 )
( 0.13030 , 0.97718 )
( 0.13333 , 0.97732 )
( 0.13636 , 0.97744 )
( 0.13939 , 0.97754 )
( 0.14242 , 0.97763 )
( 0.14545 , 0.97772 )
( 0.14848 , 0.97779 )
( 0.15152 , 0.97785 )
( 0.15455 , 0.97791 )
( 0.15758 , 0.97796 )
( 0.16061 , 0.97801 )
( 0.16364 , 0.97805 )
( 0.16667 , 0.97809 )
( 0.16970 , 0.97813 )
( 0.17273 , 0.97816 )
( 0.17576 , 0.97818 )
( 0.17879 , 0.97821 )
( 0.18182 , 0.97823 )
( 0.18485 , 0.97826 )
( 0.18788 , 0.97827 )
( 0.19091 , 0.97829 )
( 0.19394 , 0.97831 )
( 0.19697 , 0.97832 )
( 0.20000 , 0.97834 )
( 0.20303 , 0.97835 )
( 0.20606 , 0.97836 )
( 0.20909 , 0.97837 )
( 0.21212 , 0.97838 )
( 0.21515 , 0.97839 )
( 0.21818 , 0.97840 )
( 0.22121 , 0.97841 )
( 0.22424 , 0.97842 )
( 0.22727 , 0.97843 )
( 0.23030 , 0.97843 )
( 0.23333 , 0.97844 )
( 0.23636 , 0.97845 )
( 0.23939 , 0.97845 )
( 0.24242 , 0.97846 )
( 0.24545 , 0.97846 )
( 0.24848 , 0.97847 )
( 0.25152 , 0.97847 )
( 0.25455 , 0.97847 )
( 0.25758 , 0.97848 )
( 0.26061 , 0.97848 )
( 0.26364 , 0.97848 )
( 0.26667 , 0.97849 )
( 0.26970 , 0.97849 )
( 0.27273 , 0.97849 )
( 0.27576 , 0.97850 )
( 0.27879 , 0.97850 )
( 0.28182 , 0.97850 )
( 0.28485 , 0.97851 )
( 0.28788 , 0.97851 )
( 0.29091 , 0.97851 )
( 0.29394 , 0.97851 )
( 0.29697 , 0.97851 )
( 0.30000 , 0.97852 )
}

     \begin{axis}[xlabel={Cost of Swissmetro}, ylabel={Choice probability}, axis y line=left, axis x line=bottom, xmin=0, xmax=0.3, ymin=0, ymax=1,
        xtick=\empty,
	    xtick pos=bottom,
	    extra x ticks={0.1,0.2,0.3},
		extra x tick style={grid=none},
	    extra x tick labels={0.1,0.2,0.3},] 
        \addplot[smooth,blue,very thick] coordinates {\probaTrain};
       \addplot[smooth,orange,very thick] coordinates {\probaSM};
        \addplot[smooth,black!40!green,very thick] coordinates {\probaCar};
        \end{axis}
 \end{tikzpicture}
  \caption{Customer type 3}
\end{subfigure}
\caption{Different substitution patterns predicted by DeepMNL with parameters $\boldsymbol{(\ell,w)=(2,5)}$.}
\label{fig:types}
\end{figure}

\section{Conclusion}

In this paper, we introduce a new class of neural network-based choice models. Our neural network architecture, RUMnet, provides an efficient approximation of RUM discrete choice models. Unlike traditional RUM discrete choice models, which rely on specific structural and distributional assumptions, RUMnets leverage the universality of feed-forward neural networks to learn the structure of the random utility function from data. Moreover, our method is practical as RUMnets can be trained using existing open-source libraries and achieves a competitive predictive performance . By satisfying the RUM principle, we believe that our structured machine-learning approach yields better generalization to high-dimensional inputs than other model-free methods, as illustrated by our numerical findings. This work opens several intriguing research directions. 

For instance, it might be possible to exploit the flexibility of the RUMnet architecture for more complicated choice estimation problems such as multi-item and dynamic discrete choice. One drawback of RUMnets is a much larger computational cost compared to simpler neural network-based architectures; enhancing our  implementation is necessary for very large-scale applications. Another important research avenue is to embed our neural network-based choice model in assortment optimization and pricing decision problems. Additionally, even though we focus on logit-based models in our paper, one could use similar ideas to implement and generalize other choice models such as the exponomial choice model~\citep{alptekinouglu2016exponomial}.

\putbib[ref]  
\end{bibunit}


\newpage

\ECSwitch
\ECDisclaimer
\ECHead{
\begin{center}
$ $\\
{\underline{Online Appendix}} \\[8pt]
Representing Random Utility Choice Models with Neural Networks\\[8pt]
Ali Aouad and Antoine Désir\\
\end{center}}

\begin{APPENDICES}
\begin{bibunit}[ormsv080]
\OneAndAHalfSpacedXI

The Appendix is organized as follows.
\begin{itemize}
    \item Appendix~\ref{app:architecture} provides additional material for Section~\ref{sec:architecture}.
    \item Appendix~\ref{app:expressive_power} contains the proofs of Section~\ref{sec:expressive_power}.
    \item Appendix~\ref{app:curse_dimensionality} contains the proofs of Section~\ref{sec:curse_dimensionality}.
    \item Appendix~\ref{app:synthetics} provides additional material for Section~\ref{sec:synthetic}.
    \item Appendix~\ref{app:materials} provides additional material for Section~\ref{sec:numerical}.
\end{itemize}

\newpage 

\section{Additional material for Section~\ref{sec:architecture}} \label{app:architecture}

\label{app:ffnn}

A feed-forward neural network is specified by a directed acyclic graph, $G = (V,E)$. Each edge $e \in E$ is associated with a weight $w_e \in \mathbb{R}$. Each node of the graph, called a neuron, is associated with an activation function $\sigma : \mathbb{R} \to \mathbb{R}$. Each edge in the graph links the output of some neuron to the input of another neuron. The input of a neuron is obtained by taking a weighted sum of the outputs of all neuron connected to it. We assume that the nodes are organized in layers, i.e. the set of nodes is partitioned in disjoint subsets $V = \cup_{t=0}^T V_t$, such that every edge in $E$ connects some node in $V_{t-1}$ to some node in $V_t$ , for some $t\in [T]$. We denote by $v_{i,t} \in V_t$ the $i^{th}$ neuron of the $t^{th}$ layer. Moreover, for any input $\bs{x}$ to the network and $v_{i,t} \in V_t$, we let $o_{i,t}(\bs{x})$ (resp. $a_{i,t}(\bs{x})$) denotes the output (resp. input) of $v_{i,t}$. Then, 
\begin{align*}
    a_{i,t+1} = \sum \limits_{j: e= (v_{j,t},v_{i,t+1}) \in E} w_e \cdot o_{j,t}(\bs{x}), 
\end{align*}
and $o_{i,t+1}(\bs{x}) = \sigma(a_{i,t+1}(\bs{x}))$. The first layer $V_0$ is called the input layer and the last layer $V_T$ is the output layer which often contains a single neuron. The layers $V_1,\dots,V_{T-1}$ are called the hidden layers. We refer to $T$ as the depth of the network. The width of the network is $\max_{t=1,\dots,T-1} |V_t|$ and its size is $|V|$. Figure~\ref{fig:nn} illustrates a feed-forward network of depth 2.
\begin{figure}[b!]
 \centering

 \begin{tikzpicture}
[mycircle/.style={
         circle,
         thick,
         draw=black,
         fill=gray,
         fill opacity = 0.3,
         text opacity=1,
         inner sep=3pt, 
         inner ysep=3pt
         },
      myarrow/.style={-Stealth},
      node distance=0.6cm and 1.2cm
      ]

 \node[mycircle] (v01) at (0,0) {$v_{0,1}$};
 \node[mycircle] (v02) at (0,-1.5) {$v_{0,2}$};
 \node[mycircle] (v03) at (0,-3) {$v_{0,3}$};
 
\node[mycircle] (v11) at (2.5,1) {$v_{1,1}$};
 \node[mycircle] (v12) at (2.5,-0.25) {$v_{1,2}$};
 \node[mycircle] (v13) at (2.5,-1.5) {$v_{1,3}$};
 \node[mycircle] (v14) at (2.5,-2.75) {$v_{1,4}$};
 \node[mycircle] (v15) at (2.5,-4) {$v_{1,5}$};
 
 \node[mycircle] (v21) at (5,1) {$v_{2,1}$};
 \node[mycircle] (v22) at (5,-0.25) {$v_{2,2}$};
 \node[mycircle] (v23) at (5,-1.5) {$v_{2,3}$};
 \node[mycircle] (v24) at (5,-2.75) {$v_{2,4}$};
 \node[mycircle] (v25) at (5,-4) {$v_{2,5}$};

 \node[mycircle] (v31) at (7.5,-0.5) {$v_{3,1}$};
 \node[mycircle] (v32) at (7.5,-2.5) {$v_{3,2}$};

\node (x1) at (-2.5,0) {$x_1$};
\node (x2) at (-2.5,-1.5) {$x_2$};
\node (x3) at (-2.5,-3) {$x_3$};

\node (o1) at (10,-0.5) {Ouput 1};
\node (o2) at (10,-2.5) {Output 2};

\node at (-2.5,2) {Input};
\node at (0,2) {Input layer};
\node at (3.75,2) {Hidden layers};
\node at (7.5,2) {Output layer};
\node at (10,2) {Output};

\draw[->] (x1) -- (v01);
\draw[->] (x2) -- (v02);
\draw[->] (x3) -- (v03);
\draw[->] (v31) -- (o1);
\draw[->] (v32) -- (o2);
       
\foreach \i in {v01,v02,v03}
    \foreach \j in {v11,v12,v13,v14,v15}
       \draw [->] (\i) --  (\j);

\foreach \i in {v11,v12,v13,v14,v15}
    \foreach \j in {v21,v22,v23,v24,v25}
       \draw [->] (\i) --  (\j);
       
\foreach \i in {v21,v22,v23,v24,v25}
    \foreach \j in {v31,v32}
       \draw [->] (\i) --  (\j);
\end{tikzpicture}

\caption{An example of feed-forward neural network.}
\label{fig:nn}
\end{figure}
Feed-forward neural networks are able to capture complex nonlinear patterns and have been shown to be a class of universal approximators. That is, under mild conditions, the class of neural networks $\bigcup_{k\geq 1}\Theta^{1,k}_k$ with only one hidden layer and one output unit is dense in the space of continuous functions over a compact domain $X$, e.g., this result holds for any continuous, bounded and non-constant activation function~\citep{hornik1989multilayer, hornik1991approximation}. 

\newpage

\section{Proofs of Section~\ref{sec:expressive_power}} \label{app:expressive_power}

\subsection{A preliminary result}

In the RUM framework, recall that we are interested in the ordering of the utilities for different alternatives. Indeed, for every $\bs{x} \neq \bs{x}'$, the inequality $U(\bs{x},\bs{\epsilon}(\bs{x}),\bs{z},\bs{\nu}(\bs{z})) \geq U(\bs{x}',\bs{\epsilon}(\bs{x}'),\bs{z},\bs{\nu}(\bs{z}))$ implies that $\bs{x}$ is preferred to $\bs{x}'$ by the corresponding random customer. The following lemma shows that it is sufficient to approximate $U(\cdot)$ on a finite covering $C$ in order to capture the desired ordering relations of all possible pairs $(\bs{x},\bs{x}') \in \mathcal{X}^2$ and customer attribute $\bs{z}$ with high probability. Throughout our analysis, the space of product and customer attributes is endowed with the $\ell_{\infty}$-norm unless stated otherwise. We denote by $B(c,\delta)$ the open ball centred on $c$ with radius $\delta$.

\begin{lemma} [Part of Theorem 1 in \citealt{mcfadden2000mixed}] \label{lem:covering}
For every $\eta >0$, there exists  a finite covering  $(B(c, \delta(c)))_{c\in C}$ with centers $C \subset \mathcal{X}^2 \times \mathcal{Z}$, radii $(\delta(c))_{c\in C}> 0$, and integers $(n(c))_{c\in C}$  such that, given a fixed $c = (\hat{\bs{x}}, \hat{\bs{x}}', \hat{\bs{z}}) \in C$, with probability at least $1-3\eta/4\kappa$, we have for all $(\bs{x},\bs{x}',\bs{z}) \in \mathcal{X}^2 \times \mathcal{Z}$ in the neighborhood of $(\hat{\bs{x}}, \hat{\bs{x}}', \hat{\bs{z}})$ that:
\begin{enumerate}
    \item $|U(\bs{x},\bs{\epsilon}(\bs{x}),\bs{z},\bs{\nu}(\bs{z})) - U(\hat{\bs{x}},\bs{\epsilon}(\hat{\bs{x}}),\hat{\bs{z}},\bs{\nu}(\hat{\bs{z}}))| \leq 1/n(c)$,
    \item $|U(\bs{x}',\bs{\epsilon}(\bs{x}'),\bs{z},\bs{\nu}(\bs{z})) - U(\hat{\bs{x}}',\bs{\epsilon}(\hat{\bs{x}}'),\hat{\bs{z}},\bs{\nu}(\hat{\bs{z}}))| \leq 1/n(c)$.
    \item $|U({\bs{x}},\bs{\epsilon}({\bs{x}}),{\bs{z}},\bs{\nu}({\bs{z}})) - U({\bs{x}}',\bs{\epsilon}({\bs{x}}'),{\bs{z}},\bs{\nu}({\bs{z}}))| \geq 5/n(c)$.
\end{enumerate}
\end{lemma}
The above lemma formalizes the main step in the proof of the celebrated result of \cite{mcfadden2000mixed} showing that continuous mixtures of MNL models uniformly approximate the class of RUMS. The proof of this result exploits the uniform continuity of $U(\cdot)$, $\bs{\epsilon}(\bs{x})$ and $\bs{\nu}(\bs{z})$ for all $(\bs{x},\bs{z}) \in \mathcal{X} \times \mathcal{Z}$. For completeness, the proof of Lemma~\ref{lem:covering} is given below; note that the statement of this result is implicit in~\cite{mcfadden2000mixed}, who base their analysis on slightly weaker properties of the covering.

\proof{\emph{Proof.}} Recall that $\bs{\epsilon}(\cdot)$ and $\bs{\nu}(\cdot)$ are random fields over $\mathcal{X}$ and $\mathcal{Z}$ respectively. For analysis purposes, we make the underlying probability space $(\Omega, \mathcal{W},\promega)$ explicit. In particular, we  denote by $\bs{\epsilon}(w,\bs{x})$ (resp. $\bs{\nu}(w,\bs{z})$) a particular realization of $\bs{\epsilon}(\bs{x})$ (resp. $\bs{\nu}(\bs{z})$). Consequently, for every $(w,\bs{x},\bs{z}) \in \Omega \times \mathcal{X} \times \mathcal{Z}$, we use the compact notation $ F(\omega, \bs{x},\bs{z}) = U\left(\bs{x}, \bs{\eps}(\omega,\bs{x}), \bs{z}, \bs{\nu}(\omega,\bs{z})\right).$ Recall that for all choice events $(\bs{z},A)$ and $\bs{x} \in A$, we have
\begin{align*}
    \pi(\bs{x},\bs{z},A) = \promega [F(\omega,\bs{x},\bs{z}) > F(\omega,\bs{x}',\bs{z}) , \forall \bs{x}' \in A \backslash \bs{x}].
\end{align*}
The remainder of the proof proceeds in three steps.
\begin{enumerate}
    \item \textit{The utilities are sufficiently different for distinct alternatives.} For every $(\bs{x}, \bs{x}',\bs{z}) \in \mathcal{X}^2 \times \mathcal{Z}$ and $n \in {\bb N}_+$, let
\begin{align*}
    \Omega_n(\bs{x},\bs{x}',\bs{z}) = \left \{ \omega \in \Omega : \left| F(\omega, \bs{x},\bs{z}) - F(\omega, \bs{x}',\bs{z}) \right| \geq 7/n \right \} .
\end{align*}
The continuity of $U(\cdot)$ and the measurability of the random fields $\bs{\eps}(\cdot)$ and $\bs{\nu}(\cdot)$ imply that $\Omega_n(\bs{x},\bs{x}',\bs{z})$ is measurable. This set is monotone increasing as $n \to \infty$ to the set of $\omega$ for which the alternatives $\bs{x}$ and $\bs{x}'$ are not tied. By hypothesis, this set has probability one, implying that there exists $\bar{n} = n(\bs{x},\bs{x}',\bs{z}) \in {\bb N}_+$ such that we have $\promega[\Omega_{\bar{n}}(\bs{x},\bs{x}',\bs{z})] \geq 1 - \eta / 4\kappa$.
\item \textit{There exists a neighborhood where utility remains controlled for distinct alternatives.} 
Fix $(\bs{x}, \bs{x}',\bs{z}) \in \mathcal{X}^2 \times \mathcal{Z}$. The uniform continuity of $U(\cdot)$ on ${\cal X}\times [0,1]^{d_x} \times {\cal Z}\times [0,1]^{d_z}$ implies that, for every given $\bar{n}\in \mathbb{N}_+$, there exists $\bar{\delta} = \delta(\bs{x},\bs{x}',\bs{z}) \in \mathbb{R}_+$ such that in a neighborhood of size $\bar{\delta}$, $U(\cdot)$ varies by less than $1/\bar{n}$. Moreover, the almost certain continuity of $\bs{\eps}(\omega,\bs{x})$ and $\bs{\nu}(\omega,\bs{z})$ implies that
\begin{align*}
    B_m(\bs{x},\bs{z}) = \left \{ \omega \in \Omega : \sup_{|\bs{x}-\bs{x}^*| < 1/m} \left| \bs{\eps}(\omega,\bs{x}) - \bs{\eps}(\omega,\bs{x}^*)\right| + \sup_{|\bs{z}-\bs{z}^*| < 1/m} \left| \bs{\nu}(\omega,\bs{z})- \bs{\nu}(\omega,\bs{z}^*)\right| \leq \frac{\bar{\delta}}{2} \right \},
\end{align*}
and the corresponding event $B_m(\bs{x}',\bs{z})$ are monotone increasing as $m \to \infty$ to limiting events that occur with probability one. Consequently, there exists $\bar{m}= m(\bs{x},\bs{x}',\bs{z}) \in \mathbb{N}_+$ such that $\promega[B_{\bar{m}}(\bs{x},\bs{z})] \geq 1 - \eta/4\kappa$ and $\promega[B_{\bar{m}}(\bs{x}',\bs{z})] \geq 1 - \eta/4\kappa$.

\item \textit{Finite covering.} We have established so far that $\prpartwo{\omega}{\Omega_{\bar{n}}(\bs{x},\bs{x}',\bs{z}) \cap B_{\bar{m}}(\bs{x},\bs{z}) \cap B_{\bar{m}}(\bs{x}',\bs{z})}\geq 1-3\eta/4\kappa$, noting that $\omega \in \Omega_{\bar{n}}(\bs{x},\bs{x}',\bs{z}) \cap B_{\bar{m}}(\bs{x},\bs{z}) \cap B_{\bar{m}}(\bs{x}',\bs{z})$ implies:
\begin{henumerate}
    \item $|F(\omega, \bs{x},\bs{z}) - F(\omega, \bs{x}',\bs{z})| \geq 7/\bar{n}$ \ ,
    \item For all $(\hat{\bs{x}},\hat{\bs{x}}',\hat{\bs{z}})\in $ in the open ball centered on $(\bs{x},\bs{x}',\bs{z})$ with radius $\min \{ 1/\bar{m}, {\bar{\delta}}/{2} \}$, we have $|F(\omega, \hat{\bs{x}},\hat{\bs{z}}) - F(\omega, \bs{x},\bs{z}) | \leq 1/\bar{n}$ and $|F(\omega, \hat{\bs{x}}',\hat{\bs{z}}) - F(\omega, \bs{x}',\bs{z})| \leq 1/\bar{n}$.
\end{henumerate}
The above neighborhoods specify a covering of the compact set $\mathcal{X} \times \mathcal{X} \times \mathcal{Z}$, from which we extract a finite subcovering $C$. Hence, we have just shown that each  $(\hat{\bs{x}},\hat{\bs{x}}',\hat{\bs{z}}) \in \mathcal{X} \times \mathcal{X} \times \mathcal{Z}$ falls in some neighborhood centered on $({\bs{x}},{\bs{x}}',{\bs{z}}) \in C$ such that for all  $\omega \in \Omega_{\bar{n}}(\bs{x},\bs{x}',\bs{z}) \cap B_{\bar{m}}(\bs{x},\bs{z}) \cap B_{\bar{m}}(\bs{x}',\bs{z})$:
\begin{eqnarray}
    &&| F(\omega, \hat{\bs{x}}, \hat{\bs{z}})  - F(\omega, \hat{\bs{x}}', \hat{\bs{z}})| \nonumber\\
    & & \quad \geq   | F(\omega, \bs{x},\bs{z})  - F(\omega, {\bs{x}'},{\bs{z}})| - | F(\omega, \hat{\bs{x}},\hat{\bs{z}})  - F(\omega, {\bs{x}},{\bs{z}})| - | F(\omega, \hat{\bs{x}}',\hat{\bs{z}})  - F(\omega, \bs{x}',\bs{z})| \nonumber\\
    & & \quad \geq  7/n( {\bs{x}},{\bs{x}}',{\bs{z}}) - 1/n ({\bs{x}},{\bs{x}}',{\bs{z}}) - 1/n ({\bs{x}},{\bs{x}}',{\bs{z}}) \nonumber\\
    & & \quad \geq   5/n ({\bs{x}},{\bs{x}}',{\bs{z}}) \ . \nonumber
\end{eqnarray}
\end{enumerate} \hfill \halmos 
\endproof

\subsection{Proof of Proposition~\ref{prop:approx}}

We now present the proof of Proposition~\ref{prop:approx}, which proceeds linearly in showing that we can control the errors in the successive layers of approximation of the utility function. Let $C$ denote the finite covering and $\bar{N} = \max\{n(c): c\in C\}$ that satisfy the properties stated in Lemma~\ref{lem:covering}. Let $N  \geq \max \{ - \log(\eta/4\kappa) , \bar{N} \}$ be a sufficiently large integer that will be determined later on. Recall that $\bs{\epsilon}(\cdot)$ and $\bs{\nu}(\cdot)$ are random fields over $\mathcal{X}$ and $\mathcal{Z}$ respectively. As in the proof of Lemma~\ref{lem:covering}, we explicitly describe this randomness. In particular, we introduce a fundamental probability space $(\Omega, \mathcal{W},\promega)$ and denote by $\bs{\epsilon}(w,\bs{x})$ (resp. $\bs{\nu}(w,\bs{z})$) a particular realization of $\bs{\epsilon}(\bs{x})$ (resp. $\bs{\nu}(\bs{z})$). For ease of notation, we let for $(w,\bs{x},\bs{z}) \in \Omega \times \mathcal{X} \times \mathcal{Z}$,
\begin{align*}
     F(\omega, \bs{x},\bs{z}) = U\left(\bs{x}, \bs{\eps}(\omega,\bs{x}), \bs{z}, \bs{\nu}(\omega,\bs{z})\right).
\end{align*}
Recall that for all choice event $(\bs{z},A)$ and $\bs{x} \in A$, we have
\begin{align*}
     \pi(\bs{x},\bs{z},A) = \promega [F(\omega,\bs{x},\bs{z}) > F(\omega,\bs{x}',\bs{z}) , \forall \bs{x}' \in A \backslash \bs{x}].
\end{align*}

\paragraph{Step 1: Universal approximation and discretization.} Feed-forward neural networks are a known class of universal approximators. More specifically, the space of feed-forward neural networks with only one hidden layer and one output unit is dense in $C(X)$, the space of continuous function on $X$, under the conditions that $X$ is compact and the activation function is continuous, bounded and non-constant \citep{hornik1989multilayer, hornik1991approximation}. This result is extended for networks of bounded width and arbitrary depth and any nonaffine continuous activation function by \cite{kidger2020universal}.  Consequently, there exists a feed-forward neural network $N_U(\cdot)$ that approximates $U(\cdot)$ on $\mathcal{X} \times [0,1]^{d_{\eps}} \times \mathcal{Z} \times [0,1]^{d_{\nu}}$, i.e., that satisfies $\|U(\cdot) - N_U(\cdot) \|_{\infty} \leq 1/N$. 

By the uniform continuity of $U(\cdot)$, there exist $\delta >0$, such that for all $(\bs{x},\bs{z}) \in \mathcal{X} \times \mathcal{Z}$ and $(\bs{\epsilon},\bs{\epsilon}') \in [0,1]^{2d_{\epsilon}}$ and $(\bs{\nu},\bs{\nu}') \in [0,1]^{2d_{\nu}}$ such that $\|\bs{\epsilon}-\bs{\epsilon}' \|_{\infty} \leq \delta$ and $\|\bs{\nu}-\bs{\nu}' \|_{\infty} \leq \delta$, we have $| U(\bs{x},\bs{\epsilon},\bs{z},\bs{\nu}) - U(\bs{x},\bs{\epsilon}',\bs{z},\bs{\nu}')| \leq 1/N$. For every $\omega \in \Omega$, there exist feed-forward neural networks $N_{\epsilon,\omega}(\bf{x})$ and $N_{\nu,\omega}(\bf{z})$ that approximate $\bs{\eps}(\omega,\bf{x})$ and $\bs{\nu}(\omega,\bf{z})$ on $\mathcal{X}$ and $\mathcal{Z}$ respectively, i.e., $\|\bs{\eps}(\omega,\bs{x})- N_{\epsilon,\omega}(\bs{x})\|_{\infty} \leq \delta$ for all $\bf{x} \in \mathcal{X}$ and $\|\bs{\nu}(\omega,\bs{z})- N_{\nu,\omega}(\bs{z})\|_{\infty} \leq \delta$ for all $\bs{z} \in \mathcal{Z}$. Consequently, for all $\bs{x} \in \mathcal{X}$, $\omega \in \Omega$ and $\bs{z} \in \mathcal{Z}$, we have
\begin{align}  \label{eq:approx_1}
|U(\bs{x}, \bs{\eps}(\omega,\bs{x}), \bs{z},\bs{\nu}(\omega,\bs{z}))-N_U(\bs{x}, N_{\epsilon,\omega}(\bs{x}), \bs{z},N_{\nu,\omega}(\bs{z}))|  \leq \dfrac{2}{N},
\end{align}
since $|N_U(\bs{x}, N_{\epsilon,\omega}(\bs{x}), \bs{z},N_{\nu,\omega}(\bs{z})) - U(\bs{x}, N_{\epsilon,\omega}(\bs{x}), \bs{z},N_{\nu,\omega}(\bs{z}))| \leq  1/N$ in light of the uniform approximation of $U(\cdot)$ by $N_U(\cdot)$ and  $|U(\bs{x}, N_{\epsilon,\omega}(\bs{x}), \bs{z},N_{\nu,\omega}(\bs{z})) - U(\bs{x}, \bs{\eps}(\omega,\bs{x}), \bs{z},\bs{\nu}(\omega,\bs{z}))| \leq  1/N$ due to the uniform continuity of $U(\cdot)$. 

\paragraph{Step 2: Adding noise.} Throughout the remainder of the proof, we fix a choice set $A$, an alternative $\bs{x}\in A$, and customer attributes $\bs{z} \in {\cal Z}$. For each $\bs{x} \in A$, we define a utility $F^1(\cdot)$ as follows: 
\begin{align} \label{eq:Uk}
    F^1(\omega, \delta_{\bs{x}},\bs{x},\bs{z}) = N_U (\bs{x}, N_{\epsilon,\omega}(\bs{x}),\bs{z},N_{\nu,\omega}(\bs{z})) + \dfrac{\delta_{\bs{x}}}{N^2} ,
\end{align}
where $\bs{\delta} = \{\delta_{\bs{x}} \}_{\bs{x} \in A}$ is a sequence i.i.d. random variables following a standard Gumbel distribution. Let 
\begin{align*}
    \pi^1(\bs{x},\bs{z},A) = {\rm Pr}_{\omega,\bs{\delta}}[F^1(\omega,\delta_{\bs{x}},\bs{x},\bs{z}) > F^1(\omega,\delta_{\bs{x}'},\bs{x}',\bs{z}) , \forall \bs{x}' \in A \backslash \{ \bs{x} \}],
\end{align*} 
be the choice probability associated with the utility $F^1(\cdot)$. For all $\bs{x}' \in A \backslash \{ \bs{x} \}$, 
let \begin{align*}
    \Omega^1_{\bs{x},\bs{x}'} =\{\omega \in \Omega , (\delta_{\bs{x}},\delta_{\bs{x}'}) \in \mathbb{R}^2: (F(\omega,\bs{x},\bs{z}) -  F(\omega,\bs{x}',\bs{z}) ) \times (F^{1}(\omega,\delta_{\bs{x}},\bs{x},\bs{z}) -  F^{1}(\omega,\delta_{\bs{x}'},\bs{x}',\bs{z})) <0 \} 
\end{align*} 
be the collection of realisations for which the preferences over alternatives $\bs{x}$ and $\bs{x}'$ are reversed when using $F^1(\cdot)$ instead of $F(\cdot)$. Note that 
\begin{eqnarray*}
	|\pi(\bs{x}, \bs{z},A) - \pi^1(\bs{x}, \bs{z},A)| \leq {\rm Pr} \left[ \bigcup_{\bs{x}' \in A \backslash \{ \bs{x} \}} \Omega^1_{\bs{x},\bs{x}'} \right] \leq \sum_{\bs{x}' \in A \backslash \{ \bs{x} \}}  \pr{\Omega^1_{\bs{x},\bs{x}'}} \leq \kappa \cdot \pr{\Omega^1_{\bs{x},\bs{x}'}} , 
\end{eqnarray*}
where the first inequality holds follows by noting that for $F^1(\cdot)$  to designate $\bs{x}$ as the the highest-utility alternative in the choice set $A$, while $F(\cdot)$ designates a different alternative $\bs{x}'$ in the choice set $A$, there needs to be a reversal of preferences between $\bs{x}$ and $\bs{x}'$. The second inequality follows from the union bound. Without loss of generality, consider $\omega \in \Omega$ and $(\delta_{\bs{x}},\delta_{\bs{x}'})\in \mathbb{R}^2$ such that 
\begin{align*}
    F(\omega,\bs{x},\bs{z}) > F(\omega,\bs{x}',\bs{z}) \text{ and } F^1(\omega,\delta_{\bs{x}},\bs{x},\bs{z}) < F^1(\omega, \delta_{\bs{x}'}, \bs{x}',\bs{z}).
\end{align*}
If $|F(\omega,\bs{x},\bs{z}) - F(\omega,\bs{x}',\bs{z})| > 5/N$, then
\begin{align*}
     0 &> F^1(\omega,\delta_{\bs{x}},\bs{x},\bs{z}) - F^1(\omega, \delta_{\bs{x}'}, \bs{x}',\bs{z}) \\
    &= N_U(\bs{x}, N_{\epsilon,\omega}(\bs{x}), \bs{z},N_{\nu,\omega}(\bs{z})) - N_U(\bs{x}', N_{\epsilon,\omega}(\bs{x}'), \bs{z},N_{\nu,\omega}(\bs{z})) + (\delta_{\bs{x}} - \delta_{\bs{x}'}) \\
     &> F(\omega,\bs{x},\bs{z}) - F(\omega,\bs{x}',\bs{z})  - 4/N + (\delta_{\bs{x}} - \delta_{\bs{x}'})/N^{2} \\
    &\geq 1/N + (\delta_{\bs{x}} - \delta_{\bs{x}'})/N^2,
\end{align*}
where the second inequality holds by Equation~\eqref{eq:approx_1}. Consequently, $\delta_{\bs{x}} - \delta_{\bs{x}'} \leq -N$. Under the Gumbel assumption, we have for $N \geq - \log(\eta/4\kappa)$,  ${\rm{Pr}}_{\delta}[\delta_{\bs{x}} - \delta_{\bs{x}'} \leq -N]  < \eta/4\kappa$. From Lemma~\ref{lem:covering}, the probability that $|F(\omega,\bs{x},\bs{z}) - F(\omega,\bs{x}',\bs{z})| \leq 5/N$ is at most $3 \eta/4\kappa$. Consequently, $\pr{\Omega^1_{\bs{x},\bs{x}'}} \leq \eta/\kappa$ which in turn implies that
\begin{align} \label{eq:approx_2}
    |\pi(\bs{x}, \bs{z},A) - \pi^1(\bs{x}, \bs{z},A)| \leq \eta.
\end{align}

\paragraph{Step 3: Sampling error.} For purposes of analysis, we need to refine the finite covering constructed in Lemma~\ref{lem:covering}. In particular, for every $\bs{x} \in {\cal X}$, $z\in {\cal Z}$, $\omega \in \Omega$ and $\delta \in \mathbb{R}$, we define the mapping $\bar{F}^1 (\omega,\bs{x},\bs{z}) =   N_U (\bs{x}, N_{\epsilon,\omega}(\bs{x}),\bs{z},N_{\nu,\omega}(\bs{z})) $. Next, we construct a finite covering that is mutually adapted to $F$ and $\bar{F}^1$ in the sense of Lemma~\ref{lem:covering}. Specifically, there exists  a finite covering $(B(c,\delta^+(c)))_{c\in C^+}$ with centers $C^+\subseteq {\cal X}^2 \times {\cal Z}$, radii $(\delta^+(c))_{c\in C^+}$, and integers $(n^+(c))_{c\in C^+}$ such that, given a fixed  $c=(\tilde{\bs{x}},\tilde{\bs{x}}',\bs{\tilde{z}}) \in C^+$, with probability at least $1- 3\eta/4\kappa$ , we have for all $(\bs{x},\bs{x'},\bs{z}) \in {\cal X}^2 \times {\cal Z}$ in the neighborhood of $(\tilde{\bs{x}},\tilde{\bs{x}}',\bs{\tilde{z}})$ that:
\begin{enumerate}
\item $|\bar{F}^{1}(\omega ,\bs{x},\bs{z}) - \bar{F}^1(\omega,\tilde{\bs{x}},{\bs{\tilde{z}}})| \leq 1/n^+(c)$ and $|F(\omega,\bs{x},\bs{z}) - F(\omega,\tilde{\bs{x}},\tilde{\bs{z}})| \leq 1/n^+(c)$,
\item $|\bar{F}^1(\omega ,\bs{x}',\bs{z}) - \bar{F}^1(\omega ,\tilde{\bs{x}}',{\bs{\tilde{z}}})| \leq 1/n^+(c)$ and $|F(\omega,\bs{x}',\bs{z}) - F(\omega ,\tilde{\bs{x}}',\tilde{\bs{z}})| \leq 1/n^+(c)$,
\item $|\bar{F}^1(\omega ,\bs{x},\bs{\tilde{z}}) - \bar{F}^1(\omega ,\bs{x}',\bs{\tilde{z}}) | \geq 5/n^+(c)$ and $|F(\omega ,\bs{x},\bs{\tilde{z}}) - F(\omega ,\bs{x}',{\bs{\tilde{z}}})|\geq 5/n^+(c)$.
\end{enumerate}

Consequently, we set the precise value of $N$ as $N = \max\{- \log(\eta/8\kappa), \bar{N}, \max\{n^+(c):c\in C^+\}\}$. Next, we construct a mapping from any choice set $A$ to $\hat{A} = \{\bs{\hat{x}}_1,\ldots,\bs{\hat{x}}_K\}$, where each alternative $\bs{x} \in A$ is replaced by a fixed element $\hat{\bs{x}}$ of $ \bigcap_{c\in C^+(\bs{x})}\{y_1: \bs{y} \in B(c,\delta^+(c))\}$, where $C^+(\bs{x})$ denotes all the centers $c\in C^+$ such that $(\bs{x},c_2,c_3) \in B(c, \delta^+(c))$. Note that the latter intersection of open balls is non-empty since it contains $\bs{x}$. Similarly, each $\bs{z}\in {\cal Z}$ is mapped to a fixed element $\hat{\bs{z}}$ of $ \bigcap_{c\in C^+(\bs{z})}\{y_3: \bs{y} \in B(c,\delta^+(c))\}$, where $C^+(\bs{z})$ denotes all the centers $c\in C^+$ such that $(c_1,c_2,\bs{z}) \in B(c, \delta^+(c))$. We denote by ${\cal E}$ the collection of all choice events $(\hat{\bs{x}},\hat{z},\hat{A})$ generated by this mapping. Since the covering $C^+$ is finite, ${\cal E}$ is also finite.  We establish the following claim in Appendix~\ref{app:approx_3}, showing that all choice events can be approximated by those in ${\cal E}$, with only small changes in their choice probabilities $\pi,\pi^1$.

\begin{claim}\label{eq:approx_3}
For every choice event $(\bs{x},\bs{z},{A}) \in {\cal X}\times {\cal Z}\times {\cal X}^{\kappa-1}$, we have 
\begin{align} 
    |\pi(\bs{x}, \bs{z},A) - \pi(\hat{\bs{x}},\hat{\bs{z}}, \hat{A})| \leq  \eta \text{ and } |\pi^{1}(\bs{x}, \bs{z},A) - \pi^{1}(\hat{\bs{x}},\hat{\bs{z}}, \hat{A})| \leq \eta \ .
\end{align}
\end{claim}

Now, the final piece of our proof is to construct a sample average approximation of $\pi^1(\cdot)$ with respect to $\omega$. Let $\bs{\hat{\omega}} = (\hat{\omega}_1,\ldots,\hat{\omega}_{K})$ be $K$ i.i.d. samples based on the probabilistic space $(\Omega,{\cal W},{\rm Pr}_{\omega})$ where the precise value of $K$ is specified later on. We define the {\em sampled utility} function $F^{2,k} (\delta_{\bs{x}},\bs{x},\bs{z}) $ as follows:
\begin{eqnarray} \label{eq:Ukq}
	F^{2,k} (\delta_{\bs{x}},\bs{x},\bs{z}) = F^1 (\hat{\omega}_k,\delta_{\bs{x}},\bs{x},\bs{z}) = N_U (\bs{x}, N_{\epsilon,\hat{\omega}_k}(\bs{x}),\bs{z},N_{\nu,\hat{\omega}_k}(\bs{z})) + \dfrac{\delta_{\bs{x}}}{N^{2}} \ .
\end{eqnarray}
By equation~\eqref{eq:Uk}, $F^{2,k} (\delta_{\bs{x}},\bs{x},\bs{z})$ can be interpreted as the realisation of $F^1(\omega,\delta_{\bs{x}},\bs{x},\bs{z})$ with respect to the sample $\hat{\omega}_k$. However, it is worth observing that $F^{2,k} (\delta_{\bs{x}},\bs{x},\bs{z})$ is itself a random variable due to the noise term $\delta_{\bs{x}}$. Finally, we let $\pi^2(\bs{x},\bs{z},A)$ be the sample mean estimator of the choice probabilities with respect to $F^{2,1},\ldots, F^{2,K}$, i.e.,
\begin{eqnarray} \label{eq:piomega}
	\pi^2\left(\bs{x},\bs{z},A\right) = \frac{1}{K} \cdot \sum_{k=1}^{K} \prtwo{\bs{\delta}}{ F^{2,k}(\delta_{\bs{x}},\bs{x},\bs{z}) > F^{2,k}(\delta_{\bs{x}'}, \bs{x}',\bs{z}) , \ \forall \bs{x}' \in A \backslash \{ \bs{x} \}} \ .
\end{eqnarray}
By equations~\eqref{eq:Ukq} and~\eqref{eq:piomega}, it is clear that the choice model $\pi^2(\cdot)$ can be represented by an instance of the RUMnet architecture. Hence, we denote by $N(\bs{\hat{\omega}}) \in {\cal N}^{\bs{d}}(K,\Theta^{\ell,w}_M)$ the RUMnet architecture such that $\pi^2(\cdot) = \pi^{{\sf RUMnet}}_{N(\bs{\hat{\omega}})}(\cdot)$.

Next, we analyse the differences between the choice models $\pi^1(\cdot)$ and its sample mean approximation $\pi^{{\sf RUMnet}}_{N(\bs{\hat{\omega}})}(\cdot)$. First, we establish a property analogous to Claim~\ref{eq:approx_3} with respect to $\pi^2(\cdot)$. The proof is provided in Appendix~\ref{app:clm:fine}
\begin{claim} \label{clm:fine}
For any $K \geq \frac{\kappa^2\log (4|{\cal E}|)}{2\cdot \eta^2}$, with probability at least 3/4 with respect to $\hat{\bs{\omega}}$, for all choice events $(\bs{x},\bs{z},{A}) \in {\cal X}\times {\cal Z} \times {\cal X}^{\kappa-1}$, we have 
\begin{align*} 
   |\pi^{{\sf RUMnet}}_{N(\bs{\hat{\omega}})}(\bs{x}, \bs{z},A) - \pi^{{\sf RUMnet}}_{N(\bs{\hat{\omega}})}(\hat{\bs{x}},\hat{\bs{z}}, \hat{A})| \leq 2\eta \ .
\end{align*}
\end{claim}

Next, we observe that, for all $K \geq \frac{\log (4|{\cal E}|)}{2\cdot \eta^2}$,
\begin{eqnarray}
&&\prtwo{\hat{\omega}}{\max_{(\bs{x},\hat{z},\hat{A}) \in {\cal E}} \left|\pi^1(\hat{\bs{x}},\hat{\bs{z}},\hat{A}) -  \pi^{{\sf RUMnet}}_{N(\bs{\hat{\omega}})}(\hat{\bs{x}},\hat{\bs{z}},\hat{A})\right| > \eta } \nonumber\\
& & \qquad \leq \sum_{(\bs{x},\hat{z},\hat{A})\in {\cal E}} \prtwo{\hat{\omega}}{ \left|\pi^1(\hat{\bs{x}},\hat{\bs{z}},\hat{A}) -  \pi^{{\sf RUMnet}}_{N(\bs{\hat{\omega}})}(\hat{\bs{x}},\hat{\bs{z}},\hat{A})\right| > \eta } \nonumber\\
& & \qquad \leq \sum_{(\bs{x},\hat{z},\hat{A})\in {\cal E}} 2\cdot e^{-2 \cdot \eta ^2 \frac{\log (4|{\cal E}|)}{2\cdot \eta^2}} \nonumber\\
& & \qquad \leq \frac{1}{2} \label{ineq:1/2} \ ,
\end{eqnarray}
where the first inequality follows from the union bound and the second inequality from Hoeffding's inequality. Fix $K  = \lceil \frac{\kappa^2\log (4|{\cal E}|)}{2\cdot \eta^2} \rceil$. By the union bound over inequality~\eqref{ineq:1/2} and Claim~\ref{clm:fine}, there exists a realization $\hat{\bs{\omega}}^*$ of  $\hat{\bs{\omega}}$  such that, for all $(\hat{\bs{x}},\hat{\bs{z}},\hat{A}) \in {\cal E}$, we have
\begin{align} \label{eq:approx_4}
    \left|\pi^1(\hat{\bs{x}},\hat{\bs{z}},\hat{A}) -  \pi^{{\sf RUMnet}}_{N(\hat{\bs{\omega}}^*)}(\hat{\bs{x}},\hat{\bs{z}},\hat{A})\right| \leq \eta \ .
\end{align}
and for all choice events $(\bs{x},\bs{z},A) \in {\cal X}\times {\cal Z} \times {\cal X}^{\kappa-1}$, we have 
\begin{align} \label{eq:approx_5} 
   |\pi^{{\sf RUMnet}}_{N(\bs{\hat{\omega}^*})}(\bs{x}, \bs{z},A) - \pi^{{\sf RUMnet}}_{N(\bs{\hat{\omega}^*})}(\hat{\bs{x}},\hat{\bs{z}}, \hat{A})| \leq 2\eta \ .
\end{align}
Putting together Claim~\ref{eq:approx_3}, inequalities~\eqref{eq:approx_2}, \eqref{eq:approx_4} and \eqref{eq:approx_5}, we have
\begin{eqnarray*}
&&	\left|\pi(\bs{x},\bs{z},A) - \pi^{{\sf RUMnet}}_{N(\hat{\bs{\omega}}^*)}(\bs{x},\bs{z},{A})\right| \\	
&& \quad \leq \left|\pi({\bs{x}},{\bs{z}},{A}) - \pi(\hat{\bs{x}},\hat{\bs{z}},\hat{A})\right| + \left|\pi(\hat{\bs{x}},\hat{\bs{z}},\hat{A}) - \pi^1(\hat{\bs{x}},\hat{\bs{z}},\hat{A}) \right| + \left|\pi^1(\hat{\bs{x}},\hat{\bs{z}},\hat{A}) - \pi^{{\sf RUMnet}}_{N(\hat{\bs{\omega}}^*)}(\hat{\bs{x}},\hat{\bs{{z}}},\hat{A})\right|    \\
&& \qquad + \left|\pi^{{\sf RUMnet}}_{N(\hat{\bs{\omega}}^*)}(\hat{\bs{x}},\hat{\bs{{z}}},\hat{A}) - \pi^{{\sf RUMnet}}_{N(\hat{\bs{\omega}}^*)}({\bs{x}},{\bs{{z}}},{A})\right| \\
&& \quad \leq 5 \eta \ ,
\end{eqnarray*}
which yields the desired result. \hfill \halmos
\endproof 

\subsection{Proof of Claim~\ref{eq:approx_3}}\label{app:approx_3}
We establish the desired inequality for the choice model $\pi^1(\cdot)$ since the case of $\pi(\cdot)$ proceeds from an identical reasoning. As in Step~2, we define $\Omega_{\bs{x},\bs{x}'}$ for every $\bs{x}' \in A\setminus \{\bs{x}\}$ as the event where customer $\bs{z}$'s preferences over $(\bs{x},\bs{x}')$ are reversed compared to customer $\hat{\bs{z}}$'s preferences  over $(\hat{\bs{x}},\hat{\bs{x}}')$ with respect to the random utility function $F^1(\omega, \delta,\cdot,\cdot)$. By the union bound, we have 
\begin{equation}\label{ineq:union}
     |\pi(\bs{x}, \bs{z},A) - \pi(\hat{\bs{x}},\hat{\bs{z}}, \hat{A})| \leq \sum_{\bs{x}' \in A\setminus \{\bs{x}\}} \pr{\Omega_{\bs{x},\bs{x}'}} \leq \kappa \cdot \max_{\bs{x}' \in A\setminus \{\bs{x}\}} \pr{\Omega_{\bs{x},\bs{x}'}} \ .
\end{equation}
Without loss of generality, fix $\bs{x},\bs{x}'\in A^2$ and consider $\omega \in \Omega$ and $(\delta_{\bs{x}},\delta_{\bs{x}'})\in \mathbb{R}^2$ such that 
\begin{align*}
    F^1(\omega,\delta_{\bs{x}},\bs{x},\bs{z}) > F^1(\omega,\delta_{\bs{x}'},\bs{x}',\bs{z}) \text{ and } F^1(\omega,\delta_{\hat{\bs{x}}},\hat{\bs{x}},\hat{\bs{z}}) < F^1(\omega,\delta_{\hat{\bs{x}}'},\hat{\bs{x}}',\hat{\bs{z}}) .
\end{align*}
Now, let $c = (\tilde{\bs{x}},\tilde{\bs{x}}',\tilde{\bs{z}}) \in C^+$ be the center of a neighborhood that contains  $({\bs{x}},{\bs{x}}',{\bs{z}})$. By construction of our mapping, this neighborhood necessarily contains  $(\hat{\bs{x}},\hat{\bs{x}}',\hat{\bs{z}})$ based on the coordinate-wise inequalities $\delta^{+}(c) \geq \max\{|\hat{\bs{x}} -{\bs{x}}|,|\hat{\bs{x}}' -{\bs{x}}'|,|\hat{\bs{z}} -{\bs{z}}|\} $, which imply that $(\hat{\bs{x}},\hat{\bs{x}}',\hat{\bs{z}})\in B(c,\delta^+(c))$.  In what follows, suppose that $\omega$ satisfies Properties~1-3 of the covering; by construction, this event occurs with probability at least $1- 3\eta/4\kappa$. Due to properties 1-2, we have
\begin{align*}
     0 &> F^1(\omega,\delta_{\hat{\bs{x}}},\hat{\bs{x}},\hat{\bs{z}}) - F^1(\omega,\delta_{\hat{\bs{x}}'},\hat{\bs{x}}',\hat{\bs{z}}) \\
    &\geq \bar{F}^1(\omega,{\tilde{\bs{x}}},\tilde{\bs{z}}) - \bar{F}^1(\omega,\tilde{\bs{x}}',\tilde{\bs{z}}) - |\bar{F}^1(\omega,{\hat{\bs{x}}},\hat{\bs{z}}) - \bar{F}^1(\omega,\tilde{\bs{x}},\tilde{\bs{z}}) | - | \bar{F}^1(\omega,{\hat{\bs{x}}'},\hat{\bs{z}}) - \bar{F}^1(\omega,\tilde{\bs{x}}',\tilde{\bs{z}})| + (\delta_{\hat{\bs{x}}} - \delta_{\hat{\bs{x}}'})/N^{2} \\
    &\geq \bar{F}^1(\omega,{\tilde{\bs{x}}},\tilde{\bs{z}}) - \bar{F}^1(\omega,\tilde{\bs{x}}',\tilde{\bs{z}}) - 2/n^+(c) + (\delta_{\hat{\bs{x}}} - \delta_{\hat{\bs{x}}'})/N^{2} \ ,
\end{align*}
and
\begin{align*}
     0 & < F^1(\omega,\delta_{{\bs{x}}},{\bs{x}},{\bs{z}}) - F^1(\omega,\delta_{{\bs{x}}'},{\bs{x}}',{\bs{z}}) \\
    &\leq \bar{F}^1(\omega,{\tilde{\bs{x}}},\tilde{\bs{z}}) - \bar{F}^1(\omega,\tilde{\bs{x}}',\tilde{\bs{z}}) + |\bar{F}^1(\omega,{{\bs{x}}},{\bs{z}}) - \bar{F}^1(\omega,\tilde{\bs{x}},\tilde{\bs{z}}) | + | \bar{F}^1(\omega,{{\bs{x}}'},{\bs{z}}) - \bar{F}^1(\omega,\tilde{\bs{x}}',\tilde{\bs{z}})| + (\delta_{{\bs{x}}} - \delta_{{\bs{x}}'})/N^{2} \\
    &\leq \bar{F}^1(\omega,{\tilde{\bs{x}}},\tilde{\bs{z}}) - \bar{F}^1(\omega,\tilde{\bs{x}}',\tilde{\bs{z}}) + 2/n^+(c) + (\delta_{{\bs{x}}} - \delta_{{\bs{x}}'})/N^{2} \ .
\end{align*}
Combining these inequalities with property~3, we infer that either $\delta_{\hat{\bs{x}}} - \delta_{\hat{\bs{x}}'} \leq -3\frac{N^2}{n^+(c)} \leq -3 N$ or  $\delta_{{\bs{x}}} - \delta_{{\bs{x}}'} \geq 3 \frac{N^2}{n^+(c)}  \geq 3 N$. Since $N \geq - \log(\frac{\eta}{8\kappa})$ and  $\bs{\delta}$ is a collection of i.i.d. Gumbel random variables, each of these events occurs with probability at most $\frac{\eta}{8\kappa}$. By the union bound, we derive the upper bound on reversal probabilities:
\begin{eqnarray*}
\pr{\Omega_{\bs{x},\bs{x}'}} \leq \frac{3\eta}{4\kappa} + 2\frac{\eta}{8\kappa} = \frac{\eta}{\kappa} \ .
\end{eqnarray*}
The desired inequality immediately follows by plugging the above inequality into~\eqref{ineq:union}.

\subsection{Proof of Claim~\ref{clm:fine}}\label{app:clm:fine}

Fix a choice event $(\bs{x},\bs{z},A) \in {\cal X}\times {\cal Z}\times {\cal X}^{\kappa-1}$. Similarly to the proof of Claim~\ref{eq:approx_3}, for each $\bs{x}' \in A\setminus \{\bs{x}'\}$ and $k\in [K]$, we develop an upper bound on the probability of a preference reversal conditional to the unobserved attribute $\hat{\bs{\omega}} = \hat{{\omega}}_k $. Specifically, the preference reversal event $\Omega^k_{\bs{x},\bs{x}',\bs{z}}$ occurs when customer $\bs{z}$'s preferences over $(\bs{x},\bs{x}')$ are reversed compared to customer $\hat{\bs{z}}$'s preferences over $(\hat{\bs{x}},\hat{\bs{x}}')$ with respect to the $k$-th sampled random utility function $F^{2,k}(\delta,\cdot,\cdot) = F^{1}(\hat{\omega}_k,\delta,\cdot,\cdot)$.  Let $G^k_{{\bs{x}},{\bs{x}}',\bs{z}}$ denote the event for which $\hat{\omega}_k$ satisfies properties 1-3 with respect to the center $(\tilde{\bs{x}},\tilde{\bs{x}}',\tilde{\bs{z}})$. The fact that $(\hat{\bs{x}},\hat{\bs{x}}',\hat{\bs{z}})$ and $({\bs{x}},{\bs{x}}',{\bs{z}})$ are both contained in one neighborhood of the covering implies $G^k_{{\bs{x}},{\bs{x}}',\bs{z}} = G^k_{\hat{\bs{x}},\hat{\bs{x}}',\hat{\bs{z}}}$. Now, a close examination of the proof of Claim~\ref{eq:approx_3} reveals that, conditional on $G^k_{\hat{\bs{x}},\hat{\bs{x}}',\hat{\bs{z}}}$, the preference reversal occurs with probability at most $\frac{\eta}{4\kappa}$. It follows that:
\begin{eqnarray*}
&&|\pi^{{\sf RUMnet}}_{N(\bs{\hat{\omega}})}(\bs{x}, \bs{z},A) - \pi^{{\sf RUMnet}}_{N(\bs{\hat{\omega}})}(\hat{\bs{x}},\hat{\bs{z}}, \hat{A})| \\
& &\leq \frac{1}{K}\cdot \sum_{k=1}^K  \pr{\left.\bigcup_{\bs{x}' \in A\setminus \{\bs{x}\}}\Omega^k_{\bs{x},\bs{x}',\bs{z}}\right| \hat{\bs{\omega}}} \\
& &\leq \frac{1}{K}\cdot \sum_{k=1}^K  \sum_{\bs{x}' \in A\setminus \{\bs{x}\}} \prtwo{\delta}{\left.\Omega^k_{\bs{x},\bs{x}',\bs{z}}\right| \hat{\omega}_k} \\
& &\leq\frac{1}{K}\cdot \sum_{k=1}^K  \sum_{\bs{x}' \in A\setminus \{\bs{x}\}} \left( {\bb I}[\hat{\omega}_k \notin {G}^k_{\hat{\bs{x}},\hat{\bs{x}}',\hat{\bs{z}}}] +   \pr{\left.\Omega^k_{\bs{x},\bs{x}',\bs{z}} \right| \hat{\omega}_k, \hat{\omega}_k \in G^k_{\hat{\bs{x}},\hat{\bs{x}}',\hat{\bs{z}}}} \cdot {\bb I}[\hat{\omega}_k \in G^k_{\hat{\bs{x}},\hat{\bs{x}}',\hat{\bs{z}}}] \right) \\
& &\leq \frac{1}{K}\cdot \sum_{k=1}^K  \sum_{\bs{x}' \in A\setminus \{\bs{x}\}} \left( {\bb I}[\hat{\omega}_k \notin {G}^k_{\hat{\bs{x}},\hat{\bs{x}}',\hat{\bs{z}}}] +   \frac{\eta}{4\kappa}\cdot {\bb I}[\hat{\omega}_k \in G^k_{\hat{\bs{x}},\hat{\bs{x}}',\hat{\bs{z}}}] \right) \\
& &\leq \frac{\eta}{4} +  \sum_{\bs{x}' \in A\setminus \{\bs{x}\}} \frac{1}{K}\cdot \sum_{k=1}^K  {\bb I}[\hat{\omega}_k \notin {G}^k_{\hat{\bs{x}},\hat{\bs{x}}',\hat{\bs{z}}}] \ ,
\end{eqnarray*}
where the first inequality holds since, absent a preference reversal over one pair $(\bs{x},\bs{x}')$, our RUMnet model identifies the same highest-utility alternative in the assortments $A$ and $\hat{A}$. The second inequality proceed from the union bound. The next inequality follows from the formula of conditional expectations. The fourth inequality is direct consequence of our upper bound derived from the proof of Claim~\ref{eq:approx_3}. Now, we invoke Hoeffding's inequality so that, for every $\hat{\bs{x}}' \in \hat{A}$, we have
\[
\prtwo{\hat{\bs{\omega}}}{\frac{1}{K}\cdot\left(\sum_{k=1}^K  {\bb I}[\hat{\omega}_k \notin {G}^k_{\hat{\bs{x}},\hat{\bs{x}}',\hat{\bs{z}}}]\right) - \frac{3\eta}{4\kappa} > \frac{\eta}{\kappa}} \leq e^{-2\frac{\eta^2}{\kappa^2} K} \ ,
\]
where we note that $\expartwo{\hat{\bs{\omega}}}{\frac{1}{K}\sum_{k=1}^K  {\bb I}[\hat{\omega}_k \notin {G}^k_{\hat{\bs{x}},\hat{\bs{x}}',\hat{\bs{z}}}]} = \prpar{\hat{\omega}_k \notin {G}^k_{\hat{\bs{x}},\hat{\bs{x}}',\hat{\bs{z}}}]} \leq \frac{3\eta}{4\kappa}$ based on the construction of our covering. By the union bound, we have 
\begin{eqnarray*}
&&\prtwo{\hat{\bs{\omega}}}{\exists (\bs{x},\bs{z},{A}) \in {\cal X}\times {\cal Z} \times {\cal X}^{\kappa-1}\text{ s.t. } |\pi^{{\sf RUMnet}}_{N(\bs{\hat{\omega}})}(\bs{x}, \bs{z},A) - \pi^{{\sf RUMnet}}_{N(\bs{\hat{\omega}})}(\hat{\bs{x}},\hat{\bs{z}}, \hat{A})| > 2\eta }  \leq  |{\cal E}| \cdot e^{-2\frac{\eta^2}{\kappa^2} K}\leq \frac{1}{4} \ .
\end{eqnarray*}
where the first inequality holds since the number of distinct vectors $(\hat{\bs{x}},\hat{\bs{x}}',\hat{\bs{z}})$ is upper bounded by $|{\cal E}|$ and the next inequality immediately follows from the condition  $K \geq \frac{\kappa^2\log (4|{\cal E}|)}{2\cdot \eta^2}$.

\newpage 

\section{Proofs of Section~\ref{sec:curse_dimensionality}} \label{app:curse_dimensionality}

\subsection{Proof of Proposition~\ref{prop:gen-error}} \label{app:prop:gen-error}

Let ${\cal H}$ be a hypothesis class, ${\cal E}$ be a collection of data observations endowed with a distribution ${\cal D}$, and $\ell(\cdot)$ be a loss function with respect to ${\cal H}$ and ${\cal Z}$. For every hypothesis $h \in {\cal H}$, training sample $S = \{ \bs{s}_1, \dots, \bs{s}_T \}$, and loss function $\ell(\cdot)$, we define the empirical error as
\begin{align*}
   L_S(h) = \dfrac{1}{T} \cdot \sum \limits_{t=1}^T \ell (h(\bs{s}_t)),
\end{align*}
and the associated true error
\begin{align*}
 L^{\sf true}_{\mathcal{D}}(h) =  \extwo{S' \sim \mathcal{D}}{L_{S'}(h)} .
\end{align*}
For a given training sample $S$, let $h^{\sf ERM}_S$ be the hypothesis that minimizes the empirical error $L_S(h)$ over all $h \in \mathcal{H}$. We begin by invoking a classical result to bound the generalization based on the Rademacher complexity; e.g., see~\citet{shalev2014understanding,koltchinskii2000rademacher,bartlett2002rademacher}. Recall that the Rademacher complexity for a set of vectors $A \subseteq {\bb R}^T$ is 
\begin{align*}
    R(A) = \dfrac{1}{T} \cdot \extwo{\bs{\sigma}}{\sup_{a\in A} \sum_{i=1}^T \sigma_i a_i},
\end{align*}
where $\bs{\sigma} = (\sigma_1, \ldots, \sigma_m)$ is a sequence of independent Rademacher random variables defined for all $i$ by the following distribution: $\prpar{\sigma_i = 1} = \prpar{\sigma_i = -1} = \frac{1}{2}$.

\begin{theorem}[Shalev-Swartz and Ben-David (2014, Thm.~26.5)] 
	 Suppose that for every hypothesis $h \in {\cal H}$ and observation $\bs{s} \in {\cal E}$, we have $|\ell(h(\bs{s}))| \leq c$ for some $c\geq 0$. Then, with probability of at  least $(1-\delta)$ with regards to the data sample $S \sim {\cal D}^{T}$, we have
	\[
	L^{\sf true}_{\mathcal{D}} \left( h^{\sf ERM}_S \right)  \leq  L_S( h^{\sf ERM}_S ) + 2 R\left(\ell \circ{\cal H} \circ S\right) + 4c\sqrt{\frac{2\ln(4/\delta)}{T}} \ ,
	\]
	where $\ell \circ{\cal H} \circ S = \{ (\ell(h(\bs{s}_1)), \dots, \ell(h(\bs{s}_T))) : h \in \mathcal{H} \}$.
\end{theorem}
This type of bound is often referred to as a data-dependent bound since the bound depends on the specific training set $S$. It therefore suffices to bound the Rademacher complexity of the RUMnet class ${\cal N}^{\bs{d}}(K,\Theta^{\ell,w}_M)$ composed with our data generative process and the negative log-likelihood loss function. Proposition~\ref{prop:gen-error} immediately follows from Lemma~\ref{lem:rad} below.

\begin{lemma} \label{lem:rad}
Let $\ell(\cdot)$ be the negative log-likelihood loss function, $\mathcal{H}_{\sf RUMnet}$ be the class of RUMnet class and $S = ((\bs{y}_1,\bs{z}_1,A_1),\dots, (\bs{y}_T,\bs{z}_T,A_T))$ be a training sample. Then, there exists a constant $c_1>0$ such that 
\begin{align*}
    R(\ell \circ \mathcal{H}_{\sf RUMnet} \circ S) \leq c_1\cdot {\frac{\kappa 
\sqrt{\kappa }}{\sqrt{T}}}  \cdot e^{2 w \cdot M} M^{\ell}.
\end{align*}
\end{lemma}
The remainder of this section establishes Lemma~\ref{lem:rad}. The proof plugs together various notions of Rademacher calculus. We first invoke several auxiliary results established in the previous literature. We  make use of the standard lemmas; see~\citet[Chap.~26]{shalev2014understanding}.
\begin{lemma}[Contraction] \label{thm-contraction}
Let $A$ be a subset of ${\bb R}^m$. For each $i \in [m]$, let $\phi_i : \mathbb{R} \to \mathbb{R}$ be a $\rho$-Lipschitz function; namely, for all $\alpha, \beta \in \mathbb{R}$, we have $|\phi_i(\alpha) - \phi_i(\beta)| \leq \rho | \alpha- \beta|$. For $\mathbf{a} \in \mathbb{R}^m$ let $\bm{\phi(a)}$ denote the vector $(\phi_1(a_1),\dots,\phi_m(a_m))$. Let $\bm{\phi} \circ A = \{ {\boldsymbol \phi}(a) : a \in A \}$. Then,
\begin{align*}
    R(\bm{\phi} \circ A) \leq \rho R(A).
\end{align*}
\end{lemma}

\begin{lemma}[Convex combination] \label{thm-compo}
	Let $A$ be a subset of ${\bb R}^m$ and let $A' = \{\sum_{j=1}^N \alpha_j \boldsymbol{a}^{(j)}: N \in {\bb N}, \forall j, \boldsymbol{a}^{(j)} \in A, \alpha_j \geq 0, \lvert \boldsymbol{\alpha}\rvert_1 = 1\}$. Then, $R(A') = R(A)$.
\end{lemma}
We utilize a generalization of the contraction lemma for hypothesis classes formed by vector-valued functions, established in the paper by~\citet[Corollary~4]{maurer2016vector}.
\begin{lemma}[Vector-valued contraction] \label{thm-contraction2}
Let ${\cal E}$ be an arbitrary set, $(\varepsilon_1,\ldots \varepsilon_T) \in {\cal E}^T$, and let ${\cal F}$ be a class of $m$-dimensional functions $f: {\cal E} \rightarrow {\bb R}^m$ and let $h_i: {\bb R}^m \rightarrow {\bb R}$ be an $L$-Lipschitz function for each $i\in [T]$ with respect to the $\ell_2$-norm. Then,
\[
\ex{\sup_{f\in {\cal F}} \sum_{i =1}^T \sigma_i h_i\left(f\left( \varepsilon_i\right) \right)} \leq \sqrt{2} L \cdot \ex{\sup_{f\in {\cal F}} \sum_{i=1}^T \sum_{k = 1}^m \sigma_{i,k} f_k\left(\varepsilon_i\right)}  \ ,
\]

where $\{\sigma_i\}_{i\in [T]}$ and $\{\sigma_{i,k}\}_{i \in [T], k \in [m]}$ are i.i.d. Rademacher random variables.
\end{lemma}

The following claim can be found in~\citet[Cor.~2]{neyshabur2015norm}.

\begin{lemma} [Rademacher complexity of neural networks] \label{thm-NN}
Fix a neural network architecture with $\ell \geq 1$ layers and assume that (i) the weight vector $w$ for every node in the network satisfies $\Vert w \Vert_1 \leq M$, (ii) the activation functions are ReLUs, and (iii) the input vectors $S = \{x_1,\ldots,x_T\} \subseteq {\bb R}^d$ satisfy $\|\bs{x}_t\|_{\infty} \leq 1$. Then, the class of functions $\mathcal{F} \in \mathbb{R}^{\mathcal{E}}$ defined by such neural network architecture over the inputs $S$ satisfies
\begin{align*}
    R(\mathcal{F} \circ S) \leq \sqrt{\dfrac{{4\log(2d)}}{T}} (2M)^{\ell} \ .
\end{align*}
\end{lemma}

Finally, let $p_{\min}$ be the minimum choice probability attained over the compact set of RUMnet architectures $\mathcal{H}_{\sf RUMnet}$. Our analysis will make use of the following property. 
\begin{claim} \label{clm:pmin}
	$p_{\min}  \geq \frac{1}{\kappa } \cdot e^{-2M}$.
\end{claim}
The claim follows immediately by noting that there are up to $\kappa$ choice alternatives. Additionally, the MNL attractiveness weight of each alternative is in the range $[e^{-M},e^{M}]$ since the utility $u$ of each alternative is the final layer's output of a feed-forward neural network $N_U \in \Theta_M^{\ell,w}$.

Now, we break down generating the collection of vectors $\ell \circ \mathcal{H}_{\sf RUMnet} \circ S$ into four operations for each choice event (in reverse order): (1) applying a log-transformation to the choice probability of the selected alternative, (2) averaging the choice probabilities over the samples of the RUMnet architecture, (3) applying a softmax transformation of the utilities within each offered assortment, (4) computing the utility of each alternative using a feed-forward neural network. We denote by $\psi_1,\psi_2,\psi_3,\psi_4$ the class of mappings corresponding to each step, and by slightly abusing notation, we have $\psi_1 \circ \psi_2 \circ \psi_3\circ\psi_4 \circ S  =\ell \circ \mathcal{H}_{\sf RUMnet} \circ S$. Now, we bound the Rademacher complexity by considering these successive transformations:
\begin{enumerate}
\item Note that $\psi_1$ applies the transformation $p \in (0,1) \mapsto \log(p)$ to the estimated probability $p$ of the chosen alternative for each event $\bs{s}\in S$. This mapping is $(1/p_{\min})$-Lipschitz where $p_{\min}$ is the minimal choice probability attained by the RUMnet architecture. Therefore, by combining Claim~\ref{clm:pmin} and Lemma~\ref{thm-contraction}, we have:
\begin{eqnarray} \label{step1}
R( \psi_1 \circ \psi_2 \circ  \psi_3 \circ  \psi_4 \circ S) =  \kappa e^{2M}\cdot R( \psi_2 \circ \psi_3\circ \psi_4 \circ S) \ .
\end{eqnarray}

\item Next, the mapping $\psi_2$ computes the unweighted average of choice probabilities over the samples $(k_1,k_2)\in [K]^2$ of the RUMnet architecture for each choice event $\bs{s}\in S$. This transformation amount to a convex combination of the choice probability vectors. Thus, by Lemma~\ref{thm-compo}, we have
\begin{eqnarray}
  R( \psi_2\circ\psi_3 \circ \psi_4 \circ S) =  R(\psi_3 \circ \psi_4 \circ S) \ .  \label{step2}
\end{eqnarray}
\item The mapping $\psi_3$ applies a softmax to the utilities of the alternatives for each choice event $\bs{s} \in S$ and sample $k_1,k_2 \in [K]$, and then, returns the coordinate corresponding to the chosen alternative. Note that the  function $ {\rm softmax}: \bs{u} \in {\bb R}^\kappa  \mapsto \left(\frac{e^{u_j}}{\sum_{k \in [\kappa] } e^{u_k}}\right)_{j\in [\kappa ]}$ is $1$-Lipschitz. Therefore, we apply Lemma~\ref{thm-contraction2} with $T\gets T$, $\varepsilon_i \gets (\bs{z}_i, A_i)$, ${\cal E} \gets  {\cal Z}\times {\cal X}^{\kappa}$, $f \gets ({\rm utility}_k(\cdot))_{k\in \bs{A}_i}$, $m\gets \kappa$, $h_i\gets {\rm softmax}_{\bs{y}_i}(\cdot)$ and $L\gets 1$, from which it follows that
\begin{eqnarray}
R(\psi_3 \circ \psi_4 \circ S) &= & \frac{1}{T} \ex{\sup_{f\in {\cal F}} \sum_{i =1}^T \sigma_i h_i\left(f\left( \varepsilon_i\right) \right)} \nonumber\\
&\leq & \frac{\sqrt{2}}{T}  \cdot \ex{\sup_{f\in {\cal F}} \sum_{i=1}^T \sum_{k = 1}^m \sigma_{i,k} f_k\left(\varepsilon_i\right)} \nonumber\\
&= & \sqrt{2} \kappa \left( \frac{1}{\kappa T}  \cdot \ex{\sup_{f\in {\cal F}} \sum_{i=1}^T \sum_{k = 1}^m \sigma_{i,k} f_k\left(\varepsilon_i\right)} \right) \nonumber\\
&\leq & \sqrt{2} \kappa \left(\frac{1}{\kappa T}  \cdot \ex{\sup_{u_{i,k}\in {\cal U}} \sum_{i=1}^T \sum_{k = 1}^\kappa \sigma_{i,k} u_{i,k}\left(\varepsilon_i\right)}\right) \nonumber\\ 
&=& \sqrt{2}  \kappa \cdot R(\psi_4 \circ S') \label{step3} \ ,
\end{eqnarray}

where $\psi_4 \circ S'$ is the class of utility vectors computed by the RUMnet architecture from the sample $S'$ formed by the $\kappa\cdot T$ vectors of the form $(\bs{x},\bs{z})$ corresponding to each alternative and customer attribute in $S$. Here, the hypothesis class ${\cal U}$ is the class of utility functions, composed with one ``sample'' argument $k_1,k_2 \in [K]^2$, which take as argument a vector of the form $(\bs{x},\bs{z})$.

\item Finally, $\psi_4$ computes the utility associated with each of the $\kappa T$ inputted product and customer attribute vector of the form $(\bs{x},\bs{z})$. By invoking Lemma~\ref{thm-NN}, where we note that no distributional assumption is placed on how the inputs $(\bs{x},\bs{z})$ are generated, we have 
\begin{eqnarray} \label{step4}
R(\psi_1 \circ S') = O\left(\sqrt{\frac{\log(d)}{\kappa T}} \cdot (2M)^{\ell}\right) \ .
\end{eqnarray}
\end{enumerate}
By combining inequalities~\eqref{step1}-\eqref{step4}, we conclude that
\[
R( \psi_2\circ\psi_3 \circ \psi_4 \circ S) = O\left(\sqrt{{\frac{\kappa^3\log(d)}{T}}}  \cdot e^{2  M} (2M)^{\ell}\right) \ .
\]

 \hfill \halmos
 \endproof

\subsection{Proof of Proposition~\ref{prop:representation}} \label{app:prop:representation}

Fix the ground truth model $\pi = \pi^{{\sf RUMnet}}_N$, $T = \lceil\max\{\frac{c_1^2}{\eps^2}\kappa^3 \log(d) e^{4M} (2M)^{2\ell}, 128(8M+\log \kappa)^2\ln\frac{4}{\delta}\}\rceil$, and $K' = \lceil \frac{1}{2\eps^2} \cdot \log \delta \cdot (\kappa e^{2M})^2 \log(T) \rceil$.  By precisely the same line of argumentation as in the proof of Proposition~\ref{prop:gen-error},  the following inequality holds for every $\bar{N} \in {\cal N}^{\bs{d}}(K',\Theta^{\ell,w}_M)$, with probability $1-\delta$,
\begin{eqnarray}\label{ineq:conc:1}
	L_{\cal D}^{\rm true} \left( \pi^{\sf RUMnet}_{\bar{N}} \right) &\leq& L_S \left( \pi^{\sf RUMnet}_{\bar{N}} \right) + c_1\sqrt{{\frac{\kappa^3\log(d)}{T}}}  \cdot e^{2  M} (2M)^{\ell}   + (8M+\log \kappa) \sqrt{\frac{2 \ln(4/\delta)}{T}} \nonumber \\ &\leq & L_S \left( \pi^{\sf RUMnet}_{\bar{N}} \right) + \eps  \ 
\end{eqnarray}
In Proposition~\ref{prop:gen-error}, we established a similar inequality with respect the ERM estimate; however, this inequality holds for any fixed RUMnet architecture in our hypothesis class; see \citet[Thm.~26.5]{shalev2014understanding}.

Now, by applying Hoeffding's inequality with respect to the random experiment $S\sim {\cal D}^T$, while using the fact that  $\prpartwo{S \sim {\cal D}^T}{|\log(\pi^{\sf RUMnet}_{N}(\bs{s}))| \leq \log(\kappa e^{2M})} = 1$  by Claim~\ref{clm:pmin}, we have:
\begin{eqnarray}\label{ineq:conc:2}
	\prtwo{S\sim {\cal D}^T}{L_S \left( \pi^{\sf RUMnet}_{N} \right) - L_{\cal D}^{\rm true} \left( \pi^{\sf RUMnet}_{N} \right)  >\eps} \leq e^{\frac{-2 \eps^2T}{(\log(\kappa e^{2M}))^2}} \ .
\end{eqnarray}
Next, for any fixed realization $\hat{S}$ of $S$, we can bound the errors in the choice probabilities computed by $N'$ relative to $N$ using  Hoeffding's inequality with respect to the $K'$ i.i.d. samples over the unobserved attributes $\{N_{\epsilon_{k_1}}(\cdot)\}_{k_1=1}^{K}$ and $\{N_{\nu_{k_2}}(\cdot)\}_{k_2=1}^{K}$. As a result, using the union bound over the $T$ events in $\hat{S}$, we obtain
\begin{eqnarray*}
\prtwo{N'}{\left. L_{S} \left( \pi^{\sf RUMnet}_{N'} \right) > L_S \left( \pi^{\sf RUMnet}_{N} \right) + \eps \right| S = \hat{S}} \leq Te^{-\frac{2\eps^2K'}{(\kappa e^{2M})^2}}\ .
\end{eqnarray*}
By plugging the definition of $K'$, and by taking the expectation with respect to $S\sim {\cal D}^T$, we obtain
\begin{eqnarray}\label{ineq:conc:3}
\prtwo{N'}{ L_{S} \left( \pi^{\sf RUMnet}_{N'} \right) > L_S \left( \pi^{\sf RUMnet}_{N} \right) + \eps } \leq \delta \ .
\end{eqnarray}
The union bound with respect to the events of inequalities~\eqref{ineq:conc:1}-\eqref{ineq:conc:3} implies that, with probability $1-3\delta$,
\begin{eqnarray*}
	L_{\cal D}^{\rm true} \left( \pi^{\sf RUMnet}_{N'} \right) \leq L_S \left( \pi^{\sf RUMnet}_{N'} \right) + \eps \leq  L_S \left( \pi^{\sf RUMnet}_{N} \right) +2\eps \leq 	L_{\cal D}^{\rm true} \left( \pi^{\sf RUMnet}_{N} \right)+ 3\eps  \ ,
\end{eqnarray*}
where the first inequality proceeds from~\eqref{ineq:conc:1} with the instantiation $\bar{N} = N'$, the second inequality follows from~\eqref{ineq:conc:3}, and the last inequality  is a direct consequence of~\eqref{ineq:conc:2}.

By rearranging the latter inequality, we obtain the following:
\begin{eqnarray*}
	3\eps &\geq & L_{\cal D}^{\rm true}\left( \pi^{\sf RUMnet}_{N'} \right) -  	L_{\cal D}^{\rm true} \left( \pi^{\sf RUMnet}_{N} \right) \\
	&=& \extwo{(\bs{z},A)\sim {\cal D}}{\sum_{\bs{y} \in A} \pi^{\sf RUMnet}_{N}(\bs{y},\bs{z},A)	 \left(\log(\pi^{\sf RUMnet}_{N}(\bs{y},\bs{z},A)) - \log(\pi^{\sf RUMnet}_{N'}(\bs{y},\bs{z},A)	)\right) } \\
	&=&\extwo{(\bs{z},A)\sim {\cal D}}{{\sf KL} \left( \left. \pi^{{\sf RUMnet}}_{N}(\cdot,\bs{z},A)\right| \pi^{{\sf RUMnet}}_{N'}(\cdot,\bs{z},A)) \right)} \ .
\end{eqnarray*}
 
Note that, in the above proof, the sampling with respect to $S \sim {\cal D}^T$ is unnecessary; we can establish the same result as long as there exists a fixed sample $S$ such that inequalities~\eqref{ineq:conc:1}-\eqref{ineq:conc:2} hold. Nonetheless, our proof argument can be adapted to establish an {\em algorithmic version} of Proposition~\ref{prop:representation}. Specifically, with respect to the hypothesis class ${\cal N}^{\bs{d}}(K',\Theta^{\ell,w}_M)$, the ERM estimate $\pi^{\sf{ERM}}_S$  on a random sample $S \sim {\cal D}^T$ achieves a small total learning error with high probability.

\newpage

\section{Additional material for Section~\ref{sec:synthetic}} \label{app:synthetics}

\subsection{Identifiability and recovery} \label{app:uniqueness}

It is well-known that the representation uniqueness breaks even for simple LC-MNL models; see, e.g., \cite{ammar2014s}. Nevertheless, since the RUMnet architecture directly models the utility, we can numerically study the extent to which the fitted RUMnets recover the utility functions. Indeed, recall that in our architecture, there is a unique neural network $\mathcal{N}_u$ that captures the utility function (see Figure~\ref{fig:rumnet}). Here, we conduct an analysis for whether or not the fitted RUMnets can recover the true utility function in Settings 2 and 3 of the synthetic experiments in Section 5.

\begin{enumerate}
\item \textbf{Setting~2: non-linear utility.} Recall that in Setting~2, the ground truth model is a logit model with non-linear utility. In this case, there is a single sample and we can compute the output of the utility neural network $\mathcal{N}_u$. Figure~\ref{fig:utility_2} shows a scatter plot of the recovered utility as a function of the ground truth utility for the DeepMNL model with increasing complexity sizes. We can clearly see that as we increase the complexity of RUMnets, determined by the parameters $(\ell,w)$, the correlation between the predicted and ground truth utility increases, indicating that the ground truth utility is recovered as we consider more flexible specifications of the utility function. For instance, when regressing the predicted utility on the ground truth utility, we obtain, when $(\ell,w)=(2,5)$, a value of $R^2 = 0.98$ and a slope of 1.11.

\vspace{0.5cm}

\newcommand{\utilitysmall}{
( 0.2676768836317993 , 4.071454 )
( -1.3910324681900308 , 1.2703459 )
( -8.937480172661521 , -3.31874 )
( 5.749442652414051 , 7.267362 )
( 6.891810343731568 , 7.193997 )
( 6.507949783217706 , 6.3342013 )
( 6.144103035384285 , 4.8107615 )
( 6.816424821888014 , 4.3496394 )
( 3.4854345622751826 , 3.559036 )
( 1.0361119648962416 , 3.643115 )
( 8.054097235893583 , 5.8241663 )
( 4.670134201908244 , 6.415525 )
( 8.781796981838761 , 7.3830423 )
( 4.659766354476239 , 5.5818777 )
( -2.2917948105728136 , 1.1594853 )
( 2.126971429940478 , 2.4794145 )
( 0.035574980682075985 , 1.7808688 )
( 3.2741180794561346 , 3.8064742 )
( 5.417361712407309 , 6.0290966 )
( 5.706608462854018 , 4.1777744 )
( 7.7008976443924855 , 7.3362637 )
( 5.890854913827388 , 6.721058 )
( 4.635860617305445 , 3.7196808 )
( -0.00633232290850827 , -1.2861042 )
( 5.330455374610794 , 4.682168 )
( 2.241714797461916 , 2.995245 )
( -0.10010578191149244 , 2.4520526 )
( 1.4990409379348595 , 3.8657842 )
( 5.825251590519721 , 7.0551147 )
( 2.246882906138029 , 4.0090733 )
( 1.5203160341951762 , 2.4570928 )
( 0.34961138183561546 , 1.883223 )
( 1.1973893585740374 , 3.3534052 )
( 6.586282621233469 , 5.49384 )
( 8.264655767053618 , 6.416314 )
( 4.461134747589333 , 3.251299 )
( 7.262721563929104 , 6.7706885 )
( 5.013262208406412 , 7.3289995 )
( 2.3393496127544138 , 3.012609 )
( 4.91004641749204 , 3.3952198 )
( 6.990328989320571 , 5.4812927 )
( -1.2198144913926594 , 0.8061421 )
( -9.042410403820728 , -1.6560092 )
( 5.035935911046796 , 6.1685557 )
( 8.149293242002262 , 5.836069 )
( 3.5352994717246347 , 3.1880074 )
( 7.789463143742424 , 6.528879 )
( -9.67681802684643 , -3.3626184 )
( 2.990426266918986 , 3.9355602 )
( 4.770975920388231 , 4.3656282 )
( 0.4107177377514951 , 3.6286952 )
( -0.24218668137076726 , 1.9533063 )
( 7.434055839571792 , 8.0560665 )
( 5.300797028704794 , 3.7028036 )
( 3.987354219890465 , 4.8054647 )
( 6.943160816835565 , 7.0411444 )
( -0.24001589099959536 , -0.42731625 )
( 7.573566576066656 , 5.869421 )
( 8.52154058081566 , 5.547023 )
( 6.870987175435958 , 7.9886265 )
( 6.268242745235366 , 4.8284945 )
( -0.18443829083197327 , 0.4513445 )
( 0.605826515988406 , 4.352774 )
( 3.0737328698228144 , 3.7886376 )
( 3.3873604118493548 , 2.8994393 )
( 0.8845794866531771 , 2.385427 )
( 0.094090850472732 , 0.3591693 )
( 6.190662386314384 , 6.4758787 )
( -1.9781119101886349 , 0.7964649 )
( 4.286964438903956 , 5.055171 )
( 2.0371468598635616 , 3.7294884 )
( 4.835938426146559 , 3.4806519 )
( 7.575331976011167 , 8.048035 )
( 6.028099789920599 , 6.76154 )
( -2.1226508009386764 , 0.17987776 )
( 7.7974337190226715 , 7.6931715 )
( 0.7293452492793208 , 2.455689 )
( 2.0177215776243 , 2.553115 )
( 1.1658940833142832 , 2.3760138 )
( 8.514987551249266 , 7.166094 )
( 6.293848514183121 , 4.4011593 )
( 4.8429008940369895 , 7.3863206 )
( 7.405281975268689 , 8.06229 )
( 5.096440043547555 , 3.5489078 )
( 2.129283018049639 , 2.7094212 )
( 2.4003631079296395 , 5.4793468 )
( 7.754607914338543 , 8.192203 )
( 5.450374435147729 , 4.0161567 )
( 0.09806241871263455 , 2.2044759 )
( 3.4443959511103035 , 3.2039452 )
( 0.4045072959739514 , 1.8678353 )
( 3.9812004502376612 , 6.178751 )
( 2.670948639898037 , 4.2704854 )
( -6.202579021048489 , -3.2383347 )
( 0.6106154427116954 , 3.896394 )
( 5.6953031419881786 , 4.078265 )
( 6.311409638786939 , 4.6455746 )
( 7.4430003051860965 , 5.567585 )
( 1.0278914073269831 , 3.0862854 )
( -0.8911953998242435 , 0.7275804 )
( 5.06661092130334 , 3.76054 )
( 5.855617888085803 , 4.3241763 )
( 5.836307425775854 , 5.898607 )
( 4.947504092463888 , 5.3034015 )
( 1.5095068007629253 , 5.1549397 )
( 7.104821931785835 , 7.147853 )
( 0.47792012547422746 , 4.566828 )
( 0.06945965338814708 , 1.4564412 )
( 3.6280763677812384 , 2.8765721 )
( -1.5632838268450429 , -1.9889646 )
( 6.479172789397314 , 5.323649 )
( 2.52593145092612 , 4.0418124 )
( 5.633289194155285 , 4.0751753 )
( 2.1792875490410504 , 2.7792406 )
( 7.256934576450996 , 5.987919 )
( 7.223554423484355 , 6.3323593 )
( 0.011986661829006096 , 1.6486053 )
( 3.808543283570783 , 3.443954 )
( 1.5875815858092082 , 3.2954726 )
( -3.5051194536708525 , 0.42425013 )
( -2.963390022967295 , -1.983412 )
( 3.7186189185322487 , 5.077574 )
( 7.253743834414072 , 5.2400684 )
( 2.475002539405235 , 2.6808667 )
( 3.750759810389197 , 5.472553 )
( 3.3907067245023956 , 3.1678686 )
( 5.57106324599616 , 5.5760155 )
( 2.2048408643683626 , 4.5738983 )
( 3.7022400786329914 , 6.426092 )
( 2.480902656732841 , 2.739596 )
( 4.541787117090942 , 4.3735156 )
( 0.9503106433347589 , 3.5785482 )
( 4.277398412117946 , 3.7622943 )
( 3.672379100859513 , 4.9038916 )
( -3.9697186006714063 , 0.3593359 )
( 6.966574021674121 , 7.323637 )
( 4.949016346950014 , 3.853844 )
( -0.7376350298710882 , 2.6173995 )
( 7.253713327485091 , 5.503557 )
( 6.208593021962292 , 5.767669 )
( -4.113354192751081 , -2.3546891 )
( -1.9659081150789117 , 0.07700324 )
( 2.3583639629056545 , 4.1951256 )
( -0.6644182742033116 , 1.4399104 )
( 0.39096122589377647 , 1.9179225 )
( 7.7599668616618445 , 5.507905 )
( -0.8202870429949162 , 2.381721 )
( 5.440341521149703 , 5.2287917 )
( 0.4619511916524053 , 2.7558813 )
( 4.26885158670816 , 6.3671994 )
( 1.2800542348565491 , 2.426196 )
( 7.258918383760248 , 6.8648705 )
( 2.2312533517496513 , 5.017441 )
( 3.7081084510552946 , 2.9608603 )
( 5.201975406018853 , 7.2336183 )
( 1.522237457318805 , 3.7124062 )
( 2.941756254620538 , 2.9132552 )
( 5.506838240010387 , 6.333622 )
( -1.4293354467484038 , 0.86434317 )
( 2.098130895040582 , 2.733364 )
( 3.613635087880109 , 3.4820595 )
( 5.213500941601739 , 5.231406 )
( 4.2468317966711195 , 5.03775 )
( 1.9171389154471867 , 2.8533735 )
( 8.566633987378557 , 6.7145157 )
( -1.6203647536920007 , 1.1932149 )
( 1.7110126396972825 , 2.3692508 )
( 2.217116193522877 , 2.987823 )
( -2.4407617326650004 , -0.49977875 )
( -4.081875150783875 , -1.2698126 )
( 8.120655803296842 , 6.7642813 )
( 5.257586618995882 , 4.722788 )
( 1.2475351367823793 , 2.8072271 )
( 0.4016883557141555 , 2.1189609 )
( -5.729736973729846 , -3.1351473 )
( 0.5386354175986505 , 2.4382315 )
( -2.0232284892451826 , 0.64145994 )
( 8.420730543716127 , 7.803833 )
( 5.474797958984503 , 7.192106 )
( 8.806161947185453 , 6.7435465 )
( 2.928874565101916 , 3.6531792 )
( 0.9200329594311141 , 2.1389666 )
( 3.922188805469382 , 3.17061 )
( 8.180835293751983 , 5.5336084 )
( 3.7762219938519475 , 3.3683472 )
( 0.03709938490052478 , 1.1288234 )
( 3.3614352345594085 , 4.429637 )
( 2.5538050932817917 , 3.1628885 )
( 7.441796903035247 , 5.94713 )
( 4.486654636133456 , 4.5058584 )
( 0.2983752851087048 , 1.9638487 )
( 1.15412745563486 , 2.4072134 )
( 6.305311041848304 , 4.6434917 )
( -1.7354562066284753 , 1.2273579 )
( 8.32989256385774 , 5.745325 )
( 1.7851704526575878 , 3.632884 )
( 0.07424073045536095 , 1.5294778 )
( 8.182660504829279 , 6.6318083 )
( 2.8106749017164 , 4.4197493 )
( -0.3523877366456328 , 2.05028 )
( 8.21221844648461 , 7.0698104 )
( 1.6086948059849768 , 3.0371413 )
( 7.25954304084537 , 6.1458445 )
( 6.370165152554241 , 8.117838 )
( -0.42244098051768475 , 2.1932244 )
( 7.782918143338263 , 6.277953 )
( 1.283513279431482 , 2.3442717 )
( 5.108275994125956 , 4.2610245 )
( 6.69256068898618 , 5.163224 )
( 7.328771651005318 , 4.8673353 )
( 1.5180334730464906 , 3.12117 )
( 7.419994131976285 , 4.902232 )
( 6.025271582932866 , 5.286966 )
( 5.336528575210153 , 4.6728964 )
( 0.10081572525438176 , 1.7628613 )
( -3.1164905594823313 , 0.99884033 )
( -0.2109976594997009 , 3.6904569 )
( -0.24412299644987043 , 2.8194818 )
( 5.37884408193659 , 4.7049336 )
( -0.7494958829239231 , -0.8306086 )
( 2.388313142846553 , 2.6467414 )
( 1.7846513350581148 , 2.550869 )
( 5.440235792480619 , 4.2893796 )
( 0.558048374992988 , 3.7865808 )
( 8.580174717955835 , 5.88777 )
( 1.2438049862221776 , 3.171742 )
( 3.59170023561102 , 6.2581744 )
( 5.337438259524756 , 4.878681 )
( 2.4060761878287886 , 4.152919 )
( 4.446129804915182 , 4.1161737 )
( 7.235815168904421 , 5.855008 )
( 9.405199353232666 , 7.782976 )
( 0.8390495887795444 , 2.1865683 )
( 4.442173884515984 , 6.759357 )
( 6.0107328602449535 , 4.646128 )
( 2.6106269650590606 , 3.7358594 )
( 6.29003312834007 , 5.997756 )
( 1.2915725265609144 , 2.5804017 )
( 4.7527004169148075 , 6.7070236 )
( 7.475250664593411 , 8.597613 )
( 8.536817954754035 , 5.698232 )
( 2.919248408522611 , 4.2983856 )
( 0.05217678760017841 , 1.9543746 )
( 0.2509566301280288 , 1.8282535 )
( 2.5194673806306604 , 5.957506 )
( 7.719820091929492 , 7.110409 )
( 1.522949998969045 , 2.3767142 )
( 1.825643198490536 , 3.4814963 )
( 0.28426180525791667 , 2.5600135 )
( 2.43102895903682 , 4.196618 )
( 5.047594735850084 , 3.8339682 )
( 8.916845061407068 , 6.461176 )
( 6.945985617873937 , 6.8447347 )
( -0.9732080040970796 , 1.4992819 )
( 2.455856685016883 , 5.4225945 )
( -3.037478883774261 , -0.2408135 )
( 3.539002364928046 , 2.9090934 )
( 4.475667236002017 , 6.1947336 )
( 1.2995269554059536 , 2.4259782 )
( 4.335829618557123 , 6.836146 )
( -0.9387249205498538 , 1.7148952 )
( 4.162888804645874 , 3.453165 )
( -0.003695436481622849 , 1.7176151 )
( 2.545535423871721 , 3.1850595 )
( -11.117845589056639 , -4.131445 )
( 7.574677912946466 , 6.2050877 )
( 2.746049955938565 , 6.2119184 )
( -1.242157795082518 , 0.9921558 )
( -0.8618651018022133 , 1.8842164 )
( 4.856865927717134 , 7.069833 )
( -1.8399085973828004 , 1.1985893 )
( 6.9227052400387565 , 5.3004045 )
( 0.38739077290497004 , 3.3841388 )
( 0.3473328252936323 , 1.5125456 )
( 6.914238597525393 , 4.522581 )
( 0.24403380859685964 , 1.9712114 )
( 3.8628310840301365 , 3.3313155 )
( 8.548960846397414 , 5.958052 )
( 6.855049222257766 , 5.9893937 )
( 3.9817606967853756 , 6.6619005 )
( 5.7994411977760025 , 7.183496 )
( 1.2810910751368496 , 3.5745506 )
( 0.03908917180067584 , 1.0854094 )
( 3.4815359765055858 , 4.7165627 )
( 2.9721499089349046 , 5.8160324 )
( -0.696261072567994 , 1.4437811 )
( 5.513782393199161 , 5.421341 )
( 6.267688100680255 , 7.624753 )
( 2.0418777055653847 , 5.1444573 )
( -4.77565744811427 , 0.2913251 )
( 6.767686735179783 , 5.304855 )
( -1.3220716139598245 , 1.032032 )
( 6.820090380680488 , 4.2428427 )
( -2.677094435890613 , -1.6415333 )
( 1.3192856244223274 , 3.7186258 )
( 7.267337074326019 , 4.9141235 )
( -0.40808075210172867 , 0.21179557 )
( 8.206156740820592 , 7.134549 )
( 0.6322469088198712 , 2.172872 )
( 8.021103795285306 , 5.9396777 )
( 4.973118237234678 , 3.7253723 )
( 4.448385923540285 , 6.5046864 )
( -0.02199770347320107 , 1.8813531 )
( 1.7179096344313098 , 2.62507 )
( 0.832174931794368 , 4.413413 )
( 2.1511011011678045 , 4.1337967 )
( 5.970377987900401 , 4.408489 )
( 4.760904356287516 , 3.722146 )
( 3.231979654132901 , 4.362109 )
( 4.5376279352144575 , 5.163169 )
( 6.427896470127056 , 6.8060055 )
( 4.876203817844106 , 4.2744594 )
( -0.33290786170029263 , 2.3492818 )
( 0.8043722565572329 , 2.5204515 )
( 3.1019413368618833 , 3.7530901 )
( 2.7298630938542465 , 3.2695527 )
( 4.853352388767605 , 5.670052 )
( -2.5208545165025202 , -0.82279587 )
( 5.797529665575164 , 4.1166544 )
( 2.5760889487575307 , 3.0022507 )
( -0.18972927303844553 , 0.3335318 )
( 4.728681012248585 , 3.7409616 )
( -0.16517029471887434 , 2.0722208 )
( 6.15410514879552 , 4.293994 )
( 3.721256158749855 , 6.417212 )
( 3.3449144829207134 , 2.9977064 )
( 1.0373104061855787 , 3.2288136 )
( 4.585645849657365 , 4.9434557 )
( 1.1159506831087205 , 3.1397347 )
( 2.2780916114170546 , 3.3200788 )
( 4.474224694385321 , 6.83804 )
( 0.0953871201941191 , 0.4986196 )
( 3.858147853759852 , 3.055499 )
( 1.441823250724911 , 2.6951504 )
( 4.027676470627191 , 4.842287 )
( 7.137802280153975 , 6.0489807 )
( 5.533865865429053 , 3.601626 )
( 8.467344497627119 , 5.562744 )
( 3.1348093566863477 , 2.9116879 )
( 3.944689542386139 , 3.2805586 )
( 5.700390937826158 , 3.9841528 )
( 3.8001940650483252 , 3.1074781 )
( 4.2158864341280635 , 3.8478637 )
( 6.948398515323394 , 8.513425 )
( 6.06421768247725 , 4.149769 )
( 1.3968320864091814 , 2.6073663 )
( 7.236784748215651 , 6.7164564 )
( 3.998632140656552 , 3.911101 )
( 1.4733189438740606 , 3.5651505 )
( 4.182629082116691 , 3.7694879 )
( 5.7358266649504595 , 4.2913156 )
( 5.222875135796961 , 6.3474464 )
( 6.962177940908919 , 8.396099 )
( 2.0928867524727677 , 3.068779 )
( 4.339612486297609 , 4.9663563 )
( 2.1745903996841305 , 4.8357334 )
( 2.481148703852342 , 2.4669104 )
( -0.09300515726508696 , 1.6413331 )
( -0.1604761103345081 , 3.614302 )
( 8.839492147594768 , 6.856406 )
( 7.505729233239036 , 5.2124567 )
( 8.662737012893377 , 5.999463 )
( 3.503210151678705 , 5.9689565 )
( 2.3042051975153037 , 5.2884626 )
( 2.4650324492555873 , 2.9323945 )
( 5.2432588467838865 , 3.9548516 )
( 1.815629622613181 , 2.6601279 )
( 7.956636241117511 , 7.8964744 )
( 5.811371550234936 , 4.9400682 )
( -5.842394848267936 , -0.9759631 )
( 4.712009279143475 , 4.5920415 )
( -3.840615927891837 , 0.6084962 )
( -0.1668979307594185 , 3.4503076 )
( 3.5901660035451846 , 5.457752 )
( -0.7807913366435706 , -0.12477207 )
( 4.16965298960069 , 6.850938 )
( 5.069203585586157 , 4.447929 )
( 4.476676434119617 , 4.01138 )
( 4.946249997006223 , 6.141063 )
( 2.0668907741434044 , 4.00735 )
( 1.385713983881462 , 2.7386467 )
( 6.178963034821699 , 7.931977 )
( 4.283003167150721 , 5.1116285 )
( 6.835222485552192 , 7.7824802 )
( 7.0187894237663855 , 6.800687 )
( 3.0172747537474227 , 5.2032714 )
( 2.343161289625655 , 3.7016766 )
( 7.086178868707142 , 5.589501 )
( -3.4736069594393433 , -1.2493637 )
( 6.262241392965121 , 6.7714834 )
( 4.431433908418339 , 3.3224363 )
( 3.664632222983875 , 3.8209372 )
( 3.599714704665939 , 3.6553285 )
( 0.7527527328491762 , 3.1592705 )
( -0.013627553637451939 , 2.0497634 )
( 0.45242766180326965 , 2.0836256 )
( 6.360597442291068 , 5.8405795 )
( 4.722532788122859 , 4.1097965 )
( 7.428796554877485 , 7.9625883 )
( 4.686064474168175 , 5.0643826 )
( 7.144478209798695 , 5.2079134 )
( -0.5044191987529233 , 1.7714853 )
( 5.241235530229169 , 5.640241 )
( 0.6077458552489645 , 2.426997 )
( -1.7494570674221444 , 0.71230197 )
( -0.3085359025614005 , 1.4439312 )
( 3.597309139508509 , 4.5633755 )
( 6.079418824973439 , 4.838238 )
( 2.664052273559679 , 3.4800096 )
( 0.09188721535401179 , 1.7341127 )
( 3.2856718952302266 , 5.5298405 )
( 5.771290700997784 , 6.3356986 )
( 4.0357030513806835 , 3.8058634 )
( 4.336833643644864 , 3.4420848 )
( 1.9426217994612878 , 2.6329045 )
( 6.624054400583452 , 7.167512 )
( 1.8090750599968404 , 3.641635 )
( -0.07226162260812918 , 1.7552876 )
( 5.417246007865398 , 6.3225245 )
( 6.219867833521878 , 6.957234 )
( 5.265101677378989 , 7.338215 )
( 7.043692292311279 , 6.2843857 )
( 4.457690913728479 , 4.5954275 )
( 2.571705532424465 , 4.046466 )
( 3.018823359926941 , 5.512781 )
( 4.542100392209846 , 5.331935 )
( 0.013201218670070403 , 3.1481504 )
( 6.97731431598188 , 8.101243 )
( 2.7129700103713783 , 3.1410549 )
( 4.790215074046861 , 4.982917 )
( -0.6778653526631935 , 1.340801 )
( 7.203968590583353 , 4.699556 )
( -0.16742374699072693 , 1.0441445 )
( -0.9061252741010206 , 1.3932316 )
( 0.2170116340411683 , 2.00034 )
( 5.941813818408002 , 6.0021033 )
( 6.164484284048667 , 5.5768757 )
( 4.394736677669951 , 3.8717334 )
( 2.5876744961671 , 3.3365726 )
( 7.745287333339578 , 6.978314 )
( 6.464975950205047 , 7.541831 )
( 5.221598201707392 , 6.3996053 )
( 3.2746376319032793 , 4.647868 )
( 4.322557968141274 , 3.2410946 )
( 0.14107125549394084 , 1.7826772 )
( 3.981980763973077 , 4.239787 )
( -1.2359067438271043 , 0.51364326 )
( 3.3474347106638858 , 3.3331466 )
( 1.1060723748543433 , 2.040306 )
( 7.6114767610079115 , 8.006986 )
( 6.581000098121152 , 7.9842186 )
( 7.238503493573845 , 5.41924 )
( 4.269442086174138 , 5.117396 )
( 7.054061471324001 , 5.1840434 )
( -1.461811525479933 , 0.11590958 )
( 7.0034197634623006 , 5.779687 )
( 5.884257892194496 , 6.5690875 )
( 6.5217918401382144 , 8.227749 )
( 5.58326004119045 , 6.863649 )
( 0.29422834658901564 , 1.925746 )
( -0.12981704597109978 , 2.7280793 )
( 0.038041830943694066 , 1.6223831 )
( 0.40881063215363245 , 2.1893184 )
( 3.426542967294579 , 5.38407 )
( -0.9484909566289238 , 0.07422936 )
( 0.06804404806036612 , 1.0964643 )
( -1.0145524673680444 , 1.1701368 )
( 1.5015880827261952 , 2.6830406 )
( 2.8573052759412034 , 2.7322516 )
( 9.219458452935616 , 7.43336 )
( 0.5368244143942041 , 3.4621665 )
( 4.355831634649004 , 4.1927123 )
( 0.18293071013046092 , 1.9655163 )
( 3.196250422523403 , 3.816267 )
( 4.345084873172813 , 3.0403953 )
( 2.520467966738127 , 2.7848659 )
( 7.022303363600746 , 7.555588 )
( 3.241959619206911 , 2.8694892 )
( 2.260551961518866 , 3.1574326 )
( 7.599870669057844 , 7.8829665 )
( 3.1206821892354983 , 3.0728307 )
( 3.8161913075583547 , 5.161458 )
( 5.894939809451715 , 6.2929006 )
( 3.7193856645207655 , 3.7646852 )
( 6.965713092123607 , 5.5589566 )
( 5.631698208279772 , 5.5770497 )
( -0.3315919887929628 , 1.1206813 )
( 4.994235047474497 , 3.5742884 )
( 5.897214748924215 , 4.645791 )
( 4.484277854254555 , 6.2969565 )
( -1.861480115814844 , -0.42466855 )
( 8.100808604287657 , 5.692565 )
( 5.561990033500462 , 4.24537 )
( 1.1898974931511623 , 3.3743627 )
( -6.230137864499565 , -1.5236847 )
( 7.91458769948004 , 6.691156 )
( 6.562499364760522 , 7.6653037 )
( 7.924020661464057 , 7.0028963 )
( 3.4657567696548126 , 6.6852174 )
( -1.0131253469182369 , 1.5027184 )
( 8.153918173337296 , 5.6559973 )
( 8.011330087641998 , 5.9733324 )
( 5.4217228848061785 , 8.004421 )
( 6.301385079089718 , 5.341861 )
( 0.8505030277778488 , 4.9746404 )
( -0.7859233667651228 , 1.997992 )
( -1.0839744909751343 , 0.91128635 )
( -0.9320468500597869 , 1.1478071 )
( -1.782618711091846 , 0.8083806 )
( 4.50623006965518 , 4.167664 )
( 5.698790489251194 , 4.1521688 )
( 5.257528347316875 , 5.8720965 )
( -0.313603203271988 , 0.96838605 )
( 7.771614177360129 , 5.10975 )
( 1.9937111460143533 , 2.4499063 )
( 5.445762247032819 , 4.109104 )
( 5.022080931346767 , 4.2087817 )
( 0.19951050924736347 , 1.725996 )
( 6.775356556684504 , 5.7790008 )
( 0.3862617703926645 , 2.1366892 )
( 8.233215513649972 , 5.681367 )
( -0.4269751437497818 , 2.9735675 )
( 0.027307365071082 , 1.5694093 )
( -1.1343699559232412 , -1.5149075 )
( 0.4832239741317146 , 1.926414 )
( 0.9340962959134025 , 2.0264292 )
( 2.627085270190526 , 4.525205 )
( 5.379976084727163 , 6.258616 )
( 0.4772875052857728 , 2.020906 )
( 0.08905097931008821 , 3.0418382 )
( 7.207069468110452 , 5.368025 )
( 4.931528302241455 , 5.1295214 )
( 1.8369539968596635 , 2.5689282 )
( 2.909288634227189 , 5.889532 )
( -0.8457344558119377 , 1.472794 )
( 3.1456431646785417 , 5.928146 )
( 8.778435772379716 , 6.9043894 )
( 7.689663448683343 , 6.524446 )
( 3.0935712462038794 , 3.0575414 )
( 7.061850623282417 , 8.2836685 )
( 6.445507296071188 , 6.5880585 )
( 5.407126101815223 , 5.881148 )
( 7.2862028426000105 , 5.2431917 )
( 2.594359954931955 , 3.6955018 )
( 1.0909649031124928 , 5.1591344 )
( 6.142807421492974 , 4.5167284 )
( 1.1947569505749391 , 3.1250641 )
( 4.582683984229626 , 3.35111 )
( 5.513515616526465 , 5.245219 )
( 0.15173654114674306 , 1.9802918 )
( 4.361131703751097 , 6.532854 )
( 3.4214726266619833 , 5.1004996 )
( 4.589568719346355 , 5.4043612 )
( 0.671253093898023 , 3.2204444 )
( 7.514806194145972 , 5.6631436 )
( 7.124276194287938 , 6.5984645 )
( 5.585961677580761 , 6.8195934 )
( 3.8739052260086866 , 3.7430608 )
( -1.6586858297216072 , 0.9564228 )
( 4.135443786732309 , 3.7056768 )
( 3.6510572089128557 , 3.177926 )
( -0.9856076642668015 , 1.3717468 )
( 0.7860755214217638 , 2.137896 )
( -2.9881436934506818 , -0.3287487 )
( 0.2080221241290936 , 2.1429884 )
( -0.7275721409457951 , -0.50317705 )
( 7.819655320345916 , 6.5945954 )
( 3.3035475306329563 , 5.2396836 )
( -1.343147869482733 , 1.7922807 )
( 6.480415669700959 , 4.4991455 )
( 6.764791546322475 , 6.4923286 )
( 3.592512331485088 , 5.0525427 )
( 3.2411846401196414 , 3.2183197 )
( 1.3966070135792705 , 2.001822 )
( 8.098102208474536 , 5.418981 )
( 2.61185255324428 , 3.675029 )
( 1.3985140227052961 , 3.9573708 )
( 0.7805510135085597 , 2.3874798 )
( 2.482778303332326 , 2.668407 )
( 6.944341111423071 , 6.287236 )
( 5.634343785250352 , 4.337555 )
( 6.034258713263885 , 7.4735622 )
( 3.996759079656673 , 5.357402 )
( 3.009361695483859 , 2.9117951 )
( 3.15542841589615 , 6.151211 )
( 3.969255675366232 , 4.7183704 )
( 7.277163323864622 , 7.567322 )
( 6.427085901907864 , 4.429269 )
( 8.162753127607061 , 7.7165556 )
( 3.698517400206751 , 2.9587502 )
( 5.767646941102387 , 5.166655 )
( 0.8594074637564966 , 2.316609 )
( 8.580063358466644 , 6.373726 )
( 3.315873240191739 , 3.4131246 )
( 8.283795573865357 , 6.884952 )
( 1.6510101844949636 , 2.6710458 )
( 5.740889465731581 , 4.496002 )
( 6.008348913846862 , 5.586949 )
( 4.851198937158236 , 5.9045725 )
( -4.333680434259447 , -1.0714796 )
( 5.269858912259353 , 3.7768254 )
( 3.1892859827545172 , 3.5285277 )
( 5.609566610570077 , 3.6737332 )
( -0.42300459199689433 , 1.4918095 )
( -5.279697324823457 , -0.9204483 )
( 1.1402898900273768 , 2.403353 )
( 6.633971572504234 , 5.819461 )
( 5.639650980669303 , 4.2067423 )
( 1.575106667189782 , 2.6312196 )
( 8.024820045710714 , 7.107975 )
( 7.536613539384735 , 7.481207 )
( 8.668727950618546 , 7.7694955 )
( -2.9997699370174207 , 0.5231266 )
( 4.185198638723766 , 5.5316625 )
( 4.855774186888225 , 6.9607716 )
( 6.764028266025655 , 4.3962736 )
( 7.197074017778772 , 7.3210096 )
( 0.2972010925862616 , 2.0360532 )
( 5.728988040120108 , 6.305094 )
( 4.694584390778518 , 3.7923696 )
( 2.0825091754802116 , 2.9310343 )
( -0.27782473790486817 , 2.7745218 )
( 3.4407135099299007 , 4.236375 )
( 0.27044797328951997 , 1.9225578 )
( -0.10305955987728843 , 2.2018495 )
( 5.937450646185493 , 8.408633 )
( 2.245954796095088 , 4.9332333 )
( 6.468275552760025 , 5.0728645 )
( 0.10148684497548288 , 1.9280462 )
( 3.962822467698664 , 4.5411367 )
( -8.575570813006808 , -0.51136446 )
( 7.432015529573616 , 5.6309304 )
( 7.15320559444487 , 5.5680666 )
( 6.580409678493231 , 5.2431383 )
( 2.7089213812927486 , 4.11421 )
( 7.494245224364793 , 6.455785 )
( 3.971971211166494 , 3.0850968 )
( -1.310163823492471 , 0.88155675 )
( 2.195005433564788 , 5.861716 )
( 4.424420324514264 , 6.0825615 )
( 1.3250087053484476 , 2.6631045 )
( 3.6157508122114645 , 3.6774993 )
( 8.111944816892722 , 5.374503 )
( 4.809488142506297 , 3.745995 )
( 5.3528694612293615 , 3.8248253 )
( 8.074255326745986 , 5.85702 )
( -5.036652750805067 , -1.7949917 )
( 6.799023221307781 , 6.660987 )
( 0.4842734465285048 , 1.9822569 )
( -0.6807215916598376 , 1.1595752 )
( 0.3391016419927628 , 1.9167318 )
( 0.30695453329751476 , 2.1925523 )
( 1.989391232200274 , 2.7996354 )
( 5.741765168140211 , 6.7105927 )
( 2.058901163008797 , 2.7206502 )
( 5.409069259618547 , 6.1853337 )
( 7.960705380199065 , 6.309182 )
( -3.5696849425253028 , -1.125737 )
( 7.6788128693097 , 4.9908123 )
( -0.8112591349307039 , 1.5433426 )
( 1.101978508147054 , 4.270598 )
( 2.6285679810691542 , 2.6719074 )
( 1.0783652365220808 , 2.1360636 )
( 7.4732174411246906 , 5.234973 )
( 3.47231666480754 , 4.741093 )
( 7.2412005457190896 , 5.149575 )
( 8.666897534478453 , 7.3376884 )
( 8.343778302335693 , 5.728222 )
( -0.46512612394413755 , -0.55579126 )
( 8.807363703418494 , 7.2420707 )
( 6.3792336311848095 , 7.976523 )
( -10.437221645700149 , -1.8272338 )
( 3.629503518811924 , 2.8723989 )
( 4.746900388581092 , 7.251014 )
( 0.15215903926326035 , 2.2016778 )
( 0.5634830964420938 , 3.4447827 )
( 4.4001327782623605 , 3.544189 )
( 9.26421111124202 , 7.156796 )
( -0.3917148535655173 , 1.6862354 )
( 3.901191564160495 , 4.0676203 )
( 3.0996279129631175 , 3.7761471 )
( 8.51613751637073 , 7.9512196 )
( 5.733939405678729 , 6.9718814 )
( 2.4801394286970595 , 3.4505458 )
( 4.071071137260185 , 6.4537096 )
( 3.5681700925702424 , 3.4242783 )
( 5.124374737752694 , 6.851528 )
( 1.2019172880292346 , 3.3231516 )
( 6.130227380192603 , 6.2759814 )
( 6.3067799987182624 , 6.885132 )
( 3.2717690061970295 , 3.3712158 )
( 2.7349584858773444 , 2.638538 )
( 8.962820573163523 , 6.21636 )
( 2.995617081083328 , 5.2002196 )
( 2.915935001366978 , 3.2040014 )
( 1.8644158545532985 , 3.7651718 )
( 8.336129179602093 , 5.509961 )
( 6.085658676439345 , 7.2684894 )
( 0.5749252501844584 , 2.0912528 )
( 5.310152447615155 , 4.2699823 )
( 7.158445944291447 , 6.087595 )
( -1.6370201491541465 , 0.15373111 )
( 0.9986431823481587 , 3.2406664 )
( 5.1305873227938905 , 3.712431 )
( 6.155760579151626 , 5.5677395 )
( 6.769782977998894 , 8.2253475 )
( -0.9822414878575332 , 1.4209528 )
( 1.3344636175231686 , 2.9073558 )
( 4.554871489740615 , 4.9907427 )
( 2.909827855188614 , 3.0889525 )
( 4.5175169712843175 , 6.9777045 )
( 7.243242739952342 , 7.6521435 )
( -1.037852104233253 , -1.2815636 )
( 1.0532549579469026 , 2.7845535 )
( 5.01416710988768 , 6.7405906 )
( 8.511830381716779 , 6.3139215 )
( 5.896195549242847 , 4.8583565 )
( 1.1840946246274253 , 2.1764407 )
( 6.591521975808543 , 4.3692083 )
( 4.532828086221569 , 6.002968 )
( 4.166008129046396 , 3.957357 )
( 4.146551897937692 , 6.878187 )
( 2.5971589184437374 , 3.19069 )
( 1.4830701324221194 , 3.1239438 )
( 4.005073502282227 , 3.7182634 )
( 5.105853296062 , 7.6510005 )
( 0.9975795103096412 , 3.6485176 )
( 3.126753842069456 , 3.0936017 )
( -7.439218403173073 , -0.9507179 )
( 6.799888093300674 , 5.520575 )
( 3.981398885365243 , 4.57823 )
( 3.7897418956222566 , 3.3829827 )
( 6.072761189641484 , 5.56351 )
( 5.295036757229746 , 5.374308 )
( 2.6653308686888124 , 2.8266482 )
( 4.116420866323832 , 6.9977503 )
( 0.48398987865715637 , 0.024934977 )
( 6.6671629632631255 , 5.332961 )
( 1.596662606196478 , 2.9569178 )
( 0.681479149294133 , 4.4387984 )
( 1.360549839281143 , 2.5715275 )
( 5.48290609053849 , 5.2316155 )
( 3.5007163483593216 , 4.3234086 )
( 1.3658614854818878 , 4.252002 )
( 4.742767114063593 , 4.1062613 )
( 1.900833687213928 , 4.2032743 )
( 3.0900035367463516 , 2.8215237 )
( 5.334425461443245 , 7.711476 )
( -0.25184200642811927 , 1.5629473 )
( 2.471905399350013 , 5.383969 )
( 1.775837184711735 , 3.1559157 )
( 5.5819669243569985 , 4.9370008 )
( 1.3002937481563142 , 4.475806 )
( 6.225433530450647 , 5.4699144 )
( -7.047127941660372 , -0.39818573 )
( 5.801280751712154 , 7.736055 )
( 3.8385130706510076 , 4.378537 )
( 3.129081882439874 , 5.8949194 )
( 3.1381742517297635 , 3.6954541 )
( 0.911672035089985 , 2.2988315 )
( 0.39746992241505763 , 2.682351 )
( 3.683233107893736 , 5.439446 )
( -0.9879722717985144 , 0.9605515 )
( 5.552722441747106 , 4.033526 )
( 1.4879642700392357 , 3.0003412 )
( 6.493739219342513 , 6.11164 )
( 1.71769140109893 , 2.4219375 )
( -1.5754339828219046 , 0.28101254 )
( 5.997787280696301 , 6.587792 )
( 4.859847991073047 , 6.9505305 )
( 6.02204310872035 , 6.9530807 )
( 5.878205167653274 , 4.69903 )
( 8.189840437294244 , 7.1306267 )
( 3.3225445675925247 , 4.55712 )
( 2.9872772134142846 , 6.193896 )
( 1.7683474309917608 , 5.302778 )
( 7.274587817637419 , 4.9763966 )
( 5.014556318205441 , 3.9034183 )
( 0.021949093411494758 , 1.9433169 )
( 1.3791536331558263 , 4.806392 )
( 2.749995527729252 , 6.3834257 )
( 2.6807466158951563 , 2.740675 )
( -1.063042384348054 , 1.2795234 )
( 2.4163439473747848 , 4.1012464 )
( 2.1350668529301178 , 2.316762 )
( 6.771859124504322 , 5.1425495 )
( 3.8114009671871916 , 6.757035 )
( 8.595778599032936 , 6.9186707 )
( 4.331827608436718 , 3.8575869 )
( 8.694604944621343 , 6.734291 )
( 1.1174953867118749 , 2.2518125 )
( -0.8633201353072568 , -0.051421165 )
( 7.971782557729421 , 7.0354905 )
( 3.164120747466886 , 5.6713915 )
( 6.449059586758069 , 8.1714735 )
( 0.6244241847319878 , 2.067389 )
( 5.608020774523073 , 5.339887 )
( 7.279243521793322 , 6.483721 )
( 5.46935513329133 , 4.567981 )
( -5.265942177958036 , -2.8018248 )
( 0.32351073124181473 , 2.421329 )
( 6.545221502876821 , 5.8297367 )
( 6.395752252102344 , 5.085802 )
( 1.1857569710415499 , 2.21792 )
( 2.3132684916612885 , 4.8674917 )
( -1.5244140642080222 , -0.5788458 )
( 7.97112358550375 , 7.7593317 )
( 1.930134676963441 , 2.4809642 )
( 4.731446250408343 , 3.6238174 )
( 2.050535438956391 , 2.998942 )
( 8.332102242749624 , 7.7117486 )
( -0.5749081743185755 , 1.322231 )
( 4.533885930835709 , 6.956721 )
( 9.312465295026502 , 6.343188 )
( 5.863114856857048 , 6.2607727 )
( 7.546016634538164 , 6.2221355 )
( -3.9227885916115195 , 0.18187761 )
( 2.0394474555596576 , 5.056205 )
( 0.581530273852234 , -0.4608508 )
( 0.2196068267587522 , 2.1438055 )
( 2.475244755748187 , 3.4168856 )
( 8.2664019341911 , 7.0923243 )
( 2.190214619285841 , 5.8581533 )
( 7.466424746058265 , 5.4808974 )
( 5.658164198018479 , 4.8882093 )
( 4.403602631218711 , 3.7544742 )
( -5.224978802992986 , 0.46269655 )
( 8.625333882040986 , 5.841509 )
( 4.717067634553461 , 6.802379 )
( 8.701105912243985 , 6.566642 )
( 0.6829439384432998 , 0.4699755 )
( 4.275607443358914 , 6.5669994 )
( -0.08250282069655374 , 3.7135584 )
( 7.662676154006775 , 5.363049 )
( 7.6449865281401825 , 5.4821873 )
( 0.8213267527476028 , 2.8926864 )
( 7.736175343362053 , 6.1472454 )
( 8.342159482651148 , 7.5756555 )
( 2.8456036768643562 , 5.478167 )
( 2.6963482879483918 , 4.923583 )
( 6.83656710186872 , 7.7044826 )
( 4.482978576874089 , 5.1877437 )
( 5.5087617221936 , 4.1744213 )
( 1.8058079969005956 , 2.8350303 )
( 7.502734286502235 , 5.3773212 )
( 4.5511530335982915 , 4.6990657 )
( -0.3266378831588659 , 1.4565845 )
( 0.34852643591641924 , 3.1591156 )
( 4.688043987506755 , 7.3789907 )
( -0.5164141538124996 , 1.2792072 )
( 7.108832456392361 , 7.294435 )
( -0.3703417218571583 , 1.5747797 )
( 2.4701499904104214 , 2.8065367 )
( 1.344211838775722 , 5.1228642 )
( 2.5784112490213635 , 3.21601 )
( 6.070397627215414 , 6.2658944 )
( 4.08445533880795 , 3.848948 )
( -0.22542072572583272 , 1.3499446 )
( 6.466550946048175 , 5.246458 )
( 0.05979029613621627 , -0.08248179 )
( -3.8457693107436626 , -0.94929504 )
( 6.301235527873694 , 4.779604 )
( 0.8384084943593229 , 3.2316818 )
( -0.3811887327150716 , 0.41711986 )
( -0.3790195901223447 , 1.9772449 )
( 3.3648486295142765 , 2.9888268 )
( 3.1358569116311097 , 4.631921 )
( 5.426002499733469 , 4.80787 )
( 8.812234323321725 , 5.7308135 )
( 1.1408548492768396 , 2.254551 )
( 6.4719675561038015 , 4.557994 )
( 5.075685770545606 , 6.01297 )
( 0.19647626285131448 , 2.2788377 )
( 3.892928324931664 , 3.8747444 )
( 2.683407633627168 , 4.2674246 )
( 6.748026450849453 , 5.396291 )
( 6.392600500447838 , 5.810261 )
( 3.5818439435015135 , 4.168807 )
( 0.7650937026115008 , 4.417854 )
( 0.99219119179527 , 4.9071603 )
( 5.846027264664226 , 5.687986 )
( 2.884827621962347 , 4.95788 )
( 4.962376271645355 , 5.6886578 )
( -1.001470230760968 , 1.5504131 )
( -1.5929285362910726 , 1.3938766 )
( 6.571752296667191 , 5.0094414 )
( 5.426542368888581 , 4.8780484 )
( 0.5812839223241242 , 3.8773026 )
( 5.549019686858804 , 4.5951276 )
( 8.611423977018035 , 8.161648 )
( 2.8354116822532713 , 4.1456366 )
( 6.122294722659488 , 6.8097334 )
( 7.893478809954566 , 4.9820747 )
( 0.6220021003642806 , 3.4458756 )
( 0.2806964171607157 , 3.8025985 )
( -0.6954836129096513 , 1.422936 )
( 1.0186738403981115 , 2.506344 )
( -1.7419358622075105 , 1.3488331 )
( 3.416587801382687 , 3.6296933 )
( 1.4180806855703523 , 2.4103165 )
( 2.2599741183650073 , 4.524832 )
( 6.249004783412754 , 6.165996 )
( 4.397152771695386 , 7.375846 )
( 0.7803610188826315 , 2.648243 )
( 0.23949248957757596 , 3.9660969 )
( 7.259615850834216 , 8.258315 )
( 6.747536956818303 , 6.010441 )
( -9.619736345819268 , -0.2538042 )
( -1.1290545204923663 , 0.38620806 )
( 0.18453391131060626 , 0.8687905 )
( 2.7576507858120998 , 3.3358252 )
( -2.2025592371228795 , 0.6553085 )
( 7.40519542264338 , 5.0078764 )
( 2.071981326475287 , 5.0838084 )
( 5.953170096651236 , 4.4973783 )
( -1.4181221561088753 , -0.082485974 )
( 5.503889141244559 , 3.9759436 )
( 3.679803271051624 , 4.964321 )
( 5.085710288660236 , 5.242998 )
( 3.1602498312622034 , 2.9896126 )
( -0.17072681981402837 , 1.7589817 )
( 6.157057163125378 , 5.1198454 )
( 4.293359944999301 , 3.248375 )
( 5.943076598467691 , 3.9656734 )
( 6.717191927808272 , 4.4143877 )
( 5.70206384092841 , 4.603201 )
( 7.080939940798878 , 6.5349536 )
( 3.8507419056964682 , 4.131942 )
( 4.874074235990611 , 5.344499 )
( 5.995151954756908 , 4.6102247 )
( 5.562007547978609 , 4.3008804 )
( 2.0287331712064463 , 4.3404584 )
( -0.5756662140352468 , 3.3581555 )
( 7.206097375549581 , 7.5394936 )
( 5.063530258450487 , 4.0063677 )
( 7.794261899573549 , 5.707572 )
( 5.635284631293987 , 6.8274975 )
( 2.631961583243885 , 5.026061 )
( -0.13270730574559053 , 2.0508175 )
( -1.4041260499823638 , 0.33594775 )
( 4.781604518023812 , 4.135758 )
( 5.061792994362303 , 4.274206 )
( 3.7967887999214804 , 3.4643698 )
( -0.9170670861465926 , 0.6275122 )
( 2.3984446395342593 , 3.4347463 )
( 1.0403311717216241 , 2.244903 )
( 7.573532995784509 , 5.1666865 )
( 6.531591131941996 , 5.5059395 )
( 0.43287411489203625 , 4.030214 )
( 5.880681387737205 , 7.6852846 )
( 0.7457143239334482 , 2.6952207 )
( 0.8562030468105234 , 4.36595 )
( 6.251780018590224 , 5.0097566 )
( -8.951826266327188 , -0.78521395 )
( 3.7496386987232326 , 6.928675 )
( 2.3421287073115 , 3.21909 )
( 3.8528088846303286 , 3.224615 )
( 2.8423686297572637 , 2.9845443 )
( 6.908882324707088 , 4.26582 )
( 7.1431344382942985 , 7.887754 )
( -0.4952406163788685 , 3.0601583 )
( 6.425140098123507 , 4.2000275 )
( -0.28078579790237335 , 1.5130932 )
( 6.825117813014781 , 6.57559 )
( 4.467634824141519 , 3.6185923 )
( 0.7643653079046813 , 3.2254908 )
( 5.7288521996524455 , 4.727248 )
( 2.6355246828735934 , 2.9055276 )
( 1.312347594344499 , 3.554706 )
( 7.1564394875389965 , 5.8210764 )
( 5.325740843880499 , 5.694239 )
( 7.105105570822241 , 6.2249002 )
( 6.6998837701259735 , 6.0537124 )
( -7.489716220397302 , -3.6060357 )
( 7.686440170919016 , 8.398781 )
( 1.8825017002979392 , 3.5935874 )
( 8.289803656983274 , 5.419816 )
( 1.5570199862835385 , 3.3408318 )
( -11.01287074171087 , -1.4020863 )
( 7.182490697470325 , 4.675997 )
( 8.807934587908681 , 6.698331 )
( 3.582486720994801 , 3.431387 )
( 5.062651471375654 , 4.497367 )
( -3.9589630700545664 , -3.184467 )
( 0.1749816065665585 , 1.6870459 )
( 3.686370459931357 , 3.6771126 )
( 5.803068797435925 , 4.586377 )
( 4.323619852869415 , 5.6461854 )
( -0.13864754705374427 , 1.7019596 )
( 6.703282204143671 , 5.426074 )
( 1.7326988553652443 , 4.163512 )
( 0.326456935858493 , 1.8949561 )
( 6.776496382255013 , 7.1630206 )
( 7.079544498704168 , 4.7696896 )
( -3.48492228376138 , 0.7943926 )
( 1.7810780063897753 , 3.7382057 )
( 8.77903234925611 , 6.200839 )
( 5.953179207000205 , 5.5406723 )
( 0.8592441211562654 , 4.028904 )
( 7.649844541376083 , 5.6982493 )}

\newcommand{\utilitymedium}{
( 0.2676768836317993 , 2.5622826 )
( -1.3910324681900308 , 1.4765701 )
( -8.937480172661521 , -8.422862 )
( 5.749442652414051 , 8.661368 )
( 6.891810343731568 , 8.812648 )
( 6.507949783217706 , 7.9573193 )
( 6.144103035384285 , 7.833819 )
( 6.816424821888014 , 8.806866 )
( 3.4854345622751826 , 5.996467 )
( 1.0361119648962416 , 2.3643527 )
( 8.054097235893583 , 10.25498 )
( 4.670134201908244 , 7.8509045 )
( 8.781796981838761 , 10.513949 )
( 4.659766354476239 , 6.7994695 )
( -2.2917948105728136 , -0.6813234 )
( 2.126971429940478 , 3.6773896 )
( 0.035574980682075985 , 1.4506197 )
( 3.2741180794561346 , 5.602308 )
( 5.417361712407309 , 7.436546 )
( 5.706608462854018 , 8.158947 )
( 7.7008976443924855 , 9.843096 )
( 5.890854913827388 , 7.7077913 )
( 4.635860617305445 , 6.9196944 )
( -0.00633232290850827 , 0.5701002 )
( 5.330455374610794 , 7.260019 )
( 2.241714797461916 , 4.427519 )
( -0.10010578191149244 , 1.8467691 )
( 1.4990409379348595 , 3.0729551 )
( 5.825251590519721 , 8.245778 )
( 2.246882906138029 , 3.4498687 )
( 1.5203160341951762 , 3.140534 )
( 0.34961138183561546 , 1.7019879 )
( 1.1973893585740374 , 2.4812489 )
( 6.586282621233469 , 8.229291 )
( 8.264655767053618 , 10.0038395 )
( 4.461134747589333 , 5.5544477 )
( 7.262721563929104 , 9.628915 )
( 5.013262208406412 , 7.9014664 )
( 2.3393496127544138 , 4.436167 )
( 4.91004641749204 , 5.8730164 )
( 6.990328989320571 , 8.948432 )
( -1.2198144913926594 , 0.38509166 )
( -9.042410403820728 , -8.317625 )
( 5.035935911046796 , 7.335677 )
( 8.149293242002262 , 9.1433525 )
( 3.5352994717246347 , 5.4343596 )
( 7.789463143742424 , 9.543707 )
( -9.67681802684643 , -9.519167 )
( 2.990426266918986 , 4.3907285 )
( 4.770975920388231 , 6.742618 )
( 0.4107177377514951 , 2.2264667 )
( -0.24218668137076726 , 1.6686753 )
( 7.434055839571792 , 10.029157 )
( 5.300797028704794 , 7.2373056 )
( 3.987354219890465 , 6.334216 )
( 6.943160816835565 , 9.753794 )
( -0.24001589099959536 , 0.7474892 )
( 7.573566576066656 , 9.429708 )
( 8.52154058081566 , 10.348788 )
( 6.870987175435958 , 9.884443 )
( 6.268242745235366 , 8.868713 )
( -0.18443829083197327 , 1.2252283 )
( 0.605826515988406 , 2.6386728 )
( 3.0737328698228144 , 4.704367 )
( 3.3873604118493548 , 5.4952354 )
( 0.8845794866531771 , 2.5508924 )
( 0.094090850472732 , 1.2602501 )
( 6.190662386314384 , 8.085014 )
( -1.9781119101886349 , -1.3606538 )
( 4.286964438903956 , 6.211715 )
( 2.0371468598635616 , 4.392392 )
( 4.835938426146559 , 6.82415 )
( 7.575331976011167 , 10.036509 )
( 6.028099789920599 , 8.209191 )
( -2.1226508009386764 , 0.076314926 )
( 7.7974337190226715 , 10.285648 )
( 0.7293452492793208 , 2.1332917 )
( 2.0177215776243 , 3.6671822 )
( 1.1658940833142832 , 2.681981 )
( 8.514987551249266 , 10.171293 )
( 6.293848514183121 , 8.691067 )
( 4.8429008940369895 , 7.9827476 )
( 7.405281975268689 , 10.04407 )
( 5.096440043547555 , 6.833821 )
( 2.129283018049639 , 3.7316713 )
( 2.4003631079296395 , 4.271791 )
( 7.754607914338543 , 10.038371 )
( 5.450374435147729 , 7.213236 )
( 0.09806241871263455 , 2.0131464 )
( 3.4443959511103035 , 5.5043364 )
( 0.4045072959739514 , 1.450761 )
( 3.9812004502376612 , 6.6025543 )
( 2.670948639898037 , 4.514905 )
( -6.202579021048489 , -5.943217 )
( 0.6106154427116954 , 2.685628 )
( 5.6953031419881786 , 7.6034527 )
( 6.311409638786939 , 8.030728 )
( 7.4430003051860965 , 9.171265 )
( 1.0278914073269831 , 2.5001698 )
( -0.8911953998242435 , 1.0846792 )
( 5.06661092130334 , 7.315385 )
( 5.855617888085803 , 8.270513 )
( 5.836307425775854 , 8.24044 )
( 4.947504092463888 , 6.9248066 )
( 1.5095068007629253 , 3.1519256 )
( 7.104821931785835 , 9.442403 )
( 0.47792012547422746 , 3.0790868 )
( 0.06945965338814708 , 1.3523031 )
( 3.6280763677812384 , 4.5262423 )
( -1.5632838268450429 , -0.36134887 )
( 6.479172789397314 , 8.505202 )
( 2.52593145092612 , 4.296401 )
( 5.633289194155285 , 7.3708916 )
( 2.1792875490410504 , 4.4417276 )
( 7.256934576450996 , 9.151442 )
( 7.223554423484355 , 8.6711855 )
( 0.011986661829006096 , 1.8146011 )
( 3.808543283570783 , 6.297515 )
( 1.5875815858092082 , 3.159433 )
( -3.5051194536708525 , -2.3447247 )
( -2.963390022967295 , -2.9717157 )
( 3.7186189185322487 , 5.3390713 )
( 7.253743834414072 , 9.042554 )
( 2.475002539405235 , 3.9685545 )
( 3.750759810389197 , 5.7403708 )
( 3.3907067245023956 , 5.3364906 )
( 5.57106324599616 , 7.2753544 )
( 2.2048408643683626 , 3.5152974 )
( 3.7022400786329914 , 6.9101276 )
( 2.480902656732841 , 4.09972 )
( 4.541787117090942 , 6.6697707 )
( 0.9503106433347589 , 2.4040112 )
( 4.277398412117946 , 6.875869 )
( 3.672379100859513 , 5.562833 )
( -3.9697186006714063 , -3.333036 )
( 6.966574021674121 , 9.458429 )
( 4.949016346950014 , 7.1601143 )
( -0.7376350298710882 , 1.9031067 )
( 7.253713327485091 , 9.27656 )
( 6.208593021962292 , 8.711687 )
( -4.113354192751081 , -3.3053908 )
( -1.9659081150789117 , 0.9594232 )
( 2.3583639629056545 , 3.9032946 )
( -0.6644182742033116 , 1.1203384 )
( 0.39096122589377647 , 2.1874647 )
( 7.7599668616618445 , 9.22872 )
( -0.8202870429949162 , 1.8063227 )
( 5.440341521149703 , 7.104879 )
( 0.4619511916524053 , 1.8822478 )
( 4.26885158670816 , 6.6590014 )
( 1.2800542348565491 , 3.260693 )
( 7.258918383760248 , 9.260481 )
( 2.2312533517496513 , 3.9504642 )
( 3.7081084510552946 , 5.4023185 )
( 5.201975406018853 , 7.767122 )
( 1.522237457318805 , 2.9034963 )
( 2.941756254620538 , 4.959848 )
( 5.506838240010387 , 7.9794497 )
( -1.4293354467484038 , 0.8417382 )
( 2.098130895040582 , 4.6275206 )
( 3.613635087880109 , 5.815192 )
( 5.213500941601739 , 7.015784 )
( 4.2468317966711195 , 6.947009 )
( 1.9171389154471867 , 3.820949 )
( 8.566633987378557 , 10.347879 )
( -1.6203647536920007 , 0.11708474 )
( 1.7110126396972825 , 4.0257664 )
( 2.217116193522877 , 4.673812 )
( -2.4407617326650004 , -1.3359003 )
( -4.081875150783875 , -3.8210285 )
( 8.120655803296842 , 10.144467 )
( 5.257586618995882 , 7.1443763 )
( 1.2475351367823793 , 3.9071963 )
( 0.4016883557141555 , 1.8466684 )
( -5.729736973729846 , -7.265108 )
( 0.5386354175986505 , 1.9526287 )
( -2.0232284892451826 , -1.6991047 )
( 8.420730543716127 , 11.088192 )
( 5.474797958984503 , 8.4253845 )
( 8.806161947185453 , 10.5362215 )
( 2.928874565101916 , 5.0813622 )
( 0.9200329594311141 , 2.564009 )
( 3.922188805469382 , 5.977826 )
( 8.180835293751983 , 10.335552 )
( 3.7762219938519475 , 5.819904 )
( 0.03709938490052478 , 1.3267261 )
( 3.3614352345594085 , 5.136775 )
( 2.5538050932817917 , 4.6322865 )
( 7.441796903035247 , 9.305244 )
( 4.486654636133456 , 6.4365106 )
( 0.2983752851087048 , 1.683769 )
( 1.15412745563486 , 2.848538 )
( 6.305311041848304 , 8.291009 )
( -1.7354562066284753 , -0.5441513 )
( 8.32989256385774 , 10.612824 )
( 1.7851704526575878 , 3.1306589 )
( 0.07424073045536095 , 1.4541637 )
( 8.182660504829279 , 9.910279 )
( 2.8106749017164 , 4.209062 )
( -0.3523877366456328 , 1.6484103 )
( 8.21221844648461 , 10.195576 )
( 1.6086948059849768 , 3.0145898 )
( 7.25954304084537 , 8.90094 )
( 6.370165152554241 , 9.660693 )
( -0.42244098051768475 , 1.8390421 )
( 7.782918143338263 , 9.685416 )
( 1.283513279431482 , 3.3201559 )
( 5.108275994125956 , 7.229156 )
( 6.69256068898618 , 7.761413 )
( 7.328771651005318 , 9.5485935 )
( 1.5180334730464906 , 3.3078446 )
( 7.419994131976285 , 9.603724 )
( 6.025271582932866 , 7.957055 )
( 5.336528575210153 , 7.114291 )
( 0.10081572525438176 , 1.5066764 )
( -3.1164905594823313 , -1.2062477 )
( -0.2109976594997009 , 2.22581 )
( -0.24412299644987043 , 1.9525636 )
( 5.37884408193659 , 7.423831 )
( -0.7494958829239231 , 0.24672365 )
( 2.388313142846553 , 4.1451063 )
( 1.7846513350581148 , 3.6867712 )
( 5.440235792480619 , 6.7369857 )
( 0.558048374992988 , 2.4280024 )
( 8.580174717955835 , 10.328442 )
( 1.2438049862221776 , 2.5999663 )
( 3.59170023561102 , 6.2068 )
( 5.337438259524756 , 7.432332 )
( 2.4060761878287886 , 4.3556128 )
( 4.446129804915182 , 6.254244 )
( 7.235815168904421 , 9.177687 )
( 9.405199353232666 , 11.714176 )
( 0.8390495887795444 , 2.5155454 )
( 4.442173884515984 , 6.6888494 )
( 6.0107328602449535 , 7.839884 )
( 2.6106269650590606 , 4.1190195 )
( 6.29003312834007 , 7.8606606 )
( 1.2915725265609144 , 2.4247456 )
( 4.7527004169148075 , 7.2432013 )
( 7.475250664593411 , 10.737972 )
( 8.536817954754035 , 10.273896 )
( 2.919248408522611 , 4.573225 )
( 0.05217678760017841 , 1.6735373 )
( 0.2509566301280288 , 1.6755887 )
( 2.5194673806306604 , 5.627231 )
( 7.719820091929492 , 9.149604 )
( 1.522949998969045 , 3.1076744 )
( 1.825643198490536 , 2.9226909 )
( 0.28426180525791667 , 1.9740787 )
( 2.43102895903682 , 4.0108123 )
( 5.047594735850084 , 6.8203316 )
( 8.916845061407068 , 10.758611 )
( 6.945985617873937 , 9.279789 )
( -0.9732080040970796 , 0.7572824 )
( 2.455856685016883 , 4.8595476 )
( -3.037478883774261 , -1.2212486 )
( 3.539002364928046 , 4.0220885 )
( 4.475667236002017 , 6.528187 )
( 1.2995269554059536 , 3.4790907 )
( 4.335829618557123 , 7.2094173 )
( -0.9387249205498538 , 0.93967676 )
( 4.162888804645874 , 5.626299 )
( -0.003695436481622849 , 0.8313916 )
( 2.545535423871721 , 4.3958573 )
( -11.117845589056639 , -10.813119 )
( 7.574677912946466 , 9.42933 )
( 2.746049955938565 , 6.320034 )
( -1.242157795082518 , 0.8359659 )
( -0.8618651018022133 , 1.6613148 )
( 4.856865927717134 , 7.700887 )
( -1.8399085973828004 , -0.6852945 )
( 6.9227052400387565 , 8.980063 )
( 0.38739077290497004 , 2.1747038 )
( 0.3473328252936323 , 1.4849106 )
( 6.914238597525393 , 8.710922 )
( 0.24403380859685964 , 1.8385842 )
( 3.8628310840301365 , 5.832595 )
( 8.548960846397414 , 10.281851 )
( 6.855049222257766 , 9.258974 )
( 3.9817606967853756 , 7.147979 )
( 5.7994411977760025 , 8.59695 )
( 1.2810910751368496 , 2.9082737 )
( 0.03908917180067584 , 1.4759836 )
( 3.4815359765055858 , 5.388743 )
( 2.9721499089349046 , 5.633925 )
( -0.696261072567994 , 1.4955877 )
( 5.513782393199161 , 7.70063 )
( 6.267688100680255 , 9.194208 )
( 2.0418777055653847 , 4.156015 )
( -4.77565744811427 , -3.630157 )
( 6.767686735179783 , 8.867516 )
( -1.3220716139598245 , 0.95850456 )
( 6.820090380680488 , 8.054803 )
( -2.677094435890613 , -0.48947328 )
( 1.3192856244223274 , 2.644648 )
( 7.267337074326019 , 9.005049 )
( -0.40808075210172867 , 1.0079002 )
( 8.206156740820592 , 10.708302 )
( 0.6322469088198712 , 2.2146387 )
( 8.021103795285306 , 9.6274805 )
( 4.973118237234678 , 7.205188 )
( 4.448385923540285 , 6.8717594 )
( -0.02199770347320107 , 1.4296116 )
( 1.7179096344313098 , 4.062996 )
( 0.832174931794368 , 2.6458972 )
( 2.1511011011678045 , 3.2719092 )
( 5.970377987900401 , 8.01271 )
( 4.760904356287516 , 6.6030045 )
( 3.231979654132901 , 4.911503 )
( 4.5376279352144575 , 6.535248 )
( 6.427896470127056 , 8.574722 )
( 4.876203817844106 , 6.7883554 )
( -0.33290786170029263 , 1.879967 )
( 0.8043722565572329 , 2.0769417 )
( 3.1019413368618833 , 5.098349 )
( 2.7298630938542465 , 4.756868 )
( 4.853352388767605 , 7.439934 )
( -2.5208545165025202 , -0.17020953 )
( 5.797529665575164 , 7.737403 )
( 2.5760889487575307 , 4.7420073 )
( -0.18972927303844553 , 1.3559479 )
( 4.728681012248585 , 7.2596827 )
( -0.16517029471887434 , 1.6601567 )
( 6.15410514879552 , 8.15882 )
( 3.721256158749855 , 6.704047 )
( 3.3449144829207134 , 4.6337743 )
( 1.0373104061855787 , 2.9726093 )
( 4.585645849657365 , 6.8971915 )
( 1.1159506831087205 , 2.3646421 )
( 2.2780916114170546 , 3.350012 )
( 4.474224694385321 , 6.9732122 )
( 0.0953871201941191 , 1.242729 )
( 3.858147853759852 , 5.235228 )
( 1.441823250724911 , 3.322675 )
( 4.027676470627191 , 6.122693 )
( 7.137802280153975 , 8.779553 )
( 5.533865865429053 , 6.791411 )
( 8.467344497627119 , 10.269268 )
( 3.1348093566863477 , 4.9498577 )
( 3.944689542386139 , 5.961894 )
( 5.700390937826158 , 7.8537602 )
( 3.8001940650483252 , 5.60295 )
( 4.2158864341280635 , 6.8070245 )
( 6.948398515323394 , 10.034598 )
( 6.06421768247725 , 7.9500628 )
( 1.3968320864091814 , 2.8507996 )
( 7.236784748215651 , 9.577199 )
( 3.998632140656552 , 6.4400144 )
( 1.4733189438740606 , 3.4114573 )
( 4.182629082116691 , 6.482518 )
( 5.7358266649504595 , 8.013926 )
( 5.222875135796961 , 7.192989 )
( 6.962177940908919 , 10.008606 )
( 2.0928867524727677 , 4.1302843 )
( 4.339612486297609 , 6.694922 )
( 2.1745903996841305 , 4.0035233 )
( 2.481148703852342 , 2.5470326 )
( -0.09300515726508696 , 1.4101661 )
( -0.1604761103345081 , 2.153656 )
( 8.839492147594768 , 10.343256 )
( 7.505729233239036 , 9.457165 )
( 8.662737012893377 , 10.539003 )
( 3.503210151678705 , 5.7283072 )
( 2.3042051975153037 , 4.431196 )
( 2.4650324492555873 , 4.6965694 )
( 5.2432588467838865 , 7.0299664 )
( 1.815629622613181 , 3.961298 )
( 7.956636241117511 , 10.450871 )
( 5.811371550234936 , 7.9598737 )
( -5.842394848267936 , -6.045254 )
( 4.712009279143475 , 6.844334 )
( -3.840615927891837 , -2.0871115 )
( -0.1668979307594185 , 2.1387534 )
( 3.5901660035451846 , 5.7502446 )
( -0.7807913366435706 , 1.3002877 )
( 4.16965298960069 , 6.9482813 )
( 5.069203585586157 , 7.3705206 )
( 4.476676434119617 , 6.616784 )
( 4.946249997006223 , 6.8295984 )
( 2.0668907741434044 , 4.055128 )
( 1.385713983881462 , 3.1941803 )
( 6.178963034821699 , 9.376149 )
( 4.283003167150721 , 6.2217674 )
( 6.835222485552192 , 9.222966 )
( 7.0187894237663855 , 8.694145 )
( 3.0172747537474227 , 4.8743153 )
( 2.343161289625655 , 3.8634334 )
( 7.086178868707142 , 9.139832 )
( -3.4736069594393433 , -1.448369 )
( 6.262241392965121 , 8.632075 )
( 4.431433908418339 , 5.671069 )
( 3.664632222983875 , 5.676068 )
( 3.599714704665939 , 5.973357 )
( 0.7527527328491762 , 2.3725214 )
( -0.013627553637451939 , 1.6551623 )
( 0.45242766180326965 , 2.3481092 )
( 6.360597442291068 , 7.6871567 )
( 4.722532788122859 , 7.132867 )
( 7.428796554877485 , 9.96108 )
( 4.686064474168175 , 6.87542 )
( 7.144478209798695 , 8.630381 )
( -0.5044191987529233 , 1.556415 )
( 5.241235530229169 , 7.6858077 )
( 0.6077458552489645 , 3.1713498 )
( -1.7494570674221444 , -0.99214476 )
( -0.3085359025614005 , 1.4447184 )
( 3.597309139508509 , 5.4691734 )
( 6.079418824973439 , 8.017269 )
( 2.664052273559679 , 4.2141256 )
( 0.09188721535401179 , 1.6135073 )
( 3.2856718952302266 , 6.126637 )
( 5.771290700997784 , 7.7542357 )
( 4.0357030513806835 , 6.418144 )
( 4.336833643644864 , 5.82518 )
( 1.9426217994612878 , 3.9660368 )
( 6.624054400583452 , 9.1871605 )
( 1.8090750599968404 , 3.2034593 )
( -0.07226162260812918 , 1.1918788 )
( 5.417246007865398 , 7.8154383 )
( 6.219867833521878 , 8.5402 )
( 5.265101677378989 , 8.586502 )
( 7.043692292311279 , 8.817745 )
( 4.457690913728479 , 6.5237603 )
( 2.571705532424465 , 5.05926 )
( 3.018823359926941 , 5.4212213 )
( 4.542100392209846 , 7.0715966 )
( 0.013201218670070403 , 1.9537576 )
( 6.97731431598188 , 9.857474 )
( 2.7129700103713783 , 4.9024053 )
( 4.790215074046861 , 6.8389325 )
( -0.6778653526631935 , 1.2877532 )
( 7.203968590583353 , 9.335848 )
( -0.16742374699072693 , 1.329821 )
( -0.9061252741010206 , 1.3728827 )
( 0.2170116340411683 , 2.0603104 )
( 5.941813818408002 , 7.6286554 )
( 6.164484284048667 , 8.312298 )
( 4.394736677669951 , 6.5626717 )
( 2.5876744961671 , 4.4223504 )
( 7.745287333339578 , 9.75092 )
( 6.464975950205047 , 8.875286 )
( 5.221598201707392 , 7.427855 )
( 3.2746376319032793 , 4.913342 )
( 4.322557968141274 , 5.879312 )
( 0.14107125549394084 , 1.1175219 )
( 3.981980763973077 , 6.072282 )
( -1.2359067438271043 , 1.2758225 )
( 3.3474347106638858 , 5.794621 )
( 1.1060723748543433 , 2.9083219 )
( 7.6114767610079115 , 10.548443 )
( 6.581000098121152 , 9.641546 )
( 7.238503493573845 , 9.699969 )
( 4.269442086174138 , 6.0693755 )
( 7.054061471324001 , 8.61857 )
( -1.461811525479933 , 0.55265176 )
( 7.0034197634623006 , 8.839054 )
( 5.884257892194496 , 7.5304337 )
( 6.5217918401382144 , 9.797747 )
( 5.58326004119045 , 8.289443 )
( 0.29422834658901564 , 2.0754304 )
( -0.12981704597109978 , 1.8220178 )
( 0.038041830943694066 , 1.7708564 )
( 0.40881063215363245 , 2.2231474 )
( 3.426542967294579 , 5.952403 )
( -0.9484909566289238 , 0.88992345 )
( 0.06804404806036612 , 1.497661 )
( -1.0145524673680444 , 1.2989228 )
( 1.5015880827261952 , 2.4997077 )
( 2.8573052759412034 , 3.9687521 )
( 9.219458452935616 , 10.776144 )
( 0.5368244143942041 , 2.286016 )
( 4.355831634649004 , 6.5167675 )
( 0.18293071013046092 , 2.0599022 )
( 3.196250422523403 , 5.262136 )
( 4.345084873172813 , 5.51344 )
( 2.520467966738127 , 4.0567465 )
( 7.022303363600746 , 9.505751 )
( 3.241959619206911 , 4.1532245 )
( 2.260551961518866 , 3.9021554 )
( 7.599870669057844 , 9.891939 )
( 3.1206821892354983 , 5.5006857 )
( 3.8161913075583547 , 5.741485 )
( 5.894939809451715 , 8.136132 )
( 3.7193856645207655 , 6.0266275 )
( 6.965713092123607 , 8.950949 )
( 5.631698208279772 , 8.043381 )
( -0.3315919887929628 , 1.1178169 )
( 4.994235047474497 , 6.696217 )
( 5.897214748924215 , 7.8280654 )
( 4.484277854254555 , 7.075405 )
( -1.861480115814844 , 0.48557162 )
( 8.100808604287657 , 9.950532 )
( 5.561990033500462 , 8.01115 )
( 1.1898974931511623 , 2.4555683 )
( -6.230137864499565 , -5.325497 )
( 7.91458769948004 , 9.648237 )
( 6.562499364760522 , 9.308744 )
( 7.924020661464057 , 9.3966875 )
( 3.4657567696548126 , 6.2730293 )
( -1.0131253469182369 , 0.36885178 )
( 8.153918173337296 , 9.532377 )
( 8.011330087641998 , 10.189764 )
( 5.4217228848061785 , 8.671423 )
( 6.301385079089718 , 8.342666 )
( 0.8505030277778488 , 2.919078 )
( -0.7859233667651228 , 1.6453861 )
( -1.0839744909751343 , 0.8152766 )
( -0.9320468500597869 , 0.9445369 )
( -1.782618711091846 , -1.0624957 )
( 4.50623006965518 , 6.577488 )
( 5.698790489251194 , 7.6433215 )
( 5.257528347316875 , 7.870969 )
( -0.313603203271988 , 1.4062774 )
( 7.771614177360129 , 10.126511 )
( 1.9937111460143533 , 3.6364539 )
( 5.445762247032819 , 7.2669067 )
( 5.022080931346767 , 6.3428006 )
( 0.19951050924736347 , 1.4045959 )
( 6.775356556684504 , 8.667919 )
( 0.3862617703926645 , 1.6215326 )
( 8.233215513649972 , 10.110214 )
( -0.4269751437497818 , 1.9244248 )
( 0.027307365071082 , 1.4490118 )
( -1.1343699559232412 , 0.6271368 )
( 0.4832239741317146 , 1.2431858 )
( 0.9340962959134025 , 2.8040237 )
( 2.627085270190526 , 4.3114815 )
( 5.379976084727163 , 7.452113 )
( 0.4772875052857728 , 2.4177487 )
( 0.08905097931008821 , 2.164161 )
( 7.207069468110452 , 9.087418 )
( 4.931528302241455 , 6.993718 )
( 1.8369539968596635 , 4.25959 )
( 2.909288634227189 , 5.6427817 )
( -0.8457344558119377 , 0.19363767 )
( 3.1456431646785417 , 5.682217 )
( 8.778435772379716 , 10.046533 )
( 7.689663448683343 , 9.534599 )
( 3.0935712462038794 , 5.480385 )
( 7.061850623282417 , 10.066111 )
( 6.445507296071188 , 8.52535 )
( 5.407126101815223 , 8.05065 )
( 7.2862028426000105 , 9.337671 )
( 2.594359954931955 , 4.205368 )
( 1.0909649031124928 , 3.3966112 )
( 6.142807421492974 , 8.297493 )
( 1.1947569505749391 , 2.3647203 )
( 4.582683984229626 , 6.2992578 )
( 5.513515616526465 , 7.3280897 )
( 0.15173654114674306 , 1.4540832 )
( 4.361131703751097 , 7.228545 )
( 3.4214726266619833 , 5.1465726 )
( 4.589568719346355 , 6.6745167 )
( 0.671253093898023 , 2.1004138 )
( 7.514806194145972 , 9.422288 )
( 7.124276194287938 , 8.772406 )
( 5.585961677580761 , 7.5604677 )
( 3.8739052260086866 , 6.388872 )
( -1.6586858297216072 , -0.90689445 )
( 4.135443786732309 , 6.3527703 )
( 3.6510572089128557 , 5.5395656 )
( -0.9856076642668015 , 0.44380748 )
( 0.7860755214217638 , 2.0771432 )
( -2.9881436934506818 , -1.150631 )
( 0.2080221241290936 , 1.8656878 )
( -0.7275721409457951 , -0.1121788 )
( 7.819655320345916 , 9.757435 )
( 3.3035475306329563 , 5.8158484 )
( -1.343147869482733 , 0.50711644 )
( 6.480415669700959 , 8.21763 )
( 6.764791546322475 , 8.517962 )
( 3.592512331485088 , 5.5366616 )
( 3.2411846401196414 , 5.630521 )
( 1.3966070135792705 , 2.3621013 )
( 8.098102208474536 , 9.936526 )
( 2.61185255324428 , 4.72236 )
( 1.3985140227052961 , 2.7894077 )
( 0.7805510135085597 , 1.8917251 )
( 2.482778303332326 , 4.380678 )
( 6.944341111423071 , 8.4485855 )
( 5.634343785250352 , 7.9509583 )
( 6.034258713263885 , 8.932313 )
( 3.996759079656673 , 6.518733 )
( 3.009361695483859 , 4.304762 )
( 3.15542841589615 , 6.257748 )
( 3.969255675366232 , 5.741555 )
( 7.277163323864622 , 9.134391 )
( 6.427085901907864 , 8.310783 )
( 8.162753127607061 , 10.561235 )
( 3.698517400206751 , 5.2727685 )
( 5.767646941102387 , 8.006089 )
( 0.8594074637564966 , 2.0618887 )
( 8.580063358466644 , 10.149911 )
( 3.315873240191739 , 5.6367955 )
( 8.283795573865357 , 10.143237 )
( 1.6510101844949636 , 3.3936386 )
( 5.740889465731581 , 6.987086 )
( 6.008348913846862 , 8.05053 )
( 4.851198937158236 , 7.111315 )
( -4.333680434259447 , -3.1489215 )
( 5.269858912259353 , 6.8160853 )
( 3.1892859827545172 , 5.3055124 )
( 5.609566610570077 , 6.6268296 )
( -0.42300459199689433 , 1.4397886 )
( -5.279697324823457 , -4.598753 )
( 1.1402898900273768 , 3.2096515 )
( 6.633971572504234 , 8.557628 )
( 5.639650980669303 , 7.430614 )
( 1.575106667189782 , 3.3560944 )
( 8.024820045710714 , 9.928924 )
( 7.536613539384735 , 9.80437 )
( 8.668727950618546 , 10.820593 )
( -2.9997699370174207 , -2.3817093 )
( 4.185198638723766 , 6.286399 )
( 4.855774186888225 , 7.557657 )
( 6.764028266025655 , 8.673153 )
( 7.197074017778772 , 10.053829 )
( 0.2972010925862616 , 2.0379758 )
( 5.728988040120108 , 7.322678 )
( 4.694584390778518 , 6.911218 )
( 2.0825091754802116 , 3.1934557 )
( -0.27782473790486817 , 1.9942335 )
( 3.4407135099299007 , 5.317477 )
( 0.27044797328951997 , 2.1035664 )
( -0.10305955987728843 , 1.7571113 )
( 5.937450646185493 , 10.01883 )
( 2.245954796095088 , 4.2859583 )
( 6.468275552760025 , 8.122393 )
( 0.10148684497548288 , 1.4625659 )
( 3.962822467698664 , 6.0937552 )
( -8.575570813006808 , -6.5521007 )
( 7.432015529573616 , 9.128836 )
( 7.15320559444487 , 9.104468 )
( 6.580409678493231 , 8.727601 )
( 2.7089213812927486 , 4.4895196 )
( 7.494245224364793 , 9.013769 )
( 3.971971211166494 , 5.560586 )
( -1.310163823492471 , 0.67053545 )
( 2.195005433564788 , 4.7369223 )
( 4.424420324514264 , 6.7185516 )
( 1.3250087053484476 , 3.3349407 )
( 3.6157508122114645 , 5.5059547 )
( 8.111944816892722 , 9.959192 )
( 4.809488142506297 , 7.072799 )
( 5.3528694612293615 , 7.1256013 )
( 8.074255326745986 , 9.93939 )
( -5.036652750805067 , -4.9878883 )
( 6.799023221307781 , 8.953023 )
( 0.4842734465285048 , 1.9072232 )
( -0.6807215916598376 , 1.2725062 )
( 0.3391016419927628 , 1.8615284 )
( 0.30695453329751476 , 2.1497428 )
( 1.989391232200274 , 3.8536687 )
( 5.741765168140211 , 8.015368 )
( 2.058901163008797 , 4.4660263 )
( 5.409069259618547 , 7.377177 )
( 7.960705380199065 , 9.689433 )
( -3.5696849425253028 , -3.281602 )
( 7.6788128693097 , 10.008057 )
( -0.8112591349307039 , 1.0601553 )
( 1.101978508147054 , 2.6411796 )
( 2.6285679810691542 , 3.8518457 )
( 1.0783652365220808 , 2.704401 )
( 7.4732174411246906 , 9.231181 )
( 3.47231666480754 , 5.4967527 )
( 7.2412005457190896 , 9.354037 )
( 8.666897534478453 , 10.569759 )
( 8.343778302335693 , 9.891026 )
( -0.46512612394413755 , 1.0449532 )
( 8.807363703418494 , 10.668151 )
( 6.3792336311848095 , 9.187259 )
( -10.437221645700149 , -9.497181 )
( 3.629503518811924 , 5.075354 )
( 4.746900388581092 , 7.864998 )
( 0.15215903926326035 , 2.3283005 )
( 0.5634830964420938 , 2.3397207 )
( 4.4001327782623605 , 5.748893 )
( 9.26421111124202 , 10.758263 )
( -0.3917148535655173 , 0.6508542 )
( 3.901191564160495 , 5.9471426 )
( 3.0996279129631175 , 5.157268 )
( 8.51613751637073 , 10.832279 )
( 5.733939405678729 , 8.635957 )
( 2.4801394286970595 , 3.895092 )
( 4.071071137260185 , 6.2609663 )
( 3.5681700925702424 , 6.0848794 )
( 5.124374737752694 , 7.943658 )
( 1.2019172880292346 , 3.1009808 )
( 6.130227380192603 , 8.320117 )
( 6.3067799987182624 , 8.760668 )
( 3.2717690061970295 , 5.7310095 )
( 2.7349584858773444 , 3.3010454 )
( 8.962820573163523 , 10.241283 )
( 2.995617081083328 , 5.4041166 )
( 2.915935001366978 , 5.0415726 )
( 1.8644158545532985 , 3.3371067 )
( 8.336129179602093 , 10.328108 )
( 6.085658676439345 , 9.13058 )
( 0.5749252501844584 , 2.3434837 )
( 5.310152447615155 , 8.088977 )
( 7.158445944291447 , 8.982983 )
( -1.6370201491541465 , -0.06172645 )
( 0.9986431823481587 , 2.339727 )
( 5.1305873227938905 , 6.9131722 )
( 6.155760579151626 , 7.875093 )
( 6.769782977998894 , 10.072439 )
( -0.9822414878575332 , -0.035600126 )
( 1.3344636175231686 , 2.567616 )
( 4.554871489740615 , 6.7521772 )
( 2.909827855188614 , 5.2738776 )
( 4.5175169712843175 , 7.1057997 )
( 7.243242739952342 , 10.084307 )
( -1.037852104233253 , 0.61636686 )
( 1.0532549579469026 , 2.1309938 )
( 5.01416710988768 , 7.7357645 )
( 8.511830381716779 , 10.75889 )
( 5.896195549242847 , 7.677434 )
( 1.1840946246274253 , 2.937995 )
( 6.591521975808543 , 8.684051 )
( 4.532828086221569 , 6.8147655 )
( 4.166008129046396 , 6.5530057 )
( 4.146551897937692 , 7.2738767 )
( 2.5971589184437374 , 4.9140134 )
( 1.4830701324221194 , 2.9694166 )
( 4.005073502282227 , 6.4749928 )
( 5.105853296062 , 8.903135 )
( 0.9975795103096412 , 2.286749 )
( 3.126753842069456 , 4.8234534 )
( -7.439218403173073 , -6.4153023 )
( 6.799888093300674 , 8.769689 )
( 3.981398885365243 , 5.793015 )
( 3.7897418956222566 , 5.9754133 )
( 6.072761189641484 , 8.060052 )
( 5.295036757229746 , 7.36406 )
( 2.6653308686888124 , 5.021309 )
( 4.116420866323832 , 7.8984184 )
( 0.48398987865715637 , 1.240972 )
( 6.6671629632631255 , 8.49135 )
( 1.596662606196478 , 3.386372 )
( 0.681479149294133 , 2.6137824 )
( 1.360549839281143 , 3.3845143 )
( 5.48290609053849 , 7.2876215 )
( 3.5007163483593216 , 4.9428563 )
( 1.3658614854818878 , 2.863501 )
( 4.742767114063593 , 6.6162863 )
( 1.900833687213928 , 3.1569633 )
( 3.0900035367463516 , 4.2646537 )
( 5.334425461443245 , 8.193016 )
( -0.25184200642811927 , 1.3427901 )
( 2.471905399350013 , 4.834255 )
( 1.775837184711735 , 2.8768291 )
( 5.5819669243569985 , 7.793804 )
( 1.3002937481563142 , 2.854887 )
( 6.225433530450647 , 8.203783 )
( -7.047127941660372 , -6.057366 )
( 5.801280751712154 , 9.070853 )
( 3.8385130706510076 , 5.379065 )
( 3.129081882439874 , 5.6427107 )
( 3.1381742517297635 , 5.2970567 )
( 0.911672035089985 , 3.0283012 )
( 0.39746992241505763 , 1.9380429 )
( 3.683233107893736 , 5.739229 )
( -0.9879722717985144 , 1.0152361 )
( 5.552722441747106 , 7.4821067 )
( 1.4879642700392357 , 2.795986 )
( 6.493739219342513 , 8.67711 )
( 1.71769140109893 , 2.9501548 )
( -1.5754339828219046 , 0.33071506 )
( 5.997787280696301 , 8.132247 )
( 4.859847991073047 , 7.178142 )
( 6.02204310872035 , 8.375385 )
( 5.878205167653274 , 7.890745 )
( 8.189840437294244 , 9.938817 )
( 3.3225445675925247 , 4.9843187 )
( 2.9872772134142846 , 5.4799047 )
( 1.7683474309917608 , 3.9460168 )
( 7.274587817637419 , 9.420122 )
( 5.014556318205441 , 6.9483757 )
( 0.021949093411494758 , 1.7209643 )
( 1.3791536331558263 , 3.3760266 )
( 2.749995527729252 , 5.6761527 )
( 2.6807466158951563 , 3.7400718 )
( -1.063042384348054 , 0.3747679 )
( 2.4163439473747848 , 3.6549282 )
( 2.1350668529301178 , 2.2783196 )
( 6.771859124504322 , 8.87779 )
( 3.8114009671871916 , 6.9400544 )
( 8.595778599032936 , 10.374266 )
( 4.331827608436718 , 6.3896947 )
( 8.694604944621343 , 10.285299 )
( 1.1174953867118749 , 2.5928874 )
( -0.8633201353072568 , 1.0035132 )
( 7.971782557729421 , 10.477749 )
( 3.164120747466886 , 5.6735115 )
( 6.449059586758069 , 9.471424 )
( 0.6244241847319878 , 2.2266648 )
( 5.608020774523073 , 7.3438478 )
( 7.279243521793322 , 9.0357 )
( 5.46935513329133 , 6.9040194 )
( -5.265942177958036 , -4.59299 )
( 0.32351073124181473 , 2.1693354 )
( 6.545221502876821 , 8.609738 )
( 6.395752252102344 , 8.533574 )
( 1.1857569710415499 , 2.9431558 )
( 2.3132684916612885 , 3.8263893 )
( -1.5244140642080222 , 0.776014 )
( 7.97112358550375 , 10.089918 )
( 1.930134676963441 , 3.5677414 )
( 4.731446250408343 , 7.0711336 )
( 2.050535438956391 , 4.3639383 )
( 8.332102242749624 , 10.588114 )
( -0.5749081743185755 , 1.4098424 )
( 4.533885930835709 , 7.9926767 )
( 9.312465295026502 , 11.287827 )
( 5.863114856857048 , 8.623028 )
( 7.546016634538164 , 9.497785 )
( -3.9227885916115195 , -3.5067618 )
( 2.0394474555596576 , 3.623229 )
( 0.581530273852234 , 1.1085676 )
( 0.2196068267587522 , 1.8237125 )
( 2.475244755748187 , 4.140758 )
( 8.2664019341911 , 10.052313 )
( 2.190214619285841 , 5.3143663 )
( 7.466424746058265 , 9.441689 )
( 5.658164198018479 , 7.31738 )
( 4.403602631218711 , 6.9347076 )
( -5.224978802992986 , -3.6766648 )
( 8.625333882040986 , 10.498941 )
( 4.717067634553461 , 8.141573 )
( 8.701105912243985 , 10.498534 )
( 0.6829439384432998 , 1.4447472 )
( 4.275607443358914 , 7.201343 )
( -0.08250282069655374 , 2.1965523 )
( 7.662676154006775 , 9.677869 )
( 7.6449865281401825 , 9.442156 )
( 0.8213267527476028 , 2.14018 )
( 7.736175343362053 , 9.320839 )
( 8.342159482651148 , 10.262689 )
( 2.8456036768643562 , 4.5845156 )
( 2.6963482879483918 , 4.5414777 )
( 6.83656710186872 , 10.028006 )
( 4.482978576874089 , 6.2197995 )
( 5.5087617221936 , 8.041441 )
( 1.8058079969005956 , 3.787036 )
( 7.502734286502235 , 9.043807 )
( 4.5511530335982915 , 6.469219 )
( -0.3266378831588659 , 0.4795593 )
( 0.34852643591641924 , 2.2000556 )
( 4.688043987506755 , 8.108756 )
( -0.5164141538124996 , 1.4872637 )
( 7.108832456392361 , 9.506185 )
( -0.3703417218571583 , 1.3934872 )
( 2.4701499904104214 , 3.9444127 )
( 1.344211838775722 , 3.7441595 )
( 2.5784112490213635 , 4.266888 )
( 6.070397627215414 , 7.86329 )
( 4.08445533880795 , 6.0572476 )
( -0.22542072572583272 , 1.1904277 )
( 6.466550946048175 , 8.52906 )
( 0.05979029613621627 , 1.3855075 )
( -3.8457693107436626 , -1.7476132 )
( 6.301235527873694 , 8.459588 )
( 0.8384084943593229 , 2.1564941 )
( -0.3811887327150716 , 1.0909729 )
( -0.3790195901223447 , 1.7266523 )
( 3.3648486295142765 , 4.9334507 )
( 3.1358569116311097 , 5.340845 )
( 5.426002499733469 , 7.685283 )
( 8.812234323321725 , 10.499739 )
( 1.1408548492768396 , 3.3881445 )
( 6.4719675561038015 , 8.378206 )
( 5.075685770545606 , 6.890606 )
( 0.19647626285131448 , 1.8703078 )
( 3.892928324931664 , 5.7110953 )
( 2.683407633627168 , 4.636362 )
( 6.748026450849453 , 8.647892 )
( 6.392600500447838 , 8.168314 )
( 3.5818439435015135 , 5.9108253 )
( 0.7650937026115008 , 2.6420693 )
( 0.99219119179527 , 3.0105429 )
( 5.846027264664226 , 8.01601 )
( 2.884827621962347 , 5.6083593 )
( 4.962376271645355 , 6.859376 )
( -1.001470230760968 , 0.89788806 )
( -1.5929285362910726 , -0.21997638 )
( 6.571752296667191 , 8.878479 )
( 5.426542368888581 , 7.434396 )
( 0.5812839223241242 , 2.528026 )
( 5.549019686858804 , 7.872548 )
( 8.611423977018035 , 10.655916 )
( 2.8354116822532713 , 4.336446 )
( 6.122294722659488 , 8.214664 )
( 7.893478809954566 , 9.5584755 )
( 0.6220021003642806 , 2.5648837 )
( 0.2806964171607157 , 2.5040395 )
( -0.6954836129096513 , 0.47707242 )
( 1.0186738403981115 , 2.044118 )
( -1.7419358622075105 , -0.96728706 )
( 3.416587801382687 , 4.751027 )
( 1.4180806855703523 , 3.223399 )
( 2.2599741183650073 , 4.189837 )
( 6.249004783412754 , 8.234364 )
( 4.397152771695386 , 7.915432 )
( 0.7803610188826315 , 2.137908 )
( 0.23949248957757596 , 2.3680506 )
( 7.259615850834216 , 10.109549 )
( 6.747536956818303 , 8.695627 )
( -9.619736345819268 , -5.8665857 )
( -1.1290545204923663 , 0.48442948 )
( 0.18453391131060626 , 1.4341 )
( 2.7576507858120998 , 5.1920786 )
( -2.2025592371228795 , -1.8716085 )
( 7.40519542264338 , 9.502107 )
( 2.071981326475287 , 4.3595376 )
( 5.953170096651236 , 8.288518 )
( -1.4181221561088753 , 1.1524379 )
( 5.503889141244559 , 7.070734 )
( 3.679803271051624 , 5.659856 )
( 5.085710288660236 , 7.03735 )
( 3.1602498312622034 , 4.9238796 )
( -0.17072681981402837 , 1.5823537 )
( 6.157057163125378 , 8.412051 )
( 4.293359944999301 , 5.98611 )
( 5.943076598467691 , 8.118467 )
( 6.717191927808272 , 8.957792 )
( 5.70206384092841 , 7.8162074 )
( 7.080939940798878 , 9.105366 )
( 3.8507419056964682 , 5.7364807 )
( 4.874074235990611 , 6.9131327 )
( 5.995151954756908 , 7.900775 )
( 5.562007547978609 , 7.8841047 )
( 2.0287331712064463 , 3.2719543 )
( -0.5756662140352468 , 2.1219077 )
( 7.206097375549581 , 9.727852 )
( 5.063530258450487 , 7.557478 )
( 7.794261899573549 , 9.449179 )
( 5.635284631293987 , 8.108257 )
( 2.631961583243885 , 4.494267 )
( -0.13270730574559053 , 1.1123174 )
( -1.4041260499823638 , 0.74455154 )
( 4.781604518023812 , 7.184207 )
( 5.061792994362303 , 6.8520236 )
( 3.7967887999214804 , 6.040162 )
( -0.9170670861465926 , 0.8186146 )
( 2.3984446395342593 , 4.0397596 )
( 1.0403311717216241 , 2.5549567 )
( 7.573532995784509 , 9.653641 )
( 6.531591131941996 , 7.9941177 )
( 0.43287411489203625 , 2.4174829 )
( 5.880681387737205 , 9.112916 )
( 0.7457143239334482 , 2.1586378 )
( 0.8562030468105234 , 2.7199862 )
( 6.251780018590224 , 8.961838 )
( -8.951826266327188 , -7.021474 )
( 3.7496386987232326 , 7.0215816 )
( 2.3421287073115 , 3.831603 )
( 3.8528088846303286 , 5.2954817 )
( 2.8423686297572637 , 5.0642138 )
( 6.908882324707088 , 8.488002 )
( 7.1431344382942985 , 9.627097 )
( -0.4952406163788685 , 1.9252588 )
( 6.425140098123507 , 7.9000335 )
( -0.28078579790237335 , 1.2979143 )
( 6.825117813014781 , 9.10873 )
( 4.467634824141519 , 6.486085 )
( 0.7643653079046813 , 2.274494 )
( 5.7288521996524455 , 7.6728816 )
( 2.6355246828735934 , 3.718731 )
( 1.312347594344499 , 2.452824 )
( 7.1564394875389965 , 9.192915 )
( 5.325740843880499 , 7.6588173 )
( 7.105105570822241 , 8.660901 )
( 6.6998837701259735 , 8.581673 )
( -7.489716220397302 , -7.0923905 )
( 7.686440170919016 , 10.642979 )
( 1.8825017002979392 , 3.2286983 )
( 8.289803656983274 , 10.300877 )
( 1.5570199862835385 , 2.6543403 )
( -11.01287074171087 , -9.131075 )
( 7.182490697470325 , 8.914099 )
( 8.807934587908681 , 10.025459 )
( 3.582486720994801 , 5.9457307 )
( 5.062651471375654 , 7.0204043 )
( -3.9589630700545664 , -4.0070004 )
( 0.1749816065665585 , 1.5828613 )
( 3.686370459931357 , 6.1037397 )
( 5.803068797435925 , 7.6849427 )
( 4.323619852869415 , 5.9191647 )
( -0.13864754705374427 , 1.2850047 )
( 6.703282204143671 , 8.515431 )
( 1.7326988553652443 , 4.0957966 )
( 0.326456935858493 , 1.665121 )
( 6.776496382255013 , 8.940394 )
( 7.079544498704168 , 9.12663 )
( -3.48492228376138 , -2.0971804 )
( 1.7810780063897753 , 3.0512059 )
( 8.77903234925611 , 10.614533 )
( 5.953179207000205 , 7.663352 )
( 0.8592441211562654 , 2.636528 )
( 7.649844541376083 , 9.327285 )}

\newcommand{\utilitylarge}{
( 0.2676768836317993 , 1.899143 )
( -1.3910324681900308 , 0.5019625 )
( -8.937480172661521 , -8.484598 )
( 5.749442652414051 , 8.0937605 )
( 6.891810343731568 , 8.3413925 )
( 6.507949783217706 , 7.4840584 )
( 6.144103035384285 , 7.812619 )
( 6.816424821888014 , 8.340959 )
( 3.4854345622751826 , 5.259248 )
( 1.0361119648962416 , 1.5865933 )
( 8.054097235893583 , 10.346437 )
( 4.670134201908244 , 7.521714 )
( 8.781796981838761 , 10.225593 )
( 4.659766354476239 , 6.277074 )
( -2.2917948105728136 , -1.4538753 )
( 2.126971429940478 , 3.4547462 )
( 0.035574980682075985 , 0.5808505 )
( 3.2741180794561346 , 5.013838 )
( 5.417361712407309 , 6.807988 )
( 5.706608462854018 , 7.446507 )
( 7.7008976443924855 , 9.497885 )
( 5.890854913827388 , 7.100503 )
( 4.635860617305445 , 6.3949337 )
( -0.00633232290850827 , 0.013616562 )
( 5.330455374610794 , 7.2097073 )
( 2.241714797461916 , 3.527208 )
( -0.10010578191149244 , 1.0801013 )
( 1.4990409379348595 , 2.4286184 )
( 5.825251590519721 , 7.557107 )
( 2.246882906138029 , 2.9506521 )
( 1.5203160341951762 , 2.2966018 )
( 0.34961138183561546 , 0.84121275 )
( 1.1973893585740374 , 1.936652 )
( 6.586282621233469 , 7.9471107 )
( 8.264655767053618 , 9.917084 )
( 4.461134747589333 , 5.5442877 )
( 7.262721563929104 , 9.647296 )
( 5.013262208406412 , 7.1054325 )
( 2.3393496127544138 , 4.1221237 )
( 4.91004641749204 , 5.7096596 )
( 6.990328989320571 , 8.973173 )
( -1.2198144913926594 , -0.464213 )
( -9.042410403820728 , -8.709844 )
( 5.035935911046796 , 6.82432 )
( 8.149293242002262 , 9.103204 )
( 3.5352994717246347 , 5.2338676 )
( 7.789463143742424 , 9.088967 )
( -9.67681802684643 , -9.889471 )
( 2.990426266918986 , 3.8715448 )
( 4.770975920388231 , 6.5307484 )
( 0.4107177377514951 , 1.5993328 )
( -0.24218668137076726 , 1.0609442 )
( 7.434055839571792 , 9.730985 )
( 5.300797028704794 , 6.815674 )
( 3.987354219890465 , 6.1062794 )
( 6.943160816835565 , 9.355733 )
( -0.24001589099959536 , 0.11251646 )
( 7.573566576066656 , 9.360419 )
( 8.52154058081566 , 10.520954 )
( 6.870987175435958 , 9.436778 )
( 6.268242745235366 , 8.5924635 )
( -0.18443829083197327 , 0.46311343 )
( 0.605826515988406 , 1.9279222 )
( 3.0737328698228144 , 4.1290493 )
( 3.3873604118493548 , 5.086327 )
( 0.8845794866531771 , 1.9622531 )
( 0.094090850472732 , 0.40505362 )
( 6.190662386314384 , 7.5296745 )
( -1.9781119101886349 , -2.0371304 )
( 4.286964438903956 , 5.5829263 )
( 2.0371468598635616 , 4.0089593 )
( 4.835938426146559 , 6.513024 )
( 7.575331976011167 , 9.561746 )
( 6.028099789920599 , 7.820571 )
( -2.1226508009386764 , -0.8527056 )
( 7.7974337190226715 , 10.036457 )
( 0.7293452492793208 , 1.9105629 )
( 2.0177215776243 , 3.0724287 )
( 1.1658940833142832 , 2.2375145 )
( 8.514987551249266 , 10.069809 )
( 6.293848514183121 , 8.173966 )
( 4.8429008940369895 , 7.2667503 )
( 7.405281975268689 , 9.685272 )
( 5.096440043547555 , 6.2547894 )
( 2.129283018049639 , 3.4184022 )
( 2.4003631079296395 , 3.3223267 )
( 7.754607914338543 , 9.562152 )
( 5.450374435147729 , 6.779802 )
( 0.09806241871263455 , 1.8062057 )
( 3.4443959511103035 , 4.6295967 )
( 0.4045072959739514 , 1.1200302 )
( 3.9812004502376612 , 5.713035 )
( 2.670948639898037 , 3.5571446 )
( -6.202579021048489 , -6.501514 )
( 0.6106154427116954 , 1.8436233 )
( 5.6953031419881786 , 7.0670924 )
( 6.311409638786939 , 8.004592 )
( 7.4430003051860965 , 9.14087 )
( 1.0278914073269831 , 2.1959221 )
( -0.8911953998242435 , 0.38530934 )
( 5.06661092130334 , 6.6585197 )
( 5.855617888085803 , 7.988327 )
( 5.836307425775854 , 8.151611 )
( 4.947504092463888 , 6.5407343 )
( 1.5095068007629253 , 2.6629858 )
( 7.104821931785835 , 9.2661085 )
( 0.47792012547422746 , 2.284224 )
( 0.06945965338814708 , 1.1065149 )
( 3.6280763677812384 , 4.3072186 )
( -1.5632838268450429 , -0.8389395 )
( 6.479172789397314 , 8.528431 )
( 2.52593145092612 , 3.407236 )
( 5.633289194155285 , 6.9572773 )
( 2.1792875490410504 , 3.7678676 )
( 7.256934576450996 , 8.979959 )
( 7.223554423484355 , 8.348625 )
( 0.011986661829006096 , 1.2756937 )
( 3.808543283570783 , 5.4717326 )
( 1.5875815858092082 , 2.7652977 )
( -3.5051194536708525 , -3.2605999 )
( -2.963390022967295 , -3.5470316 )
( 3.7186189185322487 , 4.5556417 )
( 7.253743834414072 , 9.120669 )
( 2.475002539405235 , 3.5575624 )
( 3.750759810389197 , 5.140738 )
( 3.3907067245023956 , 4.862829 )
( 5.57106324599616 , 6.7558985 )
( 2.2048408643683626 , 2.7587843 )
( 3.7022400786329914 , 6.1330585 )
( 2.480902656732841 , 3.9247246 )
( 4.541787117090942 , 6.400218 )
( 0.9503106433347589 , 1.6901625 )
( 4.277398412117946 , 6.4676814 )
( 3.672379100859513 , 4.820248 )
( -3.9697186006714063 , -4.114892 )
( 6.966574021674121 , 9.253544 )
( 4.949016346950014 , 6.8178577 )
( -0.7376350298710882 , 1.2780836 )
( 7.253713327485091 , 9.341608 )
( 6.208593021962292 , 8.762359 )
( -4.113354192751081 , -3.8493085 )
( -1.9659081150789117 , 0.25610483 )
( 2.3583639629056545 , 3.1260214 )
( -0.6644182742033116 , 0.36647117 )
( 0.39096122589377647 , 1.328817 )
( 7.7599668616618445 , 9.246902 )
( -0.8202870429949162 , 1.1790988 )
( 5.440341521149703 , 6.647279 )
( 0.4619511916524053 , 1.02391 )
( 4.26885158670816 , 5.8589444 )
( 1.2800542348565491 , 2.816727 )
( 7.258918383760248 , 8.865874 )
( 2.2312533517496513 , 3.2824535 )
( 3.7081084510552946 , 4.9199486 )
( 5.201975406018853 , 6.9624534 )
( 1.522237457318805 , 2.3434148 )
( 2.941756254620538 , 4.0523815 )
( 5.506838240010387 , 7.3429885 )
( -1.4293354467484038 , 0.35934785 )
( 2.098130895040582 , 3.7476707 )
( 3.613635087880109 , 5.043421 )
( 5.213500941601739 , 6.4625 )
( 4.2468317966711195 , 6.4383564 )
( 1.9171389154471867 , 2.8348665 )
( 8.566633987378557 , 10.157208 )
( -1.6203647536920007 , -0.45033717 )
( 1.7110126396972825 , 3.19882 )
( 2.217116193522877 , 3.6529338 )
( -2.4407617326650004 , -2.179216 )
( -4.081875150783875 , -4.232543 )
( 8.120655803296842 , 10.091657 )
( 5.257586618995882 , 6.925396 )
( 1.2475351367823793 , 2.9911773 )
( 0.4016883557141555 , 1.381901 )
( -5.729736973729846 , -7.8183756 )
( 0.5386354175986505 , 1.638948 )
( -2.0232284892451826 , -2.6352465 )
( 8.420730543716127 , 10.800274 )
( 5.474797958984503 , 7.8112426 )
( 8.806161947185453 , 10.484318 )
( 2.928874565101916 , 4.6483536 )
( 0.9200329594311141 , 1.8006856 )
( 3.922188805469382 , 5.2482195 )
( 8.180835293751983 , 10.391851 )
( 3.7762219938519475 , 5.253971 )
( 0.03709938490052478 , 0.5678474 )
( 3.3614352345594085 , 4.446369 )
( 2.5538050932817917 , 3.7396874 )
( 7.441796903035247 , 9.153809 )
( 4.486654636133456 , 6.204171 )
( 0.2983752851087048 , 0.628242 )
( 1.15412745563486 , 1.9878316 )
( 6.305311041848304 , 8.223242 )
( -1.7354562066284753 , -1.4987011 )
( 8.32989256385774 , 10.80271 )
( 1.7851704526575878 , 2.663212 )
( 0.07424073045536095 , 0.9488364 )
( 8.182660504829279 , 9.729534 )
( 2.8106749017164 , 3.4745069 )
( -0.3523877366456328 , 0.6375266 )
( 8.21221844648461 , 9.878235 )
( 1.6086948059849768 , 2.6255124 )
( 7.25954304084537 , 8.802287 )
( 6.370165152554241 , 9.16581 )
( -0.42244098051768475 , 1.356662 )
( 7.782918143338263 , 9.451226 )
( 1.283513279431482 , 2.749512 )
( 5.108275994125956 , 7.0406632 )
( 6.69256068898618 , 7.8375216 )
( 7.328771651005318 , 9.060643 )
( 1.5180334730464906 , 2.8843124 )
( 7.419994131976285 , 9.300365 )
( 6.025271582932866 , 7.5982575 )
( 5.336528575210153 , 6.9122667 )
( 0.10081572525438176 , 0.9022456 )
( -3.1164905594823313 , -2.0745378 )
( -0.2109976594997009 , 1.3613863 )
( -0.24412299644987043 , 0.94641894 )
( 5.37884408193659 , 7.2965584 )
( -0.7494958829239231 , -0.43207222 )
( 2.388313142846553 , 3.6171727 )
( 1.7846513350581148 , 2.7792878 )
( 5.440235792480619 , 6.6054583 )
( 0.558048374992988 , 1.7912996 )
( 8.580174717955835 , 10.3846655 )
( 1.2438049862221776 , 2.232245 )
( 3.59170023561102 , 5.35089 )
( 5.337438259524756 , 7.3139205 )
( 2.4060761878287886 , 3.4831216 )
( 4.446129804915182 , 5.9101396 )
( 7.235815168904421 , 9.084971 )
( 9.405199353232666 , 11.577782 )
( 0.8390495887795444 , 2.114863 )
( 4.442173884515984 , 6.1460876 )
( 6.0107328602449535 , 7.773321 )
( 2.6106269650590606 , 3.6105723 )
( 6.29003312834007 , 7.427737 )
( 1.2915725265609144 , 2.217125 )
( 4.7527004169148075 , 6.41196 )
( 7.475250664593411 , 10.399 )
( 8.536817954754035 , 10.518187 )
( 2.919248408522611 , 3.828115 )
( 0.05217678760017841 , 1.4037151 )
( 0.2509566301280288 , 1.1941926 )
( 2.5194673806306604 , 4.723918 )
( 7.719820091929492 , 8.775702 )
( 1.522949998969045 , 2.2475371 )
( 1.825643198490536 , 2.5911508 )
( 0.28426180525791667 , 1.6468813 )
( 2.43102895903682 , 3.2169456 )
( 5.047594735850084 , 6.1823072 )
( 8.916845061407068 , 10.867254 )
( 6.945985617873937 , 8.969507 )
( -0.9732080040970796 , 0.15507674 )
( 2.455856685016883 , 3.8160095 )
( -3.037478883774261 , -1.8923697 )
( 3.539002364928046 , 3.7928157 )
( 4.475667236002017 , 5.7539773 )
( 1.2995269554059536 , 2.7738457 )
( 4.335829618557123 , 6.453164 )
( -0.9387249205498538 , 0.20673943 )
( 4.162888804645874 , 4.8799324 )
( -0.003695436481622849 , 0.2614398 )
( 2.545535423871721 , 3.8741856 )
( -11.117845589056639 , -10.693125 )
( 7.574677912946466 , 9.262185 )
( 2.746049955938565 , 5.7084546 )
( -1.242157795082518 , 0.1875163 )
( -0.8618651018022133 , 0.78772193 )
( 4.856865927717134 , 6.973886 )
( -1.8399085973828004 , -1.3200879 )
( 6.9227052400387565 , 8.963575 )
( 0.38739077290497004 , 1.4897096 )
( 0.3473328252936323 , 0.5654011 )
( 6.914238597525393 , 8.594238 )
( 0.24403380859685964 , 1.0413408 )
( 3.8628310840301365 , 5.1556144 )
( 8.548960846397414 , 10.526242 )
( 6.855049222257766 , 9.383962 )
( 3.9817606967853756 , 6.3901553 )
( 5.7994411977760025 , 7.9529877 )
( 1.2810910751368496 , 2.4724302 )
( 0.03908917180067584 , 0.4830581 )
( 3.4815359765055858 , 4.7974086 )
( 2.9721499089349046 , 4.688219 )
( -0.696261072567994 , 0.6456366 )
( 5.513782393199161 , 7.286535 )
( 6.267688100680255 , 8.776744 )
( 2.0418777055653847 , 3.22088 )
( -4.77565744811427 , -4.385228 )
( 6.767686735179783 , 8.8856125 )
( -1.3220716139598245 , 0.51784915 )
( 6.820090380680488 , 7.7115507 )
( -2.677094435890613 , -1.0415063 )
( 1.3192856244223274 , 2.0316923 )
( 7.267337074326019 , 8.837363 )
( -0.40808075210172867 , 0.26967734 )
( 8.206156740820592 , 10.638466 )
( 0.6322469088198712 , 1.6997781 )
( 8.021103795285306 , 9.602812 )
( 4.973118237234678 , 6.731815 )
( 4.448385923540285 , 6.0007234 )
( -0.02199770347320107 , 1.1215727 )
( 1.7179096344313098 , 3.0688798 )
( 0.832174931794368 , 1.8910196 )
( 2.1511011011678045 , 2.5962114 )
( 5.970377987900401 , 7.770643 )
( 4.760904356287516 , 5.8571386 )
( 3.231979654132901 , 4.088352 )
( 4.5376279352144575 , 6.1236787 )
( 6.427896470127056 , 8.108857 )
( 4.876203817844106 , 6.459942 )
( -0.33290786170029263 , 1.5119278 )
( 0.8043722565572329 , 1.7022069 )
( 3.1019413368618833 , 4.5748825 )
( 2.7298630938542465 , 4.204263 )
( 4.853352388767605 , 6.7460823 )
( -2.5208545165025202 , -0.8902466 )
( 5.797529665575164 , 7.454955 )
( 2.5760889487575307 , 3.7883666 )
( -0.18972927303844553 , 0.45109755 )
( 4.728681012248585 , 6.6440926 )
( -0.16517029471887434 , 0.64780426 )
( 6.15410514879552 , 7.651179 )
( 3.721256158749855 , 5.8905096 )
( 3.3449144829207134 , 4.4935136 )
( 1.0373104061855787 , 2.5115674 )
( 4.585645849657365 , 6.763125 )
( 1.1159506831087205 , 1.9124115 )
( 2.2780916114170546 , 2.885273 )
( 4.474224694385321 , 6.1809373 )
( 0.0953871201941191 , 0.41570318 )
( 3.858147853759852 , 4.9432673 )
( 1.441823250724911 , 2.711626 )
( 4.027676470627191 , 5.4281573 )
( 7.137802280153975 , 8.577623 )
( 5.533865865429053 , 6.6348934 )
( 8.467344497627119 , 10.398193 )
( 3.1348093566863477 , 4.32628 )
( 3.944689542386139 , 5.2063313 )
( 5.700390937826158 , 7.48201 )
( 3.8001940650483252 , 5.423455 )
( 4.2158864341280635 , 5.9882026 )
( 6.948398515323394 , 9.499909 )
( 6.06421768247725 , 7.5753913 )
( 1.3968320864091814 , 2.3398051 )
( 7.236784748215651 , 9.323372 )
( 3.998632140656552 , 5.619939 )
( 1.4733189438740606 , 3.1069145 )
( 4.182629082116691 , 5.721725 )
( 5.7358266649504595 , 7.3724356 )
( 5.222875135796961 , 6.413835 )
( 6.962177940908919 , 9.58529 )
( 2.0928867524727677 , 3.5877361 )
( 4.339612486297609 , 6.50489 )
( 2.1745903996841305 , 3.0885816 )
( 2.481148703852342 , 2.237668 )
( -0.09300515726508696 , 1.1288334 )
( -0.1604761103345081 , 1.305723 )
( 8.839492147594768 , 10.284446 )
( 7.505729233239036 , 9.618779 )
( 8.662737012893377 , 10.72677 )
( 3.503210151678705 , 4.8072166 )
( 2.3042051975153037 , 3.6362288 )
( 2.4650324492555873 , 4.0729704 )
( 5.2432588467838865 , 6.7484713 )
( 1.815629622613181 , 2.9468374 )
( 7.956636241117511 , 10.273806 )
( 5.811371550234936 , 7.837209 )
( -5.842394848267936 , -6.6420755 )
( 4.712009279143475 , 6.4151797 )
( -3.840615927891837 , -2.823606 )
( -0.1668979307594185 , 1.1742188 )
( 3.5901660035451846 , 4.91146 )
( -0.7807913366435706 , 0.4196127 )
( 4.16965298960069 , 6.067012 )
( 5.069203585586157 , 7.097879 )
( 4.476676434119617 , 6.353738 )
( 4.946249997006223 , 6.158972 )
( 2.0668907741434044 , 3.3203192 )
( 1.385713983881462 , 2.3038237 )
( 6.178963034821699 , 8.949407 )
( 4.283003167150721 , 5.5774055 )
( 6.835222485552192 , 8.779322 )
( 7.0187894237663855 , 8.270005 )
( 3.0172747537474227 , 3.9601867 )
( 2.343161289625655 , 3.2978857 )
( 7.086178868707142 , 8.997285 )
( -3.4736069594393433 , -2.2225785 )
( 6.262241392965121 , 8.272099 )
( 4.431433908418339 , 5.7135916 )
( 3.664632222983875 , 5.2804418 )
( 3.599714704665939 , 5.207717 )
( 0.7527527328491762 , 1.7600921 )
( -0.013627553637451939 , 0.6801225 )
( 0.45242766180326965 , 1.879798 )
( 6.360597442291068 , 7.2765284 )
( 4.722532788122859 , 6.5167685 )
( 7.428796554877485 , 9.586628 )
( 4.686064474168175 , 6.443986 )
( 7.144478209798695 , 8.55452 )
( -0.5044191987529233 , 0.83571553 )
( 5.241235530229169 , 7.5357914 )
( 0.6077458552489645 , 2.2190435 )
( -1.7494570674221444 , -1.697958 )
( -0.3085359025614005 , 0.9214944 )
( 3.597309139508509 , 4.8455143 )
( 6.079418824973439 , 7.9367924 )
( 2.664052273559679 , 3.7206488 )
( 0.09188721535401179 , 1.3554476 )
( 3.2856718952302266 , 5.6478834 )
( 5.771290700997784 , 7.4696107 )
( 4.0357030513806835 , 5.7028847 )
( 4.336833643644864 , 5.8193307 )
( 1.9426217994612878 , 3.4371145 )
( 6.624054400583452 , 8.915923 )
( 1.8090750599968404 , 2.8129237 )
( -0.07226162260812918 , 0.50517607 )
( 5.417246007865398 , 6.990072 )
( 6.219867833521878 , 7.9686317 )
( 5.265101677378989 , 7.9991813 )
( 7.043692292311279 , 8.529562 )
( 4.457690913728479 , 6.137648 )
( 2.571705532424465 , 4.4208574 )
( 3.018823359926941 , 4.350743 )
( 4.542100392209846 , 6.598222 )
( 0.013201218670070403 , 0.978447 )
( 6.97731431598188 , 9.432816 )
( 2.7129700103713783 , 3.9425678 )
( 4.790215074046861 , 6.5095706 )
( -0.6778653526631935 , 0.91343904 )
( 7.203968590583353 , 9.04431 )
( -0.16742374699072693 , 0.48905003 )
( -0.9061252741010206 , 0.97488195 )
( 0.2170116340411683 , 1.1725483 )
( 5.941813818408002 , 7.0181727 )
( 6.164484284048667 , 8.107263 )
( 4.394736677669951 , 5.923207 )
( 2.5876744961671 , 3.8513887 )
( 7.745287333339578 , 9.391553 )
( 6.464975950205047 , 8.599304 )
( 5.221598201707392 , 6.949786 )
( 3.2746376319032793 , 4.06941 )
( 4.322557968141274 , 5.380784 )
( 0.14107125549394084 , 0.53630304 )
( 3.981980763973077 , 5.703768 )
( -1.2359067438271043 , 0.4418288 )
( 3.3474347106638858 , 5.076431 )
( 1.1060723748543433 , 2.5312264 )
( 7.6114767610079115 , 10.251982 )
( 6.581000098121152 , 9.02239 )
( 7.238503493573845 , 9.769665 )
( 4.269442086174138 , 5.435928 )
( 7.054061471324001 , 8.522634 )
( -1.461811525479933 , -0.1718354 )
( 7.0034197634623006 , 8.64923 )
( 5.884257892194496 , 7.195285 )
( 6.5217918401382144 , 9.176818 )
( 5.58326004119045 , 7.876363 )
( 0.29422834658901564 , 1.0590836 )
( -0.12981704597109978 , 0.82612294 )
( 0.038041830943694066 , 1.172873 )
( 0.40881063215363245 , 1.5107627 )
( 3.426542967294579 , 5.196945 )
( -0.9484909566289238 , 0.24696213 )
( 0.06804404806036612 , 0.48059273 )
( -1.0145524673680444 , 0.53736067 )
( 1.5015880827261952 , 2.1889255 )
( 2.8573052759412034 , 3.78891 )
( 9.219458452935616 , 10.913517 )
( 0.5368244143942041 , 1.5808187 )
( 4.355831634649004 , 6.1878967 )
( 0.18293071013046092 , 1.0599701 )
( 3.196250422523403 , 4.8135853 )
( 4.345084873172813 , 5.080886 )
( 2.520467966738127 , 3.847038 )
( 7.022303363600746 , 9.052672 )
( 3.241959619206911 , 4.056229 )
( 2.260551961518866 , 3.3595705 )
( 7.599870669057844 , 9.652032 )
( 3.1206821892354983 , 4.883989 )
( 3.8161913075583547 , 5.012031 )
( 5.894939809451715 , 7.851505 )
( 3.7193856645207655 , 5.352621 )
( 6.965713092123607 , 9.134069 )
( 5.631698208279772 , 7.6130023 )
( -0.3315919887929628 , 0.49877876 )
( 4.994235047474497 , 6.1392 )
( 5.897214748924215 , 7.9101996 )
( 4.484277854254555 , 6.3281875 )
( -1.861480115814844 , 0.047102094 )
( 8.100808604287657 , 9.99959 )
( 5.561990033500462 , 7.7145624 )
( 1.1898974931511623 , 1.8565439 )
( -6.230137864499565 , -5.667335 )
( 7.91458769948004 , 9.694895 )
( 6.562499364760522 , 8.789988 )
( 7.924020661464057 , 9.09126 )
( 3.4657567696548126 , 5.2853813 )
( -1.0131253469182369 , -0.1689415 )
( 8.153918173337296 , 9.528039 )
( 8.011330087641998 , 10.155506 )
( 5.4217228848061785 , 7.895767 )
( 6.301385079089718 , 8.273916 )
( 0.8505030277778488 , 2.295781 )
( -0.7859233667651228 , 1.1989746 )
( -1.0839744909751343 , 0.32272607 )
( -0.9320468500597869 , 0.336591 )
( -1.782618711091846 , -1.5334588 )
( 4.50623006965518 , 6.3353157 )
( 5.698790489251194 , 7.3219485 )
( 5.257528347316875 , 7.2254233 )
( -0.313603203271988 , 0.5506372 )
( 7.771614177360129 , 9.812851 )
( 1.9937111460143533 , 2.9505084 )
( 5.445762247032819 , 6.6976905 )
( 5.022080931346767 , 6.2111754 )
( 0.19951050924736347 , 0.70367646 )
( 6.775356556684504 , 8.633563 )
( 0.3862617703926645 , 1.0311837 )
( 8.233215513649972 , 10.201825 )
( -0.4269751437497818 , 0.9941239 )
( 0.027307365071082 , 0.9104445 )
( -1.1343699559232412 , 0.102680326 )
( 0.4832239741317146 , 0.70949984 )
( 0.9340962959134025 , 2.0319138 )
( 2.627085270190526 , 3.5218341 )
( 5.379976084727163 , 6.7426467 )
( 0.4772875052857728 , 1.4583254 )
( 0.08905097931008821 , 1.3490825 )
( 7.207069468110452 , 9.33982 )
( 4.931528302241455 , 6.5388317 )
( 1.8369539968596635 , 3.5222898 )
( 2.909288634227189 , 4.685118 )
( -0.8457344558119377 , -0.65126634 )
( 3.1456431646785417 , 4.8633494 )
( 8.778435772379716 , 9.796442 )
( 7.689663448683343 , 9.223524 )
( 3.0935712462038794 , 4.66642 )
( 7.061850623282417 , 9.614056 )
( 6.445507296071188 , 8.02202 )
( 5.407126101815223 , 7.6577306 )
( 7.2862028426000105 , 9.44777 )
( 2.594359954931955 , 3.6694894 )
( 1.0909649031124928 , 2.555791 )
( 6.142807421492974 , 7.923347 )
( 1.1947569505749391 , 1.8381046 )
( 4.582683984229626 , 5.657216 )
( 5.513515616526465 , 7.107692 )
( 0.15173654114674306 , 1.0062413 )
( 4.361131703751097 , 6.400421 )
( 3.4214726266619833 , 4.21144 )
( 4.589568719346355 , 5.9606147 )
( 0.671253093898023 , 1.4884343 )
( 7.514806194145972 , 9.327931 )
( 7.124276194287938 , 8.566337 )
( 5.585961677580761 , 6.905182 )
( 3.8739052260086866 , 5.728567 )
( -1.6586858297216072 , -1.4095528 )
( 4.135443786732309 , 5.5878873 )
( 3.6510572089128557 , 4.998114 )
( -0.9856076642668015 , -0.42387938 )
( 0.7860755214217638 , 1.6472435 )
( -2.9881436934506818 , -1.896416 )
( 0.2080221241290936 , 1.5711105 )
( -0.7275721409457951 , -0.8399503 )
( 7.819655320345916 , 9.576131 )
( 3.3035475306329563 , 4.9431977 )
( -1.343147869482733 , -0.26665354 )
( 6.480415669700959 , 7.9725604 )
( 6.764791546322475 , 8.144485 )
( 3.592512331485088 , 4.867998 )
( 3.2411846401196414 , 5.0109653 )
( 1.3966070135792705 , 2.1064196 )
( 8.098102208474536 , 10.085353 )
( 2.61185255324428 , 4.1654463 )
( 1.3985140227052961 , 2.1210394 )
( 0.7805510135085597 , 1.308913 )
( 2.482778303332326 , 3.9398165 )
( 6.944341111423071 , 8.083297 )
( 5.634343785250352 , 7.455042 )
( 6.034258713263885 , 8.342058 )
( 3.996759079656673 , 5.6876707 )
( 3.009361695483859 , 3.7328715 )
( 3.15542841589615 , 5.170604 )
( 3.969255675366232 , 5.4043875 )
( 7.277163323864622 , 8.676061 )
( 6.427085901907864 , 8.1939335 )
( 8.162753127607061 , 10.322455 )
( 3.698517400206751 , 4.5092506 )
( 5.767646941102387 , 7.8254046 )
( 0.8594074637564966 , 1.5703603 )
( 8.580063358466644 , 10.031574 )
( 3.315873240191739 , 4.873382 )
( 8.283795573865357 , 10.073459 )
( 1.6510101844949636 , 2.4518495 )
( 5.740889465731581 , 6.5904484 )
( 6.008348913846862 , 7.8307333 )
( 4.851198937158236 , 6.4543347 )
( -4.333680434259447 , -3.8320556 )
( 5.269858912259353 , 6.7236767 )
( 3.1892859827545172 , 4.8709664 )
( 5.609566610570077 , 6.03663 )
( -0.42300459199689433 , 0.89383155 )
( -5.279697324823457 , -4.74514 )
( 1.1402898900273768 , 2.0073986 )
( 6.633971572504234 , 8.283981 )
( 5.639650980669303 , 7.2790217 )
( 1.575106667189782 , 2.4830508 )
( 8.024820045710714 , 9.639199 )
( 7.536613539384735 , 9.4245825 )
( 8.668727950618546 , 10.745632 )
( -2.9997699370174207 , -3.5822296 )
( 4.185198638723766 , 5.537021 )
( 4.855774186888225 , 6.8458185 )
( 6.764028266025655 , 8.183995 )
( 7.197074017778772 , 10.032653 )
( 0.2972010925862616 , 1.5329676 )
( 5.728988040120108 , 6.430538 )
( 4.694584390778518 , 6.441453 )
( 2.0825091754802116 , 2.7708013 )
( -0.27782473790486817 , 1.4516331 )
( 3.4407135099299007 , 4.76183 )
( 0.27044797328951997 , 0.9807755 )
( -0.10305955987728843 , 1.1259053 )
( 5.937450646185493 , 9.433977 )
( 2.245954796095088 , 3.5046504 )
( 6.468275552760025 , 7.9441957 )
( 0.10148684497548288 , 0.9569385 )
( 3.962822467698664 , 5.641126 )
( -8.575570813006808 , -7.349297 )
( 7.432015529573616 , 9.141075 )
( 7.15320559444487 , 8.960428 )
( 6.580409678493231 , 8.561462 )
( 2.7089213812927486 , 3.797906 )
( 7.494245224364793 , 8.673066 )
( 3.971971211166494 , 4.7984643 )
( -1.310163823492471 , 0.13895118 )
( 2.195005433564788 , 3.9641333 )
( 4.424420324514264 , 5.9727325 )
( 1.3250087053484476 , 2.4132133 )
( 3.6157508122114645 , 4.9767666 )
( 8.111944816892722 , 9.885482 )
( 4.809488142506297 , 6.638986 )
( 5.3528694612293615 , 6.8915753 )
( 8.074255326745986 , 10.021231 )
( -5.036652750805067 , -5.580194 )
( 6.799023221307781 , 8.47645 )
( 0.4842734465285048 , 1.3736691 )
( -0.6807215916598376 , 0.55143136 )
( 0.3391016419927628 , 1.1714258 )
( 0.30695453329751476 , 1.8619852 )
( 1.989391232200274 , 3.1552227 )
( 5.741765168140211 , 7.4429584 )
( 2.058901163008797 , 3.458054 )
( 5.409069259618547 , 6.965061 )
( 7.960705380199065 , 9.544696 )
( -3.5696849425253028 , -3.908278 )
( 7.6788128693097 , 9.626372 )
( -0.8112591349307039 , 0.42775127 )
( 1.101978508147054 , 1.8342234 )
( 2.6285679810691542 , 3.4513948 )
( 1.0783652365220808 , 2.3542283 )
( 7.4732174411246906 , 9.346743 )
( 3.47231666480754 , 4.677791 )
( 7.2412005457190896 , 9.401794 )
( 8.666897534478453 , 10.496367 )
( 8.343778302335693 , 9.915554 )
( -0.46512612394413755 , 0.3119849 )
( 8.807363703418494 , 10.666014 )
( 6.3792336311848095 , 8.607623 )
( -10.437221645700149 , -9.512594 )
( 3.629503518811924 , 4.769514 )
( 4.746900388581092 , 7.1124916 )
( 0.15215903926326035 , 1.8463972 )
( 0.5634830964420938 , 1.6217558 )
( 4.4001327782623605 , 5.711671 )
( 9.26421111124202 , 10.950724 )
( -0.3917148535655173 , -0.09691739 )
( 3.901191564160495 , 5.631542 )
( 3.0996279129631175 , 4.6958733 )
( 8.51613751637073 , 10.468479 )
( 5.733939405678729 , 8.121033 )
( 2.4801394286970595 , 3.2964654 )
( 4.071071137260185 , 5.1250086 )
( 3.5681700925702424 , 5.299407 )
( 5.124374737752694 , 7.2887306 )
( 1.2019172880292346 , 2.6959643 )
( 6.130227380192603 , 7.9779882 )
( 6.3067799987182624 , 8.27833 )
( 3.2717690061970295 , 5.400794 )
( 2.7349584858773444 , 3.0897903 )
( 8.962820573163523 , 10.434887 )
( 2.995617081083328 , 4.5721583 )
( 2.915935001366978 , 4.401074 )
( 1.8644158545532985 , 2.7941983 )
( 8.336129179602093 , 10.411146 )
( 6.085658676439345 , 8.711027 )
( 0.5749252501844584 , 1.8674817 )
( 5.310152447615155 , 7.631142 )
( 7.158445944291447 , 9.019493 )
( -1.6370201491541465 , -1.0713904 )
( 0.9986431823481587 , 1.7642884 )
( 5.1305873227938905 , 6.4016023 )
( 6.155760579151626 , 7.651522 )
( 6.769782977998894 , 9.817339 )
( -0.9822414878575332 , -0.8825338 )
( 1.3344636175231686 , 2.213159 )
( 4.554871489740615 , 6.1787863 )
( 2.909827855188614 , 4.5196137 )
( 4.5175169712843175 , 6.2072735 )
( 7.243242739952342 , 9.749432 )
( -1.037852104233253 , 0.097059906 )
( 1.0532549579469026 , 1.6514772 )
( 5.01416710988768 , 7.057122 )
( 8.511830381716779 , 10.929638 )
( 5.896195549242847 , 7.5646095 )
( 1.1840946246274253 , 2.4983988 )
( 6.591521975808543 , 8.587136 )
( 4.532828086221569 , 6.2286553 )
( 4.166008129046396 , 6.1900983 )
( 4.146551897937692 , 6.4091043 )
( 2.5971589184437374 , 4.240612 )
( 1.4830701324221194 , 2.5726438 )
( 4.005073502282227 , 5.9282646 )
( 5.105853296062 , 8.197611 )
( 0.9975795103096412 , 1.8423183 )
( 3.126753842069456 , 4.6980743 )
( -7.439218403173073 , -7.2728834 )
( 6.799888093300674 , 8.640316 )
( 3.981398885365243 , 5.3578506 )
( 3.7897418956222566 , 5.3267164 )
( 6.072761189641484 , 7.872425 )
( 5.295036757229746 , 6.803442 )
( 2.6653308686888124 , 4.305586 )
( 4.116420866323832 , 7.4705024 )
( 0.48398987865715637 , 0.3933615 )
( 6.6671629632631255 , 8.365522 )
( 1.596662606196478 , 2.92678 )
( 0.681479149294133 , 1.886038 )
( 1.360549839281143 , 3.0775719 )
( 5.48290609053849 , 6.8013663 )
( 3.5007163483593216 , 4.3054886 )
( 1.3658614854818878 , 2.2588525 )
( 4.742767114063593 , 6.225522 )
( 1.900833687213928 , 2.5301516 )
( 3.0900035367463516 , 3.9230368 )
( 5.334425461443245 , 7.5539913 )
( -0.25184200642811927 , 0.75870204 )
( 2.471905399350013 , 3.9170275 )
( 1.775837184711735 , 2.6150498 )
( 5.5819669243569985 , 7.6564207 )
( 1.3002937481563142 , 2.1384823 )
( 6.225433530450647 , 7.8690925 )
( -7.047127941660372 , -7.149742 )
( 5.801280751712154 , 8.514827 )
( 3.8385130706510076 , 4.8022738 )
( 3.129081882439874 , 4.8223343 )
( 3.1381742517297635 , 4.7090807 )
( 0.911672035089985 , 1.9850185 )
( 0.39746992241505763 , 1.3560821 )
( 3.683233107893736 , 4.9445634 )
( -0.9879722717985144 , 0.55870366 )
( 5.552722441747106 , 7.054687 )
( 1.4879642700392357 , 2.5156062 )
( 6.493739219342513 , 8.403737 )
( 1.71769140109893 , 2.3557234 )
( -1.5754339828219046 , -0.44533587 )
( 5.997787280696301 , 7.691699 )
( 4.859847991073047 , 6.141047 )
( 6.02204310872035 , 7.834529 )
( 5.878205167653274 , 7.729244 )
( 8.189840437294244 , 9.790419 )
( 3.3225445675925247 , 4.169211 )
( 2.9872772134142846 , 4.495748 )
( 1.7683474309917608 , 3.2223136 )
( 7.274587817637419 , 9.105763 )
( 5.014556318205441 , 6.3743606 )
( 0.021949093411494758 , 1.0624621 )
( 1.3791536331558263 , 2.6446786 )
( 2.749995527729252 , 4.647053 )
( 2.6807466158951563 , 3.4686284 )
( -1.063042384348054 , -0.24170268 )
( 2.4163439473747848 , 3.1278849 )
( 2.1350668529301178 , 1.9179707 )
( 6.771859124504322 , 8.879998 )
( 3.8114009671871916 , 6.1051006 )
( 8.595778599032936 , 10.35103 )
( 4.331827608436718 , 6.137731 )
( 8.694604944621343 , 10.465017 )
( 1.1174953867118749 , 2.282011 )
( -0.8633201353072568 , 0.29281056 )
( 7.971782557729421 , 10.258798 )
( 3.164120747466886 , 4.6860228 )
( 6.449059586758069 , 8.867858 )
( 0.6244241847319878 , 1.1120393 )
( 5.608020774523073 , 7.058623 )
( 7.279243521793322 , 8.626764 )
( 5.46935513329133 , 6.7042975 )
( -5.265942177958036 , -5.374528 )
( 0.32351073124181473 , 1.9681541 )
( 6.545221502876821 , 8.222654 )
( 6.395752252102344 , 8.408594 )
( 1.1857569710415499 , 1.9011587 )
( 2.3132684916612885 , 3.0723312 )
( -1.5244140642080222 , 0.16095328 )
( 7.97112358550375 , 9.668508 )
( 1.930134676963441 , 2.964134 )
( 4.731446250408343 , 6.3767886 )
( 2.050535438956391 , 3.3821435 )
( 8.332102242749624 , 10.443482 )
( -0.5749081743185755 , 0.6052397 )
( 4.533885930835709 , 7.415987 )
( 9.312465295026502 , 11.490305 )
( 5.863114856857048 , 8.509095 )
( 7.546016634538164 , 9.336941 )
( -3.9227885916115195 , -4.5061045 )
( 2.0394474555596576 , 2.8121474 )
( 0.581530273852234 , 0.3506434 )
( 0.2196068267587522 , 1.312078 )
( 2.475244755748187 , 3.6291537 )
( 8.2664019341911 , 9.852113 )
( 2.190214619285841 , 4.1777797 )
( 7.466424746058265 , 9.38368 )
( 5.658164198018479 , 7.021668 )
( 4.403602631218711 , 6.2141824 )
( -5.224978802992986 , -4.712556 )
( 8.625333882040986 , 10.596866 )
( 4.717067634553461 , 7.798277 )
( 8.701105912243985 , 10.37694 )
( 0.6829439384432998 , 0.4519192 )
( 4.275607443358914 , 6.1924176 )
( -0.08250282069655374 , 1.403069 )
( 7.662676154006775 , 9.503456 )
( 7.6449865281401825 , 9.519916 )
( 0.8213267527476028 , 1.6698444 )
( 7.736175343362053 , 9.121113 )
( 8.342159482651148 , 10.095975 )
( 2.8456036768643562 , 3.4633482 )
( 2.6963482879483918 , 3.9074116 )
( 6.83656710186872 , 9.465315 )
( 4.482978576874089 , 5.5610647 )
( 5.5087617221936 , 7.4296346 )
( 1.8058079969005956 , 2.858903 )
( 7.502734286502235 , 9.184642 )
( 4.5511530335982915 , 6.1560035 )
( -0.3266378831588659 , 0.03709817 )
( 0.34852643591641924 , 1.4426656 )
( 4.688043987506755 , 7.4175186 )
( -0.5164141538124996 , 0.57351905 )
( 7.108832456392361 , 9.036374 )
( -0.3703417218571583 , 0.8698418 )
( 2.4701499904104214 , 3.8560715 )
( 1.344211838775722 , 2.90164 )
( 2.5784112490213635 , 3.8877738 )
( 6.070397627215414 , 7.3908195 )
( 4.08445533880795 , 5.7498474 )
( -0.22542072572583272 , 0.66859674 )
( 6.466550946048175 , 8.3320675 )
( 0.05979029613621627 , 0.4230423 )
( -3.8457693107436626 , -2.1874125 )
( 6.301235527873694 , 8.423102 )
( 0.8384084943593229 , 1.4753226 )
( -0.3811887327150716 , 0.3607782 )
( -0.3790195901223447 , 0.748851 )
( 3.3648486295142765 , 4.6410894 )
( 3.1358569116311097 , 4.6588254 )
( 5.426002499733469 , 7.4945297 )
( 8.812234323321725 , 10.765465 )
( 1.1408548492768396 , 2.4802475 )
( 6.4719675561038015 , 7.9388347 )
( 5.075685770545606 , 6.258258 )
( 0.19647626285131448 , 1.7330976 )
( 3.892928324931664 , 5.304279 )
( 2.683407633627168 , 3.9343948 )
( 6.748026450849453 , 8.531408 )
( 6.392600500447838 , 7.726942 )
( 3.5818439435015135 , 5.3521857 )
( 0.7650937026115008 , 1.9290948 )
( 0.99219119179527 , 2.2214026 )
( 5.846027264664226 , 7.6599116 )
( 2.884827621962347 , 5.132779 )
( 4.962376271645355 , 6.146397 )
( -1.001470230760968 , 0.18493712 )
( -1.5929285362910726 , -0.8954649 )
( 6.571752296667191 , 9.03205 )
( 5.426542368888581 , 7.286674 )
( 0.5812839223241242 , 1.7635815 )
( 5.549019686858804 , 7.5543227 )
( 8.611423977018035 , 10.638132 )
( 2.8354116822532713 , 3.652286 )
( 6.122294722659488 , 7.621224 )
( 7.893478809954566 , 9.734747 )
( 0.6220021003642806 , 1.9805627 )
( 0.2806964171607157 , 1.6235415 )
( -0.6954836129096513 , -0.052595615 )
( 1.0186738403981115 , 1.6009264 )
( -1.7419358622075105 , -1.6521168 )
( 3.416587801382687 , 4.443193 )
( 1.4180806855703523 , 2.8981185 )
( 2.2599741183650073 , 3.412416 )
( 6.249004783412754 , 7.8123355 )
( 4.397152771695386 , 7.226096 )
( 0.7803610188826315 , 1.7506806 )
( 0.23949248957757596 , 1.5457382 )
( 7.259615850834216 , 9.737318 )
( 6.747536956818303 , 8.607792 )
( -9.619736345819268 , -7.1122656 )
( -1.1290545204923663 , -0.39345258 )
( 0.18453391131060626 , 0.4759572 )
( 2.7576507858120998 , 4.1812983 )
( -2.2025592371228795 , -2.2968948 )
( 7.40519542264338 , 9.558481 )
( 2.071981326475287 , 3.445491 )
( 5.953170096651236 , 8.14747 )
( -1.4181221561088753 , 0.34232968 )
( 5.503889141244559 , 6.93388 )
( 3.679803271051624 , 4.825202 )
( 5.085710288660236 , 6.483948 )
( 3.1602498312622034 , 4.722772 )
( -0.17072681981402837 , 0.67893744 )
( 6.157057163125378 , 8.3339815 )
( 4.293359944999301 , 5.878449 )
( 5.943076598467691 , 7.7129116 )
( 6.717191927808272 , 8.641236 )
( 5.70206384092841 , 7.679452 )
( 7.080939940798878 , 8.771531 )
( 3.8507419056964682 , 5.3132353 )
( 4.874074235990611 , 6.271104 )
( 5.995151954756908 , 7.7982187 )
( 5.562007547978609 , 7.4959016 )
( 2.0287331712064463 , 2.814353 )
( -0.5756662140352468 , 1.3655889 )
( 7.206097375549581 , 9.432768 )
( 5.063530258450487 , 7.082176 )
( 7.794261899573549 , 9.449181 )
( 5.635284631293987 , 7.5277042 )
( 2.631961583243885 , 3.6621895 )
( -0.13270730574559053 , 0.61609936 )
( -1.4041260499823638 , 0.1299783 )
( 4.781604518023812 , 6.947441 )
( 5.061792994362303 , 6.71346 )
( 3.7967887999214804 , 5.382408 )
( -0.9170670861465926 , 0.21177414 )
( 2.3984446395342593 , 3.481049 )
( 1.0403311717216241 , 1.7049823 )
( 7.573532995784509 , 9.256842 )
( 6.531591131941996 , 7.744784 )
( 0.43287411489203625 , 1.7345955 )
( 5.880681387737205 , 8.738031 )
( 0.7457143239334482 , 1.7718668 )
( 0.8562030468105234 , 2.0526993 )
( 6.251780018590224 , 8.938464 )
( -8.951826266327188 , -8.082057 )
( 3.7496386987232326 , 6.1861835 )
( 2.3421287073115 , 3.3471243 )
( 3.8528088846303286 , 4.9067206 )
( 2.8423686297572637 , 4.3536315 )
( 6.908882324707088 , 8.318051 )
( 7.1431344382942985 , 9.333752 )
( -0.4952406163788685 , 1.0723045 )
( 6.425140098123507 , 7.419941 )
( -0.28078579790237335 , 0.8828956 )
( 6.825117813014781 , 8.782778 )
( 4.467634824141519 , 5.8876543 )
( 0.7643653079046813 , 1.6764507 )
( 5.7288521996524455 , 7.440816 )
( 2.6355246828735934 , 3.361199 )
( 1.312347594344499 , 1.845495 )
( 7.1564394875389965 , 8.985819 )
( 5.325740843880499 , 7.2922444 )
( 7.105105570822241 , 8.295276 )
( 6.6998837701259735 , 8.272135 )
( -7.489716220397302 , -7.266631 )
( 7.686440170919016 , 10.398749 )
( 1.8825017002979392 , 2.6841364 )
( 8.289803656983274 , 10.37216 )
( 1.5570199862835385 , 2.240616 )
( -11.01287074171087 , -9.850235 )
( 7.182490697470325 , 8.915426 )
( 8.807934587908681 , 9.820602 )
( 3.582486720994801 , 5.4086347 )
( 5.062651471375654 , 6.7014484 )
( -3.9589630700545664 , -4.221526 )
( 0.1749816065665585 , 0.653003 )
( 3.686370459931357 , 5.7016783 )
( 5.803068797435925 , 7.5569043 )
( 4.323619852869415 , 5.1573853 )
( -0.13864754705374427 , 0.90867895 )
( 6.703282204143671 , 8.367184 )
( 1.7326988553652443 , 3.6003003 )
( 0.326456935858493 , 0.67352015 )
( 6.776496382255013 , 8.627266 )
( 7.079544498704168 , 8.752037 )
( -3.48492228376138 , -3.2100248 )
( 1.7810780063897753 , 2.4661255 )
( 8.77903234925611 , 10.780855 )
( 5.953179207000205 , 7.3264046 )
( 0.8592441211562654 , 2.1115713 )
( 7.649844541376083 , 9.299493 )}

\begin{figure}[h!]
\centering
\begin{subfigure}{.3\textwidth}
  \centering
  \begin{tikzpicture}[scale=0.6]
  \begin{axis}[xlabel={True utility}, ylabel={Predicted utility}, axis y line=left, axis x line=bottom, xmin = -10, xmax = 10, ymin = -10, ymax = 10] 
  \addplot[scatter,only marks] coordinates {\utilitysmall};
   \addplot[dashed, very thick, samples=100, smooth, domain=-30:40] expression{0.57*\x + 2.13};
  \node at (axis cs:5,-2) {\sf Best fitted line:};
  \node at (axis cs:5,-4.2) {$y = 0.57x - 2.13$};
  \node at (axis cs:5,-6) {$R^2 = 0.75$};
  \end{axis}
 \end{tikzpicture}
  \caption{$(\ell,w)=(0,0)$.}
\end{subfigure}
\begin{subfigure}{.3\textwidth}
  \centering
  \begin{tikzpicture}[scale=0.6]
   \begin{axis}[xlabel={True utility}, ylabel={Predicted utility}, axis y line=left, axis x line=bottom, xmin = -10, xmax = 10, ymin = -10, ymax = 10] 
  \addplot[scatter,only marks] coordinates {\utilitymedium};
   \addplot[dashed, very thick, samples=100, smooth, domain=-30:40] expression{1.06*\x + 1.68};
  \node at (axis cs:5,-2) {\sf Best fitted line:};
  \node at (axis cs:5,-4.2) {$y = 1.06x + 1.68$};
  \node at (axis cs:5,-6) {$R^2 = 0.97$};
        \end{axis}
 \end{tikzpicture}
  \caption{$(\ell,w)=(1,3)$.}
\end{subfigure}
\begin{subfigure}{.3\textwidth}
  \centering
  \begin{tikzpicture}[scale=0.6]
  \begin{axis}[xlabel={True utility}, ylabel={Predicted utility}, axis y line=left, axis x line=bottom, xmin = -10, xmax = 10, ymin = -10, ymax = 10] 
        \addplot[scatter,only marks] coordinates {\utilitylarge};
          \addplot[dashed, very thick, samples=100, smooth, domain=-30:40] expression{1.11*\x + 1.03};
  \node at (axis cs:5,-2) {\sf Best fitted line:};
  \node at (axis cs:5,-4.2) {$y = 1.11x + 1.03$};
  \node at (axis cs:5,-6) {$R^2 = 0.98$};
        \end{axis}
 \end{tikzpicture}
  \caption{$(\ell,w)=(2,5)$.}
\end{subfigure}
\caption{Out-of-sample predicted utility versus true utility for Setting~2 of Section~5 for different DeepMNL models. Recall that in Setting~2, the utility in the ground truth model is non-linear.}
\label{fig:utility_2}
\end{figure}

\vspace{0.5cm}

\item \textbf{Setting~3: latent class MNL.} In Setting~3, the ground truth model is a latent class MNL model with two customer classes. For each (out-of-sample) customer feature vector, we determine the ground truth utility for each customer class. Similarly, for our RUMnet model with $K=2$, we compute the recovered utility for each of the $K$ samples. Naturally, we do not know ex-ante how to label these $K$ samples relative to the $K$ customer classes. However, we can compute the utility for each sample, i.e., before averaging. Hence, in Figure~\ref{fig:utility_3}, we construct scatter plots of the predicted utility (for each sample) as a function of the true utility (for each customer class). Among these four scatter plots, we observe that two of them actually concentrate around a line, suggesting that customer class 1 corresponds to Sample 1 and customer class 2 corresponds to Sample 2.  If we regress the predicted utility on the true utility in these two quadrants, we obtain $R^2$ values of 0.87 and 0.99, with slopes of 1.28 and 0.56. These high $R^2$ values indicate that the heterogeneity in our RUMnet model captures and recovers the underlying heterogeneity in the ground truth model. Note, however, that the ground truth model is a non-uniform mixture of two customer classes, whereas our RUMnet is a uniform mixture over the two samples. The latter restriction may explain why the slopes are not close to one. 
\end{enumerate}

\newcommand{\utilityA}{
( -0.37513290753492834 , -9.661038 )
( -4.4756778362243015 , -5.207035 )
( -12.955751524089276 , -10.960983 )
( 2.2594629568305407 , -5.4371595 )
( -9.363319513009078 , -15.241807 )
( 3.3714726908889396 , -6.5213327 )
( 3.9848515396424764 , -5.9856014 )
( 24.627006325746763 , 6.6638317 )
( 2.9667489169882897 , -6.876933 )
( 5.372542325595924 , -2.804696 )
( -7.319050367127131 , -7.6446466 )
( -1.9212755572910325 , -2.9422755 )
( 23.28771502357361 , 5.4764605 )
( -0.6618809378310787 , -8.066754 )
( 26.226532989784058 , 8.126583 )
( 4.329692556274253 , -3.7475462 )
( -2.4878443417469267 , -3.4794183 )
( 1.5626732350581154 , -6.074254 )
( -7.261973260637454 , -17.030182 )
( 39.67254991022935 , 17.182384 )
( -4.324927613710429 , -5.086295 )
( 1.8620334192062409 , -7.894448 )
( -5.382943865164341 , -15.444013 )
( -18.50877480328292 , -15.782425 )
( 1.5892076789424165 , -6.0455317 )
( -5.748864000689925 , -15.042258 )
( -6.222931791522218 , -6.725687 )
( 1.347687202807038 , -8.233293 )
( -16.83518336342517 , -14.377267 )
( 26.808809343199183 , 8.588201 )
( -2.8627027813719597 , -11.785353 )
( 25.678879824998134 , 7.4946165 )
( -5.836786196106895 , -6.3399563 )
( -6.826177572635219 , -12.935963 )
( 41.58945732148878 , 18.870756 )
( -16.524385355179 , -14.008721 )
( 3.5287228448825063 , -6.421456 )
( -2.2127953475323707 , -3.1746478 )
( 0.05240143960428967 , -8.311514 )
( 4.563173064765126 , -4.3390064 )
( -0.5709987755064919 , -7.9685197 )
( -4.435849304125644 , -13.218806 )
( 4.012934724974026 , -4.112586 )
( 6.385262475399307 , -3.8488083 )
( 4.1875410060157625 , -3.921465 )
( -12.954497830151613 , -11.026434 )
( 3.9026150345215385 , -5.913568 )
( -4.379290593126323 , -13.124132 )
( 1.6493026229300636 , -8.036342 )
( 1.5177597118334367 , -8.175153 )
( 28.932690250529355 , 10.326015 )
( 3.9319449279350507 , -6.000219 )
( -6.792213811267834 , -7.0708804 )
( 4.023771381092057 , -4.0663433 )
( -0.7714267361472669 , -10.062152 )
( 29.048250097401116 , 9.420021 )
( -0.22829522722298434 , -7.6706624 )
( -5.221123957613866 , -11.692188 )
( -1.6425804497446865 , -10.720003 )
( -5.79172207743037 , -6.273054 )
( 4.374827595975566 , -5.7325153 )
( 1.1338339012569718 , -6.434931 )
( -2.308285372230004 , -11.310661 )
( 3.6274468576063432 , -6.2284317 )
( 3.2853107780034083 , -6.584287 )
( 2.2120061420248676 , -5.6885867 )
( 25.301599330029426 , 7.2462096 )
( 27.480971591502115 , 9.029857 )
( 29.46440674560889 , 10.854541 )
( -0.6882492159339195 , -10.0602455 )
( 29.04773459095626 , 10.5707035 )
( 0.4605892341305434 , -7.0189853 )
( 2.7724849277070427 , -5.165095 )
( -9.606811283123438 , -15.472147 )
( 3.8076693960274737 , -6.1379743 )
( -5.430970241680317 , -5.844674 )
( -7.404156245680586 , -17.28572 )
( -5.603669924995814 , -6.0063205 )
( 25.937940942646236 , 7.7621775 )
( -6.256321508891226 , -12.433112 )
( -4.287894226852044 , -4.9993734 )
( -0.26768113034742747 , -7.773208 )
( 33.35185206591384 , 11.81553 )
( -4.75910493977257 , -5.4104095 )
( 25.6670549356605 , 7.5377784 )
( 5.1095919554512435 , -5.0432024 )
( 29.03300987140308 , 10.444172 )
( 2.594828311238421 , -15.6052 )
( -15.034544077241167 , -12.817329 )
( -8.096427939297486 , -14.06328 )
( -1.3954581336180603 , -9.513443 )
( -13.656389229084807 , -11.666554 )
( -6.509169903006968 , -6.8649282 )
( 2.545766310242424 , -5.3869963 )
( -5.794257336100853 , -6.3495455 )
( -1.9493454495629372 , -10.952775 )
( -0.34850014875493385 , -7.856394 )
( -4.0404480543161085 , -4.7432775 )
( -2.3725192914653075 , -3.3701897 )
( 29.11348330369907 , 10.60185 )
( -3.09155277223511 , -4.0004535 )
( 3.4328376532717506 , -6.4860134 )
( 28.679978460799184 , 10.233255 )
( -3.1758372576561595 , -12.034469 )
( -1.7273143495709569 , -10.753862 )
( 1.800110559881793 , -7.828721 )
( -0.26553640811603374 , -7.7933974 )
( 4.258338962343654 , -4.721988 )
( -4.28634846341409 , -4.925024 )
( -17.359879511570323 , -14.878536 )
( 2.2406905639915777 , -5.4495306 )
( 26.054467045055574 , 7.9773746 )
( 0.0986898241311771 , -9.2252 )
( 0.37293770650773483 , -9.112598 )
( 1.455336796791996 , -6.2144604 )
( -0.19779704437725154 , -7.7052817 )
( -1.3859390967681726 , -2.4187298 )
( 2.978012035878329 , -4.8506346 )
( 2.9679756497385203 , -5.0320153 )
( 1.8718188756716994 , -7.766234 )
( -6.123602212558979 , -6.4625688 )
( 30.162861669896948 , 10.420643 )
( -1.3450862278325078 , -10.544964 )
( -29.221138423652057 , -18.80501 )
( -0.082048968459977 , -9.322459 )
( 1.1060294596577878 , -8.338282 )
( -4.308822866363755 , -13.069373 )
( 4.6135275773623174 , -5.490714 )
( 23.182380836187225 , 5.3700857 )
( 1.861071667952459 , -7.7801847 )
( 5.097722356624128 , -3.0110335 )
( 2.8894690212558864 , -5.030721 )
( -15.40479428287016 , -13.062927 )
( -3.214154916694754 , -12.025969 )
( 2.6788968185364403 , -7.0550036 )
( -4.827101775898062 , -13.5544405 )
( 2.2752608006908304 , -5.447981 )
( 3.6288299078960717 , -4.4593744 )
( -3.7431577023115628 , -12.470754 )
( 1.378639653246807 , -8.106198 )
( 25.354024933687253 , 7.275524 )
( 4.86964147650852 , -5.132082 )
( 4.610601854273379 , -5.4956408 )
( 2.6732349762428504 , -5.1503 )
( 3.3512618560631884 , -4.585529 )
( -2.188357830365444 , -3.2276163 )
( 1.9038018459498423 , -7.783107 )
( -0.6795200956670128 , -8.104245 )
( -5.286080576033016 , -11.732021 )
( 4.631465183014006 , -5.2790046 )
( 39.18854011028482 , 16.919842 )
( -3.9523750873830736 , -12.649929 )
( 0.8294539679560815 , -6.755002 )
( 5.080552546594476 , -5.084824 )
( -17.21173700037463 , -14.749554 )
( 3.7565672992232493 , -4.988292 )
( -5.593458299517989 , -15.741139 )
( 27.24456647612582 , 9.032762 )
( -17.883737687513648 , -15.242584 )
( -16.02293573209296 , -13.645465 )
( -0.30991637104190906 , -9.66586 )
( 0.3278306439438936 , -8.937808 )
( -5.508003141973301 , -11.791669 )
( 4.627319431687126 , -5.4492226 )
( -7.207955804049962 , -7.53856 )
( -3.750172853981446 , -4.4249954 )
( 27.460476222700365 , 9.183624 )
( 26.522240507974143 , 8.203971 )
( -7.328400726164517 , -7.683052 )
( 6.072341108186702 , -4.0955997 )
( 2.704202068953242 , -7.134927 )
( 2.037036206585555 , -5.6597853 )
( 3.3181294886155897 , -6.438158 )
( 4.799059169111444 , -5.409939 )
( -4.685096756788186 , -5.2966475 )
( 3.763013496907226 , -4.293587 )
( 28.983554625216815 , 10.4810705 )
( -7.770639777616417 , -7.9941573 )
( 27.45403859562484 , 9.008017 )
( 3.0257740374098576 , -6.8761578 )
( 0.25647011967840017 , -7.326561 )
( 3.272752459952607 , -5.453801 )
( -5.9272216688387225 , -12.213893 )
( 3.0747765404096454 , -4.8026505 )
( -8.038569241966702 , -16.281727 )
( -5.050541274535516 , -5.707135 )
( 0.34600595641583676 , -9.031828 )
( 28.673132361515542 , 10.237376 )
( -5.940311628745455 , -15.1379795 )
( -7.993433898029448 , -17.765602 )
( 4.86724649983393 , -3.2058022 )
( -6.286200530195909 , -12.488496 )
( 4.607480126728213 , -5.3138347 )
( -0.3220136286931321 , -9.753966 )
( 4.304927980611079 , -3.7900343 )
( 1.5448401819701143 , -6.8733716 )
( -1.243613978005303 , -16.903696 )
( -6.492846752996463 , -6.7959595 )
( -4.371450245977409 , -13.12211 )
( 29.292636044241142 , 10.752769 )
( 4.340962795431471 , -5.7459393 )
( -3.6055625052763522 , -12.358425 )
( 3.307942025732043 , -5.5107317 )
( 41.71605270303151 , 18.95273 )
( -9.75825799393571 , -15.518086 )
( 3.1137936036477067 , -6.632189 )
( 0.749527433582081 , -8.674794 )
( -4.712378944304795 , -11.267805 )
( 2.33499182975675 , -7.3087177 )
( 26.07741710924019 , 7.886932 )
( 4.350337015857328 , -3.849785 )
( -7.072632660735184 , -17.009533 )
( 1.5885030577036807 , -6.9945908 )
( 3.358999200951279 , -4.6007633 )
( -7.243656911213327 , -7.5597134 )
( 28.16460746931054 , 9.643274 )
( 2.9476043793671636 , -4.923463 )
( -0.07839158336669882 , -15.830524 )
( -0.2573700197857307 , -9.61285 )
( 25.781080302201456 , 7.612752 )
( 1.7924330199408818 , -7.795311 )
( -3.033547743406295 , -3.8719916 )
( -5.651592348771004 , -16.373795 )
( -4.848260266172271 , -11.225523 )
( 3.657926263806515 , -6.228513 )
( -5.7275204712465175 , -6.1019354 )
( 0.39054285033325187 , -7.102794 )
( -7.070880066448955 , -7.4361515 )
( 4.551197423320765 , -3.575603 )
( -3.376362237148901 , -12.286161 )
( -4.006914503736079 , -10.490519 )
( -6.383671666703839 , -6.6624093 )
( 36.40246355764393 , 14.41259 )
( -8.605498544062526 , -14.446651 )
( -1.9874702197765348 , -11.050585 )
( -12.592806729541639 , -10.672408 )
( 0.22994967924095194 , -9.017608 )
( -3.9246200538839755 , -14.92982 )
( 27.634462477577955 , 9.316537 )
( -2.7904523577570925 , -3.717733 )
( 3.0698862944811616 , -6.643015 )
( -3.1640763915907217 , -4.092356 )
( 0.39793826102693397 , -9.042595 )
( 3.676392893471476 , -6.142051 )
( 28.83238522792501 , 10.238077 )
( 24.287224451661068 , 6.4567065 )
( -7.296232763317365 , -7.520169 )
( -13.478731406415502 , -11.487602 )
( 3.1302041233692215 , -6.715288 )
( 3.2575479920709336 , -4.6978436 )
( 26.548206213550372 , 8.3807745 )
( -4.998077137917798 , -13.630967 )
( 0.6300744088395285 , -6.827143 )
( 1.0728121816095395 , -8.549881 )
( 27.047347584951662 , 7.7307124 )
( 39.65500958119758 , 17.188173 )
( -3.8021681819800115 , -14.058999 )
( -5.638237075720608 , -6.018161 )
( 5.391921514257042 , -4.7801476 )
( 3.76359751327836 , -6.1872115 )
( 0.4375329902270777 , -9.022945 )
( -1.8038124482295501 , -10.865513 )
( -5.678760618200772 , -12.11315 )
( -17.31519344262353 , -14.738144 )
( -4.912901626085253 , -5.5693426 )
( 3.0785379955798744 , -5.7108974 )
( -7.002891610789505 , -7.4044166 )
( -9.77042116993057 , -15.538696 )
( -7.468568395570703 , -7.7299213 )
( 29.06191005929157 , 10.471966 )
( -7.162703410326941 , -17.055937 )
( -4.756995161342651 , -5.4405875 )
( 1.4246313920229121 , -6.1729593 )
( -5.020137111330444 , -11.553379 )
( -5.462358918385754 , -6.0385513 )
( -3.5994501658228657 , -4.348161 )
( -16.8963700936613 , -14.415786 )
( 37.015181645161746 , 14.91942 )
( 24.20909228874411 , 6.3105545 )
( 13.686501465044556 , 2.3814921 )
( 1.1015688097319556 , -7.3779483 )
( 0.2903348603685959 , -7.127053 )
( -5.444269437582314 , -5.865841 )
( -6.093792214168757 , -6.5802174 )
( 0.6180036475955978 , -7.0200424 )
( -4.36882385079552 , -4.9921217 )
( -5.824608676722514 , -14.354057 )
( 1.6345173793976744 , -6.099955 )
( 4.421730077523113 , -5.673976 )
( -4.9063355538026565 , -13.559995 )
( 3.602905505761588 , -6.306274 )
( -18.745198290052 , -16.063211 )
( -8.50385906226788 , -18.202671 )
( -7.236955966072974 , -7.5333853 )
( -0.9369924359706691 , -10.1147175 )
( 3.308854811213803 , -5.37916 )
( 1.7867671482778746 , -11.5755205 )
( -2.3745076560747354 , -11.37314 )
( 36.31686725073701 , 14.332441 )
( 36.010476453851815 , 14.063114 )
( -4.895920032366863 , -5.4213686 )
( 28.13985453954783 , 9.780101 )
( 2.243323291581172 , -5.587113 )
( 4.004292853653088 , -6.040084 )
( 42.58813713423385 , 20.143606 )
( -3.6062896223049816 , -4.513036 )
( -5.755666507495902 , -6.1554413 )
( -2.9445370724727495 , -3.8754442 )
( -15.230440871518676 , -13.09363 )
( -13.328262948482662 , -11.406705 )
( -15.769579660406867 , -13.410984 )
( 2.3038483089284463 , -7.3599024 )
( -6.575023601775807 , -6.8944316 )
( 4.288954507396213 , -5.6276875 )
( -0.9441069026076495 , -8.367856 )
( 3.7672636197743543 , -4.1442575 )
( 0.6099486144663137 , -8.770909 )
( 13.196528237497636 , 2.031159 )
( -8.68554392992226 , -14.523313 )
( 29.9827958467547 , 10.202996 )
( 3.462606124527012 , -6.4328747 )
( -7.3479258091309685 , -13.540163 )
( 4.9774182278158206 , -5.095708 )
( 23.183036835776512 , 5.372976 )
( 2.5654652816801695 , -7.1282873 )
( 2.2782901576965786 , -5.590254 )
( -7.583076658048427 , -17.315695 )
( -17.48762755408501 , -14.989201 )
( -5.76707975216587 , -6.1736875 )
( -5.149957916241362 , -13.6720915 )
( 3.6449551478141404 , -6.2686453 )
( 4.749772822012873 , -5.2365966 )
( 1.8665764233152506 , -7.7633634 )
( 2.2524162547819246 , -7.439589 )
( 0.3844938667898464 , -8.990867 )
( -5.959033858275541 , -12.318352 )
( -4.422777490411916 , -15.370251 )
( 3.0216097605762107 , -6.783477 )
( -2.7801664273247795 , -13.144209 )
( 28.328132104414127 , 9.908262 )
( -6.818294314569681 , -7.205009 )
( -3.2252848495045097 , -4.0441313 )
( 1.3985206716847665 , -6.1937647 )
( -6.89161581861339 , -12.97393 )
( 0.1562032418578967 , -9.303354 )
( 2.7542831576188833 , -5.8942676 )
( 5.837000510813783 , -4.384678 )
( -4.246140545230042 , -14.466449 )
( -2.4853613346705377 , -3.4335127 )
( -4.730095804192757 , -5.260125 )
( -4.02198571624416 , -12.748839 )
( -3.887162363070385 , -4.7268486 )
( 3.0580957626475196 , -6.672926 )
( -6.4203813069769495 , -6.71612 )
( -5.188414144657724 , -11.708014 )
( 41.15105665404912 , 18.523832 )
( -2.1014116857757 , -11.081899 )
( 1.5019234618037993 , -14.4758 )
( -18.521407058784302 , -15.754671 )
( -3.3905481022508352 , -4.2854857 )
( 0.03714196415189597 , -9.2910385 )
( 40.71749486432457 , 18.068075 )
( -17.114955687786832 , -14.61874 )
( 2.663154175028249 , -5.286018 )
( -14.781541359376238 , -12.70093 )
( 3.6551475350014044 , -6.0902834 )
( 26.906876785044638 , 8.678341 )
( -17.926939156376967 , -15.320283 )
( -9.671204370713015 , -15.4822 )
( 1.6436051171871697 , -6.7351437 )
( 3.636822594486676 , -6.092821 )
( 32.23448322527591 , 12.2913885 )
( -5.890544443093427 , -12.152317 )
( 25.57182049348597 , 7.467173 )
( -14.591012087461893 , -12.350724 )
( -6.522446667412835 , -12.812398 )
( 0.26818130943996654 , -7.97974 )
( 25.75444095556324 , 7.6812596 )
( -0.6608149282753726 , -8.130095 )
( 27.696445032533898 , 9.263458 )
( 0.1537869736242592 , -7.259136 )
( 2.578916768016769 , -5.270219 )
( -6.7519016376136625 , -17.192465 )
( 3.91236539782608 , -4.8107843 )
( 1.1169769921144033 , -8.247723 )
( 1.925515998828521 , -7.706348 )
( 27.786384048825067 , 9.447005 )
( -8.494040216564967 , -14.428933 )
( 29.216948894844407 , 10.539953 )
( 29.098894058585238 , 9.503365 )
( -2.7494244517267474 , -3.575105 )
( 26.476144030990042 , 8.366704 )
( -8.140556005147703 , -8.367523 )
( 4.183160855518565 , -5.8757925 )
( 5.099141561442456 , -3.0236945 )
( -0.781332160967925 , -8.165685 )
( 4.4107375715566866 , -4.517605 )
( -8.267456851116869 , -14.332022 )
( 5.4388182737085335 , -3.5112476 )
( 0.16570039107644086 , -9.136511 )
( -2.8408009214492216 , -3.7795115 )
( 26.171191471599357 , 8.072614 )
( -14.473009102877892 , -12.2765665 )
( 25.569188621037135 , 7.4776425 )
( -3.1893069459915493 , -13.561352 )
( 3.683153809958865 , -6.2603974 )
( 4.54189211285806 , -5.3198705 )
( 28.281342300978654 , 9.8770895 )
( 1.0545206321049703 , -12.328381 )
( -7.056040541960307 , -7.324956 )
( 28.44092616763587 , 10.043753 )
( -17.385738081776363 , -14.902921 )
( 4.591517058412605 , -3.4714885 )
( 3.746318308308042 , -5.174675 )
( 1.4282087506501169 , -14.67624 )
( -6.111134539945006 , -6.5394635 )
( 0.42932714104570124 , -7.036692 )
( 9.236546043196796 , -1.4244118 )
( 0.5306666048330058 , -8.975592 )
( -5.466704421523893 , -15.439581 )
( 1.481899388380282 , -8.126427 )
( -3.648497039275275 , -4.336553 )
( -0.7014985658366878 , -9.858268 )
( 3.500259522015294 , -6.4395556 )
( -7.329335159387352 , -13.488724 )
( 3.0520695768548447 , -6.84824 )
( 2.207911215297216 , -6.3132405 )
( -8.045068171877206 , -13.984383 )
( -4.214770275386082 , -4.8330727 )
( 4.024919984368261 , -4.073023 )
( 1.1647719574135098 , -6.5402713 )
( 2.449515718102006 , -7.1945696 )
( 28.95830011092621 , 10.348471 )
( 2.2390213958613465 , -7.4401383 )
( 4.251861930639272 , -5.7360263 )
( 4.261740405361794 , -3.7194297 )
( -10.334212926261952 , -16.039078 )
( -3.069691934674182 , -4.0230646 )
( -7.197358768276895 , -7.4543104 )
( -5.35760206917578 , -11.762748 )
( 5.402233562091136 , -2.83413 )
( 0.09236553467451558 , -9.40405 )
( 24.435467200405196 , 6.597245 )
( -18.139985072557096 , -15.423223 )
( -2.599975337403807 , -11.649824 )
( 4.048992937736649 , -3.912674 )
( 4.412513767972631 , -3.7129514 )
( -7.451455797456186 , -7.714012 )
( -6.454539821282772 , -6.7831755 )
( 5.0567257120060205 , -5.159898 )
( -4.824585717999195 , -11.270037 )
( 3.2644214315641875 , -6.6084666 )
( -6.768729524868202 , -7.130002 )
( -5.131392841842633 , -11.461806 )
( -6.884391763250255 , -13.162588 )
( -6.248782464055826 , -6.6678324 )
( -5.173883397860826 , -5.8350983 )
( 25.438016454548798 , 7.430523 )
( 3.7263657496884894 , -6.1345873 )
( 4.142480091093228 , -4.0397735 )
( 3.593981530588276 , -4.306045 )
( 2.831545062893946 , -5.0312967 )
( -7.181712279855059 , -15.522822 )
( -0.19200386988015516 , -9.511755 )
( 29.283937576301923 , 10.762413 )
( 28.65643404531277 , 10.1547575 )
( 2.987414105922891 , -6.916917 )
( 3.735042479340909 , -4.982245 )
( 0.8170177431901178 , -8.486688 )
( -5.5394617164628785 , -6.0178494 )
( 1.5333000926032083 , -6.267022 )
( 3.8256080755158512 , -5.9880986 )
( -2.6776618102144965 , -18.048267 )
( 38.28472570117899 , 16.044659 )
( 1.2028328714263181 , -8.234768 )
( -3.634311930925045 , -12.479593 )
( 2.916823527558439 , -5.6625166 )
( 5.690648471516558 , -4.5076857 )
( 26.24973517229346 , 8.068145 )
( 24.070242497472407 , 6.167318 )
( 2.5619542912634383 , -7.223472 )
( -3.383329134020242 , -13.827731 )
( 0.43429611655229294 , -7.838048 )
( -6.834102513280582 , -7.209876 )
( 3.746324425548132 , -6.129756 )
( -4.924417873843612 , -5.5598717 )
( -17.046049321077643 , -14.502774 )
( 25.099748743790823 , 7.1034126 )
( 1.70765052694141 , -7.7351513 )
( 1.7230210919009417 , -6.7673903 )
( -4.954566290212696 , -5.5560417 )
( 0.828170250744505 , -8.674606 )
( 2.7257033574478706 , -6.90397 )
( 0.5917403365937548 , -8.9235325 )
( 0.49613137819088937 , -15.417576 )
( -3.2038338925840675 , -4.1206856 )
( 11.679672154586893 , 0.5788151 )
( 27.49248095317935 , 9.224079 )
( 1.3777028967021867 , -8.08418 )
( 0.9429381403833497 , -8.593141 )
( 4.5215737939594005 , -4.3738456 )
( -17.51389400617633 , -15.056053 )
( -3.6933514589345466 , -4.430229 )
( 0.7575054044845271 , -8.821707 )
( 28.619486266314148 , 10.021659 )
( 4.391060777999038 , -5.582555 )
( -0.7707643610161211 , -8.164225 )
( 3.5568963633201807 , -6.36208 )
( 28.65480286043933 , 9.121892 )
( 0.6200070329056891 , -6.9203067 )
( 1.1630522038911022 , -8.499506 )
( -0.07290451630059769 , -9.542443 )
( -7.094639936437648 , -15.432123 )
( 0.01135821245542512 , -7.5202203 )
( -0.5876798359584836 , -8.008498 )
( 2.151154664667139 , -7.5336776 )
( -4.045969544741744 , -4.8565407 )
( -31.963986691057187 , -21.1783 )
( 5.270500403615802 , -2.9096074 )
( 4.833007116651967 , -3.2574878 )
( 0.33040758193618003 , -9.1209755 )
( 4.678869807818497 , -3.5397303 )
( 3.72255471164918 , -6.1437435 )
( -7.21706790569163 , -7.565748 )
( 4.668096698925162 , -5.4721694 )
( 1.4669215032884617 , -6.1397567 )
( -2.1964101560671168 , -11.188883 )
( -16.00126685615424 , -13.692698 )
( 1.84711867429494 , -7.6305356 )
( 28.73824751307284 , 10.311522 )
( 0.015416296277850572 , -7.4870996 )
( -3.859657948889683 , -4.50616 )
( -16.54611860199655 , -14.190331 )
( 4.765780355604369 , -3.4304113 )
( -6.623510913291959 , -6.859576 )
( 0.9944501187987387 , -6.593591 )
( -7.35324830125218 , -15.58271 )
( -4.245120161576652 , -5.027862 )
( 0.9824396788699158 , -6.6518617 )
( -1.0101009312664737 , -8.403458 )
( -0.26441726365205964 , -7.6913056 )
( -6.804601975867051 , -15.136467 )
( 3.6678374629051773 , -6.107295 )
( 1.5200992551864259 , -6.082721 )
( -6.640442287069489 , -12.754278 )
( -17.611791199317036 , -15.10163 )
( -18.301008014046303 , -15.719962 )
( 1.4822617500715247 , -6.1009626 )
( 26.65738729251772 , 8.367191 )
( 4.758169932681535 , -5.4176683 )
( 4.229806788321008 , -3.8843997 )
( 4.409213079265041 , -5.614603 )
( 1.9684101197481665 , -5.890006 )
( -5.41655310340806 , -5.8500667 )
( 5.001200826881892 , -5.147388 )
( -5.538656610594286 , -14.07947 )
( -2.5326405400216743 , -3.5197294 )
( -13.665908979148861 , -11.598055 )
( -4.007195725220788 , -12.685447 )
( 1.8691190617086977 , -7.6176205 )
( -6.2575051540347575 , -6.570481 )
( 3.219591789060618 , -6.6442924 )
( -2.7890098880469925 , -3.8077796 )
( -0.7917174793611809 , -8.17874 )
( 38.29734358176616 , 16.13898 )
( 4.006973499363639 , -5.862313 )
( -3.8688935791148458 , -12.709575 )
( -19.181897066205515 , -16.450962 )
( -5.316942434976272 , -11.7264805 )
( 0.6840614569378541 , -6.902215 )
( -13.18663441345779 , -11.225505 )
( -8.273093351006004 , -14.197372 )
( 4.963183265470731 , -5.1898446 )
( -2.912502025523548 , -11.864471 )
( 2.038734060396331 , -7.570003 )
( -5.6596961733590225 , -6.0853887 )
( -2.6068297471467847 , -3.5667853 )
( -17.027896744025917 , -14.537728 )
( 4.626302937446828 , -3.5994563 )
( -10.384356559937995 , -16.10277 )
( 31.941156426255226 , 11.973918 )
( 29.503522176341242 , 10.877116 )
( -5.810343093045505 , -6.2069483 )
( -5.464872436535039 , -6.087069 )
( -2.337355020270669 , -3.300333 )
( 0.8379530925403071 , -8.60255 )
( -3.296738937570999 , -4.1123495 )
( -6.354279539627521 , -6.756185 )
( 1.9423264542244754 , -7.7471304 )
( 0.1513386172541208 , -7.412763 )
( 28.527490537240737 , 9.976872 )
( -3.5186832907819845 , -13.784222 )
( -14.161025195925566 , -12.215408 )
( 31.193255856031794 , 11.402729 )
( -6.8327763523995895 , -7.158149 )
( 1.9622771151763954 , -5.847143 )
( 2.6957250093086604 , -5.2015514 )
( -1.0477950551020636 , -8.437412 )
( 3.4458482622990108 , -6.3548765 )
( 28.054247092015483 , 8.556139 )
( 5.425816640820411 , -4.7114353 )
( 2.0650568913522687 , -7.5818653 )
( -6.91014468842319 , -7.298506 )
( 0.08694153646173786 , -9.377256 )
( -16.27521438002538 , -13.80834 )
( -7.269320341850966 , -13.320117 )
( 23.11388801366832 , 5.3080907 )
( 42.64145837555966 , 19.892746 )
( 1.5084081594706746 , -8.071563 )
( 0.7861593595076398 , -6.901784 )
( 3.9487420221971217 , -5.9202776 )
( -0.14251596396028116 , -9.309264 )
( 0.1045719695026629 , -8.2145605 )
( -4.053130929284488 , -14.437194 )
( -7.868567139935692 , -8.0764475 )
( 15.913582179264258 , 3.1421707 )
( 4.152984474812349 , -5.7882323 )
( 25.8364127222377 , 7.8341026 )
( 1.0025193699001047 , -8.4237 )
( 2.365250240545956 , -7.4748864 )
( 24.21803850692122 , 6.368123 )
( -6.232390259023143 , -6.6558776 )
( 4.159638913737969 , -5.801836 )
( 26.50931744286942 , 8.408279 )
( 3.2717394304857668 , -5.545498 )
( 2.1818561552932065 , -5.4997554 )
( 28.26282148933871 , 9.684173 )
( 26.129646472842097 , 7.8849745 )
( -12.606131243179075 , -10.687131 )
( -2.301154228229576 , -11.376796 )
( 33.60124531926155 , 12.117475 )
( 5.352166435460048 , -2.8327122 )
( -6.0565001753488605 , -6.5497074 )
( -2.4048822287984724 , -3.4787145 )
( 27.85942666441861 , 9.389284 )
( -4.26214450974834 , -10.822712 )
( 38.94064862528118 , 16.589445 )
( -14.095711471975356 , -11.945796 )
( -5.240586824869009 , -11.641887 )
( -5.755004881535787 , -6.1378303 )
( 27.40505518547066 , 8.972687 )
( -6.978534189446581 , -15.361569 )
( 26.45868003168448 , 7.238616 )
( -0.26455539625096947 , -9.411212 )
( -7.065145972867826 , -7.2929306 )
( 27.872123972874565 , 9.52585 )
( 1.8731667824956926 , -7.759322 )
( 1.8736731473031414 , -7.724966 )
( -16.57579628358038 , -14.203669 )
( 13.345515146911751 , 1.9709857 )
( 0.4362260795403329 , -7.854273 )
( -6.893492857799388 , -16.107807 )
( 1.4532656963828732 , -6.2875237 )
( -0.0848452699779021 , -9.435914 )
( 2.5069549519272627 , -5.3673277 )
( -2.5632039887006153 , -3.4624734 )
( 4.079116046313742 , -5.8814545 )
( 25.844800684621543 , 7.737664 )
( -6.447528731370772 , -6.9103622 )
( -7.20498655130829 , -7.4672074 )
( -0.002367384625082125 , -7.5231576 )
( -2.684433804401487 , -3.5803075 )
( 0.6402228361483312 , -6.9914675 )
( 0.5104158586024399 , -8.939097 )
( -5.44322006186124 , -5.929719 )
( 4.76572818129957 , -5.4383473 )
( 4.060358779322164 , -4.089585 )
( 4.76036387606734 , -5.29569 )
( 4.044633999102522 , -5.8821564 )
( 28.403631002739456 , 8.963632 )
( 1.8491070771338451 , -5.7942924 )
( -14.054821641877744 , -11.906437 )
( 4.311481804649467 , -5.705115 )
( -6.910989789460332 , -17.49092 )
( 3.180194830972323 , -5.4272814 )
( 5.181646897621209 , -3.7620761 )
( 2.2532983345077615 , -5.65211 )
( -3.765352720627778 , -14.768337 )
( -6.807855315232329 , -7.169405 )
( 2.8688533507973073 , -5.807884 )
( 23.68011157272641 , 5.8668523 )
( -6.526974741174766 , -12.841042 )
( 4.120819755353602 , -4.725824 )
( 27.70999962523249 , 9.23748 )
( 27.956460619299225 , 9.598842 )
( -4.921769173905849 , -5.4820123 )
( 2.9547672779428913 , -5.8456945 )
( -7.122898927497952 , -7.40718 )
( -8.140326354838548 , -8.341388 )
( -6.7849415018609385 , -15.212117 )
( -1.8603217610859168 , -10.927504 )
( 0.4280701024451027 , -8.995624 )
( 5.282286948952137 , -4.7048287 )
( -4.643156934273097 , -5.2935266 )
( 6.1231064172039575 , -4.10439 )
( 3.541010521173916 , -6.346133 )
( -7.600000846277389 , -7.8364964 )
( 5.2865503823612015 , -4.7864084 )
( 0.9597955647529667 , -6.6203403 )
( 1.0294338138423016 , -8.462491 )
( 23.747482707801122 , 5.83974 )
( -3.954905109855241 , -4.7893877 )
( 3.942963479367927 , -4.0970016 )
( 29.37383466820609 , 9.80772 )
( 27.075435861863504 , 8.823834 )
( 3.333489889105123 , -10.236821 )
( -6.292511355630256 , -6.765739 )
( 4.17517767123178 , -5.801771 )
( 4.524379087004291 , -3.6677177 )
( 0.9505328958849821 , -6.6545024 )
( -4.734188502327113 , -11.283291 )
( 2.060763619146849 , -16.019272 )
( -0.301809148168086 , -7.777925 )
( -2.2443936330900014 , -3.258086 )
( 3.644219168923745 , -6.275248 )
( 24.95312084458878 , 6.9408393 )
( 5.212385825868532 , -4.979027 )
( -6.367327632651129 , -6.84559 )
( -0.09091191997368553 , -8.380713 )
( 1.345663969065874 , -8.228157 )
( -0.9494849904339464 , -2.0574632 )
( -8.932169436586655 , -14.874353 )
( 28.29320613025725 , 8.7493305 )
( 26.99631167223887 , 8.701223 )
( 2.3662159245458483 , -7.3699055 )
( 28.988096430939002 , 10.340494 )
( 4.012529683719399 , -4.158914 )
( -3.929462170478843 , -4.6233683 )
( 30.16281552579482 , 11.470796 )
( 27.3863861880228 , 9.119246 )
( 28.108435551901735 , 9.559869 )
( 1.5476597748278356 , -6.2058063 )
( 0.37828325382011485 , -7.9150267 )
( -3.9925915151921214 , -12.648194 )
( 4.994829705669159 , -4.9762077 )
( 5.015611122557139 , -4.932096 )
( 27.535709642548348 , 9.2860155 )
( 1.2983190190358282 , -8.285376 )
( 2.27196808779932 , -7.4164653 )
( 0.07603708474583737 , -9.397164 )
( 28.709747507188887 , 10.120183 )
( 3.66751946964645 , -5.0510283 )
( -17.868621373890043 , -15.34692 )
( -3.7821408907213634 , -10.365394 )
( -4.835991987422307 , -13.481489 )
( -5.517488487992516 , -6.0185103 )
( -5.581693077580376 , -6.0717583 )
( 28.25617161644313 , 9.689868 )
( 2.875180967844578 , -4.995784 )
( 41.4699358031516 , 18.72502 )
( 37.7568685869549 , 15.528465 )
( 2.9302964723710545 , -6.973382 )
( 25.838613188115843 , 7.7312665 )
( 28.288954805734537 , 9.756302 )
( 37.462085578874806 , 15.346285 )
( -2.435727117683876 , -3.3676481 )
( 3.541665740883162 , -6.343174 )
( 5.013263232770628 , -3.1669807 )
( -2.3360693253180624 , -3.2663815 )
( 4.907029560033907 , -3.2379866 )
( 4.149849820343989 , -3.8842158 )
( -1.4805529975790996 , -17.111 )
( -0.5010722003808228 , -9.911394 )
( 27.008597027314757 , 8.723086 )
( 2.9501712340922475 , -4.9427385 )
( 4.48591237533181 , -5.5611324 )
( 1.5629521916279852 , -8.0616865 )
( 2.136715750243579 , -7.6124077 )
( 4.243044327651825 , -5.78602 )
( 5.775095567587289 , -2.4446435 )
( -5.054479622661488 , -15.8874855 )
( 2.2358976156500323 , -6.282707 )
( -4.86429978767251 , -15.13942 )
( 4.438618081924648 , -3.705066 )
( -2.9368157397150796 , -11.786312 )
( 5.35042806803198 , -2.8043272 )
( 2.5903117267787876 , -6.131446 )
( -3.5742571814756303 , -4.312094 )
( -1.4083748049945863 , -10.5708065 )
( 28.579673486283188 , 10.0916195 )
( -7.454228374338955 , -13.651838 )
( 0.6871985359805078 , -8.711714 )
( 4.582940010688571 , -5.53742 )
( -3.61760844990468 , -4.4813905 )
( 25.712190385284533 , 7.5079713 )
( 26.048807425570068 , 7.918631 )
( 1.175810782484588 , -6.389774 )
( 2.1153489963348178 , -5.6061163 )
( 26.70319877520511 , 8.418203 )
( -5.3564381312076925 , -11.825421 )
( 0.4617314968695978 , -8.988271 )
( -2.1851840178912134 , -3.0903997 )
( -0.18653194916055826 , -8.466262 )
( -0.3580390464991162 , -8.5756645 )
( -6.88963612152839 , -16.73021 )
( 24.607304887990043 , 6.7333326 )
( 5.367044362804423 , -4.726308 )
( 0.6133870990284244 , -7.7664795 )
( 38.18302469735766 , 15.947941 )
( 26.81042994319711 , 8.470539 )
( -8.072231413961456 , -8.305355 )
( 3.3544964138311615 , -6.478014 )
( 4.703237800473055 , -3.4706779 )
( 1.617281598122137 , -14.458218 )
( -6.215436493591991 , -16.134691 )
( -4.6029958217738205 , -15.393894 )
( -0.27046693525310095 , -7.6462417 )
( -6.05550000588457 , -16.030663 )
( -18.640342871074495 , -15.871736 )
( 3.5737278928608904 , -4.3033695 )
( 4.377083163375056 , -5.604987 )
( -8.928292500975154 , -14.814927 )
( -2.993089675595109 , -3.9106586 )
( -5.91129161137945 , -6.273827 )
( 0.07777440229650301 , -9.170161 )
( -4.563503816599706 , -5.2565236 )
( 4.5479025744505055 , -5.5330687 )
( -5.747666169120658 , -6.2261806 )
( 3.3464539037280083 , -4.5915947 )
( 5.36466317756775 , -4.804898 )
( 1.7498882807587182 , -7.8872867 )
( -6.102023345186945 , -12.4918995 )
( 4.862722192134202 , -3.2642727 )
( 29.833750518813776 , 11.207285 )
( -2.749722685471707 , -3.5985126 )
( 25.814323936796473 , 7.633175 )
( 30.10550657472474 , 11.458857 )
( -2.013451872203583 , -3.075007 )
( 39.089912332050034 , 16.843348 )
( 29.35211764968348 , 10.723595 )
( 1.331635876177745 , -6.251796 )
( -4.513937957937506 , -5.2334356 )
( 32.34214722525529 , 12.348341 )
( -2.2482789170329394 , -12.742494 )
( -9.175351352035662 , -15.003475 )
( 26.738721427630843 , 8.5260105 )
( 2.404775553886237 , -5.375407 )
( -7.345467407766396 , -17.164787 )
( 4.350811652230849 , -3.6561542 )
( 2.4653829812405395 , -7.265791 )
( 2.1445807422805925 , -7.51372 )
( 3.9800334133454776 , -4.0880513 )
( 26.05907671262275 , 7.857216 )
( 3.565618252276166 , -6.4384546 )
( 10.150652034300071 , -0.75438726 )
( -6.0281011405649885 , -6.543644 )
( -18.075837581126383 , -15.386632 )
( -0.594503112917149 , -8.774491 )
( -2.021413012660318 , -2.9886947 )
( 2.161927562154982 , -5.7109804 )
( -3.3297237870664844 , -4.142879 )
( 3.8275126141486746 , -6.1296425 )
( 2.5256612553294024 , -5.295642 )
( 0.2930820248539542 , -7.0964713 )
( 0.4115296153322139 , -9.126928 )
( -3.3982637157392643 , -18.810442 )
( 5.211538125064196 , -4.8776293 )
( 26.00433116647994 , 7.824921 )
( 28.80044277379544 , 10.213755 )
( -3.3681727656153955 , -12.266691 )
( -3.0898043559386705 , -3.9975789 )
( -0.04493622574912115 , -7.6050034 )
( 25.676035992322273 , 7.526033 )
( -0.9137059039220332 , -10.16214 )
( 5.186072119196561 , -4.960434 )
( 3.0539471984017967 , -4.75851 )
( 30.672533999654107 , 10.80728 )
( -2.147028338466029 , -3.053193 )
( -3.9206907255308088 , -12.718622 )
( 5.991744084732627 , -2.26307 )
( -10.140166464697986 , -15.921195 )
( -29.125382892296972 , -18.642538 )
( 25.595317478862242 , 7.4504275 )
( -4.258971591302181 , -4.8677154 )
( 0.9488058781723643 , -6.6202593 )
( -3.8286977656838364 , -4.6962543 )
( -3.016579824561195 , -3.8789046 )
( 26.829911386780736 , 8.556135 )
( -0.1344390666528572 , -9.563635 )
( -0.1493012415939271 , -8.48921 )
( -14.030090010810195 , -12.05608 )
( 27.66998069231172 , 9.250923 )
( 4.404555865561246 , -5.6987305 )
( 27.5661525023216 , 8.234638 )
( 24.362120472168506 , 6.5338 )
( -17.632339564327463 , -15.087212 )
( 24.43159701498322 , 6.580624 )
( 1.9093348965550496 , -7.661863 )
( 3.9085840431873486 , -4.1323795 )
( 4.945005371639416 , -5.2355685 )
( -18.922503146412016 , -16.229599 )
( 4.621461542836914 , -5.4021134 )
( -3.408959432756644 , -4.2713056 )
( -4.883786669628401 , -5.53385 )
( 2.0667769312781052 , -7.5650992 )
( -3.276507559917107 , -4.056751 )
( -6.281471623732172 , -6.7733974 )
( 1.6232059675017563 , -7.935751 )
( 3.3494164776066935 , -4.7266455 )
( 26.367892556596207 , 8.14577 )
( -6.673093181834586 , -12.845053 )
( -5.064504400929529 , -13.586773 )
( -5.9576582680552495 , -6.453798 )
( 2.8171689378143423 , -4.962522 )
( -3.56466429887997 , -13.257122 )
( 41.34610604788379 , 18.757183 )
( -3.879956661684385 , -4.6196494 )
( 4.646635466876867 , -5.3091226 )
( 0.10993052194986408 , -9.299065 )
( -16.29835151716548 , -13.992435 )
( 26.591937205952693 , 8.347476 )
( -18.054252718351034 , -15.502033 )
( -2.926494428353991 , -3.9050162 )
( 40.16103170623735 , 17.610987 )
( 1.743805132498711 , -7.8395915 )
( -3.669567000212038 , -4.4934893 )
( -0.32016493842772453 , -7.7779446 )
( -4.637230707830507 , -11.187366 )
( 27.544909238036293 , 9.253354 )
( 4.133051035470853 , -5.9194455 )
( 4.660243546709455 , -4.3149652 )
( -10.037363965267035 , -15.76145 )
( 26.69418381042812 , 8.467854 )
( 26.490542824408564 , 7.315309 )
( 2.431255341708849 , -7.382926 )
( 2.6043881903947694 , -5.320357 )
( -2.154502139737051 , -3.1832461 )
( -5.337260884707278 , -15.529906 )
( 1.863296502217234 , -7.6924047 )
( 3.4356950041295145 , -5.3961573 )
( 26.98151997162104 , 8.701686 )
( 0.7569397154703847 , -8.679827 )
( -1.9983037850961693 , -3.0354013 )
( 4.2201682831261405 , -3.851843 )
( 0.6889678202960818 , -8.822645 )
( 3.6953767531576722 , -6.280512 )
( -13.362360944075819 , -11.333585 )
( 4.863357063774673 , -3.2216172 )
( 26.672129264282983 , 8.54246 )
( -16.879429009398017 , -14.305501 )
( 2.1206020428948604 , -5.675569 )
( -19.115868222206863 , -16.405563 )
( -8.070801562121188 , -8.314745 )
( -13.879476393008247 , -11.810243 )
( -16.294627310246064 , -13.996075 )
( 2.807564595586782 , -4.9523363 )
( 1.3584284120611836 , -8.161961 )
( 40.966375180506134 , 18.3857 )
( -0.7101576659375551 , -8.097512 )
( -18.440394541227697 , -15.836866 )
( 1.6185146444403973 , -8.029251 )
( -4.865605223013 , -11.361472 )
( 2.09845034747214 , -7.5123434 )
( -4.838426672216019 , -5.4115644 )
( -2.5434625028514297 , -3.416066 )
( 3.803843058881105 , -5.966516 )
( 3.1481359642183344 , -6.6188297 )
( -2.3601812676393115 , -3.2976735 )
( 29.08364075441755 , 10.522272 )
( 3.233176186765286 , -6.7059493 )
( 1.282478991059858 , -6.3146715 )
( -3.8563324038527016 , -4.5057387 )
( 24.600708390054137 , 6.639674 )
( -4.264380023235855 , -5.0606775 )
( -0.797976621750714 , -10.059666 )
( -6.8273603678774695 , -7.205263 )
( 3.9181519854685187 , -6.1003127 )
( 2.8751477584248866 , -6.9509597 )
( 25.841286493128727 , 7.7973895 )
( 3.4931725761336683 , -6.3092065 )
( 3.1688456958247944 , -4.7989883 )
( -4.494910598820105 , -10.972187 )
( -1.149821651321784 , -10.369938 )
( 0.027447892876183655 , -9.3551 )
( 4.771497260721855 , -3.336261 )
( 33.078514280475694 , 11.573475 )
( 0.3400702726877992 , -7.2449827 )
( 28.90702558303202 , 10.390784 )
( -4.862291010376643 , -15.102772 )
( -0.7886614529098784 , -8.157232 )
( -15.160440740977998 , -12.997223 )
( -5.43224813170333 , -11.824487 )
( 1.464659466104001 , -8.191924 )
( -2.90037503747201 , -3.889411 )
( -7.863119345463469 , -13.883218 )
( 24.667062139718993 , 6.7201943 )
( -6.592208435490847 , -6.8317313 )
( 26.959656240910554 , 7.693486 )
( -3.7962641022073362 , -4.4943476 )
( -4.717934043013626 , -5.2650776 )
( 3.203500068793475 , -4.820799 )
( 1.769759389383288 , -7.686323 )
( 3.6245112510749946 , -4.3606052 )
( 25.950358516044467 , 7.9078765 )
( -17.852103954497018 , -15.345373 )
( -17.16066747474745 , -14.694075 )
( -9.093575508316974 , -14.872782 )
( 28.075711812232115 , 9.634254 )
( -3.13937222419788 , -4.0509405 )
( -3.134356831824396 , -4.1225505 )}

\newcommand{\utilityB}{
( -0.37513290753492834 , 8.357604 )
( -4.4756778362243015 , 24.20777 )
( -12.955751524089276 , 3.0985262 )
( 2.2594629568305407 , 12.598244 )
( -9.363319513009078 , 7.0449595 )
( 3.3714726908889396 , 11.657818 )
( 3.9848515396424764 , 15.652487 )
( 24.627006325746763 , 13.012259 )
( 2.9667489169882897 , 11.7737465 )
( 5.372542325595924 , 14.660285 )
( -7.319050367127131 , 23.279106 )
( -1.9212755572910325 , 22.889668 )
( 23.28771502357361 , 13.701676 )
( -0.6618809378310787 , 15.241055 )
( 26.226532989784058 , 10.905085 )
( 4.329692556274253 , 15.722944 )
( -2.4878443417469267 , 24.182808 )
( 1.5626732350581154 , 13.512923 )
( -7.261973260637454 , -3.66957 )
( 39.67254991022935 , -4.3639197 )
( -4.324927613710429 , 24.501253 )
( 1.8620334192062409 , 17.516306 )
( -5.382943865164341 , -2.3454947 )
( -18.50877480328292 , 3.0341988 )
( 1.5892076789424165 , 13.349851 )
( -5.748864000689925 , -8.342764 )
( -6.222931791522218 , 24.233213 )
( 1.347687202807038 , 14.413517 )
( -16.83518336342517 , 4.4317513 )
( 26.808809343199183 , 12.175035 )
( -2.8627027813719597 , 2.4423993 )
( 25.678879824998134 , 15.396093 )
( -5.836786196106895 , 22.787346 )
( -6.826177572635219 , 4.1150155 )
( 41.58945732148878 , -5.0317607 )
( -16.524385355179 , 1.5984886 )
( 3.5287228448825063 , 16.791733 )
( -2.2127953475323707 , 22.290127 )
( 0.05240143960428967 , 2.4503422 )
( 4.563173064765126 , 0.8928215 )
( -0.5709987755064919 , 14.686629 )
( -4.435849304125644 , 4.3661013 )
( 4.012934724974026 , 18.306591 )
( 6.385262475399307 , 14.167708 )
( 4.1875410060157625 , 17.172712 )
( -12.954497830151613 , 5.0111346 )
( 3.9026150345215385 , 7.450217 )
( -4.379290593126323 , 3.05701 )
( 1.6493026229300636 , 16.283289 )
( 1.5177597118334367 , 16.988836 )
( 28.932690250529355 , 15.253812 )
( 3.9319449279350507 , 14.751655 )
( -6.792213811267834 , 19.941248 )
( 4.023771381092057 , 17.247972 )
( -0.7714267361472669 , 9.99221 )
( 29.048250097401116 , -1.0212834 )
( -0.22829522722298434 , 14.681875 )
( -5.221123957613866 , 8.440013 )
( -1.6425804497446865 , 2.2848501 )
( -5.79172207743037 , 21.989494 )
( 4.374827595975566 , 18.11496 )
( 1.1338339012569718 , 13.172646 )
( -2.308285372230004 , 2.6409483 )
( 3.6274468576063432 , 13.708494 )
( 3.2853107780034083 , 11.316001 )
( 2.2120061420248676 , 18.640255 )
( 25.301599330029426 , 13.116119 )
( 27.480971591502115 , 16.263063 )
( 29.46440674560889 , 13.338202 )
( -0.6882492159339195 , 16.092697 )
( 29.04773459095626 , 11.093094 )
( 0.4605892341305434 , 13.150663 )
( 2.7724849277070427 , 17.587463 )
( -9.606811283123438 , 7.586171 )
( 3.8076693960274737 , 11.529827 )
( -5.430970241680317 , 18.683699 )
( -7.404156245680586 , 0.12544215 )
( -5.603669924995814 , 19.017967 )
( 25.937940942646236 , 14.173782 )
( -6.256321508891226 , 3.8896139 )
( -4.287894226852044 , 22.92746 )
( -0.26768113034742747 , 16.646023 )
( 33.35185206591384 , -6.3127193 )
( -4.75910493977257 , 22.97676 )
( 25.6670549356605 , 13.860216 )
( 5.1095919554512435 , 16.647491 )
( 29.03300987140308 , 14.362402 )
( 2.594828311238421 , -19.48729 )
( -15.034544077241167 , 4.552505 )
( -8.096427939297486 , 4.802056 )
( -1.3954581336180603 , 0.84742403 )
( -13.656389229084807 , 5.8854504 )
( -6.509169903006968 , 21.088593 )
( 2.545766310242424 , 18.304903 )
( -5.794257336100853 , 24.124851 )
( -1.9493454495629372 , 1.3161626 )
( -0.34850014875493385 , 17.019165 )
( -4.0404480543161085 , 21.74469 )
( -2.3725192914653075 , 23.922653 )
( 29.11348330369907 , 11.839474 )
( -3.09155277223511 , 24.085615 )
( 3.4328376532717506 , 12.174804 )
( 28.679978460799184 , 11.610272 )
( -3.1758372576561595 , 1.7844274 )
( -1.7273143495709569 , 1.1424515 )
( 1.800110559881793 , 14.080949 )
( -0.26553640811603374 , 17.279898 )
( 4.258338962343654 , 4.2898803 )
( -4.28634846341409 , 20.828999 )
( -17.359879511570323 , 5.7392063 )
( 2.2406905639915777 , 12.48512 )
( 26.054467045055574 , 10.897665 )
( 0.0986898241311771 , 7.6605873 )
( 0.37293770650773483 , 15.350027 )
( 1.455336796791996 , 14.862959 )
( -0.19779704437725154 , 16.438456 )
( -1.3859390967681726 , 21.207586 )
( 2.978012035878329 , 13.6804085 )
( 2.9679756497385203 , 18.643368 )
( 1.8718188756716994 , 14.075289 )
( -6.123602212558979 , 19.150307 )
( 30.162861669896948 , -1.9527446 )
( -1.3450862278325078 , 9.546655 )
( -29.221138423652057 , -14.854486 )
( -0.082048968459977 , 5.9434485 )
( 1.1060294596577878 , 7.318995 )
( -4.308822866363755 , 3.2425196 )
( 4.6135275773623174 , 17.12469 )
( 23.182380836187225 , 14.129254 )
( 1.861071667952459 , 14.207953 )
( 5.097722356624128 , 13.729351 )
( 2.8894690212558864 , 16.645962 )
( -15.40479428287016 , 2.3673697 )
( -3.214154916694754 , 0.5833692 )
( 2.6788968185364403 , 9.704934 )
( -4.827101775898062 , 4.244535 )
( 2.2752608006908304 , 13.303724 )
( 3.6288299078960717 , 18.684052 )
( -3.7431577023115628 , 0.1597448 )
( 1.378639653246807 , 7.454747 )
( 25.354024933687253 , 13.582498 )
( 4.86964147650852 , 13.211018 )
( 4.610601854273379 , 17.193241 )
( 2.6732349762428504 , 14.684083 )
( 3.3512618560631884 , 15.379757 )
( -2.188357830365444 , 24.422798 )
( 1.9038018459498423 , 15.3588505 )
( -0.6795200956670128 , 15.878275 )
( -5.286080576033016 , 7.9630966 )
( 4.631465183014006 , 7.4087725 )
( 39.18854011028482 , -7.29641 )
( -3.9523750873830736 , 0.086081445 )
( 0.8294539679560815 , 14.772791 )
( 5.080552546594476 , 17.11879 )
( -17.21173700037463 , 5.730657 )
( 3.7565672992232493 , -0.5840122 )
( -5.593458299517989 , 0.9388166 )
( 27.24456647612582 , 10.276867 )
( -17.883737687513648 , 3.1233628 )
( -16.02293573209296 , 3.677618 )
( -0.30991637104190906 , 10.1245575 )
( 0.3278306439438936 , 5.121175 )
( -5.508003141973301 , 4.1364937 )
( 4.627319431687126 , 16.276423 )
( -7.207955804049962 , 23.003668 )
( -3.750172853981446 , 19.843784 )
( 27.460476222700365 , 11.331486 )
( 26.522240507974143 , 16.064041 )
( -7.328400726164517 , 24.14957 )
( 6.072341108186702 , 13.4482565 )
( 2.704202068953242 , 16.71252 )
( 2.037036206585555 , 13.443653 )
( 3.3181294886155897 , 12.013626 )
( 4.799059169111444 , 19.43533 )
( -4.685096756788186 , 21.5547 )
( 3.763013496907226 , 17.26909 )
( 28.983554625216815 , 12.066975 )
( -7.770639777616417 , 22.049843 )
( 27.45403859562484 , 16.218342 )
( 3.0257740374098576 , 17.303696 )
( 0.25647011967840017 , 16.894981 )
( 3.272752459952607 , 0.71631515 )
( -5.9272216688387225 , 5.805554 )
( 3.0747765404096454 , 14.7172365 )
( -8.038569241966702 , 2.451426 )
( -5.050541274535516 , 24.229044 )
( 0.34600595641583676 , 12.355936 )
( 28.673132361515542 , 11.320878 )
( -5.940311628745455 , -10.37151 )
( -7.993433898029448 , -0.79433644 )
( 4.86724649983393 , 13.573116 )
( -6.286200530195909 , 4.735517 )
( 4.607480126728213 , 7.8110547 )
( -0.3220136286931321 , 16.433441 )
( 4.304927980611079 , 16.325882 )
( 1.5448401819701143 , -1.6232729 )
( -1.243613978005303 , -6.790854 )
( -6.492846752996463 , 19.513756 )
( -4.371450245977409 , 3.1946373 )
( 29.292636044241142 , 11.974651 )
( 4.340962795431471 , 17.655283 )
( -3.6055625052763522 , 0.36652732 )
( 3.307942025732043 , 3.2313437 )
( 41.71605270303151 , -4.2270794 )
( -9.75825799393571 , 5.125232 )
( 3.1137936036477067 , 12.488823 )
( 0.749527433582081 , 8.09031 )
( -4.712378944304795 , 8.944268 )
( 2.33499182975675 , 12.489309 )
( 26.07741710924019 , 14.070363 )
( 4.350337015857328 , 19.177124 )
( -7.072632660735184 , 0.4644866 )
( 1.5885030577036807 , 2.9512036 )
( 3.358999200951279 , 16.010838 )
( -7.243656911213327 , 22.720268 )
( 28.16460746931054 , 15.700479 )
( 2.9476043793671636 , 15.01453 )
( -0.07839158336669882 , -8.543936 )
( -0.2573700197857307 , 13.9913025 )
( 25.781080302201456 , 14.552242 )
( 1.7924330199408818 , 12.928931 )
( -3.033547743406295 , 21.84145 )
( -5.651592348771004 , -4.1331034 )
( -4.848260266172271 , 4.3361535 )
( 3.657926263806515 , 14.471839 )
( -5.7275204712465175 , 18.673943 )
( 0.39054285033325187 , 13.810682 )
( -7.070880066448955 , 23.482626 )
( 4.551197423320765 , 16.311295 )
( -3.376362237148901 , 4.012179 )
( -4.006914503736079 , 4.2165327 )
( -6.383671666703839 , 18.400951 )
( 36.40246355764393 , -4.7927885 )
( -8.605498544062526 , 3.1108847 )
( -1.9874702197765348 , 3.175664 )
( -12.592806729541639 , 3.8659997 )
( 0.22994967924095194 , 4.971009 )
( -3.9246200538839755 , -2.5184631 )
( 27.634462477577955 , 11.855224 )
( -2.7904523577570925 , 23.4772 )
( 3.0698862944811616 , 7.625256 )
( -3.1640763915907217 , 24.916452 )
( 0.39793826102693397 , 9.883662 )
( 3.676392893471476 , 12.447698 )
( 28.83238522792501 , 15.27705 )
( 24.287224451661068 , 10.481979 )
( -7.296232763317365 , 20.270893 )
( -13.478731406415502 , 5.1775255 )
( 3.1302041233692215 , 15.287179 )
( 3.2575479920709336 , 16.268219 )
( 26.548206213550372 , 11.630351 )
( -4.998077137917798 , 2.1751547 )
( 0.6300744088395285 , 11.868187 )
( 1.0728121816095395 , 16.650234 )
( 27.047347584951662 , -2.4237576 )
( 39.65500958119758 , -4.968237 )
( -3.8021681819800115 , -2.686357 )
( -5.638237075720608 , 18.49522 )
( 5.391921514257042 , 16.135664 )
( 3.76359751327836 , 15.923148 )
( 0.4375329902270777 , 14.385931 )
( -1.8038124482295501 , 2.4416633 )
( -5.678760618200772 , 9.113539 )
( -17.31519344262353 , 2.819564 )
( -4.912901626085253 , 23.705038 )
( 3.0785379955798744 , 3.2569919 )
( -7.002891610789505 , 24.268017 )
( -9.77042116993057 , 5.413951 )
( -7.468568395570703 , 21.997005 )
( 29.06191005929157 , 14.285132 )
( -7.162703410326941 , -0.45071578 )
( -4.756995161342651 , 23.896866 )
( 1.4246313920229121 , 12.903381 )
( -5.020137111330444 , 9.468448 )
( -5.462358918385754 , 23.472717 )
( -3.5994501658228657 , 21.398619 )
( -16.8963700936613 , 4.0111732 )
( 37.015181645161746 , -4.0623193 )
( 24.20909228874411 , 12.732091 )
( 13.686501465044556 , -14.806321 )
( 1.1015688097319556 , 1.8123193 )
( 0.2903348603685959 , 12.005934 )
( -5.444269437582314 , 18.96006 )
( -6.093792214168757 , 23.276295 )
( 0.6180036475955978 , 17.111431 )
( -4.36882385079552 , 20.69835 )
( -5.824608676722514 , 2.3222237 )
( 1.6345173793976744 , 16.04553 )
( 4.421730077523113 , 17.60342 )
( -4.9063355538026565 , 2.4257922 )
( 3.602905505761588 , 15.333195 )
( -18.745198290052 , 5.201923 )
( -8.50385906226788 , -0.97590166 )
( -7.236955966072974 , 22.130833 )
( -0.9369924359706691 , 7.3692436 )
( 3.308854811213803 , -0.5277147 )
( 1.7867671482778746 , -18.012852 )
( -2.3745076560747354 , 2.7833686 )
( 36.31686725073701 , -4.626198 )
( 36.010476453851815 , -4.534833 )
( -4.895920032366863 , 19.875711 )
( 28.13985453954783 , 11.149555 )
( 2.243323291581172 , 16.505466 )
( 4.004292853653088 , 17.703943 )
( 42.58813713423385 , -23.759453 )
( -3.6062896223049816 , 25.966959 )
( -5.755666507495902 , 19.509125 )
( -2.9445370724727495 , 24.163145 )
( -15.230440871518676 , 7.6031804 )
( -13.328262948482662 , 6.609211 )
( -15.769579660406867 , 3.2636864 )
( 2.3038483089284463 , 13.182924 )
( -6.575023601775807 , 20.292364 )
( 4.288954507396213 , 12.957685 )
( -0.9441069026076495 , 16.8491 )
( 3.7672636197743543 , 13.0829 )
( 0.6099486144663137 , 11.446442 )
( 13.196528237497636 , -16.970324 )
( -8.68554392992226 , 3.315791 )
( 29.9827958467547 , -0.19269866 )
( 3.462606124527012 , 11.390644 )
( -7.3479258091309685 , 8.4547 )
( 4.9774182278158206 , 14.856525 )
( 23.183036835776512 , 14.062548 )
( 2.5654652816801695 , 13.057624 )
( 2.2782901576965786 , 17.468838 )
( -7.583076658048427 , -3.4803143 )
( -17.48762755408501 , 5.730436 )
( -5.76707975216587 , 19.74862 )
( -5.149957916241362 , -0.43497854 )
( 3.6449551478141404 , 15.301523 )
( 4.749772822012873 , 13.222235 )
( 1.8665764233152506 , 9.783434 )
( 2.2524162547819246 , 14.189244 )
( 0.3844938667898464 , 12.139551 )
( -5.959033858275541 , 8.013816 )
( -4.422777490411916 , -2.2970712 )
( 3.0216097605762107 , 14.535737 )
( -2.7801664273247795 , -3.4630165 )
( 28.328132104414127 , 12.166704 )
( -6.818294314569681 , 23.145422 )
( -3.2252848495045097 , 22.001993 )
( 1.3985206716847665 , 12.849472 )
( -6.89161581861339 , 3.5724 )
( 0.1562032418578967 , 15.421529 )
( 2.7542831576188833 , 0.43157342 )
( 5.837000510813783 , 15.88132 )
( -4.246140545230042 , -2.0601082 )
( -2.4853613346705377 , 22.925318 )
( -4.730095804192757 , 19.381344 )
( -4.02198571624416 , 1.1910486 )
( -3.887162363070385 , 25.099766 )
( 3.0580957626475196 , 12.269091 )
( -6.4203813069769495 , 19.028208 )
( -5.188414144657724 , 9.711583 )
( 41.15105665404912 , -6.006052 )
( -2.1014116857757 , 1.2307824 )
( 1.5019234618037993 , -8.0257845 )
( -18.521407058784302 , 1.9209898 )
( -3.3905481022508352 , 24.813053 )
( 0.03714196415189597 , 8.0162945 )
( 40.71749486432457 , -3.7312622 )
( -17.114955687786832 , 4.3870883 )
( 2.663154175028249 , 18.333412 )
( -14.781541359376238 , 7.5238914 )
( 3.6551475350014044 , 6.3507776 )
( 26.906876785044638 , 12.032644 )
( -17.926939156376967 , 4.278068 )
( -9.671204370713015 , 6.267301 )
( 1.6436051171871697 , -3.1304603 )
( 3.636822594486676 , 5.966183 )
( 32.23448322527591 , -3.9996765 )
( -5.890544443093427 , 4.9514284 )
( 25.57182049348597 , 13.51182 )
( -14.591012087461893 , 2.214879 )
( -6.522446667412835 , 8.146985 )
( 0.26818130943996654 , -1.6983593 )
( 25.75444095556324 , 11.917927 )
( -0.6608149282753726 , 17.088299 )
( 27.696445032533898 , 14.928519 )
( 0.1537869736242592 , 12.393101 )
( 2.578916768016769 , 15.776031 )
( -6.7519016376136625 , -8.074551 )
( 3.91236539782608 , -1.7962053 )
( 1.1169769921144033 , 4.9893465 )
( 1.925515998828521 , 9.616219 )
( 27.786384048825067 , 11.898298 )
( -8.494040216564967 , 5.3844995 )
( 29.216948894844407 , 16.201965 )
( 29.098894058585238 , -2.1523893 )
( -2.7494244517267474 , 20.401949 )
( 26.476144030990042 , 10.235523 )
( -8.140556005147703 , 23.545536 )
( 4.183160855518565 , 17.447666 )
( 5.099141561442456 , 14.128732 )
( -0.781332160967925 , 15.102195 )
( 4.4107375715566866 , 2.2203128 )
( -8.267456851116869 , 8.256359 )
( 5.4388182737085335 , -1.0365299 )
( 0.16570039107644086 , 6.784501 )
( -2.8408009214492216 , 23.995811 )
( 26.171191471599357 , 11.074578 )
( -14.473009102877892 , 3.02969 )
( 25.569188621037135 , 13.14517 )
( -3.1893069459915493 , -1.6883907 )
( 3.683153809958865 , 16.01823 )
( 4.54189211285806 , 6.346904 )
( 28.281342300978654 , 11.894477 )
( 1.0545206321049703 , -14.655942 )
( -7.056040541960307 , 20.656982 )
( 28.44092616763587 , 11.088493 )
( -17.385738081776363 , 5.7944856 )
( 4.591517058412605 , 14.325372 )
( 3.746318308308042 , 4.517403 )
( 1.4282087506501169 , -4.104861 )
( -6.111134539945006 , 21.67187 )
( 0.42932714104570124 , 12.879049 )
( 9.236546043196796 , -29.099161 )
( 0.5306666048330058 , 15.3502 )
( -5.466704421523893 , -4.5642347 )
( 1.481899388380282 , 10.614396 )
( -3.648497039275275 , 19.840336 )
( -0.7014985658366878 , 5.877848 )
( 3.500259522015294 , 16.601307 )
( -7.329335159387352 , 7.4403152 )
( 3.0520695768548447 , 17.157795 )
( 2.207911215297216 , -1.16766 )
( -8.045068171877206 , 3.8166494 )
( -4.214770275386082 , 19.973164 )
( 4.024919984368261 , 17.468649 )
( 1.1647719574135098 , 16.972996 )
( 2.449515718102006 , 12.067758 )
( 28.95830011092621 , 15.247845 )
( 2.2390213958613465 , 13.870586 )
( 4.251861930639272 , 15.145624 )
( 4.261740405361794 , 13.218116 )
( -10.334212926261952 , 5.71974 )
( -3.069691934674182 , 25.281385 )
( -7.197358768276895 , 20.846598 )
( -5.35760206917578 , 7.0605087 )
( 5.402233562091136 , 16.24768 )
( 0.09236553467451558 , 16.72198 )
( 24.435467200405196 , 10.143778 )
( -18.139985072557096 , 1.9174817 )
( -2.599975337403807 , 5.106702 )
( 4.048992937736649 , 13.4606905 )
( 4.412513767972631 , 16.796484 )
( -7.451455797456186 , 21.966984 )
( -6.454539821282772 , 20.102755 )
( 5.0567257120060205 , 18.681782 )
( -4.824585717999195 , 6.206756 )
( 3.2644214315641875 , 15.567921 )
( -6.768729524868202 , 22.226992 )
( -5.131392841842633 , 4.0584354 )
( -6.884391763250255 , 9.175532 )
( -6.248782464055826 , 21.924797 )
( -5.173883397860826 , 24.827509 )
( 25.438016454548798 , 11.224398 )
( 3.7263657496884894 , 13.480915 )
( 4.142480091093228 , 19.448221 )
( 3.593981530588276 , 13.406689 )
( 2.831545062893946 , 15.216255 )
( -7.181712279855059 , 2.0320878 )
( -0.19200386988015516 , 8.639085 )
( 29.283937576301923 , 11.480232 )
( 28.65643404531277 , 13.278749 )
( 2.987414105922891 , 17.517483 )
( 3.735042479340909 , -1.2952816 )
( 0.8170177431901178 , 4.3685913 )
( -5.5394617164628785 , 20.952517 )
( 1.5333000926032083 , 18.320395 )
( 3.8256080755158512 , 7.6697335 )
( -2.6776618102144965 , -9.697562 )
( 38.28472570117899 , -4.707278 )
( 1.2028328714263181 , 10.839153 )
( -3.634311930925045 , 3.1315417 )
( 2.916823527558439 , -2.1714172 )
( 5.690648471516558 , 15.762848 )
( 26.24973517229346 , 13.164131 )
( 24.070242497472407 , 13.382364 )
( 2.5619542912634383 , 15.705927 )
( -3.383329134020242 , 1.1239134 )
( 0.43429611655229294 , -1.6234739 )
( -6.834102513280582 , 22.890585 )
( 3.746324425548132 , 13.840374 )
( -4.924417873843612 , 23.14526 )
( -17.046049321077643 , 2.7742877 )
( 25.099748743790823 , 12.18073 )
( 1.70765052694141 , 5.004367 )
( 1.7230210919009417 , -0.22068036 )
( -4.954566290212696 , 22.282393 )
( 0.828170250744505 , 14.126984 )
( 2.7257033574478706 , 6.532351 )
( 0.5917403365937548 , 15.378736 )
( 0.49613137819088937 , -6.068693 )
( -3.2038338925840675 , 24.738037 )
( 11.679672154586893 , -13.09815 )
( 27.49248095317935 , 10.96779 )
( 1.3777028967021867 , 10.87694 )
( 0.9429381403833497 , 14.650936 )
( 4.5215737939594005 , 0.855538 )
( -17.51389400617633 , 6.9962173 )
( -3.6933514589345466 , 21.412983 )
( 0.7575054044845271 , 16.590786 )
( 28.619486266314148 , 16.181963 )
( 4.391060777999038 , 14.209862 )
( -0.7707643610161211 , 15.324049 )
( 3.5568963633201807 , 11.71004 )
( 28.65480286043933 , -2.2757006 )
( 0.6200070329056891 , 14.294681 )
( 1.1630522038911022 , 17.45543 )
( -0.07290451630059769 , 16.573362 )
( -7.094639936437648 , 1.599102 )
( 0.01135821245542512 , 16.34139 )
( -0.5876798359584836 , 15.41927 )
( 2.151154664667139 , 10.286877 )
( -4.045969544741744 , 24.86242 )
( -31.963986691057187 , -15.12203 )
( 5.270500403615802 , 15.127997 )
( 4.833007116651967 , 14.203836 )
( 0.33040758193618003 , 14.528918 )
( 4.678869807818497 , 18.467957 )
( 3.72255471164918 , 9.570494 )
( -7.21706790569163 , 23.55762 )
( 4.668096698925162 , 17.954147 )
( 1.4669215032884617 , 13.004921 )
( -2.1964101560671168 , 1.9339352 )
( -16.00126685615424 , 5.57627 )
( 1.84711867429494 , 5.479637 )
( 28.73824751307284 , 10.815502 )
( 0.015416296277850572 , 15.490712 )
( -3.859657948889683 , 19.443079 )
( -16.54611860199655 , 6.276024 )
( 4.765780355604369 , 17.495754 )
( -6.623510913291959 , 18.079845 )
( 0.9944501187987387 , 14.252942 )
( -7.35324830125218 , -0.52952445 )
( -4.245120161576652 , 24.814348 )
( 0.9824396788699158 , 15.627944 )
( -1.0101009312664737 , 16.22466 )
( -0.26441726365205964 , 14.373338 )
( -6.804601975867051 , -12.574562 )
( 3.6678374629051773 , 7.1565886 )
( 1.5200992551864259 , 12.693302 )
( -6.640442287069489 , 3.5302463 )
( -17.611791199317036 , 5.8619266 )
( -18.301008014046303 , 6.426389 )
( 1.4822617500715247 , 12.27287 )
( 26.65738729251772 , 14.746913 )
( 4.758169932681535 , 18.636576 )
( 4.229806788321008 , 17.162626 )
( 4.409213079265041 , 15.5842705 )
( 1.9684101197481665 , 18.347574 )
( -5.41655310340806 , 19.198662 )
( 5.001200826881892 , 16.935812 )
( -5.538656610594286 , 1.5693768 )
( -2.5326405400216743 , 24.223019 )
( -13.665908979148861 , 3.6787927 )
( -4.007195725220788 , -0.26181048 )
( 1.8691190617086977 , 5.6577125 )
( -6.2575051540347575 , 18.908772 )
( 3.219591789060618 , 11.399899 )
( -2.7890098880469925 , 26.101488 )
( -0.7917174793611809 , 15.218109 )
( 38.29734358176616 , -8.70938 )
( 4.006973499363639 , 12.661075 )
( -3.8688935791148458 , 3.8849387 )
( -19.181897066205515 , 5.443616 )
( -5.316942434976272 , 7.0332794 )
( 0.6840614569378541 , 15.373999 )
( -13.18663441345779 , 4.9370875 )
( -8.273093351006004 , 4.2452927 )
( 4.963183265470731 , 17.206932 )
( -2.912502025523548 , 3.4731398 )
( 2.038734060396331 , 12.602498 )
( -5.6596961733590225 , 19.89177 )
( -2.6068297471467847 , 23.723179 )
( -17.027896744025917 , 4.2322817 )
( 4.626302937446828 , 18.87218 )
( -10.384356559937995 , 6.2985435 )
( 31.941156426255226 , -2.1983113 )
( 29.503522176341242 , 13.665943 )
( -5.810343093045505 , 19.624435 )
( -5.464872436535039 , 24.804531 )
( -2.337355020270669 , 22.792736 )
( 0.8379530925403071 , 8.22161 )
( -3.296738937570999 , 22.178736 )
( -6.354279539627521 , 21.830273 )
( 1.9423264542244754 , 15.286669 )
( 0.1513386172541208 , 16.747795 )
( 28.527490537240737 , 15.17228 )
( -3.5186832907819845 , -3.5063362 )
( -14.161025195925566 , 9.061553 )
( 31.193255856031794 , -4.4541407 )
( -6.8327763523995895 , 21.436878 )
( 1.9622771151763954 , 16.962399 )
( 2.6957250093086604 , 16.718784 )
( -1.0477950551020636 , 16.259468 )
( 3.4458482622990108 , 8.730289 )
( 28.054247092015483 , -1.0088215 )
( 5.425816640820411 , 15.006917 )
( 2.0650568913522687 , 9.522255 )
( -6.91014468842319 , 23.539484 )
( 0.08694153646173786 , 15.816394 )
( -16.27521438002538 , 2.0602458 )
( -7.269320341850966 , 4.0924425 )
( 23.11388801366832 , 14.201027 )
( 42.64145837555966 , -8.140654 )
( 1.5084081594706746 , 13.777756 )
( 0.7861593595076398 , 17.910849 )
( 3.9487420221971217 , 12.873241 )
( -0.14251596396028116 , 4.054401 )
( 0.1045719695026629 , 0.9661728 )
( -4.053130929284488 , 1.9181933 )
( -7.868567139935692 , 21.970081 )
( 15.913582179264258 , -27.581038 )
( 4.152984474812349 , 10.098958 )
( 25.8364127222377 , 9.571371 )
( 1.0025193699001047 , 11.268239 )
( 2.365250240545956 , 18.021124 )
( 24.21803850692122 , 11.300731 )
( -6.232390259023143 , 21.990454 )
( 4.159638913737969 , 14.734558 )
( 26.50931744286942 , 9.868794 )
( 3.2717394304857668 , 3.3267186 )
( 2.1818561552932065 , 12.459769 )
( 28.26282148933871 , 16.977179 )
( 26.129646472842097 , 15.430689 )
( -12.606131243179075 , 3.9565046 )
( -2.301154228229576 , 4.71998 )
( 33.60124531926155 , -8.764754 )
( 5.352166435460048 , 14.956823 )
( -6.0565001753488605 , 23.330444 )
( -2.4048822287984724 , 26.234016 )
( 27.85942666441861 , 15.381207 )
( -4.26214450974834 , 7.3922973 )
( 38.94064862528118 , -5.595059 )
( -14.095711471975356 , 2.9426649 )
( -5.240586824869009 , 6.508191 )
( -5.755004881535787 , 19.019463 )
( 27.40505518547066 , 16.010817 )
( -6.978534189446581 , 2.4701536 )
( 26.45868003168448 , -2.9771888 )
( -0.26455539625096947 , 3.9376187 )
( -7.065145972867826 , 19.509104 )
( 27.872123972874565 , 11.77281 )
( 1.8731667824956926 , 9.831822 )
( 1.8736731473031414 , 8.856937 )
( -16.57579628358038 , 5.9184074 )
( 13.345515146911751 , -16.756569 )
( 0.4362260795403329 , -1.1089445 )
( -6.893492857799388 , -6.294544 )
( 1.4532656963828732 , 16.911343 )
( -0.0848452699779021 , 9.134692 )
( 2.5069549519272627 , 16.770496 )
( -2.5632039887006153 , 21.814146 )
( 4.079116046313742 , 15.012569 )
( 25.844800684621543 , 12.552791 )
( -6.447528731370772 , 23.933628 )
( -7.20498655130829 , 21.026844 )
( -0.002367384625082125 , 16.083124 )
( -2.684433804401487 , 22.174208 )
( 0.6402228361483312 , 16.844862 )
( 0.5104158586024399 , 9.717101 )
( -5.44322006186124 , 20.822336 )
( 4.76572818129957 , 19.419676 )
( 4.060358779322164 , 18.829557 )
( 4.76036387606734 , 15.185248 )
( 4.044633999102522 , 14.171773 )
( 28.403631002739456 , -3.9980922 )
( 1.8491070771338451 , 12.617598 )
( -14.054821641877744 , 2.8323147 )
( 4.311481804649467 , 15.745733 )
( -6.910989789460332 , -3.4680405 )
( 3.180194830972323 , -2.3569808 )
( 5.181646897621209 , -0.24796924 )
( 2.2532983345077615 , 18.622805 )
( -3.765352720627778 , -3.1834004 )
( -6.807855315232329 , 22.382664 )
( 2.8688533507973073 , 0.80923665 )
( 23.68011157272641 , 12.277896 )
( -6.526974741174766 , 8.8572035 )
( 4.120819755353602 , 0.9664871 )
( 27.70999962523249 , 16.013672 )
( 27.956460619299225 , 11.780547 )
( -4.921769173905849 , 20.973425 )
( 2.9547672779428913 , 4.0411987 )
( -7.122898927497952 , 21.351042 )
( -8.140326354838548 , 22.80006 )
( -6.7849415018609385 , 3.0080383 )
( -1.8603217610859168 , 2.8125322 )
( 0.4280701024451027 , 13.364315 )
( 5.282286948952137 , 7.1548586 )
( -4.643156934273097 , 22.512173 )
( 6.1231064172039575 , 14.968452 )
( 3.541010521173916 , 10.855015 )
( -7.600000846277389 , 21.77867 )
( 5.2865503823612015 , 13.684662 )
( 0.9597955647529667 , 14.156546 )
( 1.0294338138423016 , 13.055252 )
( 23.747482707801122 , 14.739368 )
( -3.954905109855241 , 25.205927 )
( 3.942963479367927 , 16.111557 )
( 29.37383466820609 , -4.0358143 )
( 27.075435861863504 , 12.059294 )
( 3.333489889105123 , -17.872751 )
( -6.292511355630256 , 23.647158 )
( 4.17517767123178 , 11.04221 )
( 4.524379087004291 , 18.289396 )
( 0.9505328958849821 , 14.907201 )
( -4.734188502327113 , 8.844828 )
( 2.060763619146849 , -20.920118 )
( -0.301809148168086 , 15.929481 )
( -2.2443936330900014 , 23.899494 )
( 3.644219168923745 , 15.472953 )
( 24.95312084458878 , 13.192617 )
( 5.212385825868532 , 17.369453 )
( -6.367327632651129 , 24.07433 )
( -0.09091191997368553 , 0.861111 )
( 1.345663969065874 , 10.136938 )
( -0.9494849904339464 , 21.721016 )
( -8.932169436586655 , 7.2481194 )
( 28.29320613025725 , -0.5954139 )
( 26.99631167223887 , 13.607996 )
( 2.3662159245458483 , 15.027676 )
( 28.988096430939002 , 16.221046 )
( 4.012529683719399 , 19.62809 )
( -3.929462170478843 , 21.069193 )
( 30.16281552579482 , 13.062942 )
( 27.3863861880228 , 11.332045 )
( 28.108435551901735 , 16.695375 )
( 1.5476597748278356 , 16.919361 )
( 0.37828325382011485 , -0.80937725 )
( -3.9925915151921214 , -0.96793205 )
( 4.994829705669159 , 7.77792 )
( 5.015611122557139 , 7.02886 )
( 27.535709642548348 , 10.266836 )
( 1.2983190190358282 , 14.677933 )
( 2.27196808779932 , 14.012766 )
( 0.07603708474583737 , 16.116386 )
( 28.709747507188887 , 15.603689 )
( 3.66751946964645 , -1.0041457 )
( -17.868621373890043 , 6.4998045 )
( -3.7821408907213634 , 6.232221 )
( -4.835991987422307 , 1.9256041 )
( -5.517488487992516 , 21.520157 )
( -5.581693077580376 , 21.447632 )
( 28.25617161644313 , 16.64745 )
( 2.875180967844578 , 15.284994 )
( 41.4699358031516 , -3.8269367 )
( 37.7568685869549 , -3.0497303 )
( 2.9302964723710545 , 17.714354 )
( 25.838613188115843 , 12.582178 )
( 28.288954805734537 , 15.556423 )
( 37.462085578874806 , -6.779497 )
( -2.435727117683876 , 22.271406 )
( 3.541665740883162 , 14.864766 )
( 5.013263232770628 , 16.103043 )
( -2.3360693253180624 , 21.848946 )
( 4.907029560033907 , 15.491532 )
( 4.149849820343989 , 15.160934 )
( -1.4805529975790996 , -6.748151 )
( -0.5010722003808228 , 16.48769 )
( 27.008597027314757 , 13.286335 )
( 2.9501712340922475 , 15.632674 )
( 4.48591237533181 , 15.962394 )
( 1.5629521916279852 , 14.855759 )
( 2.136715750243579 , 16.267818 )
( 4.243044327651825 , 16.362473 )
( 5.775095567587289 , 14.362188 )
( -5.054479622661488 , -3.202423 )
( 2.2358976156500323 , -1.3465362 )
( -4.86429978767251 , 1.849119 )
( 4.438618081924648 , 17.221601 )
( -2.9368157397150796 , 0.61949265 )
( 5.35042806803198 , 14.097537 )
( 2.5903117267787876 , 3.154826 )
( -3.5742571814756303 , 20.990936 )
( -1.4083748049945863 , 8.709232 )
( 28.579673486283188 , 13.176985 )
( -7.454228374338955 , 9.010463 )
( 0.6871985359805078 , 11.673742 )
( 4.582940010688571 , 17.703442 )
( -3.61760844990468 , 24.774736 )
( 25.712190385284533 , 15.843943 )
( 26.048807425570068 , 12.444868 )
( 1.175810782484588 , 12.922755 )
( 2.1153489963348178 , 13.856351 )
( 26.70319877520511 , 14.424445 )
( -5.3564381312076925 , 8.89105 )
( 0.4617314968695978 , 9.914978 )
( -2.1851840178912134 , 20.557951 )
( -0.18653194916055826 , 0.9325872 )
( -0.3580390464991162 , -0.20503479 )
( -6.88963612152839 , -2.9952154 )
( 24.607304887990043 , 10.522627 )
( 5.367044362804423 , 13.966955 )
( 0.6133870990284244 , 0.7910034 )
( 38.18302469735766 , -6.0725584 )
( 26.81042994319711 , 15.597525 )
( -8.072231413961456 , 23.464546 )
( 3.3544964138311615 , 14.067265 )
( 4.703237800473055 , 17.091572 )
( 1.617281598122137 , -5.650796 )
( -6.215436493591991 , -3.2788935 )
( -4.6029958217738205 , -6.117243 )
( -0.27046693525310095 , 12.926975 )
( -6.05550000588457 , -2.2756863 )
( -18.640342871074495 , 2.3162167 )
( 3.5737278928608904 , 12.824089 )
( 4.377083163375056 , 14.505676 )
( -8.928292500975154 , 5.636819 )
( -2.993089675595109 , 23.963066 )
( -5.91129161137945 , 19.02627 )
( 0.07777440229650301 , 5.5563807 )
( -4.563503816599706 , 23.437366 )
( 4.5479025744505055 , 16.703535 )
( -5.747666169120658 , 21.742184 )
( 3.3464539037280083 , 15.434068 )
( 5.36466317756775 , 16.166485 )
( 1.7498882807587182 , 14.510377 )
( -6.102023345186945 , 9.431999 )
( 4.862722192134202 , 15.140809 )
( 29.833750518813776 , 12.42098 )
( -2.749722685471707 , 21.06732 )
( 25.814323936796473 , 14.795253 )
( 30.10550657472474 , 11.975216 )
( -2.013451872203583 , 24.40339 )
( 39.089912332050034 , -7.560315 )
( 29.35211764968348 , 14.298393 )
( 1.331635876177745 , 12.8474865 )
( -4.513937957937506 , 24.011318 )
( 32.34214722525529 , -2.948512 )
( -2.2482789170329394 , -1.7293882 )
( -9.175351352035662 , 4.887692 )
( 26.738721427630843 , 12.2126465 )
( 2.404775553886237 , 14.451481 )
( -7.345467407766396 , -1.8852648 )
( 4.350811652230849 , 13.623304 )
( 2.4653829812405395 , 14.511081 )
( 2.1445807422805925 , 13.627555 )
( 3.9800334133454776 , 16.77987 )
( 26.05907671262275 , 14.466573 )
( 3.565618252276166 , 18.20156 )
( 10.150652034300071 , -18.190084 )
( -6.0281011405649885 , 23.865221 )
( -18.075837581126383 , 2.4673712 )
( -0.594503112917149 , -0.39415175 )
( -2.021413012660318 , 21.72368 )
( 2.161927562154982 , 18.033548 )
( -3.3297237870664844 , 22.232697 )
( 3.8275126141486746 , 11.785784 )
( 2.5256612553294024 , 15.177071 )
( 0.2930820248539542 , 11.195489 )
( 0.4115296153322139 , 16.72552 )
( -3.3982637157392643 , -5.78218 )
( 5.211538125064196 , 14.433734 )
( 26.00433116647994 , 14.027939 )
( 28.80044277379544 , 15.178604 )
( -3.3681727656153955 , 3.6570263 )
( -3.0898043559386705 , 24.046646 )
( -0.04493622574912115 , 17.372799 )
( 25.676035992322273 , 14.422069 )
( -0.9137059039220332 , 9.313782 )
( 5.186072119196561 , 16.178026 )
( 3.0539471984017967 , 12.928395 )
( 30.672533999654107 , -0.34032747 )
( -2.147028338466029 , 20.441166 )
( -3.9206907255308088 , 2.8516731 )
( 5.991744084732627 , 14.552465 )
( -10.140166464697986 , 7.176433 )
( -29.125382892296972 , -17.133633 )
( 25.595317478862242 , 14.579843 )
( -4.258971591302181 , 19.865286 )
( 0.9488058781723643 , 13.879799 )
( -3.8286977656838364 , 25.680134 )
( -3.016579824561195 , 22.46381 )
( 26.829911386780736 , 13.623598 )
( -0.1344390666528572 , 15.646084 )
( -0.1493012415939271 , 2.5217972 )
( -14.030090010810195 , 7.7510777 )
( 27.66998069231172 , 14.62806 )
( 4.404555865561246 , 17.886139 )
( 27.5661525023216 , -3.9546902 )
( 24.362120472168506 , 10.136066 )
( -17.632339564327463 , 4.93441 )
( 24.43159701498322 , 10.524884 )
( 1.9093348965550496 , 12.011999 )
( 3.9085840431873486 , 16.270061 )
( 4.945005371639416 , 18.067324 )
( -18.922503146412016 , 5.557452 )
( 4.621461542836914 , 14.776062 )
( -3.408959432756644 , 23.945751 )
( -4.883786669628401 , 23.4118 )
( 2.0667769312781052 , 13.161731 )
( -3.276507559917107 , 21.085772 )
( -6.281471623732172 , 24.142944 )
( 1.6232059675017563 , 8.661879 )
( 3.3494164776066935 , 19.389893 )
( 26.367892556596207 , 13.883112 )
( -6.673093181834586 , 5.3242025 )
( -5.064504400929529 , -0.75369555 )
( -5.9576582680552495 , 23.041573 )
( 2.8171689378143423 , 12.880468 )
( -3.56466429887997 , -5.1180816 )
( 41.34610604788379 , -7.843302 )
( -3.879956661684385 , 22.198362 )
( 4.646635466876867 , 12.731707 )
( 0.10993052194986408 , 10.064414 )
( -16.29835151716548 , 6.7741036 )
( 26.591937205952693 , 13.679375 )
( -18.054252718351034 , 6.3233857 )
( -2.926494428353991 , 25.463646 )
( 40.16103170623735 , -4.486885 )
( 1.743805132498711 , 12.98754 )
( -3.669567000212038 , 23.825172 )
( -0.32016493842772453 , 15.47171 )
( -4.637230707830507 , 8.508449 )
( 27.544909238036293 , 11.435336 )
( 4.133051035470853 , 17.451225 )
( 4.660243546709455 , 2.6254876 )
( -10.037363965267035 , 5.1515403 )
( 26.69418381042812 , 12.772246 )
( 26.490542824408564 , -4.3860626 )
( 2.431255341708849 , 17.025873 )
( 2.6043881903947694 , 17.853151 )
( -2.154502139737051 , 23.992765 )
( -5.337260884707278 , 1.2640486 )
( 1.863296502217234 , 7.661909 )
( 3.4356950041295145 , 3.1278517 )
( 26.98151997162104 , 13.225368 )
( 0.7569397154703847 , 12.498531 )
( -1.9983037850961693 , 23.6432 )
( 4.2201682831261405 , 15.986164 )
( 0.6889678202960818 , 14.906505 )
( 3.6953767531576722 , 16.901571 )
( -13.362360944075819 , 3.6561027 )
( 4.863357063774673 , 13.930591 )
( 26.672129264282983 , 10.077072 )
( -16.879429009398017 , 1.2641838 )
( 2.1206020428948604 , 15.983851 )
( -19.115868222206863 , 5.787288 )
( -8.070801562121188 , 23.770164 )
( -13.879476393008247 , 4.445417 )
( -16.294627310246064 , 6.9717197 )
( 2.807564595586782 , 12.347886 )
( 1.3584284120611836 , 8.552917 )
( 40.966375180506134 , -6.646805 )
( -0.7101576659375551 , 14.919757 )
( -18.440394541227697 , 6.306399 )
( 1.6185146444403973 , 11.232298 )
( -4.865605223013 , 7.8107576 )
( 2.09845034747214 , 12.436176 )
( -4.838426672216019 , 21.029432 )
( -2.5434625028514297 , 20.97312 )
( 3.803843058881105 , 6.50594 )
( 3.1481359642183344 , 12.962329 )
( -2.3601812676393115 , 22.14635 )
( 29.08364075441755 , 13.381701 )
( 3.233176186765286 , 17.589773 )
( 1.282478991059858 , 13.427404 )
( -3.8563324038527016 , 19.514029 )
( 24.600708390054137 , 13.050034 )
( -4.264380023235855 , 25.276703 )
( -0.797976621750714 , 9.257862 )
( -6.8273603678774695 , 22.92636 )
( 3.9181519854685187 , 17.284332 )
( 2.8751477584248866 , 11.614393 )
( 25.841286493128727 , 10.748316 )
( 3.4931725761336683 , 12.677643 )
( 3.1688456958247944 , 16.960722 )
( -4.494910598820105 , 5.876935 )
( -1.149821651321784 , 9.391215 )
( 0.027447892876183655 , 9.615611 )
( 4.771497260721855 , 14.932258 )
( 33.078514280475694 , -6.1799984 )
( 0.3400702726877992 , 16.637497 )
( 28.90702558303202 , 12.751327 )
( -4.862291010376643 , 0.8458451 )
( -0.7886614529098784 , 14.676233 )
( -15.160440740977998 , 6.5799193 )
( -5.43224813170333 , 6.9713306 )
( 1.464659466104001 , 16.145061 )
( -2.90037503747201 , 25.66727 )
( -7.863119345463469 , 5.451754 )
( 24.667062139718993 , 12.392312 )
( -6.592208435490847 , 18.061062 )
( 26.959656240910554 , -3.5432465 )
( -3.7962641022073362 , 20.686415 )
( -4.717934043013626 , 19.827364 )
( 3.203500068793475 , 18.452911 )
( 1.769759389383288 , 5.1516256 )
( 3.6245112510749946 , 15.737261 )
( 25.950358516044467 , 10.295878 )
( -17.852103954497018 , 6.867753 )
( -17.16066747474745 , 5.411112 )
( -9.093575508316974 , 3.1728792 )
( 28.075711812232115 , 13.74016 )
( -3.13937222419788 , 24.342842 )
( -3.134356831824396 , 26.526392 )}

\newcommand{\utilityC}{
( 28.641607118358092 , -9.661038 )
( 40.583312343429085 , -5.207035 )
( 24.94821810861823 , -10.960983 )
( 31.817277174366406 , -5.4371595 )
( 27.538644376105765 , -15.241807 )
( 31.279277615094994 , -6.5213327 )
( 34.213060040501865 , -5.9856014 )
( 32.377360774699696 , 6.6638317 )
( 31.3650167922573 , -6.876933 )
( 33.475303613174574 , -2.804696 )
( 39.818018124568766 , -7.6446466 )
( 39.58205812831336 , -2.9422755 )
( 32.90101267879367 , 5.4764605 )
( 33.85317264925551 , -8.066754 )
( 30.74468641807341 , 8.126583 )
( 34.295625120561816 , -3.7475462 )
( 40.589358212733174 , -3.4794183 )
( 32.525955268268554 , -6.074254 )
( 18.86352175801707 , -17.030182 )
( 19.60774895313541 , 17.182384 )
( 40.815525623123605 , -5.086295 )
( 35.64805003248654 , -7.894448 )
( 19.926624604688875 , -15.444013 )
( 24.826147770557995 , -15.782425 )
( 32.39834976381803 , -6.0455317 )
( 15.707485599480762 , -15.042258 )
( 40.580749410792365 , -6.725687 )
( 33.20695513890671 , -8.233293 )
( 25.944252101466684 , -14.377267 )
( 31.74860304421942 , 8.588201 )
( 24.09962872470965 , -11.785353 )
( 34.261290426851076 , 7.4946165 )
( 39.451290850386755 , -6.3399563 )
( 25.272509996319567 , -12.935963 )
( 19.108473910585555 , 18.870756 )
( 23.725269073944414 , -14.008721 )
( 35.10103617967763 , -6.421456 )
( 39.107897532247925 , -3.1746478 )
( 24.124318025808865 , -8.311514 )
( 22.960440202107804 , -4.3390064 )
( 33.419339527345485 , -7.9685197 )
( 25.588690295793285 , -13.218806 )
( 36.318682250215744 , -4.112586 )
( 33.07904400906489 , -3.8488083 )
( 35.43128698699658 , -3.921465 )
( 26.448875586881154 , -11.026434 )
( 27.98482192713519 , -5.913568 )
( 24.56230051769584 , -13.124132 )
( 34.67787540021606 , -8.036342 )
( 35.229756283174474 , -8.175153 )
( 34.191609332074464 , 10.326015 )
( 33.50557895899502 , -6.000219 )
( 37.205912546696105 , -7.0708804 )
( 35.48822572650235 , -4.0663433 )
( 29.919016605881637 , -10.062152 )
( 21.280912787649825 , 9.420021 )
( 33.42002933210724 , -7.6706624 )
( 28.68661776308989 , -11.692188 )
( 23.991746512622306 , -10.720003 )
( 38.825876362387596 , -6.273054 )
( 36.15015354077819 , -5.7325153 )
( 32.25344242481992 , -6.434931 )
( 24.262559748847096 , -11.310661 )
( 32.68318527742466 , -6.2284317 )
( 31.00997416552923 , -6.584287 )
( 36.55725777809617 , -5.6885867 )
( 32.467547977392805 , 7.2462096 )
( 34.9647559988635 , 9.029857 )
( 32.69546884916133 , 10.854541 )
( 34.49819908140167 , -10.0602455 )
( 30.928574151100356 , 10.5707035 )
( 32.22751431213917 , -7.0189853 )
( 35.738455368152586 , -5.165095 )
( 27.960141829191798 , -15.472147 )
( 31.184479008523393 , -6.1379743 )
( 36.2367840749645 , -5.844674 )
( 21.839272336283013 , -17.28572 )
( 36.49682630879627 , -6.0063205 )
( 33.305600850112555 , 7.7621775 )
( 25.103006035887724 , -12.433112 )
( 39.581198209182254 , -4.9993734 )
( 34.96060090052615 , -7.773208 )
( 17.020164268251865 , 11.81553 )
( 39.61380476783444 , -5.4104095 )
( 33.05608202260174 , 7.5377784 )
( 35.00824624051854 , -5.0432024 )
( 33.4934992968809 , 10.444172 )
( 6.751039349052832 , -15.6052 )
( 26.062213612820642 , -12.817329 )
( 25.795186670179028 , -14.06328 )
( 22.84799269202622 , -9.513443 )
( 27.1258182169797 , -11.666554 )
( 38.10977281250606 , -6.8649282 )
( 36.29843964875715 , -5.3869963 )
( 40.50125285875163 , -6.3495455 )
( 23.22775448452817 , -10.952775 )
( 35.252327536827764 , -7.856394 )
( 38.65637995429246 , -4.7432775 )
( 40.38672525533176 , -3.3701897 )
( 31.515035556150767 , 10.60185 )
( 40.50531541869345 , -4.0004535 )
( 31.685698700674042 , -6.4860134 )
( 31.32961328869699 , 10.233255 )
( 23.57934328136322 , -12.034469 )
( 23.09432297176417 , -10.753862 )
( 32.9518541502414 , -7.828721 )
( 35.45796907774144 , -7.7933974 )
( 25.621857528373184 , -4.721988 )
( 37.934753810862794 , -4.925024 )
( 26.96332280971096 , -14.878536 )
( 31.72827829273838 , -5.4495306 )
( 30.736646043831232 , 7.9773746 )
( 28.100833564011165 , -9.2252 )
( 33.92917668812341 , -9.112598 )
( 33.58381527184225 , -6.2144604 )
( 34.79864331667709 , -7.7052817 )
( 38.269193413211376 , -2.4187298 )
( 32.67561474967453 , -4.8506346 )
( 36.569444633053976 , -5.0320153 )
( 32.948338395006665 , -7.766234 )
( 36.593957168095464 , -6.4625688 )
( 20.564455092036333 , 10.420643 )
( 29.562035142059035 , -10.544964 )
( 9.023836881887856 , -18.80501 )
( 26.75122813642974 , -9.322459 )
( 27.845806710223904 , -8.338282 )
( 24.70876108899276 , -13.069373 )
( 35.37626103276529 , -5.490714 )
( 33.235135034692306 , 5.3700857 )
( 33.05228789006134 , -7.7801847 )
( 32.74134573069266 , -3.0110335 )
( 35.00126004808046 , -5.030721 )
( 24.342972353587754 , -13.062927 )
( 22.636493761541963 , -12.025969 )
( 29.7381057884806 , -7.0550036 )
( 25.488264801117538 , -13.5544405 )
( 32.37100585383353 , -5.447981 )
( 36.609887235242795 , -4.4593744 )
( 22.2972952483215 , -12.470754 )
( 27.95583343743546 , -8.106198 )
( 32.83414737075985 , 7.275524 )
( 32.30888030213087 , -5.132082 )
( 35.43001082231518 , -5.4956408 )
( 33.459172755431 , -5.1503 )
( 34.01374343632955 , -4.585529 )
( 40.78151746890384 , -3.2276163 )
( 33.95583892188105 , -7.783107 )
( 34.35291199530505 , -8.104245 )
( 28.311588634666087 , -11.732021 )
( 27.96170226954663 , -5.2790046 )
( 16.323611430210466 , 16.919842 )
( 22.236801118687687 , -12.649929 )
( 33.50499969143415 , -6.755002 )
( 35.37765466146984 , -5.084824 )
( 26.958524814870948 , -14.749554 )
( 21.7913095552882 , -4.988292 )
( 22.50079611400843 , -15.741139 )
( 30.264910302575 , 9.032762 )
( 24.90416511890277 , -15.242584 )
( 25.36302901246181 , -13.645465 )
( 30.028807100713365 , -9.66586 )
( 26.11135317812774 , -8.937808 )
( 25.306358012979487 , -11.791669 )
( 34.71088541407079 , -5.4492226 )
( 39.60334031055676 , -7.53856 )
( 37.168664302711136 , -4.4249954 )
( 31.09515219323982 , 9.183624 )
( 34.79623962095831 , 8.203971 )
( 40.50086872403076 , -7.683052 )
( 32.51052576147439 , -4.0955997 )
( 35.02825431029954 , -7.134927 )
( 32.47772202979256 , -5.6597853 )
( 31.34939530777636 , -6.438158 )
( 37.191593702816384 , -5.409939 )
( 38.49900208899635 , -5.2966475 )
( 35.5014313753338 , -4.293587 )
( 31.69185857121992 , 10.4810705 )
( 38.847708226471575 , -7.9941573 )
( 34.92931993468824 , 9.008017 )
( 35.4962394498725 , -6.8761578 )
( 35.16269285155097 , -7.326561 )
( 22.805314220538836 , -5.453801 )
( 26.61050464226176 , -12.213893 )
( 33.49036217819625 , -4.8026505 )
( 24.039975728360066 , -16.281727 )
( 40.59259218956224 , -5.707135 )
( 31.579652441160896 , -9.031828 )
( 31.102465261804035 , 10.237376 )
( 14.113253421715811 , -15.1379795 )
( 21.11001202656732 , -17.765602 )
( 32.61579137796287 , -3.2058022 )
( 25.76632054963112 , -12.488496 )
( 28.277025798494726 , -5.3138347 )
( 34.77026906586744 , -9.753966 )
( 34.76837359274902 , -3.7900343 )
( 20.94738390324752 , -6.8733716 )
( 16.32267357891193 , -16.903696 )
( 36.87435971009155 , -6.7959595 )
( 24.670384683917693 , -13.12211 )
( 31.623407071690117 , 10.752769 )
( 35.78905515566422 , -5.7459393 )
( 22.461311955133844 , -12.358425 )
( 24.779070815207895 , -5.5107317 )
( 19.741462932539388 , 18.95273 )
( 26.02732535929103 , -15.518086 )
( 31.71960327269034 , -6.632189 )
( 28.44638821413367 , -8.674794 )
( 29.08881793320301 , -11.267805 )
( 31.709943466710055 , -7.3087177 )
( 33.226255927035425 , 7.886932 )
( 37.00605641779078 , -3.849785 )
( 22.109562843530558 , -17.009533 )
( 24.5371018484743 , -6.9945908 )
( 34.50899226493017 , -4.6007633 )
( 39.38052146945568 , -7.5597134 )
( 34.53216344690215 , 9.643274 )
( 33.72198058399747 , -4.923463 )
( 14.962221165981976 , -15.830524 )
( 32.854988395325805 , -9.61285 )
( 33.600519530588514 , 7.612752 )
( 32.04787708882429 , -7.795311 )
( 38.74528184459432 , -3.8719916 )
( 18.35983383291153 , -16.373795 )
( 25.47151809050091 , -11.225523 )
( 33.28250299036509 , -6.228513 )
( 36.225304795751335 , -6.1019354 )
( 32.74446515525911 , -7.102794 )
( 39.98090043399772 , -7.4361515 )
( 34.76010460864118 , -3.575603 )
( 25.32466176800243 , -12.286161 )
( 25.388511206551374 , -10.490519 )
( 36.00265580715479 , -6.6624093 )
( 18.2520418628653 , 14.41259 )
( 24.461722226216285 , -14.446651 )
( 24.686236582459543 , -11.050585 )
( 25.555060719145427 , -10.672408 )
( 25.992270130674235 , -9.017608 )
( 19.648955084894165 , -14.92982 )
( 31.50832233792505 , 9.316537 )
( 40.03183231202101 , -3.717733 )
( 28.111421991878725 , -6.643015 )
( 41.15626003111627 , -4.092356 )
( 29.848926058867143 , -9.042595 )
( 31.69459061692184 , -6.142051 )
( 34.20854915803302 , 10.238077 )
( 30.387711532073588 , 6.4567065 )
( 37.458053245486774 , -7.520169 )
( 26.57266717263867 , -11.487602 )
( 33.91541806232324 , -6.715288 )
( 34.70962500488928 , -4.6978436 )
( 31.31788173302038 , 8.3807745 )
( 23.86241457291046 , -13.630967 )
( 31.223463335824334 , -6.827143 )
( 34.958349901544835 , -8.549881 )
( 20.154725947400998 , 7.7307124 )
( 19.133372934377253 , 17.188173 )
( 19.679564508859265 , -14.058999 )
( 36.08622953946408 , -6.018161 )
( 34.6103063469385 , -4.7801476 )
( 34.42256843795616 , -6.1872115 )
( 33.17357554111332 , -9.022945 )
( 24.11270429131364 , -10.865513 )
( 29.20916833992609 , -12.11315 )
( 24.673133814083965 , -14.738144 )
( 40.18323041688037 , -5.5693426 )
( 24.796236735300972 , -5.7108974 )
( 40.5979988727476 , -7.4044166 )
( 26.253698818102926 , -15.538696 )
( 38.81014754090073 , -7.7299213 )
( 33.43324500679887 , 10.471966 )
( 21.390329547349776 , -17.055937 )
( 40.3357482316534 , -5.4405875 )
( 32.0459259834398 , -6.1729593 )
( 29.49612433087185 , -11.553379 )
( 39.99386742386043 , -6.0385513 )
( 38.39053632061206 , -4.348161 )
( 25.613476365121887 , -14.415786 )
( 18.8330713273474 , 14.91942 )
( 32.15215203891515 , 6.3105545 )
( 10.348562659450403 , 2.3814921 )
( 23.63724975232735 , -7.3779483 )
( 31.32716034703509 , -7.127053 )
( 36.453445835216954 , -5.865841 )
( 39.83160988658322 , -6.5802174 )
( 35.337180742516836 , -7.0200424 )
( 37.83118459504663 , -4.9921217 )
( 23.96714872444963 , -14.354057 )
( 34.51397771143954 , -6.099955 )
( 35.74940417271092 , -5.673976 )
( 24.060249069042115 , -13.559995 )
( 33.95761598006834 , -6.306274 )
( 26.523905802813392 , -16.063211 )
( 20.960973876754597 , -18.202671 )
( 38.918134842064404 , -7.5333853 )
( 27.858890006895543 , -10.1147175 )
( 21.829708193060746 , -5.37916 )
( 7.429668087233259 , -11.5755205 )
( 24.37344925866834 , -11.37314 )
( 18.38164557029281 , 14.332441 )
( 18.44938088774829 , 14.063114 )
( 37.17894047951803 , -5.4213686 )
( 30.96116914732167 , 9.780101 )
( 34.882695171877046 , -5.587113 )
( 35.822892380697354 , -6.040084 )
( 2.5947272540503867 , 20.143606 )
( 41.97479051054419 , -4.513036 )
( 36.880230269043565 , -6.1554413 )
( 40.568041772164314 , -3.8754442 )
( 28.45326140699919 , -13.09363 )
( 27.697914244032436 , -11.406705 )
( 25.041523075179086 , -13.410984 )
( 32.25375575835115 , -7.3599024 )
( 37.48419951271355 , -6.8944316 )
( 32.10262724999731 , -5.6276875 )
( 35.11121367845425 , -8.367856 )
( 32.2169833139254 , -4.1442575 )
( 30.869461768037613 , -8.770909 )
( 8.644358299235458 , 2.031159 )
( 24.621460295921032 , -14.523313 )
( 21.94307352500145 , 10.202996 )
( 31.0708267871276 , -6.4328747 )
( 28.67071935006296 , -13.540163 )
( 33.60134171701842 , -5.095708 )
( 33.18280555612799 , 5.372976 )
( 32.158817799572816 , -7.1282873 )
( 35.63901139664339 , -5.590254 )
( 19.007872622210105 , -17.315695 )
( 26.954794337368917 , -14.989201 )
( 37.067991174865384 , -6.1736875 )
( 21.81253398647727 , -13.6720915 )
( 33.93330899390357 , -6.2686453 )
( 32.31613499890445 , -5.2365966 )
( 29.789223308038167 , -7.7633634 )
( 33.04265494014401 , -7.439589 )
( 31.41037279776537 , -8.990867 )
( 28.342707154324163 , -12.318352 )
( 19.816237084498447 , -15.370251 )
( 33.32443330561487 , -6.783477 )
( 19.083371137402136 , -13.144209 )
( 31.761656728487843 , 9.908262 )
( 39.719584620602355 , -7.205009 )
( 38.86877148699883 , -4.0441313 )
( 32.003293001639435 , -6.1937647 )
( 24.845927676985763 , -12.97393 )
( 33.98248399578574 , -9.303354 )
( 22.575219664100743 , -5.8942676 )
( 34.4164835788749 , -4.384678 )
( 20.165197857339656 , -14.466449 )
( 39.602757256240594 , -3.4335127 )
( 36.79319745618599 , -5.260125 )
( 23.102865444160475 , -12.748839 )
( 41.290766065022325 , -4.7268486 )
( 31.546481599690733 , -6.672926 )
( 36.494330833018815 , -6.71612 )
( 29.68471974016409 , -11.708014 )
( 18.338387557368304 , 18.523832 )
( 23.158804471435452 , -11.081899 )
( 15.389141629896931 , -14.4758 )
( 23.952556289430103 , -15.754671 )
( 41.07221245385001 , -4.2854857 )
( 28.379130127483997 , -9.2910385 )
( 20.11760799147358 , 18.068075 )
( 25.905602548589005 , -14.61874 )
( 36.322322573517795 , -5.286018 )
( 28.396838916468436 , -12.70093 )
( 27.119006368860383 , -6.0902834 )
( 31.63814673100157 , 8.678341 )
( 25.809595467561145 , -15.320283 )
( 26.924519799459354 , -15.4822 )
( 19.7661111547486 , -6.7351437 )
( 26.817015549219917 , -6.092821 )
( 18.985133572167484 , 12.2913885 )
( 25.940826982768485 , -12.152317 )
( 32.78150171784585 , 7.467173 )
( 24.23381984773596 , -12.350724 )
( 28.439927818518832 , -12.812398 )
( 20.872009953460513 , -7.97974 )
( 31.533280299015757 , 7.6812596 )
( 35.3025431308904 , -8.130095 )
( 33.92044462259322 , 9.263458 )
( 31.629171656048126 , -7.259136 )
( 34.3147028812722 , -5.270219 )
( 15.253169459062523 , -17.192465 )
( 20.84222593367418 , -4.8107843 )
( 26.01809483100192 , -8.247723 )
( 29.658785949391586 , -7.706348 )
( 31.54407881207591 , 9.447005 )
( 26.247047541406967 , -14.428933 )
( 34.93919984679106 , 10.539953 )
( 20.394094691434702 , 9.503365 )
( 37.61950630539822 , -3.575105 )
( 30.222563758235854 , 8.366704 )
( 40.01646997994432 , -8.367523 )
( 35.624122738419764 , -5.8757925 )
( 33.054719474782864 , -3.0236945 )
( 33.74268247459917 , -8.165685 )
( 24.000030726751522 , -4.517605 )
( 28.503243822112744 , -14.332022 )
( 21.457952488648026 , -3.5112476 )
( 27.414315684271777 , -9.136511 )
( 40.438088443679085 , -3.7795115 )
( 30.876957317491957 , 8.072614 )
( 24.87464599107572 , -12.2765665 )
( 32.49379172844906 , 7.4776425 )
( 20.470475619239274 , -13.561352 )
( 34.49613358611962 , -6.2603974 )
( 27.127399974887524 , -5.3198705 )
( 31.547462257027277 , 9.8770895 )
( 10.054073135484398 , -12.328381 )
( 37.76407685581014 , -7.324956 )
( 30.917141929929198 , 10.043753 )
( 27.00636169945976 , -14.902921 )
( 33.20245933176197 , -3.4714885 )
( 25.79377126922423 , -5.174675 )
( 18.464565928138676 , -14.67624 )
( 38.57254679049623 , -6.5394635 )
( 32.014001373460836 , -7.036692 )
( 3.785493766411116 , -1.4244118 )
( 33.931346476504956 , -8.975592 )
( 18.184710874633893 , -15.439581 )
( 30.436239225483174 , -8.126427 )
( 37.16726933596548 , -4.336553 )
( 26.691770627528182 , -9.858268 )
( 34.95126052643242 , -6.4395556 )
( 27.87506817041995 , -13.488724 )
( 35.382103578995455 , -6.84824 )
( 21.313409255775614 , -6.3132405 )
( 25.02269453893141 , -13.984383 )
( 37.2641846351513 , -4.8330727 )
( 35.66138516037682 , -4.073023 )
( 35.23561420697994 , -6.5402713 )
( 31.38066780305833 , -7.1945696 )
( 34.18725747677574 , 10.348471 )
( 32.7924619551558 , -7.4401383 )
( 33.81881503564231 , -5.7360263 )
( 32.32944887484828 , -3.7194297 )
( 26.486352337815912 , -16.039078 )
( 41.44380500559915 , -4.0230646 )
( 37.91102848536405 , -7.4543104 )
( 27.60249213824098 , -11.762748 )
( 34.72116527024053 , -2.83413 )
( 35.002000577095565 , -9.40405 )
( 30.12426795694947 , 6.597245 )
( 23.954721417130173 , -15.423223 )
( 26.193440207055975 , -11.649824 )
( 32.517031002639634 , -3.912674 )
( 35.13899720304801 , -3.7129514 )
( 38.78681550316604 , -7.714012 )
( 37.33698395987212 , -6.7831755 )
( 36.60367859298953 , -5.159898 )
( 26.939506914932657 , -11.270037 )
( 34.137418606843255 , -6.6084666 )
( 38.99962037975017 , -7.130002 )
( 25.24996876352917 , -11.461806 )
( 29.242263800738268 , -13.162588 )
( 38.76921986615886 , -6.6678324 )
( 41.06056018344069 , -5.8350983 )
( 30.98505391184146 , 7.430523 )
( 32.505900956152516 , -6.1345873 )
( 37.216080922128526 , -4.0397735 )
( 32.468794399869225 , -4.306045 )
( 33.87875778277504 , -5.0312967 )
( 23.722008723169417 , -15.522822 )
( 28.86481929092303 , -9.511755 )
( 31.23537106647529 , 10.762413 )
( 32.638404309490745 , 10.1547575 )
( 35.66348320691342 , -6.916917 )
( 21.23296673332983 , -4.982245 )
( 25.527179897439726 , -8.486688 )
( 38.01551031403763 , -6.0178494 )
( 36.297541440527795 , -6.267022 )
( 28.156063350895103 , -5.9880986 )
( 14.023566713520452 , -18.048267 )
( 18.343403073175047 , 16.044659 )
( 30.40062480441057 , -8.234768 )
( 24.63038441926774 , -12.479593 )
( 20.534997029175578 , -5.6625166 )
( 34.32164218938149 , -4.5076857 )
( 32.51744432029379 , 8.068145 )
( 32.66056854593617 , 6.167318 )
( 34.23664201329531 , -7.223472 )
( 22.67452034910909 , -13.827731 )
( 20.932906983547603 , -7.838048 )
( 39.51943551708676 , -7.209876 )
( 32.78819157127425 , -6.129756 )
( 39.743877828333694 , -5.5598717 )
( 24.641080469848713 , -14.502774 )
( 31.73103461292598 , 7.1034126 )
( 26.03749537597519 , -7.7351513 )
( 22.05016171563699 , -6.7673903 )
( 39.06647910841185 , -5.5560417 )
( 32.975441454919164 , -8.674606 )
( 27.249485571742333 , -6.90397 )
( 33.95452230987607 , -8.9235325 )
( 16.91171570851234 , -15.417576 )
( 41.015761581507036 , -4.1206856 )
( 11.662925873980623 , 0.5788151 )
( 30.81020748305378 , 9.224079 )
( 30.43252808850635 , -8.08418 )
( 33.38801627782733 , -8.593141 )
( 22.930651143635004 , -4.3738456 )
( 27.947593233061863 , -15.056053 )
( 38.400597989543854 , -4.430229 )
( 34.90764042973128 , -8.821707 )
( 34.91580464822609 , 10.021659 )
( 33.08640609095581 , -5.582555 )
( 33.91688687672751 , -8.164225 )
( 31.32264194559882 , -6.36208 )
( 20.291617811355636 , 9.121892 )
( 33.12717104206742 , -6.9203067 )
( 35.59127253446876 , -8.499506 )
( 34.88326300727184 , -9.542443 )
( 23.38340906703761 , -15.432123 )
( 34.72518261051753 , -7.5202203 )
( 33.993957178471675 , -8.008498 )
( 30.187895881994507 , -7.5336776 )
( 41.10249416101675 , -4.8565407 )
( 8.778555407727653 , -21.1783 )
( 33.84095823536287 , -2.9096074 )
( 33.11021512034317 , -3.2574878 )
( 33.2843822431534 , -9.1209755 )
( 36.45387666153501 , -3.5397303 )
( 29.646079382596998 , -6.1437435 )
( 40.03785660610753 , -7.565748 )
( 36.027761594385304 , -5.4721694 )
( 32.12613997999067 , -6.1397567 )
( 23.709276187918075 , -11.188883 )
( 26.853000146280177 , -13.692698 )
( 26.412193758884065 , -7.6305356 )
( 30.70678533760118 , 10.311522 )
( 34.05778780843064 , -7.4870996 )
( 36.852856752021594 , -4.50616 )
( 27.395004717160077 , -14.190331 )
( 35.6922034085208 , -3.4304113 )
( 35.747622096811995 , -6.859576 )
( 33.0992506171165 , -6.593591 )
( 21.70994413115192 , -15.58271 )
( 41.06221016819851 , -5.027862 )
( 34.17793002020716 , -6.6518617 )
( 34.620424378045236 , -8.403458 )
( 33.17748340765239 , -7.6913056 )
( 11.116016965508706 , -15.136467 )
( 27.75141271468486 , -6.107295 )
( 31.882327968439355 , -6.082721 )
( 24.816091781999244 , -12.754278 )
( 27.056361282025996 , -15.10163 )
( 27.490355073441712 , -15.719962 )
( 31.55196658946628 , -6.1009626 )
( 33.764558052662835 , 8.367191 )
( 36.56435948021629 , -5.4176683 )
( 35.42391821408809 , -3.8843997 )
( 34.16500752153817 , -5.614603 )
( 36.324476318053314 , -5.890006 )
( 36.64101254071015 , -5.8500667 )
( 35.23306638300326 , -5.147388 )
( 23.380148879898144 , -14.07947 )
( 40.6203293135466 , -3.5197294 )
( 25.394341446189706 , -11.598055 )
( 21.963136740633857 , -12.685447 )
( 26.552195797174665 , -7.6176205 )
( 36.4027187063488 , -6.570481 )
( 31.074953945807547 , -6.6442924 )
( 42.090881228688225 , -3.8077796 )
( 33.833496324130905 , -8.17874 )
( 16.18054899890647 , 16.13898 )
( 31.86626833313723 , -5.862313 )
( 25.21847873417681 , -12.709575 )
( 26.707908603188653 , -16.450962 )
( 27.581652591268 , -11.7264805 )
( 33.97483453287975 , -6.902215 )
( 26.38778459686693 , -11.225505 )
( 25.356069592761735 , -14.197372 )
( 35.44529957595143 , -5.1898446 )
( 24.90770979039573 , -11.864471 )
( 31.794932241823243 , -7.570003 )
( 37.18169305611916 , -6.0853887 )
( 40.227197761586595 , -3.5667853 )
( 25.785263098392637 , -14.537728 )
( 36.77035549473477 , -3.5994563 )
( 26.93983763859024 , -16.10277 )
( 20.39471120565266 , 11.973918 )
( 32.95312073790831 , 10.877116 )
( 36.96999981803036 , -6.2069483 )
( 41.038782537324245 , -6.087069 )
( 39.50064085215665 , -3.300333 )
( 28.550546484030757 , -8.60255 )
( 39.00652438451735 , -4.1123495 )
( 38.69369711905328 , -6.756185 )
( 33.899701520576166 , -7.7471304 )
( 35.045854060365485 , -7.412763 )
( 34.12241550676127 , 9.976872 )
( 19.03986027713293 , -13.784222 )
( 29.611296001889617 , -12.215408 )
( 18.61513362042907 , 11.402729 )
( 38.37886562965609 , -7.158149 )
( 35.23758267531952 , -5.847143 )
( 35.055898007894605 , -5.2015514 )
( 34.64724875050398 , -8.437412 )
( 28.983283584742495 , -6.3548765 )
( 21.277873739885255 , 8.556139 )
( 33.72512286733978 , -4.7114353 )
( 29.58685963120332 , -7.5818653 )
( 40.02758363152913 , -7.298506 )
( 34.29140332523713 , -9.377256 )
( 24.090778366845687 , -13.80834 )
( 25.24908497025491 , -13.320117 )
( 33.2905644626415 , 5.3080907 )
( 16.68278596206827 , 19.892746 )
( 32.710206108631816 , -8.071563 )
( 35.96657565333173 , -6.901784 )
( 32.03198469205631 , -5.9202776 )
( 25.268292986926244 , -9.309264 )
( 22.960504413197626 , -8.2145605 )
( 23.289079996527168 , -14.437194 )
( 38.78386529780876 , -8.0764475 )
( -12.112849035657948 , 3.1421707 )
( 30.066264625999636 , -5.7882323 )
( 29.693218354703312 , 7.8341026 )
( 30.734705097410732 , -8.4237 )
( 36.05062035350515 , -7.4748864 )
( 31.029214951299863 , 6.368123 )
( 38.820944790526354 , -6.6558776 )
( 33.49510208678044 , -5.801836 )
( 29.935254695875358 , 8.408279 )
( 24.85343569402169 , -5.545498 )
( 31.707629543201897 , -5.4997554 )
( 35.53513459798149 , 9.684173 )
( 34.29424751448113 , 7.8849745 )
( 25.62589940873335 , -10.687131 )
( 25.89386967119173 , -11.376796 )
( 15.09950157390881 , 12.117475 )
( 33.70770700693911 , -2.8327122 )
( 39.874577480804184 , -6.5497074 )
( 42.19981676023478 , -3.4787145 )
( 34.277726511016375 , 9.389284 )
( 27.87693991131234 , -10.822712 )
( 18.632354886523185 , 16.589445 )
( 24.811230186899344 , -11.945796 )
( 27.17065052000396 , -11.641887 )
( 36.496046982359914 , -6.1378303 )
( 34.76586338701323 , 8.972687 )
( 24.06833655850569 , -15.361569 )
( 19.712911504690155 , 7.238616 )
( 25.175091663214552 , -9.411212 )
( 36.863329817429246 , -7.2929306 )
( 31.446725767307132 , 9.52585 )
( 29.827274166984154 , -7.759322 )
( 29.062380916701382 , -7.724966 )
( 27.114034297865878 , -14.203669 )
( 9.790241233307121 , 1.9709857 )
( 21.33663410371555 , -7.854273 )
( 17.29976906796546 , -16.107807 )
( 35.19096029679361 , -6.2875237 )
( 29.25505632333622 , -9.435914 )
( 35.09403761041487 , -5.3673277 )
( 38.72992286626984 , -3.4624734 )
( 33.71219279588793 , -5.8814545 )
( 32.03256233161337 , 7.737664 )
( 40.34279449889793 , -6.9103622 )
( 38.05235265139269 , -7.4672074 )
( 34.52236816460027 , -7.5231576 )
( 39.0108650092447 , -3.5803075 )
( 35.12831669053984 , -6.9914675 )
( 29.719692826002184 , -8.939097 )
( 37.91461000891694 , -5.929719 )
( 37.17888009149164 , -5.4383473 )
( 36.72961602366098 , -4.089585 )
( 33.856461017753766 , -5.29569 )
( 33.05205429459912 , -5.8821564 )
( 18.936983416423406 , 8.963632 )
( 31.827171887974245 , -5.7942924 )
( 24.72517630859962 , -11.906437 )
( 34.29043252951146 , -5.705115 )
( 18.865407403033014 , -17.49092 )
( 20.392798868788965 , -5.4272814 )
( 22.073344913463433 , -3.7620761 )
( 36.54409662149129 , -5.65211 )
( 19.129295548669788 , -14.768337 )
( 39.12125529228621 , -7.169405 )
( 22.87301297092151 , -5.807884 )
( 31.788967800730145 , 5.8668523 )
( 28.997109424188785 , -12.841042 )
( 23.012535037745977 , -4.725824 )
( 34.77203416373149 , 9.23748 )
( 31.45388367610416 , 9.598842 )
( 38.03987891370384 , -5.4820123 )
( 25.40993310138083 , -5.8456945 )
( 38.30777788079774 , -7.40718 )
( 39.431566108116016 , -8.341388 )
( 24.49285990038878 , -15.212117 )
( 24.402961181745944 , -10.927504 )
( 32.3718896591699 , -8.995624 )
( 27.77087158021384 , -4.7048287 )
( 39.250779942040566 , -5.2935266 )
( 33.70393231732394 , -4.10439 )
( 30.651579647220785 , -6.346133 )
( 38.637146778290095 , -7.8364964 )
( 32.68587820878182 , -4.7864084 )
( 33.02317179390761 , -6.6203403 )
( 32.1371507176413 , -8.462491 )
( 33.72111953346118 , 5.83974 )
( 41.373186249322266 , -4.7893877 )
( 34.595545299305186 , -4.0970016 )
( 18.91989571636589 , 9.80772 )
( 31.66122946541526 , 8.823834 )
( 7.5595321395558015 , -10.236821 )
( 40.12002774838881 , -6.765739 )
( 30.806631490014492 , -5.801771 )
( 36.3117856533565 , -3.6677177 )
( 33.61201984587737 , -6.6545024 )
( 29.01051573598371 , -11.283291 )
( 5.619950811670716 , -16.019272 )
( 34.397958227730136 , -7.777925 )
( 40.37020908981519 , -3.258086 )
( 34.06780427706178 , -6.275248 )
( 32.523076628595874 , 6.9408393 )
( 35.576027104197586 , -4.979027 )
( 40.45422394987088 , -6.84559 )
( 22.875551743583358 , -8.380713 )
( 30.059867496793174 , -8.228157 )
( 38.67766022133556 , -2.0574632 )
( 27.703603648988032 , -14.874353 )
( 21.605316558822523 , 8.7493305 )
( 32.87532844222325 , 8.701223 )
( 33.70195836038554 , -7.3699055 )
( 34.951221169055316 , 10.340494 )
( 37.35553154475986 , -4.158914 )
( 38.12781448951637 , -4.6233683 )
( 32.48850435107181 , 11.470796 )
( 31.09463613727953 , 9.119246 )
( 35.31203908698735 , 9.559869 )
( 35.19847005667082 , -6.2058063 )
( 21.570928684476677 , -7.9150267 )
( 21.409298886240542 , -12.648194 )
( 28.25602286425895 , -4.9762077 )
( 27.66857419368931 , -4.932096 )
( 30.260794232402557 , 9.2860155 )
( 33.41378030785581 , -8.285376 )
( 32.90444217846883 , -7.4164653 )
( 34.52663781985207 , -9.397164 )
( 34.46325072790401 , 10.120183 )
( 21.46052288965662 , -5.0510283 )
( 27.55353176997407 , -15.34692 )
( 26.972927200508927 , -10.365394 )
( 23.668705860541994 , -13.481489 )
( 38.46116690067567 , -6.0185103 )
( 38.40343416312591 , -6.0717583 )
( 35.27634142960076 , 9.689868 )
( 33.93325359374165 , -4.995784 )
( 20.052243596811557 , 18.72502 )
( 19.637116435607012 , 15.528465 )
( 35.81721350899269 , -6.973382 )
( 32.05554002245933 , 7.7312665 )
( 34.42073986631071 , 9.756302 )
( 17.683974941346463 , 15.346285 )
( 39.090336959028775 , -3.3676481 )
( 33.589296878596365 , -6.343174 )
( 34.60266545449186 , -3.1669807 )
( 38.76015746305716 , -3.2663815 )
( 34.12150215764184 , -3.2379866 )
( 33.85235050799966 , -3.8842158 )
( 16.353123684369283 , -17.111 )
( 34.810524121345566 , -9.911394 )
( 32.62310981192874 , 8.723086 )
( 34.20701123173998 , -4.9427385 )
( 34.46267298360046 , -5.5611324 )
( 33.55671592117271 , -8.0616865 )
( 34.67202284890875 , -7.6124077 )
( 34.77344680418227 , -5.78602 )
( 33.24660697333295 , -2.4446435 )
( 19.097748782692204 , -15.8874855 )
( 21.17342298622756 , -6.282707 )
( 23.22442499180483 , -15.13942 )
( 35.472883662478274 , -3.705066 )
( 22.668412083858776 , -11.786312 )
( 33.03348379616022 , -2.8043272 )
( 24.709781937120937 , -6.131446 )
( 38.07099307926448 , -4.312094 )
( 28.904173360967782 , -10.5708065 )
( 32.557569734630434 , 10.0916195 )
( 29.10540202563562 , -13.651838 )
( 31.048799173398823 , -8.711714 )
( 35.82995943736652 , -5.53742 )
( 41.03922124767695 , -4.4813905 )
( 34.61310546669014 , 7.5079713 )
( 31.950515602031878 , 7.918631 )
( 32.05791896825756 , -6.389774 )
( 32.80253741477757 , -5.6061163 )
( 33.51213960260061 , 8.418203 )
( 29.03875903653404 , -11.825421 )
( 29.874319117103738 , -8.988271 )
( 37.749180529885564 , -3.0903997 )
( 22.930399692761174 , -8.466262 )
( 22.035605023921896 , -8.5756645 )
( 19.397424326091098 , -16.73021 )
( 30.42372970296535 , 6.7333326 )
( 32.90840461354012 , -4.726308 )
( 22.829626223772703 , -7.7664795 )
( 18.247937997414564 , 15.947941 )
( 34.43392472084263 , 8.470539 )
( 39.953804243277226 , -8.305355 )
( 32.96115926944776 , -6.478014 )
( 35.37427362194359 , -3.4706779 )
( 17.254055826993298 , -14.458218 )
( 19.183541900657833 , -16.134691 )
( 16.816588992557357 , -15.393894 )
( 32.04258226146042 , -7.6462417 )
( 19.972725394025634 , -16.030663 )
( 24.261119499512077 , -15.871736 )
( 32.01142236919534 , -4.3033695 )
( 33.31832460516379 , -5.604987 )
( 26.439419160323098 , -14.814927 )
( 40.41043175585792 , -3.9106586 )
( 36.49937231188628 , -6.273827 )
( 26.449593471675154 , -9.170161 )
( 39.97771727616839 , -5.2565236 )
( 35.04497584743132 , -5.5330687 )
( 38.63240365336764 , -6.2261806 )
( 34.05629392712132 , -4.5915947 )
( 34.63413554853643 , -4.804898 )
( 33.288138570356004 , -7.8872867 )
( 29.453576238329525 , -12.4918995 )
( 33.84575149298101 , -3.2642727 )
( 31.980574869152985 , 11.207285 )
( 38.14155588986045 , -3.5985126 )
( 33.79161511190438 , 7.633175 )
( 31.634330457085394 , 11.458857 )
( 40.76854708700893 , -3.075007 )
( 16.11527894582526 , 16.843348 )
( 33.4473916947134 , 10.723595 )
( 32.000871776647436 , -6.251796 )
( 40.42868279154715 , -5.2334356 )
( 19.811270398907077 , 12.348341 )
( 20.450441889854183 , -12.742494 )
( 25.84846610016251 , -15.003475 )
( 31.777209066276846 , 8.5260105 )
( 33.27321039867952 , -5.375407 )
( 20.262419305308416 , -17.164787 )
( 32.64850967252891 , -3.6561542 )
( 33.29791567461419 , -7.265791 )
( 32.60056077307632 , -7.51372 )
( 35.12038440950698 , -4.0880513 )
( 33.536887963979055 , 7.857216 )
( 36.20766819371841 , -6.4384546 )
( 8.624306219404923 , -0.75438726 )
( 40.29453145452019 , -6.543644 )
( 24.38699446985504 , -15.386632 )
( 21.884173750875917 , -8.774491 )
( 38.66593023341042 , -2.9886947 )
( 36.08058482078245 , -5.7109804 )
( 39.04843787112041 , -4.142879 )
( 31.385560289115457 , -6.1296425 )
( 33.84407023758092 , -5.295642 )
( 30.6913163453293 , -7.0964713 )
( 35.00889308406855 , -9.126928 )
( 17.08630269963148 , -18.810442 )
( 33.27263567002342 , -4.8776293 )
( 33.19202611105419 , 7.824921 )
( 34.130896033279434 , 10.213755 )
( 25.046113697930473 , -12.266691 )
( 40.474762101582286 , -3.9975789 )
( 35.53370370084878 , -7.6050034 )
( 33.49703094187659 , 7.526033 )
( 29.384883522199015 , -10.16214 )
( 34.640887523164615 , -4.960434 )
( 32.08656005235964 , -4.75851 )
( 21.83613664646608 , 10.80728 )
( 37.658044110241306 , -3.053193 )
( 24.407105542830656 , -12.718622 )
( 33.39869197113705 , -2.26307 )
( 27.631782827215034 , -15.921195 )
( 7.236842689027008 , -18.642538 )
( 33.61977979776889 , 7.4504275 )
( 37.178974583343475 , -4.8677154 )
( 32.805893635116476 , -6.6202593 )
( 41.74687900957899 , -4.6962543 )
( 39.23380775937681 , -3.8789046 )
( 32.885423063505435 , 8.556135 )
( 34.15492268957931 , -9.563635 )
( 24.177781227718366 , -8.48921 )
( 28.584779119746848 , -12.05608 )
( 33.68436179570615 , 9.250923 )
( 35.97100456351283 , -5.6987305 )
( 18.96023875303978 , 8.234638 )
( 30.117272464382832 , 6.5338 )
( 26.328362335442783 , -15.087212 )
( 30.423235310332004 , 6.580624 )
( 31.32995545635916 , -7.661863 )
( 34.71946610965902 , -4.1323795 )
( 36.120131332239254 , -5.2355685 )
( 26.800570083312362 , -16.229599 )
( 33.53362016505358 , -5.4021134 )
( 40.39148438352227 , -4.2713056 )
( 39.953529688576964 , -5.53385 )
( 32.23406937506621 , -7.5650992 )
( 38.1492396936721 , -4.056751 )
( 40.50916837338535 , -6.7733974 )
( 28.906108160749824 , -7.935751 )
( 37.160089095219064 , -4.7266455 )
( 33.08308223955535 , 8.14577 )
( 26.223216885150407 , -12.845053 )
( 21.563568913658663 , -13.586773 )
( 39.6492019373186 , -6.453798 )
( 32.04590339502899 , -4.962522 )
( 18.26574769086919 , -13.257122 )
( 16.899387804164594 , 18.757183 )
( 39.014403448518266 , -4.6196494 )
( 31.929934537167206 , -5.3091226 )
( 29.987032387170842 , -9.299065 )
( 27.78899485128985 , -13.992435 )
( 32.9261189488422 , 8.347476 )
( 27.412719480200042 , -15.502033 )
( 41.58865278719463 , -3.9050162 )
( 19.51756847428361 , 17.610987 )
( 32.0932349160175 , -7.8395915 )
( 40.2935179069426 , -4.4934893 )
( 34.03855309627093 , -7.7779446 )
( 28.747841461227285 , -11.187366 )
( 31.1777221057431 , 9.253354 )
( 35.62626905896162 , -5.9194455 )
( 24.321149569050405 , -4.3149652 )
( 26.044368218485474 , -15.76145 )
( 32.215699413283765 , 8.467854 )
( 18.607913280773847 , 7.315309 )
( 35.2705941155071 , -7.382926 )
( 35.94474934304075 , -5.320357 )
( 40.4445477296077 , -3.1832461 )
( 22.75927799930691 , -15.529906 )
( 28.124622557908665 , -7.6924047 )
( 24.699518007158154 , -5.3961573 )
( 32.574925052318505 , 8.701686 )
( 31.6968317399523 , -8.679827 )
( 40.17229189977225 , -3.0354013 )
( 34.50073747465127 , -3.851843 )
( 33.58526202331896 , -8.822645 )
( 35.18936451119475 , -6.280512 )
( 25.380452083813687 , -11.333585 )
( 32.89621667942565 , -3.2216172 )
( 30.10076944966183 , 8.54246 )
( 23.458393873997657 , -14.305501 )
( 34.47185105313723 , -5.675569 )
( 26.978406817448466 , -16.405563 )
( 40.193612069910166 , -8.314745 )
( 25.993084642688476 , -11.810243 )
( 27.94409323434084 , -13.996075 )
( 31.627913519731923 , -4.9523363 )
( 28.817201114821447 , -8.161961 )
( 17.833269046889306 , 18.3857 )
( 33.600457921978546 , -8.097512 )
( 27.39441271606376 , -15.836866 )
( 30.922810103168487 , -8.029251 )
( 28.19748472470411 , -11.361472 )
( 31.66520415892485 , -7.5123434 )
( 38.08489799526574 , -5.4115644 )
( 38.0703044394377 , -3.416066 )
( 27.242664174945915 , -5.966516 )
( 32.09156069419752 , -6.6188297 )
( 38.99319121544138 , -3.2976735 )
( 32.7246893912801 , 10.522272 )
( 35.723372462541775 , -6.7059493 )
( 32.455243103031876 , -6.3146715 )
( 36.90856651694638 , -4.5057387 )
( 32.406659765592735 , 6.639674 )
( 41.42472656216824 , -5.0606775 )
( 29.342501257468694 , -10.059666 )
( 39.54758862625337 , -7.205263 )
( 35.49255278356418 , -6.1003127 )
( 31.238806102051253 , -6.9509597 )
( 30.616716397104064 , 7.7973895 )
( 31.872642714049825 , -6.3092065 )
( 35.25182404254292 , -4.7989883 )
( 26.684977672863695 , -10.972187 )
( 29.442594179201034 , -10.369938 )
( 29.63383532142278 , -9.3551 )
( 33.68094566227282 , -3.336261 )
( 17.120772559663383 , 11.573475 )
( 34.961746888187164 , -7.2449827 )
( 32.22781723914764 , 10.390784 )
( 22.437277723715535 , -15.102772 )
( 33.40837643694654 , -8.157232 )
( 27.65130889670058 , -12.997223 )
( 27.531560457911617 , -11.824487 )
( 34.5670411836849 , -8.191924 )
( 41.74875610565822 , -3.889411 )
( 26.307951435292587 , -13.883218 )
( 31.89146502086846 , 6.7201943 )
( 35.73328722236937 , -6.8317313 )
( 19.27523933036466 , 7.693486 )
( 37.8292008691768 , -4.4943476 )
( 37.14330271421103 , -5.2650776 )
( 36.42304922828001 , -4.820799 )
( 26.153836779974434 , -7.686323 )
( 34.297764785947955 , -4.3606052 )
( 30.263140203069927 , 7.9078765 )
( 27.842439401276955 , -15.345373 )
( 26.708466702897375 , -14.694075 )
( 24.504070379347926 , -14.872782 )
( 32.99294213118022 , 9.634254 )
( 40.70652150422813 , -4.0509405 )
( 42.41980809170193 , -4.1225505 )}

\newcommand{\utilityD}{
( 28.641607118358092 , 8.357604 )
( 40.583312343429085 , 24.20777 )
( 24.94821810861823 , 3.0985262 )
( 31.817277174366406 , 12.598244 )
( 27.538644376105765 , 7.0449595 )
( 31.279277615094994 , 11.657818 )
( 34.213060040501865 , 15.652487 )
( 32.377360774699696 , 13.012259 )
( 31.3650167922573 , 11.7737465 )
( 33.475303613174574 , 14.660285 )
( 39.818018124568766 , 23.279106 )
( 39.58205812831336 , 22.889668 )
( 32.90101267879367 , 13.701676 )
( 33.85317264925551 , 15.241055 )
( 30.74468641807341 , 10.905085 )
( 34.295625120561816 , 15.722944 )
( 40.589358212733174 , 24.182808 )
( 32.525955268268554 , 13.512923 )
( 18.86352175801707 , -3.66957 )
( 19.60774895313541 , -4.3639197 )
( 40.815525623123605 , 24.501253 )
( 35.64805003248654 , 17.516306 )
( 19.926624604688875 , -2.3454947 )
( 24.826147770557995 , 3.0341988 )
( 32.39834976381803 , 13.349851 )
( 15.707485599480762 , -8.342764 )
( 40.580749410792365 , 24.233213 )
( 33.20695513890671 , 14.413517 )
( 25.944252101466684 , 4.4317513 )
( 31.74860304421942 , 12.175035 )
( 24.09962872470965 , 2.4423993 )
( 34.261290426851076 , 15.396093 )
( 39.451290850386755 , 22.787346 )
( 25.272509996319567 , 4.1150155 )
( 19.108473910585555 , -5.0317607 )
( 23.725269073944414 , 1.5984886 )
( 35.10103617967763 , 16.791733 )
( 39.107897532247925 , 22.290127 )
( 24.124318025808865 , 2.4503422 )
( 22.960440202107804 , 0.8928215 )
( 33.419339527345485 , 14.686629 )
( 25.588690295793285 , 4.3661013 )
( 36.318682250215744 , 18.306591 )
( 33.07904400906489 , 14.167708 )
( 35.43128698699658 , 17.172712 )
( 26.448875586881154 , 5.0111346 )
( 27.98482192713519 , 7.450217 )
( 24.56230051769584 , 3.05701 )
( 34.67787540021606 , 16.283289 )
( 35.229756283174474 , 16.988836 )
( 34.191609332074464 , 15.253812 )
( 33.50557895899502 , 14.751655 )
( 37.205912546696105 , 19.941248 )
( 35.48822572650235 , 17.247972 )
( 29.919016605881637 , 9.99221 )
( 21.280912787649825 , -1.0212834 )
( 33.42002933210724 , 14.681875 )
( 28.68661776308989 , 8.440013 )
( 23.991746512622306 , 2.2848501 )
( 38.825876362387596 , 21.989494 )
( 36.15015354077819 , 18.11496 )
( 32.25344242481992 , 13.172646 )
( 24.262559748847096 , 2.6409483 )
( 32.68318527742466 , 13.708494 )
( 31.00997416552923 , 11.316001 )
( 36.55725777809617 , 18.640255 )
( 32.467547977392805 , 13.116119 )
( 34.9647559988635 , 16.263063 )
( 32.69546884916133 , 13.338202 )
( 34.49819908140167 , 16.092697 )
( 30.928574151100356 , 11.093094 )
( 32.22751431213917 , 13.150663 )
( 35.738455368152586 , 17.587463 )
( 27.960141829191798 , 7.586171 )
( 31.184479008523393 , 11.529827 )
( 36.2367840749645 , 18.683699 )
( 21.839272336283013 , 0.12544215 )
( 36.49682630879627 , 19.017967 )
( 33.305600850112555 , 14.173782 )
( 25.103006035887724 , 3.8896139 )
( 39.581198209182254 , 22.92746 )
( 34.96060090052615 , 16.646023 )
( 17.020164268251865 , -6.3127193 )
( 39.61380476783444 , 22.97676 )
( 33.05608202260174 , 13.860216 )
( 35.00824624051854 , 16.647491 )
( 33.4934992968809 , 14.362402 )
( 6.751039349052832 , -19.48729 )
( 26.062213612820642 , 4.552505 )
( 25.795186670179028 , 4.802056 )
( 22.84799269202622 , 0.84742403 )
( 27.1258182169797 , 5.8854504 )
( 38.10977281250606 , 21.088593 )
( 36.29843964875715 , 18.304903 )
( 40.50125285875163 , 24.124851 )
( 23.22775448452817 , 1.3161626 )
( 35.252327536827764 , 17.019165 )
( 38.65637995429246 , 21.74469 )
( 40.38672525533176 , 23.922653 )
( 31.515035556150767 , 11.839474 )
( 40.50531541869345 , 24.085615 )
( 31.685698700674042 , 12.174804 )
( 31.32961328869699 , 11.610272 )
( 23.57934328136322 , 1.7844274 )
( 23.09432297176417 , 1.1424515 )
( 32.9518541502414 , 14.080949 )
( 35.45796907774144 , 17.279898 )
( 25.621857528373184 , 4.2898803 )
( 37.934753810862794 , 20.828999 )
( 26.96332280971096 , 5.7392063 )
( 31.72827829273838 , 12.48512 )
( 30.736646043831232 , 10.897665 )
( 28.100833564011165 , 7.6605873 )
( 33.92917668812341 , 15.350027 )
( 33.58381527184225 , 14.862959 )
( 34.79864331667709 , 16.438456 )
( 38.269193413211376 , 21.207586 )
( 32.67561474967453 , 13.6804085 )
( 36.569444633053976 , 18.643368 )
( 32.948338395006665 , 14.075289 )
( 36.593957168095464 , 19.150307 )
( 20.564455092036333 , -1.9527446 )
( 29.562035142059035 , 9.546655 )
( 9.023836881887856 , -14.854486 )
( 26.75122813642974 , 5.9434485 )
( 27.845806710223904 , 7.318995 )
( 24.70876108899276 , 3.2425196 )
( 35.37626103276529 , 17.12469 )
( 33.235135034692306 , 14.129254 )
( 33.05228789006134 , 14.207953 )
( 32.74134573069266 , 13.729351 )
( 35.00126004808046 , 16.645962 )
( 24.342972353587754 , 2.3673697 )
( 22.636493761541963 , 0.5833692 )
( 29.7381057884806 , 9.704934 )
( 25.488264801117538 , 4.244535 )
( 32.37100585383353 , 13.303724 )
( 36.609887235242795 , 18.684052 )
( 22.2972952483215 , 0.1597448 )
( 27.95583343743546 , 7.454747 )
( 32.83414737075985 , 13.582498 )
( 32.30888030213087 , 13.211018 )
( 35.43001082231518 , 17.193241 )
( 33.459172755431 , 14.684083 )
( 34.01374343632955 , 15.379757 )
( 40.78151746890384 , 24.422798 )
( 33.95583892188105 , 15.3588505 )
( 34.35291199530505 , 15.878275 )
( 28.311588634666087 , 7.9630966 )
( 27.96170226954663 , 7.4087725 )
( 16.323611430210466 , -7.29641 )
( 22.236801118687687 , 0.086081445 )
( 33.50499969143415 , 14.772791 )
( 35.37765466146984 , 17.11879 )
( 26.958524814870948 , 5.730657 )
( 21.7913095552882 , -0.5840122 )
( 22.50079611400843 , 0.9388166 )
( 30.264910302575 , 10.276867 )
( 24.90416511890277 , 3.1233628 )
( 25.36302901246181 , 3.677618 )
( 30.028807100713365 , 10.1245575 )
( 26.11135317812774 , 5.121175 )
( 25.306358012979487 , 4.1364937 )
( 34.71088541407079 , 16.276423 )
( 39.60334031055676 , 23.003668 )
( 37.168664302711136 , 19.843784 )
( 31.09515219323982 , 11.331486 )
( 34.79623962095831 , 16.064041 )
( 40.50086872403076 , 24.14957 )
( 32.51052576147439 , 13.4482565 )
( 35.02825431029954 , 16.71252 )
( 32.47772202979256 , 13.443653 )
( 31.34939530777636 , 12.013626 )
( 37.191593702816384 , 19.43533 )
( 38.49900208899635 , 21.5547 )
( 35.5014313753338 , 17.26909 )
( 31.69185857121992 , 12.066975 )
( 38.847708226471575 , 22.049843 )
( 34.92931993468824 , 16.218342 )
( 35.4962394498725 , 17.303696 )
( 35.16269285155097 , 16.894981 )
( 22.805314220538836 , 0.71631515 )
( 26.61050464226176 , 5.805554 )
( 33.49036217819625 , 14.7172365 )
( 24.039975728360066 , 2.451426 )
( 40.59259218956224 , 24.229044 )
( 31.579652441160896 , 12.355936 )
( 31.102465261804035 , 11.320878 )
( 14.113253421715811 , -10.37151 )
( 21.11001202656732 , -0.79433644 )
( 32.61579137796287 , 13.573116 )
( 25.76632054963112 , 4.735517 )
( 28.277025798494726 , 7.8110547 )
( 34.77026906586744 , 16.433441 )
( 34.76837359274902 , 16.325882 )
( 20.94738390324752 , -1.6232729 )
( 16.32267357891193 , -6.790854 )
( 36.87435971009155 , 19.513756 )
( 24.670384683917693 , 3.1946373 )
( 31.623407071690117 , 11.974651 )
( 35.78905515566422 , 17.655283 )
( 22.461311955133844 , 0.36652732 )
( 24.779070815207895 , 3.2313437 )
( 19.741462932539388 , -4.2270794 )
( 26.02732535929103 , 5.125232 )
( 31.71960327269034 , 12.488823 )
( 28.44638821413367 , 8.09031 )
( 29.08881793320301 , 8.944268 )
( 31.709943466710055 , 12.489309 )
( 33.226255927035425 , 14.070363 )
( 37.00605641779078 , 19.177124 )
( 22.109562843530558 , 0.4644866 )
( 24.5371018484743 , 2.9512036 )
( 34.50899226493017 , 16.010838 )
( 39.38052146945568 , 22.720268 )
( 34.53216344690215 , 15.700479 )
( 33.72198058399747 , 15.01453 )
( 14.962221165981976 , -8.543936 )
( 32.854988395325805 , 13.9913025 )
( 33.600519530588514 , 14.552242 )
( 32.04787708882429 , 12.928931 )
( 38.74528184459432 , 21.84145 )
( 18.35983383291153 , -4.1331034 )
( 25.47151809050091 , 4.3361535 )
( 33.28250299036509 , 14.471839 )
( 36.225304795751335 , 18.673943 )
( 32.74446515525911 , 13.810682 )
( 39.98090043399772 , 23.482626 )
( 34.76010460864118 , 16.311295 )
( 25.32466176800243 , 4.012179 )
( 25.388511206551374 , 4.2165327 )
( 36.00265580715479 , 18.400951 )
( 18.2520418628653 , -4.7927885 )
( 24.461722226216285 , 3.1108847 )
( 24.686236582459543 , 3.175664 )
( 25.555060719145427 , 3.8659997 )
( 25.992270130674235 , 4.971009 )
( 19.648955084894165 , -2.5184631 )
( 31.50832233792505 , 11.855224 )
( 40.03183231202101 , 23.4772 )
( 28.111421991878725 , 7.625256 )
( 41.15626003111627 , 24.916452 )
( 29.848926058867143 , 9.883662 )
( 31.69459061692184 , 12.447698 )
( 34.20854915803302 , 15.27705 )
( 30.387711532073588 , 10.481979 )
( 37.458053245486774 , 20.270893 )
( 26.57266717263867 , 5.1775255 )
( 33.91541806232324 , 15.287179 )
( 34.70962500488928 , 16.268219 )
( 31.31788173302038 , 11.630351 )
( 23.86241457291046 , 2.1751547 )
( 31.223463335824334 , 11.868187 )
( 34.958349901544835 , 16.650234 )
( 20.154725947400998 , -2.4237576 )
( 19.133372934377253 , -4.968237 )
( 19.679564508859265 , -2.686357 )
( 36.08622953946408 , 18.49522 )
( 34.6103063469385 , 16.135664 )
( 34.42256843795616 , 15.923148 )
( 33.17357554111332 , 14.385931 )
( 24.11270429131364 , 2.4416633 )
( 29.20916833992609 , 9.113539 )
( 24.673133814083965 , 2.819564 )
( 40.18323041688037 , 23.705038 )
( 24.796236735300972 , 3.2569919 )
( 40.5979988727476 , 24.268017 )
( 26.253698818102926 , 5.413951 )
( 38.81014754090073 , 21.997005 )
( 33.43324500679887 , 14.285132 )
( 21.390329547349776 , -0.45071578 )
( 40.3357482316534 , 23.896866 )
( 32.0459259834398 , 12.903381 )
( 29.49612433087185 , 9.468448 )
( 39.99386742386043 , 23.472717 )
( 38.39053632061206 , 21.398619 )
( 25.613476365121887 , 4.0111732 )
( 18.8330713273474 , -4.0623193 )
( 32.15215203891515 , 12.732091 )
( 10.348562659450403 , -14.806321 )
( 23.63724975232735 , 1.8123193 )
( 31.32716034703509 , 12.005934 )
( 36.453445835216954 , 18.96006 )
( 39.83160988658322 , 23.276295 )
( 35.337180742516836 , 17.111431 )
( 37.83118459504663 , 20.69835 )
( 23.96714872444963 , 2.3222237 )
( 34.51397771143954 , 16.04553 )
( 35.74940417271092 , 17.60342 )
( 24.060249069042115 , 2.4257922 )
( 33.95761598006834 , 15.333195 )
( 26.523905802813392 , 5.201923 )
( 20.960973876754597 , -0.97590166 )
( 38.918134842064404 , 22.130833 )
( 27.858890006895543 , 7.3692436 )
( 21.829708193060746 , -0.5277147 )
( 7.429668087233259 , -18.012852 )
( 24.37344925866834 , 2.7833686 )
( 18.38164557029281 , -4.626198 )
( 18.44938088774829 , -4.534833 )
( 37.17894047951803 , 19.875711 )
( 30.96116914732167 , 11.149555 )
( 34.882695171877046 , 16.505466 )
( 35.822892380697354 , 17.703943 )
( 2.5947272540503867 , -23.759453 )
( 41.97479051054419 , 25.966959 )
( 36.880230269043565 , 19.509125 )
( 40.568041772164314 , 24.163145 )
( 28.45326140699919 , 7.6031804 )
( 27.697914244032436 , 6.609211 )
( 25.041523075179086 , 3.2636864 )
( 32.25375575835115 , 13.182924 )
( 37.48419951271355 , 20.292364 )
( 32.10262724999731 , 12.957685 )
( 35.11121367845425 , 16.8491 )
( 32.2169833139254 , 13.0829 )
( 30.869461768037613 , 11.446442 )
( 8.644358299235458 , -16.970324 )
( 24.621460295921032 , 3.315791 )
( 21.94307352500145 , -0.19269866 )
( 31.0708267871276 , 11.390644 )
( 28.67071935006296 , 8.4547 )
( 33.60134171701842 , 14.856525 )
( 33.18280555612799 , 14.062548 )
( 32.158817799572816 , 13.057624 )
( 35.63901139664339 , 17.468838 )
( 19.007872622210105 , -3.4803143 )
( 26.954794337368917 , 5.730436 )
( 37.067991174865384 , 19.74862 )
( 21.81253398647727 , -0.43497854 )
( 33.93330899390357 , 15.301523 )
( 32.31613499890445 , 13.222235 )
( 29.789223308038167 , 9.783434 )
( 33.04265494014401 , 14.189244 )
( 31.41037279776537 , 12.139551 )
( 28.342707154324163 , 8.013816 )
( 19.816237084498447 , -2.2970712 )
( 33.32443330561487 , 14.535737 )
( 19.083371137402136 , -3.4630165 )
( 31.761656728487843 , 12.166704 )
( 39.719584620602355 , 23.145422 )
( 38.86877148699883 , 22.001993 )
( 32.003293001639435 , 12.849472 )
( 24.845927676985763 , 3.5724 )
( 33.98248399578574 , 15.421529 )
( 22.575219664100743 , 0.43157342 )
( 34.4164835788749 , 15.88132 )
( 20.165197857339656 , -2.0601082 )
( 39.602757256240594 , 22.925318 )
( 36.79319745618599 , 19.381344 )
( 23.102865444160475 , 1.1910486 )
( 41.290766065022325 , 25.099766 )
( 31.546481599690733 , 12.269091 )
( 36.494330833018815 , 19.028208 )
( 29.68471974016409 , 9.711583 )
( 18.338387557368304 , -6.006052 )
( 23.158804471435452 , 1.2307824 )
( 15.389141629896931 , -8.0257845 )
( 23.952556289430103 , 1.9209898 )
( 41.07221245385001 , 24.813053 )
( 28.379130127483997 , 8.0162945 )
( 20.11760799147358 , -3.7312622 )
( 25.905602548589005 , 4.3870883 )
( 36.322322573517795 , 18.333412 )
( 28.396838916468436 , 7.5238914 )
( 27.119006368860383 , 6.3507776 )
( 31.63814673100157 , 12.032644 )
( 25.809595467561145 , 4.278068 )
( 26.924519799459354 , 6.267301 )
( 19.7661111547486 , -3.1304603 )
( 26.817015549219917 , 5.966183 )
( 18.985133572167484 , -3.9996765 )
( 25.940826982768485 , 4.9514284 )
( 32.78150171784585 , 13.51182 )
( 24.23381984773596 , 2.214879 )
( 28.439927818518832 , 8.146985 )
( 20.872009953460513 , -1.6983593 )
( 31.533280299015757 , 11.917927 )
( 35.3025431308904 , 17.088299 )
( 33.92044462259322 , 14.928519 )
( 31.629171656048126 , 12.393101 )
( 34.3147028812722 , 15.776031 )
( 15.253169459062523 , -8.074551 )
( 20.84222593367418 , -1.7962053 )
( 26.01809483100192 , 4.9893465 )
( 29.658785949391586 , 9.616219 )
( 31.54407881207591 , 11.898298 )
( 26.247047541406967 , 5.3844995 )
( 34.93919984679106 , 16.201965 )
( 20.394094691434702 , -2.1523893 )
( 37.61950630539822 , 20.401949 )
( 30.222563758235854 , 10.235523 )
( 40.01646997994432 , 23.545536 )
( 35.624122738419764 , 17.447666 )
( 33.054719474782864 , 14.128732 )
( 33.74268247459917 , 15.102195 )
( 24.000030726751522 , 2.2203128 )
( 28.503243822112744 , 8.256359 )
( 21.457952488648026 , -1.0365299 )
( 27.414315684271777 , 6.784501 )
( 40.438088443679085 , 23.995811 )
( 30.876957317491957 , 11.074578 )
( 24.87464599107572 , 3.02969 )
( 32.49379172844906 , 13.14517 )
( 20.470475619239274 , -1.6883907 )
( 34.49613358611962 , 16.01823 )
( 27.127399974887524 , 6.346904 )
( 31.547462257027277 , 11.894477 )
( 10.054073135484398 , -14.655942 )
( 37.76407685581014 , 20.656982 )
( 30.917141929929198 , 11.088493 )
( 27.00636169945976 , 5.7944856 )
( 33.20245933176197 , 14.325372 )
( 25.79377126922423 , 4.517403 )
( 18.464565928138676 , -4.104861 )
( 38.57254679049623 , 21.67187 )
( 32.014001373460836 , 12.879049 )
( 3.785493766411116 , -29.099161 )
( 33.931346476504956 , 15.3502 )
( 18.184710874633893 , -4.5642347 )
( 30.436239225483174 , 10.614396 )
( 37.16726933596548 , 19.840336 )
( 26.691770627528182 , 5.877848 )
( 34.95126052643242 , 16.601307 )
( 27.87506817041995 , 7.4403152 )
( 35.382103578995455 , 17.157795 )
( 21.313409255775614 , -1.16766 )
( 25.02269453893141 , 3.8166494 )
( 37.2641846351513 , 19.973164 )
( 35.66138516037682 , 17.468649 )
( 35.23561420697994 , 16.972996 )
( 31.38066780305833 , 12.067758 )
( 34.18725747677574 , 15.247845 )
( 32.7924619551558 , 13.870586 )
( 33.81881503564231 , 15.145624 )
( 32.32944887484828 , 13.218116 )
( 26.486352337815912 , 5.71974 )
( 41.44380500559915 , 25.281385 )
( 37.91102848536405 , 20.846598 )
( 27.60249213824098 , 7.0605087 )
( 34.72116527024053 , 16.24768 )
( 35.002000577095565 , 16.72198 )
( 30.12426795694947 , 10.143778 )
( 23.954721417130173 , 1.9174817 )
( 26.193440207055975 , 5.106702 )
( 32.517031002639634 , 13.4606905 )
( 35.13899720304801 , 16.796484 )
( 38.78681550316604 , 21.966984 )
( 37.33698395987212 , 20.102755 )
( 36.60367859298953 , 18.681782 )
( 26.939506914932657 , 6.206756 )
( 34.137418606843255 , 15.567921 )
( 38.99962037975017 , 22.226992 )
( 25.24996876352917 , 4.0584354 )
( 29.242263800738268 , 9.175532 )
( 38.76921986615886 , 21.924797 )
( 41.06056018344069 , 24.827509 )
( 30.98505391184146 , 11.224398 )
( 32.505900956152516 , 13.480915 )
( 37.216080922128526 , 19.448221 )
( 32.468794399869225 , 13.406689 )
( 33.87875778277504 , 15.216255 )
( 23.722008723169417 , 2.0320878 )
( 28.86481929092303 , 8.639085 )
( 31.23537106647529 , 11.480232 )
( 32.638404309490745 , 13.278749 )
( 35.66348320691342 , 17.517483 )
( 21.23296673332983 , -1.2952816 )
( 25.527179897439726 , 4.3685913 )
( 38.01551031403763 , 20.952517 )
( 36.297541440527795 , 18.320395 )
( 28.156063350895103 , 7.6697335 )
( 14.023566713520452 , -9.697562 )
( 18.343403073175047 , -4.707278 )
( 30.40062480441057 , 10.839153 )
( 24.63038441926774 , 3.1315417 )
( 20.534997029175578 , -2.1714172 )
( 34.32164218938149 , 15.762848 )
( 32.51744432029379 , 13.164131 )
( 32.66056854593617 , 13.382364 )
( 34.23664201329531 , 15.705927 )
( 22.67452034910909 , 1.1239134 )
( 20.932906983547603 , -1.6234739 )
( 39.51943551708676 , 22.890585 )
( 32.78819157127425 , 13.840374 )
( 39.743877828333694 , 23.14526 )
( 24.641080469848713 , 2.7742877 )
( 31.73103461292598 , 12.18073 )
( 26.03749537597519 , 5.004367 )
( 22.05016171563699 , -0.22068036 )
( 39.06647910841185 , 22.282393 )
( 32.975441454919164 , 14.126984 )
( 27.249485571742333 , 6.532351 )
( 33.95452230987607 , 15.378736 )
( 16.91171570851234 , -6.068693 )
( 41.015761581507036 , 24.738037 )
( 11.662925873980623 , -13.09815 )
( 30.81020748305378 , 10.96779 )
( 30.43252808850635 , 10.87694 )
( 33.38801627782733 , 14.650936 )
( 22.930651143635004 , 0.855538 )
( 27.947593233061863 , 6.9962173 )
( 38.400597989543854 , 21.412983 )
( 34.90764042973128 , 16.590786 )
( 34.91580464822609 , 16.181963 )
( 33.08640609095581 , 14.209862 )
( 33.91688687672751 , 15.324049 )
( 31.32264194559882 , 11.71004 )
( 20.291617811355636 , -2.2757006 )
( 33.12717104206742 , 14.294681 )
( 35.59127253446876 , 17.45543 )
( 34.88326300727184 , 16.573362 )
( 23.38340906703761 , 1.599102 )
( 34.72518261051753 , 16.34139 )
( 33.993957178471675 , 15.41927 )
( 30.187895881994507 , 10.286877 )
( 41.10249416101675 , 24.86242 )
( 8.778555407727653 , -15.12203 )
( 33.84095823536287 , 15.127997 )
( 33.11021512034317 , 14.203836 )
( 33.2843822431534 , 14.528918 )
( 36.45387666153501 , 18.467957 )
( 29.646079382596998 , 9.570494 )
( 40.03785660610753 , 23.55762 )
( 36.027761594385304 , 17.954147 )
( 32.12613997999067 , 13.004921 )
( 23.709276187918075 , 1.9339352 )
( 26.853000146280177 , 5.57627 )
( 26.412193758884065 , 5.479637 )
( 30.70678533760118 , 10.815502 )
( 34.05778780843064 , 15.490712 )
( 36.852856752021594 , 19.443079 )
( 27.395004717160077 , 6.276024 )
( 35.6922034085208 , 17.495754 )
( 35.747622096811995 , 18.079845 )
( 33.0992506171165 , 14.252942 )
( 21.70994413115192 , -0.52952445 )
( 41.06221016819851 , 24.814348 )
( 34.17793002020716 , 15.627944 )
( 34.620424378045236 , 16.22466 )
( 33.17748340765239 , 14.373338 )
( 11.116016965508706 , -12.574562 )
( 27.75141271468486 , 7.1565886 )
( 31.882327968439355 , 12.693302 )
( 24.816091781999244 , 3.5302463 )
( 27.056361282025996 , 5.8619266 )
( 27.490355073441712 , 6.426389 )
( 31.55196658946628 , 12.27287 )
( 33.764558052662835 , 14.746913 )
( 36.56435948021629 , 18.636576 )
( 35.42391821408809 , 17.162626 )
( 34.16500752153817 , 15.5842705 )
( 36.324476318053314 , 18.347574 )
( 36.64101254071015 , 19.198662 )
( 35.23306638300326 , 16.935812 )
( 23.380148879898144 , 1.5693768 )
( 40.6203293135466 , 24.223019 )
( 25.394341446189706 , 3.6787927 )
( 21.963136740633857 , -0.26181048 )
( 26.552195797174665 , 5.6577125 )
( 36.4027187063488 , 18.908772 )
( 31.074953945807547 , 11.399899 )
( 42.090881228688225 , 26.101488 )
( 33.833496324130905 , 15.218109 )
( 16.18054899890647 , -8.70938 )
( 31.86626833313723 , 12.661075 )
( 25.21847873417681 , 3.8849387 )
( 26.707908603188653 , 5.443616 )
( 27.581652591268 , 7.0332794 )
( 33.97483453287975 , 15.373999 )
( 26.38778459686693 , 4.9370875 )
( 25.356069592761735 , 4.2452927 )
( 35.44529957595143 , 17.206932 )
( 24.90770979039573 , 3.4731398 )
( 31.794932241823243 , 12.602498 )
( 37.18169305611916 , 19.89177 )
( 40.227197761586595 , 23.723179 )
( 25.785263098392637 , 4.2322817 )
( 36.77035549473477 , 18.87218 )
( 26.93983763859024 , 6.2985435 )
( 20.39471120565266 , -2.1983113 )
( 32.95312073790831 , 13.665943 )
( 36.96999981803036 , 19.624435 )
( 41.038782537324245 , 24.804531 )
( 39.50064085215665 , 22.792736 )
( 28.550546484030757 , 8.22161 )
( 39.00652438451735 , 22.178736 )
( 38.69369711905328 , 21.830273 )
( 33.899701520576166 , 15.286669 )
( 35.045854060365485 , 16.747795 )
( 34.12241550676127 , 15.17228 )
( 19.03986027713293 , -3.5063362 )
( 29.611296001889617 , 9.061553 )
( 18.61513362042907 , -4.4541407 )
( 38.37886562965609 , 21.436878 )
( 35.23758267531952 , 16.962399 )
( 35.055898007894605 , 16.718784 )
( 34.64724875050398 , 16.259468 )
( 28.983283584742495 , 8.730289 )
( 21.277873739885255 , -1.0088215 )
( 33.72512286733978 , 15.006917 )
( 29.58685963120332 , 9.522255 )
( 40.02758363152913 , 23.539484 )
( 34.29140332523713 , 15.816394 )
( 24.090778366845687 , 2.0602458 )
( 25.24908497025491 , 4.0924425 )
( 33.2905644626415 , 14.201027 )
( 16.68278596206827 , -8.140654 )
( 32.710206108631816 , 13.777756 )
( 35.96657565333173 , 17.910849 )
( 32.03198469205631 , 12.873241 )
( 25.268292986926244 , 4.054401 )
( 22.960504413197626 , 0.9661728 )
( 23.289079996527168 , 1.9181933 )
( 38.78386529780876 , 21.970081 )
( -12.112849035657948 , -27.581038 )
( 30.066264625999636 , 10.098958 )
( 29.693218354703312 , 9.571371 )
( 30.734705097410732 , 11.268239 )
( 36.05062035350515 , 18.021124 )
( 31.029214951299863 , 11.300731 )
( 38.820944790526354 , 21.990454 )
( 33.49510208678044 , 14.734558 )
( 29.935254695875358 , 9.868794 )
( 24.85343569402169 , 3.3267186 )
( 31.707629543201897 , 12.459769 )
( 35.53513459798149 , 16.977179 )
( 34.29424751448113 , 15.430689 )
( 25.62589940873335 , 3.9565046 )
( 25.89386967119173 , 4.71998 )
( 15.09950157390881 , -8.764754 )
( 33.70770700693911 , 14.956823 )
( 39.874577480804184 , 23.330444 )
( 42.19981676023478 , 26.234016 )
( 34.277726511016375 , 15.381207 )
( 27.87693991131234 , 7.3922973 )
( 18.632354886523185 , -5.595059 )
( 24.811230186899344 , 2.9426649 )
( 27.17065052000396 , 6.508191 )
( 36.496046982359914 , 19.019463 )
( 34.76586338701323 , 16.010817 )
( 24.06833655850569 , 2.4701536 )
( 19.712911504690155 , -2.9771888 )
( 25.175091663214552 , 3.9376187 )
( 36.863329817429246 , 19.509104 )
( 31.446725767307132 , 11.77281 )
( 29.827274166984154 , 9.831822 )
( 29.062380916701382 , 8.856937 )
( 27.114034297865878 , 5.9184074 )
( 9.790241233307121 , -16.756569 )
( 21.33663410371555 , -1.1089445 )
( 17.29976906796546 , -6.294544 )
( 35.19096029679361 , 16.911343 )
( 29.25505632333622 , 9.134692 )
( 35.09403761041487 , 16.770496 )
( 38.72992286626984 , 21.814146 )
( 33.71219279588793 , 15.012569 )
( 32.03256233161337 , 12.552791 )
( 40.34279449889793 , 23.933628 )
( 38.05235265139269 , 21.026844 )
( 34.52236816460027 , 16.083124 )
( 39.0108650092447 , 22.174208 )
( 35.12831669053984 , 16.844862 )
( 29.719692826002184 , 9.717101 )
( 37.91461000891694 , 20.822336 )
( 37.17888009149164 , 19.419676 )
( 36.72961602366098 , 18.829557 )
( 33.856461017753766 , 15.185248 )
( 33.05205429459912 , 14.171773 )
( 18.936983416423406 , -3.9980922 )
( 31.827171887974245 , 12.617598 )
( 24.72517630859962 , 2.8323147 )
( 34.29043252951146 , 15.745733 )
( 18.865407403033014 , -3.4680405 )
( 20.392798868788965 , -2.3569808 )
( 22.073344913463433 , -0.24796924 )
( 36.54409662149129 , 18.622805 )
( 19.129295548669788 , -3.1834004 )
( 39.12125529228621 , 22.382664 )
( 22.87301297092151 , 0.80923665 )
( 31.788967800730145 , 12.277896 )
( 28.997109424188785 , 8.8572035 )
( 23.012535037745977 , 0.9664871 )
( 34.77203416373149 , 16.013672 )
( 31.45388367610416 , 11.780547 )
( 38.03987891370384 , 20.973425 )
( 25.40993310138083 , 4.0411987 )
( 38.30777788079774 , 21.351042 )
( 39.431566108116016 , 22.80006 )
( 24.49285990038878 , 3.0080383 )
( 24.402961181745944 , 2.8125322 )
( 32.3718896591699 , 13.364315 )
( 27.77087158021384 , 7.1548586 )
( 39.250779942040566 , 22.512173 )
( 33.70393231732394 , 14.968452 )
( 30.651579647220785 , 10.855015 )
( 38.637146778290095 , 21.77867 )
( 32.68587820878182 , 13.684662 )
( 33.02317179390761 , 14.156546 )
( 32.1371507176413 , 13.055252 )
( 33.72111953346118 , 14.739368 )
( 41.373186249322266 , 25.205927 )
( 34.595545299305186 , 16.111557 )
( 18.91989571636589 , -4.0358143 )
( 31.66122946541526 , 12.059294 )
( 7.5595321395558015 , -17.872751 )
( 40.12002774838881 , 23.647158 )
( 30.806631490014492 , 11.04221 )
( 36.3117856533565 , 18.289396 )
( 33.61201984587737 , 14.907201 )
( 29.01051573598371 , 8.844828 )
( 5.619950811670716 , -20.920118 )
( 34.397958227730136 , 15.929481 )
( 40.37020908981519 , 23.899494 )
( 34.06780427706178 , 15.472953 )
( 32.523076628595874 , 13.192617 )
( 35.576027104197586 , 17.369453 )
( 40.45422394987088 , 24.07433 )
( 22.875551743583358 , 0.861111 )
( 30.059867496793174 , 10.136938 )
( 38.67766022133556 , 21.721016 )
( 27.703603648988032 , 7.2481194 )
( 21.605316558822523 , -0.5954139 )
( 32.87532844222325 , 13.607996 )
( 33.70195836038554 , 15.027676 )
( 34.951221169055316 , 16.221046 )
( 37.35553154475986 , 19.62809 )
( 38.12781448951637 , 21.069193 )
( 32.48850435107181 , 13.062942 )
( 31.09463613727953 , 11.332045 )
( 35.31203908698735 , 16.695375 )
( 35.19847005667082 , 16.919361 )
( 21.570928684476677 , -0.80937725 )
( 21.409298886240542 , -0.96793205 )
( 28.25602286425895 , 7.77792 )
( 27.66857419368931 , 7.02886 )
( 30.260794232402557 , 10.266836 )
( 33.41378030785581 , 14.677933 )
( 32.90444217846883 , 14.012766 )
( 34.52663781985207 , 16.116386 )
( 34.46325072790401 , 15.603689 )
( 21.46052288965662 , -1.0041457 )
( 27.55353176997407 , 6.4998045 )
( 26.972927200508927 , 6.232221 )
( 23.668705860541994 , 1.9256041 )
( 38.46116690067567 , 21.520157 )
( 38.40343416312591 , 21.447632 )
( 35.27634142960076 , 16.64745 )
( 33.93325359374165 , 15.284994 )
( 20.052243596811557 , -3.8269367 )
( 19.637116435607012 , -3.0497303 )
( 35.81721350899269 , 17.714354 )
( 32.05554002245933 , 12.582178 )
( 34.42073986631071 , 15.556423 )
( 17.683974941346463 , -6.779497 )
( 39.090336959028775 , 22.271406 )
( 33.589296878596365 , 14.864766 )
( 34.60266545449186 , 16.103043 )
( 38.76015746305716 , 21.848946 )
( 34.12150215764184 , 15.491532 )
( 33.85235050799966 , 15.160934 )
( 16.353123684369283 , -6.748151 )
( 34.810524121345566 , 16.48769 )
( 32.62310981192874 , 13.286335 )
( 34.20701123173998 , 15.632674 )
( 34.46267298360046 , 15.962394 )
( 33.55671592117271 , 14.855759 )
( 34.67202284890875 , 16.267818 )
( 34.77344680418227 , 16.362473 )
( 33.24660697333295 , 14.362188 )
( 19.097748782692204 , -3.202423 )
( 21.17342298622756 , -1.3465362 )
( 23.22442499180483 , 1.849119 )
( 35.472883662478274 , 17.221601 )
( 22.668412083858776 , 0.61949265 )
( 33.03348379616022 , 14.097537 )
( 24.709781937120937 , 3.154826 )
( 38.07099307926448 , 20.990936 )
( 28.904173360967782 , 8.709232 )
( 32.557569734630434 , 13.176985 )
( 29.10540202563562 , 9.010463 )
( 31.048799173398823 , 11.673742 )
( 35.82995943736652 , 17.703442 )
( 41.03922124767695 , 24.774736 )
( 34.61310546669014 , 15.843943 )
( 31.950515602031878 , 12.444868 )
( 32.05791896825756 , 12.922755 )
( 32.80253741477757 , 13.856351 )
( 33.51213960260061 , 14.424445 )
( 29.03875903653404 , 8.89105 )
( 29.874319117103738 , 9.914978 )
( 37.749180529885564 , 20.557951 )
( 22.930399692761174 , 0.9325872 )
( 22.035605023921896 , -0.20503479 )
( 19.397424326091098 , -2.9952154 )
( 30.42372970296535 , 10.522627 )
( 32.90840461354012 , 13.966955 )
( 22.829626223772703 , 0.7910034 )
( 18.247937997414564 , -6.0725584 )
( 34.43392472084263 , 15.597525 )
( 39.953804243277226 , 23.464546 )
( 32.96115926944776 , 14.067265 )
( 35.37427362194359 , 17.091572 )
( 17.254055826993298 , -5.650796 )
( 19.183541900657833 , -3.2788935 )
( 16.816588992557357 , -6.117243 )
( 32.04258226146042 , 12.926975 )
( 19.972725394025634 , -2.2756863 )
( 24.261119499512077 , 2.3162167 )
( 32.01142236919534 , 12.824089 )
( 33.31832460516379 , 14.505676 )
( 26.439419160323098 , 5.636819 )
( 40.41043175585792 , 23.963066 )
( 36.49937231188628 , 19.02627 )
( 26.449593471675154 , 5.5563807 )
( 39.97771727616839 , 23.437366 )
( 35.04497584743132 , 16.703535 )
( 38.63240365336764 , 21.742184 )
( 34.05629392712132 , 15.434068 )
( 34.63413554853643 , 16.166485 )
( 33.288138570356004 , 14.510377 )
( 29.453576238329525 , 9.431999 )
( 33.84575149298101 , 15.140809 )
( 31.980574869152985 , 12.42098 )
( 38.14155588986045 , 21.06732 )
( 33.79161511190438 , 14.795253 )
( 31.634330457085394 , 11.975216 )
( 40.76854708700893 , 24.40339 )
( 16.11527894582526 , -7.560315 )
( 33.4473916947134 , 14.298393 )
( 32.000871776647436 , 12.8474865 )
( 40.42868279154715 , 24.011318 )
( 19.811270398907077 , -2.948512 )
( 20.450441889854183 , -1.7293882 )
( 25.84846610016251 , 4.887692 )
( 31.777209066276846 , 12.2126465 )
( 33.27321039867952 , 14.451481 )
( 20.262419305308416 , -1.8852648 )
( 32.64850967252891 , 13.623304 )
( 33.29791567461419 , 14.511081 )
( 32.60056077307632 , 13.627555 )
( 35.12038440950698 , 16.77987 )
( 33.536887963979055 , 14.466573 )
( 36.20766819371841 , 18.20156 )
( 8.624306219404923 , -18.190084 )
( 40.29453145452019 , 23.865221 )
( 24.38699446985504 , 2.4673712 )
( 21.884173750875917 , -0.39415175 )
( 38.66593023341042 , 21.72368 )
( 36.08058482078245 , 18.033548 )
( 39.04843787112041 , 22.232697 )
( 31.385560289115457 , 11.785784 )
( 33.84407023758092 , 15.177071 )
( 30.6913163453293 , 11.195489 )
( 35.00889308406855 , 16.72552 )
( 17.08630269963148 , -5.78218 )
( 33.27263567002342 , 14.433734 )
( 33.19202611105419 , 14.027939 )
( 34.130896033279434 , 15.178604 )
( 25.046113697930473 , 3.6570263 )
( 40.474762101582286 , 24.046646 )
( 35.53370370084878 , 17.372799 )
( 33.49703094187659 , 14.422069 )
( 29.384883522199015 , 9.313782 )
( 34.640887523164615 , 16.178026 )
( 32.08656005235964 , 12.928395 )
( 21.83613664646608 , -0.34032747 )
( 37.658044110241306 , 20.441166 )
( 24.407105542830656 , 2.8516731 )
( 33.39869197113705 , 14.552465 )
( 27.631782827215034 , 7.176433 )
( 7.236842689027008 , -17.133633 )
( 33.61977979776889 , 14.579843 )
( 37.178974583343475 , 19.865286 )
( 32.805893635116476 , 13.879799 )
( 41.74687900957899 , 25.680134 )
( 39.23380775937681 , 22.46381 )
( 32.885423063505435 , 13.623598 )
( 34.15492268957931 , 15.646084 )
( 24.177781227718366 , 2.5217972 )
( 28.584779119746848 , 7.7510777 )
( 33.68436179570615 , 14.62806 )
( 35.97100456351283 , 17.886139 )
( 18.96023875303978 , -3.9546902 )
( 30.117272464382832 , 10.136066 )
( 26.328362335442783 , 4.93441 )
( 30.423235310332004 , 10.524884 )
( 31.32995545635916 , 12.011999 )
( 34.71946610965902 , 16.270061 )
( 36.120131332239254 , 18.067324 )
( 26.800570083312362 , 5.557452 )
( 33.53362016505358 , 14.776062 )
( 40.39148438352227 , 23.945751 )
( 39.953529688576964 , 23.4118 )
( 32.23406937506621 , 13.161731 )
( 38.1492396936721 , 21.085772 )
( 40.50916837338535 , 24.142944 )
( 28.906108160749824 , 8.661879 )
( 37.160089095219064 , 19.389893 )
( 33.08308223955535 , 13.883112 )
( 26.223216885150407 , 5.3242025 )
( 21.563568913658663 , -0.75369555 )
( 39.6492019373186 , 23.041573 )
( 32.04590339502899 , 12.880468 )
( 18.26574769086919 , -5.1180816 )
( 16.899387804164594 , -7.843302 )
( 39.014403448518266 , 22.198362 )
( 31.929934537167206 , 12.731707 )
( 29.987032387170842 , 10.064414 )
( 27.78899485128985 , 6.7741036 )
( 32.9261189488422 , 13.679375 )
( 27.412719480200042 , 6.3233857 )
( 41.58865278719463 , 25.463646 )
( 19.51756847428361 , -4.486885 )
( 32.0932349160175 , 12.98754 )
( 40.2935179069426 , 23.825172 )
( 34.03855309627093 , 15.47171 )
( 28.747841461227285 , 8.508449 )
( 31.1777221057431 , 11.435336 )
( 35.62626905896162 , 17.451225 )
( 24.321149569050405 , 2.6254876 )
( 26.044368218485474 , 5.1515403 )
( 32.215699413283765 , 12.772246 )
( 18.607913280773847 , -4.3860626 )
( 35.2705941155071 , 17.025873 )
( 35.94474934304075 , 17.853151 )
( 40.4445477296077 , 23.992765 )
( 22.75927799930691 , 1.2640486 )
( 28.124622557908665 , 7.661909 )
( 24.699518007158154 , 3.1278517 )
( 32.574925052318505 , 13.225368 )
( 31.6968317399523 , 12.498531 )
( 40.17229189977225 , 23.6432 )
( 34.50073747465127 , 15.986164 )
( 33.58526202331896 , 14.906505 )
( 35.18936451119475 , 16.901571 )
( 25.380452083813687 , 3.6561027 )
( 32.89621667942565 , 13.930591 )
( 30.10076944966183 , 10.077072 )
( 23.458393873997657 , 1.2641838 )
( 34.47185105313723 , 15.983851 )
( 26.978406817448466 , 5.787288 )
( 40.193612069910166 , 23.770164 )
( 25.993084642688476 , 4.445417 )
( 27.94409323434084 , 6.9717197 )
( 31.627913519731923 , 12.347886 )
( 28.817201114821447 , 8.552917 )
( 17.833269046889306 , -6.646805 )
( 33.600457921978546 , 14.919757 )
( 27.39441271606376 , 6.306399 )
( 30.922810103168487 , 11.232298 )
( 28.19748472470411 , 7.8107576 )
( 31.66520415892485 , 12.436176 )
( 38.08489799526574 , 21.029432 )
( 38.0703044394377 , 20.97312 )
( 27.242664174945915 , 6.50594 )
( 32.09156069419752 , 12.962329 )
( 38.99319121544138 , 22.14635 )
( 32.7246893912801 , 13.381701 )
( 35.723372462541775 , 17.589773 )
( 32.455243103031876 , 13.427404 )
( 36.90856651694638 , 19.514029 )
( 32.406659765592735 , 13.050034 )
( 41.42472656216824 , 25.276703 )
( 29.342501257468694 , 9.257862 )
( 39.54758862625337 , 22.92636 )
( 35.49255278356418 , 17.284332 )
( 31.238806102051253 , 11.614393 )
( 30.616716397104064 , 10.748316 )
( 31.872642714049825 , 12.677643 )
( 35.25182404254292 , 16.960722 )
( 26.684977672863695 , 5.876935 )
( 29.442594179201034 , 9.391215 )
( 29.63383532142278 , 9.615611 )
( 33.68094566227282 , 14.932258 )
( 17.120772559663383 , -6.1799984 )
( 34.961746888187164 , 16.637497 )
( 32.22781723914764 , 12.751327 )
( 22.437277723715535 , 0.8458451 )
( 33.40837643694654 , 14.676233 )
( 27.65130889670058 , 6.5799193 )
( 27.531560457911617 , 6.9713306 )
( 34.5670411836849 , 16.145061 )
( 41.74875610565822 , 25.66727 )
( 26.307951435292587 , 5.451754 )
( 31.89146502086846 , 12.392312 )
( 35.73328722236937 , 18.061062 )
( 19.27523933036466 , -3.5432465 )
( 37.8292008691768 , 20.686415 )
( 37.14330271421103 , 19.827364 )
( 36.42304922828001 , 18.452911 )
( 26.153836779974434 , 5.1516256 )
( 34.297764785947955 , 15.737261 )
( 30.263140203069927 , 10.295878 )
( 27.842439401276955 , 6.867753 )
( 26.708466702897375 , 5.411112 )
( 24.504070379347926 , 3.1728792 )
( 32.99294213118022 , 13.74016 )
( 40.70652150422813 , 24.342842 )
( 42.41980809170193 , 26.526392 )
}

\begin{figure}[h]
\centering
\begin{subfigure}{.45\textwidth}
  \centering
  \begin{tikzpicture}[scale=0.6]
  \begin{axis}[xlabel={True utility}, ylabel={Predicted utility}, axis y line=left, axis x line=bottom, xmin = -30, xmax = 40, ymin = -30, ymax = 20]
  \addplot[scatter,only marks] coordinates {\utilityA};
  \addplot[dashed, very thick, samples=100, smooth, domain=-30:40] expression{0.56*\x - 6.99};
  \node at (axis cs:-8,15) {\sf Best fitted line:};
  \node at (axis cs:-8,9.5) {$y = 0.56x - 6.99$};
  \node at (axis cs:-8,5) {$R^2 = 0.87$};
  \end{axis}
 \end{tikzpicture}
  \caption{Customer class 1 versus Sample 1.}
\end{subfigure}
\begin{subfigure}{.45\textwidth}
  \centering
  \begin{tikzpicture}[scale=0.6]
  \begin{axis}[xlabel={True utility}, ylabel={Predicted utility}, axis y line=left, axis x line=bottom, xmin = -30, xmax = 40, ymin = -30, ymax = 20]
  \addplot[scatter,only marks] coordinates {\utilityB};
        \end{axis}
 \end{tikzpicture}
  \caption{Customer class 1 versus Sample 2.}
\end{subfigure}
\begin{subfigure}{.45\textwidth}
  \centering
  \begin{tikzpicture}[scale=0.6]
  \begin{axis}[xlabel={True utility}, ylabel={Predicted utility}, axis y line=left, axis x line=bottom, xmin = -30, xmax = 40, ymin = -30, ymax = 20]
  \addplot[scatter,only marks] coordinates {\utilityC};
        \end{axis}
 \end{tikzpicture}
  \caption{Customer class 2 versus Sample 1.}
\end{subfigure}
\begin{subfigure}{.45\textwidth}
  \centering
  \begin{tikzpicture}[scale=0.6]
  \begin{axis}[xlabel={True utility}, ylabel={Predicted utility}, axis y line=left, axis x line=bottom, xmin = -30, xmax = 40, ymin = -30, ymax = 20]
  \addplot[scatter,only marks] coordinates {\utilityD};
  \addplot[dashed, very thick, samples=100, smooth, domain=-30:40] expression{1.286*\x - 28.39};
  \node at (axis cs:-8,5) {\sf Best fitted line:};
  \node at (axis cs:-8,-0.5) {$y = 1.28x - 28.39$};
  \node at (axis cs:-8,-5) {$R^2 = 0.99$};
        \end{axis}
 \end{tikzpicture}
  \caption{Customer class 2 versus Sample 2.}
\end{subfigure}
\caption{Out-of-sample predicted utility (for each sample) versus true utility (for each customer class) for Setting~3 of Section~5 for RUMnet model with $K=2$ and $(\ell,w)=(0,0)$. Recall that in Setting~3, the ground truth model a latent class MNL with two customer classes.}
\label{fig:utility_3}
\end{figure}

In these experiments, we see that there is a strong correlation between the learned utility and the ground truth utility, indicating that despite the non-identifiability of the model class, our method can, to some extent, recover the underlying utility.

\subsection{Experiment using ranking-based non-contextual data} \label{app:rank}

\subsubsection{Experiment setup.} We generate synthetic choice data where the ground truth is a nonparametric ranking-based choice model. Specifically, the ground truth model is a distribution over 10 ranked lists. Each ranked list in the support is generated as an independent uniform permutation of the products. Additionally, for each ranked lists $i$, we generate an independent random variable $X_i$ in $[0,1]$ and assign to each ranked list the probability $X_i / \sum_{j=1}^{10} X_j$.  For the training set, we generate a sequence of $T=10,000$ customers, each being presented with an assortment of 5 products chosen uniformly at random from a universe of 10 products. 

Note that in this case, there are no product or customer features, and the input vector is simply, for each product, an indicator vector $\boldsymbol{x} \in \{0,1\}^{10}$. We compare the performance of various RUMnet models with a ranking-based model, which we estimate using the code of~\cite{berbeglia2022comparative}. In particular, we experiment with $(\ell,w) \in \{(0,0),(1,3)\}$ and $K \in \{2,5,10,20\}$. To assess the model's performance, we generate an additional $1,000$ customers as a test set to compute an out-of-sample log-likelihood loss and accuracy. 

\subsubsection{Results.} Table~\ref{tab:rankings} shows the average performance metrics over 10 repetitions of the experiment.
\begin{table}[t]\centering
\caption{Average out-of-sample log-likelihood loss and accuracy over 10 experiments using a ranking-based ranking-based model. We report the performance of ranking-based model estimated using an EM algorithm and a direct maximum likelihood optimization algorithm~\cite{berbeglia2022comparative}.} \label{tab:rankings}
\setlength\extrarowheight{2pt} 
\resizebox{15cm}{!}{
\begin{tabular}{ccccccccc}
  \multicolumn{3}{c}{Model}  & \multicolumn{2}{c}{Train} & \multicolumn{2}{c}{Test}     \\
\cmidrule[1pt](lr){1-3} \cmidrule[1pt](lr){4-5}  \cmidrule[1pt](lr){6-7} 
 Type & $(\ell,w)$ & $K$  &  Loss & Accuracy & Loss & Accuracy & Time to fit (s) & Number of parameters \\
\cmidrule[1pt](lr){1-9}
 MNL & - & -  & 1.4760 & 0.3570 & 1.4808 & 0.354 & 8.9 & 10\\
\cmidrule(lr){1-9} 
 \multirow{8}{*}{RUMnet} 
& (0,0) & 2  & 1.4552 & 0.3601 & 1.4591 & 0.3611 & 11.0 & 125\\
 & (1,3) & 2 & 1.4529 & 0.3604 & 1.4572 & 0.3575 & 13.8 & 157\\
 \cmidrule(lr){2-9} 
  &  (0,0) & 5  & 1.4037 & 0.3841 & 1.4076 & 0.3844 & 24.5 & 290 \\
 & (1,3) & 5   & 1.4026 & 0.3836 & 1.4088 & 0.3841 & 23.3 & 316\\
 \cmidrule[1pt](lr){2-9}
 &  (0,0) & 10  & 1.3796 & 0.3883 & 1.3895 & 0.3858 & 35.6 & 565\\
 & (1,3) & 10   & 1.3801 & 0.3876 & 1.3883 & 0.3835 & 43.5 & 581 \\
 \cmidrule[1pt](lr){2-9}
 &  (0,0) & 20  & 1.3738 & 0.3919 & 1.3840 & 0.3867 & 60.5 & 1,115\\
 & (1,3) & 20   & 1.3758 & 0.3913 & 1.3862 & 0.3859 & 61.9 & 1,111\\
 \cmidrule[1pt](lr){1-9}
 \multicolumn{2}{c}{\multirow{4}{*}{Latent MNL}} & 2 & 1.4525 & 0.3602 & 1.4571 & 0.3601 & 38.9 & 22\\\
  & & 5 & 1.4035 & 0.3823 & 1.4093 & 0.3821 & 99.5 & 55\\
  & & 10 & 1.3834 & 0.3875 & 1.3916 & 0.3842 & 127.4 & 110\\
  & & 20 & 1.3784 & 0.3894 & 1.3866 & 0.3846 & 184.5 & 220\\
 \cmidrule[1pt](lr){1-9}
 \multicolumn{3}{c}{Ranked list (EM)} &  1.3785 & 0.3886	 &	1.4628 &	0.3818 & 1,802.4	 & 486	\\
 \multicolumn{3}{c}{Ranked list (Max)} &  1.3926 &	0.3811 &	1.4029 &	0.3741 & 1,851.1	 &	367\\
 \cmidrule[1pt](lr){1-9}
 \multicolumn{3}{c}{Ground Truth} & 1.3706 &	0.3914 &	1.3743 &	0.3868 &	- &	100\\
 \cmidrule[1pt](lr){1-9} 
\end{tabular}
}
\end{table} 
We find that the fitted RUMnets attain an out-of-sample performance that approaches that of the ground truth model. Interestingly, the complexity parameters $(\ell,w)$ of the neural network building blocks do not affect  predictive performance, i.e., increasing the non-linearity of the utility function does not help achieve better accuracy.\footnote{It may seem counter-intuitive that, for $K=20$, increasing the complexity parameters $(\ell,w)$ from $(0,0)$ to $(1,3)$ actually decreases the number of parameters.  When $(\ell,w)=(0,0)$, the number of parameters is equal to $55\cdot K + 15$. On the other hand, when $(\ell,w)=(1,3)$, the number of parameters is equal to $53\cdot K + 48$. This is due to the convention we follow: we hard code the size of the unobserved vector to 5.} 
This is expected, as there are no features to leverage. By contrast, we yet again observe the importance of adding latent heterogeneity in the model as the performance of the RUMnet models increases with $K$ (i.e., we can interpret each realized ``sample'' of the RUMnet architecture as analogous to a ranking sampling from the nonparametric distribution).

We now turn our attention to the ranking-based models. In terms of performance, we see that the training losses for both the EM algorithm and the MLE approach are higher than those achieved by RUMnets. There is also a greater extent of overfitting, considering the performance gap between the training and test sets. Quoting the authors in~\cite{berbeglia2022comparative}, ``[their numerical] results highlight the need to implement additional methods to reduce overfitting [...] when dealing with a relatively small historical data set''. However, integrating early stopping, or any alternative regularization technique, is not  straightforward, whereas this is just a parameter to pass to the $\textsf{fit}$ method of Keras. 

Finally, there is an important difference in running times between our Keras-based implementation and the mixed-integer programming-based code used to fit ranking-based choice models (see Table~\ref{tab:rankings}).  Note that one could speed up the column generation subproblem step when fitting a ranking-based choice model by using a heuristic instead~\citep[see e.g.,][]{mivsic2016data,chen2022decision}. Our method scales more efficiently for a large number of products, as it leverages highly optimized deep-learning open-source libraries.

\newpage 

\section{Additional material for Section~\ref{sec:numerical}} \label{app:materials}

\subsection{Implementation details} \label{app:implementation}
 We utilize the same Keras implementation and specify similar hyper-parameters for all neural network-based models. We select the standard ADAM optimizer with respect to the categorical cross-entropy loss function. Each dataset is split into training, validation, and testing sets. The validation set is utilized to tune the hyper-parameters on each split.  Since our real-world datasets differ in size, we choose slightly different sets of hyper-parameters for each dataset, as detailed in Table~\ref{tab:param}.
 
 \begin{table}[H]\centering
\caption{Training parameters}\label{tab:param}
\setlength\extrarowheight{2pt} 
\resizebox{8cm}{!}{
\begin{tabular}{ccccc}
 Parameter & Swissmetro & Expedia & Synthetic & Synthetic (4) \\
 \cmidrule[1pt](lr){1-1} \cmidrule[1pt](lr){2-2}  \cmidrule[1pt](lr){3-3} \cmidrule[1pt](lr){4-4} \cmidrule[1pt](lr){5-5}
 Activation function & ELU & ELU & ELU & ELU\\
 Number of epochs & 1,000 & 100 & 100 & 1000/500\\
Early stopping & 200 & 10 & 10 & 1000/500\\
Regularization & None & None & None & None\\
Batch size & 32 & 32 & 32 & 32\\
Learning rate & 0.001 & 0.001 & 0.001 & 0.001\\
Label smoothing & 0.01 & 0.0001 & 0 & 0 \\
\cmidrule[1pt](lr){1-5}
\end{tabular}
}
\end{table}

Regarding regularization, we implement the label smoothing method to alleviate peaky distributions~\citep{szegedy2016rethinking,muller2019does}. For each model, we introduce a small perturbation of the classification labels that assigns a small probability to the unchosen products. Note that  RFs cannot be trained with custom regularization; however, tuning the number of trees in the forest is a direct alternative to alleviate small choice probabilities. Finally, we use an early stopping criterion: the gradient-descent terminates when the loss on the validation set does not improve for more than a given number of epochs. The weights from the epoch with the best validation loss are then restored. A typical profile of the loss function during the training phase is shown in Figure~\ref{fig:sample_train}.

\newcommand{\trainLoss}{
( 0 , 0.88503 )
( 1 , 0.74039 )
( 2 , 0.69590 )
( 3 , 0.67219 )
( 4 , 0.66132 )
( 5 , 0.65508 )
( 6 , 0.64739 )
( 7 , 0.64309 )
( 8 , 0.63857 )
( 9 , 0.63600 )
( 10 , 0.63150 )
( 11 , 0.62658 )
( 12 , 0.62385 )
( 13 , 0.62181 )
( 14 , 0.61832 )
( 15 , 0.61362 )
( 16 , 0.61168 )
( 17 , 0.60765 )
( 18 , 0.60493 )
( 19 , 0.60095 )
( 20 , 0.59710 )
( 21 , 0.59384 )
( 22 , 0.59265 )
( 23 , 0.58877 )
( 24 , 0.58575 )
( 25 , 0.58100 )
( 26 , 0.58033 )
( 27 , 0.57718 )
( 28 , 0.57371 )
( 29 , 0.57228 )
( 30 , 0.56929 )
( 31 , 0.56543 )
( 32 , 0.56362 )
( 33 , 0.56183 )
( 34 , 0.55930 )
( 35 , 0.55798 )
( 36 , 0.55494 )
( 37 , 0.55451 )
( 38 , 0.55163 )
( 39 , 0.54867 )
( 40 , 0.54742 )
( 41 , 0.54806 )
( 42 , 0.54416 )
( 43 , 0.54330 )
( 44 , 0.54084 )
( 45 , 0.54118 )
( 46 , 0.53894 )
( 47 , 0.53703 )
( 48 , 0.53624 )
( 49 , 0.53257 )
( 50 , 0.53411 )
( 51 , 0.53254 )
( 52 , 0.52988 )
( 53 , 0.53022 )
( 54 , 0.53032 )
( 55 , 0.52627 )
( 56 , 0.52575 )
( 57 , 0.52559 )
( 58 , 0.52339 )
( 59 , 0.52379 )
( 60 , 0.52372 )
( 61 , 0.52073 )
( 62 , 0.52093 )
( 63 , 0.52032 )
( 64 , 0.51772 )
( 65 , 0.51914 )
( 66 , 0.52030 )
( 67 , 0.51490 )
( 68 , 0.51379 )
( 69 , 0.51332 )
( 70 , 0.51315 )
( 71 , 0.51381 )
( 72 , 0.51193 )
( 73 , 0.51032 )
( 74 , 0.51198 )
( 75 , 0.50772 )
( 76 , 0.51032 )
( 77 , 0.51041 )
( 78 , 0.50725 )
( 79 , 0.50862 )
( 80 , 0.50660 )
( 81 , 0.50655 )
( 82 , 0.50800 )
( 83 , 0.50324 )
( 84 , 0.50423 )
( 85 , 0.50165 )
( 86 , 0.50204 )
( 87 , 0.50095 )
( 88 , 0.50220 )
( 89 , 0.49982 )
( 90 , 0.49994 )
( 91 , 0.50012 )
( 92 , 0.49913 )
( 93 , 0.49952 )
( 94 , 0.49747 )
( 95 , 0.49750 )
( 96 , 0.49715 )
( 97 , 0.49705 )
( 98 , 0.49638 )
( 99 , 0.49444 )
( 100 , 0.49229 )
( 101 , 0.49263 )
( 102 , 0.49215 )
( 103 , 0.49198 )
( 104 , 0.49161 )
( 105 , 0.48965 )
( 106 , 0.48896 )
( 107 , 0.49166 )
( 108 , 0.48825 )
( 109 , 0.48914 )
( 110 , 0.48972 )
( 111 , 0.48706 )
( 112 , 0.48774 )
( 113 , 0.48892 )
( 114 , 0.48516 )
( 115 , 0.48557 )
( 116 , 0.48581 )
( 117 , 0.48181 )
( 118 , 0.48495 )
( 119 , 0.48401 )
( 120 , 0.48101 )
( 121 , 0.48409 )
( 122 , 0.48429 )
( 123 , 0.48227 )
( 124 , 0.48120 )
( 125 , 0.48201 )
( 126 , 0.48068 )
( 127 , 0.47944 )
( 128 , 0.47929 )
( 129 , 0.47960 )
( 130 , 0.47775 )
( 131 , 0.47931 )
( 132 , 0.47993 )
( 133 , 0.47798 )
( 134 , 0.47732 )
( 135 , 0.47598 )
( 136 , 0.47879 )
( 137 , 0.47443 )
( 138 , 0.47552 )
( 139 , 0.47390 )
( 140 , 0.47448 )
( 141 , 0.47460 )
( 142 , 0.47421 )
( 143 , 0.47137 )
( 144 , 0.47088 )
( 145 , 0.47293 )
( 146 , 0.47487 )
( 147 , 0.47305 )
( 148 , 0.47213 )
( 149 , 0.47323 )
( 150 , 0.47388 )
( 151 , 0.47075 )
( 152 , 0.46971 )
( 153 , 0.47149 )
( 154 , 0.47174 )
( 155 , 0.46709 )
( 156 , 0.46801 )
( 157 , 0.46759 )
( 158 , 0.46767 )
( 159 , 0.47142 )
( 160 , 0.46732 )
( 161 , 0.46611 )
( 162 , 0.46685 )
( 163 , 0.46627 )
( 164 , 0.46714 )
( 165 , 0.46490 )
( 166 , 0.46548 )
( 167 , 0.46417 )
( 168 , 0.46566 )
( 169 , 0.46500 )
( 170 , 0.46822 )
( 171 , 0.46399 )
( 172 , 0.46277 )
( 173 , 0.46270 )
( 174 , 0.46163 )
( 175 , 0.46084 )
( 176 , 0.46518 )
( 177 , 0.46307 )
( 178 , 0.46225 )
( 179 , 0.46159 )
( 180 , 0.46361 )
( 181 , 0.46324 )
( 182 , 0.45939 )
( 183 , 0.46088 )
( 184 , 0.46308 )
( 185 , 0.46019 )
( 186 , 0.45904 )
( 187 , 0.46185 )
( 188 , 0.45669 )
( 189 , 0.45966 )
( 190 , 0.45838 )
( 191 , 0.46059 )
( 192 , 0.45908 )
( 193 , 0.46106 )
( 194 , 0.45844 )
( 195 , 0.45609 )
( 196 , 0.45749 )
( 197 , 0.45775 )
( 198 , 0.45868 )
( 199 , 0.45797 )
( 200 , 0.45707 )
( 201 , 0.45528 )
( 202 , 0.45599 )
( 203 , 0.45550 )
( 204 , 0.45801 )
( 205 , 0.45659 )
( 206 , 0.45641 )
( 207 , 0.45343 )
( 208 , 0.45455 )
( 209 , 0.45486 )
( 210 , 0.45446 )
( 211 , 0.46056 )
( 212 , 0.45502 )
( 213 , 0.45418 )
( 214 , 0.45583 )
( 215 , 0.45405 )
( 216 , 0.45526 )
( 217 , 0.45424 )
( 218 , 0.45172 )
( 219 , 0.45466 )
( 220 , 0.45253 )
( 221 , 0.45380 )
( 222 , 0.45396 )
( 223 , 0.45308 )
( 224 , 0.45201 )
( 225 , 0.45297 )
( 226 , 0.45177 )
( 227 , 0.45186 )
( 228 , 0.44842 )
( 229 , 0.45195 )
( 230 , 0.45116 )
( 231 , 0.45213 )
( 232 , 0.45232 )
( 233 , 0.45004 )
( 234 , 0.44903 )
( 235 , 0.44958 )
( 236 , 0.45036 )
( 237 , 0.45028 )
( 238 , 0.44959 )
( 239 , 0.45204 )
( 240 , 0.45355 )
( 241 , 0.44957 )
( 242 , 0.45207 )
( 243 , 0.44821 )
( 244 , 0.44745 )
( 245 , 0.45256 )
( 246 , 0.45052 )
( 247 , 0.44939 )
( 248 , 0.44774 )
( 249 , 0.44775 )
( 250 , 0.44933 )
( 251 , 0.44744 )
( 252 , 0.45039 )
( 253 , 0.44744 )
( 254 , 0.44733 )
( 255 , 0.44882 )
( 256 , 0.44746 )
( 257 , 0.44787 )
( 258 , 0.44683 )
( 259 , 0.44380 )
( 260 , 0.44716 )
( 261 , 0.44558 )
( 262 , 0.44631 )
( 263 , 0.44424 )
( 264 , 0.44516 )
( 265 , 0.44665 )
( 266 , 0.44420 )
( 267 , 0.44407 )
( 268 , 0.44388 )
( 269 , 0.44278 )
( 270 , 0.44613 )
( 271 , 0.44734 )
( 272 , 0.44287 )
( 273 , 0.44445 )
( 274 , 0.44627 )
( 275 , 0.44361 )
( 276 , 0.44653 )
( 277 , 0.44389 )
( 278 , 0.44328 )
( 279 , 0.44156 )
( 280 , 0.44406 )
( 281 , 0.44024 )
( 282 , 0.44386 )
( 283 , 0.44346 )
( 284 , 0.44108 )
( 285 , 0.44207 )
( 286 , 0.44174 )
( 287 , 0.44250 )
( 288 , 0.44194 )
( 289 , 0.44131 )
( 290 , 0.44123 )
( 291 , 0.44240 )
( 292 , 0.44181 )
( 293 , 0.44229 )
( 294 , 0.44023 )
( 295 , 0.44268 )
( 296 , 0.44082 )
( 297 , 0.43937 )
( 298 , 0.44135 )
( 299 , 0.43924 )
( 300 , 0.44134 )
( 301 , 0.44049 )
( 302 , 0.44239 )
( 303 , 0.44146 )
( 304 , 0.43972 )
( 305 , 0.43948 )
( 306 , 0.44007 )
( 307 , 0.43984 )
( 308 , 0.44015 )
( 309 , 0.44083 )
( 310 , 0.43981 )
( 311 , 0.43900 )
( 312 , 0.44016 )
( 313 , 0.43711 )
( 314 , 0.44273 )
( 315 , 0.43768 )
( 316 , 0.43826 )
( 317 , 0.43973 )
( 318 , 0.44004 )
( 319 , 0.44046 )
( 320 , 0.43665 )
( 321 , 0.43935 )
( 322 , 0.43770 )
( 323 , 0.43816 )
( 324 , 0.43889 )
( 325 , 0.43877 )
( 326 , 0.43764 )
( 327 , 0.43618 )
( 328 , 0.43758 )
( 329 , 0.43614 )
( 330 , 0.43590 )
( 331 , 0.44149 )
( 332 , 0.43707 )
( 333 , 0.43716 )
( 334 , 0.43593 )
( 335 , 0.43580 )
( 336 , 0.43726 )
( 337 , 0.43642 )
( 338 , 0.43590 )
( 339 , 0.43370 )
( 340 , 0.43662 )
( 341 , 0.43439 )
( 342 , 0.43689 )
( 343 , 0.43959 )
( 344 , 0.43439 )
( 345 , 0.43554 )
( 346 , 0.43567 )
( 347 , 0.43525 )
( 348 , 0.43418 )
( 349 , 0.43519 )
( 350 , 0.43663 )
( 351 , 0.43409 )
( 352 , 0.43708 )
( 353 , 0.43642 )
( 354 , 0.43577 )
( 355 , 0.43572 )
( 356 , 0.43652 )
( 357 , 0.43545 )
( 358 , 0.43402 )
( 359 , 0.43454 )
( 360 , 0.43418 )
( 361 , 0.43298 )
( 362 , 0.43430 )
( 363 , 0.43358 )
( 364 , 0.43560 )
( 365 , 0.43354 )
( 366 , 0.43331 )
( 367 , 0.43463 )
( 368 , 0.43159 )
( 369 , 0.43304 )
( 370 , 0.43237 )
( 371 , 0.43395 )
( 372 , 0.43577 )
( 373 , 0.43204 )
( 374 , 0.43303 )
( 375 , 0.43296 )
( 376 , 0.43209 )
( 377 , 0.43147 )
( 378 , 0.43074 )
( 379 , 0.43385 )
( 380 , 0.43110 )
( 381 , 0.43303 )
( 382 , 0.43141 )
( 383 , 0.43030 )
( 384 , 0.42982 )
( 385 , 0.43429 )
( 386 , 0.43364 )
( 387 , 0.43013 )
( 388 , 0.43170 )
( 389 , 0.43164 )
( 390 , 0.43256 )
( 391 , 0.43099 )
( 392 , 0.43261 )
( 393 , 0.43165 )
( 394 , 0.42950 )
( 395 , 0.43272 )
( 396 , 0.42960 )
( 397 , 0.43072 )
( 398 , 0.43441 )
( 399 , 0.43266 )
( 400 , 0.43017 )
( 401 , 0.43142 )
( 402 , 0.43050 )
( 403 , 0.43269 )
( 404 , 0.43174 )
( 405 , 0.43307 )
( 406 , 0.43157 )
( 407 , 0.42809 )
( 408 , 0.42998 )
( 409 , 0.43339 )
( 410 , 0.42883 )
( 411 , 0.43150 )
( 412 , 0.43151 )
( 413 , 0.43106 )
( 414 , 0.42722 )
( 415 , 0.42840 )
( 416 , 0.42985 )
( 417 , 0.42783 )
( 418 , 0.43107 )
( 419 , 0.42894 )
( 420 , 0.42815 )
( 421 , 0.43115 )
( 422 , 0.42715 )
( 423 , 0.42911 )
( 424 , 0.42839 )
( 425 , 0.42935 )
( 426 , 0.43132 )
( 427 , 0.42869 )
( 428 , 0.43068 )
( 429 , 0.42806 )
( 430 , 0.43013 )
( 431 , 0.42881 )
( 432 , 0.42801 )
( 433 , 0.43310 )
( 434 , 0.42691 )
( 435 , 0.42821 )
( 436 , 0.42638 )
( 437 , 0.42890 )
( 438 , 0.42971 )
( 439 , 0.42761 )
( 440 , 0.42932 )
( 441 , 0.42931 )
( 442 , 0.42965 )
( 443 , 0.42827 )
( 444 , 0.43132 )
( 445 , 0.43005 )
( 446 , 0.42782 )
( 447 , 0.42815 )
( 448 , 0.42846 )
( 449 , 0.42704 )
( 450 , 0.42832 )
( 451 , 0.42822 )
( 452 , 0.42808 )
( 453 , 0.43030 )
( 454 , 0.42809 )
( 455 , 0.42714 )
( 456 , 0.42688 )
( 457 , 0.42677 )
( 458 , 0.42706 )
( 459 , 0.42550 )
( 460 , 0.42321 )
( 461 , 0.42410 )
( 462 , 0.42582 )
( 463 , 0.42723 )
( 464 , 0.42501 )
( 465 , 0.42771 )
( 466 , 0.42649 )
( 467 , 0.42814 )
( 468 , 0.42547 )
( 469 , 0.42713 )
( 470 , 0.42764 )
( 471 , 0.42487 )
( 472 , 0.42290 )
( 473 , 0.42466 )
( 474 , 0.42640 )
( 475 , 0.42566 )
( 476 , 0.42581 )
( 477 , 0.42737 )
( 478 , 0.42638 )
( 479 , 0.42612 )
( 480 , 0.42388 )
( 481 , 0.42568 )
( 482 , 0.42740 )
( 483 , 0.42567 )
( 484 , 0.42363 )
( 485 , 0.42530 )
( 486 , 0.42516 )
( 487 , 0.42649 )
( 488 , 0.42564 )
( 489 , 0.42497 )
( 490 , 0.42482 )
( 491 , 0.42519 )
( 492 , 0.42484 )
( 493 , 0.42516 )
( 494 , 0.42580 )
( 495 , 0.42532 )
( 496 , 0.42554 )
( 497 , 0.42342 )
( 498 , 0.42525 )
( 499 , 0.42521 )
( 500 , 0.42427 )
( 501 , 0.42333 )
( 502 , 0.42407 )
( 503 , 0.42285 )
( 504 , 0.42345 )
( 505 , 0.42438 )
( 506 , 0.42467 )
( 507 , 0.42813 )
( 508 , 0.42869 )
( 509 , 0.42567 )
( 510 , 0.42348 )
( 511 , 0.42383 )
( 512 , 0.42473 )
( 513 , 0.42325 )
( 514 , 0.42219 )
( 515 , 0.42524 )
( 516 , 0.42366 )
( 517 , 0.42402 )
( 518 , 0.42137 )
( 519 , 0.42287 )
( 520 , 0.42464 )
( 521 , 0.42423 )
( 522 , 0.42137 )
( 523 , 0.42383 )
( 524 , 0.42131 )
( 525 , 0.42330 )
( 526 , 0.42393 )
( 527 , 0.42187 )
( 528 , 0.42368 )
( 529 , 0.42326 )
( 530 , 0.42274 )
( 531 , 0.42200 )
( 532 , 0.42140 )
( 533 , 0.42320 )}
\newcommand{\valLoss}{( 0 , 0.77419 )
( 1 , 0.68746 )
( 2 , 0.65426 )
( 3 , 0.64073 )
( 4 , 0.63434 )
( 5 , 0.62918 )
( 6 , 0.63200 )
( 7 , 0.62357 )
( 8 , 0.62257 )
( 9 , 0.61858 )
( 10 , 0.61288 )
( 11 , 0.62368 )
( 12 , 0.61080 )
( 13 , 0.60819 )
( 14 , 0.60910 )
( 15 , 0.60629 )
( 16 , 0.60944 )
( 17 , 0.60495 )
( 18 , 0.60678 )
( 19 , 0.59861 )
( 20 , 0.59453 )
( 21 , 0.59608 )
( 22 , 0.60357 )
( 23 , 0.59722 )
( 24 , 0.58820 )
( 25 , 0.58521 )
( 26 , 0.58314 )
( 27 , 0.58347 )
( 28 , 0.59065 )
( 29 , 0.58127 )
( 30 , 0.57989 )
( 31 , 0.58501 )
( 32 , 0.57765 )
( 33 , 0.57167 )
( 34 , 0.57616 )
( 35 , 0.57074 )
( 36 , 0.56757 )
( 37 , 0.56984 )
( 38 , 0.57674 )
( 39 , 0.56671 )
( 40 , 0.57991 )
( 41 , 0.56980 )
( 42 , 0.56345 )
( 43 , 0.56313 )
( 44 , 0.58203 )
( 45 , 0.61766 )
( 46 , 0.56392 )
( 47 , 0.55865 )
( 48 , 0.55530 )
( 49 , 0.56637 )
( 50 , 0.55908 )
( 51 , 0.55922 )
( 52 , 0.55145 )
( 53 , 0.55805 )
( 54 , 0.56605 )
( 55 , 0.55877 )
( 56 , 0.56038 )
( 57 , 0.54911 )
( 58 , 0.56739 )
( 59 , 0.56017 )
( 60 , 0.55860 )
( 61 , 0.55284 )
( 62 , 0.56062 )
( 63 , 0.55358 )
( 64 , 0.55800 )
( 65 , 0.55562 )
( 66 , 0.55986 )
( 67 , 0.54913 )
( 68 , 0.55000 )
( 69 , 0.56539 )
( 70 , 0.55160 )
( 71 , 0.56113 )
( 72 , 0.54825 )
( 73 , 0.56385 )
( 74 , 0.54763 )
( 75 , 0.54570 )
( 76 , 0.54951 )
( 77 , 0.54705 )
( 78 , 0.55782 )
( 79 , 0.54804 )
( 80 , 0.54822 )
( 81 , 0.54605 )
( 82 , 0.55634 )
( 83 , 0.54095 )
( 84 , 0.55072 )
( 85 , 0.54279 )
( 86 , 0.55777 )
( 87 , 0.55979 )
( 88 , 0.54436 )
( 89 , 0.55593 )
( 90 , 0.54439 )
( 91 , 0.54522 )
( 92 , 0.54027 )
( 93 , 0.55169 )
( 94 , 0.53941 )
( 95 , 0.55662 )
( 96 , 0.55341 )
( 97 , 0.56445 )
( 98 , 0.54103 )
( 99 , 0.54172 )
( 100 , 0.54159 )
( 101 , 0.54279 )
( 102 , 0.54366 )
( 103 , 0.54844 )
( 104 , 0.56759 )
( 105 , 0.54413 )
( 106 , 0.54118 )
( 107 , 0.54885 )
( 108 , 0.54484 )
( 109 , 0.55200 )
( 110 , 0.53974 )
( 111 , 0.55558 )
( 112 , 0.54072 )
( 113 , 0.53861 )
( 114 , 0.54868 )
( 115 , 0.55486 )
( 116 , 0.54276 )
( 117 , 0.54633 )
( 118 , 0.54455 )
( 119 , 0.54168 )
( 120 , 0.56616 )
( 121 , 0.53932 )
( 122 , 0.55697 )
( 123 , 0.54046 )
( 124 , 0.53920 )
( 125 , 0.54891 )
( 126 , 0.53391 )
( 127 , 0.53595 )
( 128 , 0.54302 )
( 129 , 0.53717 )
( 130 , 0.53940 )
( 131 , 0.53681 )
( 132 , 0.54545 )
( 133 , 0.53897 )
( 134 , 0.54104 )
( 135 , 0.54370 )
( 136 , 0.53334 )
( 137 , 0.54056 )
( 138 , 0.54591 )
( 139 , 0.53601 )
( 140 , 0.53804 )
( 141 , 0.53921 )
( 142 , 0.53920 )
( 143 , 0.54030 )
( 144 , 0.53909 )
( 145 , 0.54100 )
( 146 , 0.54755 )
( 147 , 0.53606 )
( 148 , 0.53669 )
( 149 , 0.54038 )
( 150 , 0.53490 )
( 151 , 0.54163 )
( 152 , 0.53834 )
( 153 , 0.54019 )
( 154 , 0.53962 )
( 155 , 0.54010 )
( 156 , 0.53179 )
( 157 , 0.54081 )
( 158 , 0.53552 )
( 159 , 0.53806 )
( 160 , 0.53861 )
( 161 , 0.54674 )
( 162 , 0.54720 )
( 163 , 0.54573 )
( 164 , 0.53850 )
( 165 , 0.54151 )
( 166 , 0.54404 )
( 167 , 0.57238 )
( 168 , 0.56572 )
( 169 , 0.54329 )
( 170 , 0.53853 )
( 171 , 0.54637 )
( 172 , 0.53500 )
( 173 , 0.53801 )
( 174 , 0.53892 )
( 175 , 0.54291 )
( 176 , 0.53837 )
( 177 , 0.54184 )
( 178 , 0.54204 )
( 179 , 0.54941 )
( 180 , 0.54466 )
( 181 , 0.55748 )
( 182 , 0.54724 )
( 183 , 0.54718 )
( 184 , 0.55072 )
( 185 , 0.54332 )
( 186 , 0.54216 )
( 187 , 0.53204 )
( 188 , 0.53219 )
( 189 , 0.53795 )
( 190 , 0.54061 )
( 191 , 0.54567 )
( 192 , 0.54144 )
( 193 , 0.53351 )
( 194 , 0.56218 )
( 195 , 0.55147 )
( 196 , 0.54300 )
( 197 , 0.54533 )
( 198 , 0.53670 )
( 199 , 0.54597 )
( 200 , 0.54175 )
( 201 , 0.53636 )
( 202 , 0.53349 )
( 203 , 0.54414 )
( 204 , 0.54330 )
( 205 , 0.53955 )
( 206 , 0.55047 )
( 207 , 0.53758 )
( 208 , 0.53535 )
( 209 , 0.53453 )
( 210 , 0.57576 )
( 211 , 0.53898 )
( 212 , 0.53769 )
( 213 , 0.53354 )
( 214 , 0.53415 )
( 215 , 0.54183 )
( 216 , 0.53153 )
( 217 , 0.53611 )
( 218 , 0.54271 )
( 219 , 0.54567 )
( 220 , 0.54204 )
( 221 , 0.54274 )
( 222 , 0.54419 )
( 223 , 0.54232 )
( 224 , 0.54858 )
( 225 , 0.53941 )
( 226 , 0.53827 )
( 227 , 0.53882 )
( 228 , 0.53926 )
( 229 , 0.55127 )
( 230 , 0.54147 )
( 231 , 0.53302 )
( 232 , 0.53645 )
( 233 , 0.53924 )
( 234 , 0.53715 )
( 235 , 0.54963 )
( 236 , 0.53551 )
( 237 , 0.53294 )
( 238 , 0.53472 )
( 239 , 0.54001 )
( 240 , 0.54357 )
( 241 , 0.53119 )
( 242 , 0.53537 )
( 243 , 0.54309 )
( 244 , 0.52980 )
( 245 , 0.54604 )
( 246 , 0.53570 )
( 247 , 0.53422 )
( 248 , 0.53162 )
( 249 , 0.53477 )
( 250 , 0.54331 )
( 251 , 0.53989 )
( 252 , 0.53386 )
( 253 , 0.54793 )
( 254 , 0.52730 )
( 255 , 0.53077 )
( 256 , 0.53031 )
( 257 , 0.54199 )
( 258 , 0.53087 )
( 259 , 0.53169 )
( 260 , 0.53861 )
( 261 , 0.53460 )
( 262 , 0.53172 )
( 263 , 0.53279 )
( 264 , 0.52559 )
( 265 , 0.53081 )
( 266 , 0.54087 )
( 267 , 0.52874 )
( 268 , 0.54008 )
( 269 , 0.52655 )
( 270 , 0.54057 )
( 271 , 0.53605 )
( 272 , 0.52492 )
( 273 , 0.53178 )
( 274 , 0.52020 )
( 275 , 0.53594 )
( 276 , 0.52990 )
( 277 , 0.52516 )
( 278 , 0.53131 )
( 279 , 0.52732 )
( 280 , 0.53636 )
( 281 , 0.53834 )
( 282 , 0.52922 )
( 283 , 0.54278 )
( 284 , 0.52933 )
( 285 , 0.52505 )
( 286 , 0.52911 )
( 287 , 0.52465 )
( 288 , 0.55346 )
( 289 , 0.53734 )
( 290 , 0.52959 )
( 291 , 0.53891 )
( 292 , 0.53378 )
( 293 , 0.52491 )
( 294 , 0.53461 )
( 295 , 0.52313 )
( 296 , 0.54307 )
( 297 , 0.52894 )
( 298 , 0.52821 )
( 299 , 0.53519 )
( 300 , 0.53418 )
( 301 , 0.52294 )
( 302 , 0.53936 )
( 303 , 0.52391 )
( 304 , 0.53168 )
( 305 , 0.53150 )
( 306 , 0.52497 )
( 307 , 0.52335 )
( 308 , 0.53755 )
( 309 , 0.52873 )
( 310 , 0.52193 )
( 311 , 0.53126 )
( 312 , 0.52771 )
( 313 , 0.52254 )
( 314 , 0.52300 )
( 315 , 0.52451 )
( 316 , 0.54803 )
( 317 , 0.52707 )
( 318 , 0.52875 )
( 319 , 0.54111 )
( 320 , 0.54334 )
( 321 , 0.52778 )
( 322 , 0.52715 )
( 323 , 0.51998 )
( 324 , 0.52220 )
( 325 , 0.52521 )
( 326 , 0.53612 )
( 327 , 0.52822 )
( 328 , 0.51909 )
( 329 , 0.54097 )
( 330 , 0.52892 )
( 331 , 0.53496 )
( 332 , 0.52386 )
( 333 , 0.52413 )
( 334 , 0.52107 )
( 335 , 0.52399 )
( 336 , 0.52335 )
( 337 , 0.52734 )
( 338 , 0.53343 )
( 339 , 0.52731 )
( 340 , 0.53505 )
( 341 , 0.54381 )
( 342 , 0.52441 )
( 343 , 0.53754 )
( 344 , 0.51441 )
( 345 , 0.53013 )
( 346 , 0.54012 )
( 347 , 0.51449 )
( 348 , 0.51800 )
( 349 , 0.52664 )
( 350 , 0.53595 )
( 351 , 0.52158 )
( 352 , 0.52793 )
( 353 , 0.52865 )
( 354 , 0.52004 )
( 355 , 0.52316 )
( 356 , 0.53236 )
( 357 , 0.52126 )
( 358 , 0.53006 )
( 359 , 0.51458 )
( 360 , 0.51151 )
( 361 , 0.52238 )
( 362 , 0.52755 )
( 363 , 0.52405 )
( 364 , 0.52238 )
( 365 , 0.51897 )
( 366 , 0.52211 )
( 367 , 0.52013 )
( 368 , 0.52667 )
( 369 , 0.52908 )
( 370 , 0.52204 )
( 371 , 0.52872 )
( 372 , 0.52786 )
( 373 , 0.52176 )
( 374 , 0.52598 )
( 375 , 0.52006 )
( 376 , 0.52039 )
( 377 , 0.52035 )
( 378 , 0.52785 )
( 379 , 0.52808 )
( 380 , 0.55030 )
( 381 , 0.53140 )
( 382 , 0.52534 )
( 383 , 0.52100 )
( 384 , 0.53331 )
( 385 , 0.52285 )
( 386 , 0.52146 )
( 387 , 0.52133 )
( 388 , 0.51516 )
( 389 , 0.51948 )
( 390 , 0.52685 )
( 391 , 0.52104 )
( 392 , 0.51621 )
( 393 , 0.51813 )
( 394 , 0.51443 )
( 395 , 0.51956 )
( 396 , 0.51824 )
( 397 , 0.51504 )
( 398 , 0.51528 )
( 399 , 0.52390 )
( 400 , 0.52678 )
( 401 , 0.51968 )
( 402 , 0.51832 )
( 403 , 0.52060 )
( 404 , 0.52511 )
( 405 , 0.51607 )
( 406 , 0.51767 )
( 407 , 0.51947 )
( 408 , 0.53661 )
( 409 , 0.51931 )
( 410 , 0.52716 )
( 411 , 0.52026 )
( 412 , 0.53305 )
( 413 , 0.52533 )
( 414 , 0.51985 )
( 415 , 0.52666 )
( 416 , 0.51783 )
( 417 , 0.51790 )
( 418 , 0.51876 )
( 419 , 0.53108 )
( 420 , 0.51921 )
( 421 , 0.52317 )
( 422 , 0.52491 )
( 423 , 0.52770 )
( 424 , 0.51996 )
( 425 , 0.51970 )
( 426 , 0.52193 )
( 427 , 0.52202 )
( 428 , 0.52863 )
( 429 , 0.53155 )
( 430 , 0.51175 )
( 431 , 0.52192 )
( 432 , 0.52579 )
( 433 , 0.51049 )
( 434 , 0.51668 )
( 435 , 0.51757 )
( 436 , 0.52275 )
( 437 , 0.52729 )
( 438 , 0.51873 )
( 439 , 0.51850 )
( 440 , 0.52124 )
( 441 , 0.52421 )
( 442 , 0.52744 )
( 443 , 0.52404 )
( 444 , 0.53484 )
( 445 , 0.53329 )
( 446 , 0.52358 )
( 447 , 0.51074 )
( 448 , 0.51820 )
( 449 , 0.51999 )
( 450 , 0.52408 )
( 451 , 0.52480 )
( 452 , 0.52717 )
( 453 , 0.52181 )
( 454 , 0.52702 )
( 455 , 0.52078 )
( 456 , 0.51956 )
( 457 , 0.54001 )
( 458 , 0.52653 )
( 459 , 0.52078 )
( 460 , 0.52136 )
( 461 , 0.52850 )
( 462 , 0.52431 )
( 463 , 0.52107 )
( 464 , 0.52114 )
( 465 , 0.51480 )
( 466 , 0.52287 )
( 467 , 0.52212 )
( 468 , 0.52480 )
( 469 , 0.53792 )
( 470 , 0.52205 )
( 471 , 0.51605 )
( 472 , 0.51996 )
( 473 , 0.52001 )
( 474 , 0.52401 )
( 475 , 0.52737 )
( 476 , 0.52051 )
( 477 , 0.52957 )
( 478 , 0.53312 )
( 479 , 0.52504 )
( 480 , 0.51445 )
( 481 , 0.51523 )
( 482 , 0.51788 )
( 483 , 0.52365 )
( 484 , 0.52333 )
( 485 , 0.52295 )
( 486 , 0.52815 )
( 487 , 0.51538 )
( 488 , 0.51697 )
( 489 , 0.53216 )
( 490 , 0.51972 )
( 491 , 0.51912 )
( 492 , 0.51995 )
( 493 , 0.52624 )
( 494 , 0.51365 )
( 495 , 0.51260 )
( 496 , 0.51678 )
( 497 , 0.52678 )
( 498 , 0.51234 )
( 499 , 0.52910 )
( 500 , 0.53421 )
( 501 , 0.52232 )
( 502 , 0.51859 )
( 503 , 0.51727 )
( 504 , 0.52799 )
( 505 , 0.52967 )
( 506 , 0.51977 )
( 507 , 0.52590 )
( 508 , 0.54530 )
( 509 , 0.52071 )
( 510 , 0.52498 )
( 511 , 0.52362 )
( 512 , 0.52491 )
( 513 , 0.51983 )
( 514 , 0.52086 )
( 515 , 0.51615 )
( 516 , 0.51381 )
( 517 , 0.51926 )
( 518 , 0.52780 )
( 519 , 0.54601 )
( 520 , 0.52277 )
( 521 , 0.51571 )
( 522 , 0.52779 )
( 523 , 0.52563 )
( 524 , 0.53294 )
( 525 , 0.53083 )
( 526 , 0.52234 )
( 527 , 0.52018 )
( 528 , 0.52210 )
( 529 , 0.52609 )
( 530 , 0.51784 )
( 531 , 0.53836 )
( 532 , 0.51775 )
( 533 , 0.52165 )}

\begin{figure}[H]
\begin{center}
\begin{tikzpicture}[scale=0.8]
        \begin{axis}[xlabel={Epochs}, ylabel={Loss}, axis y line=left, axis x line=bottom, xmin=0, xmax=500,legend pos=north east] 
        \addplot[smooth,blue,thick] coordinates {\trainLoss};
         \addlegendentry{Train loss}
      \addplot[smooth,orange,thick] coordinates {\valLoss};
        \addlegendentry{Val Loss}
        \end{axis}
\end{tikzpicture}
\end{center}
\caption{Sample loss during the training phase.}
\label{fig:sample_train}
\end{figure}

To get an external reference point, we compared our implementation of TasteNet to that of~\citet{han2020neural} in Pytorch. Using the same train/validation/testing split of Swissmetro data, the out-of-sample normalized negative log-likelihoods achieved by the latter are higher (on average, $0.69$) than those obtained from our implementation (see Table~\ref{tab:numerics_summary}). These differences are explained by our choice of larger neural networks and possibly by other aspects of the estimation process (initialization and regularization). As a result, we use our implementation framework uniformly across all neural-network-based models, to enable a fair comparison of the model classes.

Regarding model size, Table~\ref{tab:params} summarizes the number of parameters for each model. Note that it takes on average approximately 45 min to fit the largest RUMnet on Swissmetro data and 36 hours on Expedia data.  Moderate-size RUMnets are trained faster. For instance, for $(\ell,w)=(3,10)$ and $K=5$, it takes on average 10 minutes on the Swissmetro dataset and 4 hours on the Expedia dataset.  Note that the running time for estimating RUMnets is an order of magnitude larger than the running time of random forests, which is the second most computationally intensive method (up to 1 hour on Expedia dataset).  It is unclear whether this is an inherent property of the architecture or if it is due to our implementation. We do not attempt to use large-scale distributed computations. As we combine different computational resources to run our experiments (both desktops and a CPU cluster), the reported running times are only indicative.

\begin{table}[H]\centering
\caption{Number of parameters for the different models.} \label{tab:params}
\setlength\extrarowheight{2pt} 
\resizebox{6cm}{!}{
\begin{tabular}{ccccc}
Model  & $(\ell,w)$ & $K$  &  Swissmetro & Expedia\\
\cmidrule[1pt](lr){1-3} \cmidrule[1pt](lr){4-4} \cmidrule[1pt](lr){5-5}
 MNL & - & -  & 87 & 92 \\
\cmidrule(lr){1-5} 
 \multirow{3}{*}{Latent Class MNL} & - &  5 & 450 &  660\\
& -  & 10   & 900   &   1,320\\
& -  & 20   & 1,800  & 2,640\\
\cmidrule(lr){1-5} 

 \multirow{3}{*}{TasteNet} & (3,10) &  - & 1,099 & 1,633 \\
& (5,20)  &-   & 3,429 &  4,413 \\
& (10,30)  & -   & 11,019 &  12,423\\
\cmidrule(lr){1-5} 
 \multirow{3}{*}{DeepMNL} 
& (3,10) & -  & 1,101 &   1,550  \\
 &  (5,20) & - & 3,441 & 4,340  \\
 & (10,30) & -  & 11,041 & 12,360\\
 \cmidrule[1pt](lr){1-5} 
\multirow{4}{*}{RUMnet} 
 & (3,10) & 5  & 8,140 & 10,600\\
 & (3,10) & 10   & 14,940 &  19,450\\
\cmidrule(lr){2-5} 
  &  (5,20) & 5  & 30,320 & 35,240\\
 & (5,20) & 10   & 56,320& - \\
 \cmidrule[1pt](lr){1-5} 
 \multirow{3}{*}{VNN} & (3,10)  &-   & 1,200 & 19,410\\
 &  (5,20) & -  & 3,640 & 40,060\\
    &  (10,30) & -  & 11,340 &  65,940\\
 \cmidrule(lr){1-5} 
\end{tabular}
}
\end{table}

As our main goal is to show a proof of concept, we do not attempt to optimize the running time and estimation process. Nonetheless, several practical tricks could be used to boost the predictive and computational performance, including parallelization, norm-based regularization, dropout layers, scheduled learning rates, etc.

\subsection{Case study 1: Swissmetro dataset} \label{app:swissmetro}

The Swissmetro is a proposed revolutionary underground system connecting major cities in Switzerland. To assess potential demand, a survey collected data from 1,192 respondents (441 rail-based travellers and 751 car users), with 9 choice events from each respondent. Each respondent was asked to choose one mode out of a set of alternatives for inter-city travel given the features of each mode such as travel time or cost. The choice set includes train, Swissmetro, and car ($\kappa = 3$). For individuals without a car, the choice set only includes train and Swissmetro. Each alternative has 4 features ($d_x=4$) and each choice has 29 categorical features such as information about the respondent or the origin-destination pair which we transform using a simple binary encoding into a vector of size $d_z=83$.  Table~\ref{tab:sm_variables} details the different features present in the data.
\begin{table}[H]\centering
\caption{Variable description}\label{tab:sm_variables}
\setlength\extrarowheight{2pt} 
\resizebox{10cm}{!}{
\begin{tabular}{cc}
Product features & Context features \\
\cmidrule[1pt](lr){1-1} \cmidrule[1pt](lr){2-2}
 Availability dummy & User group (current road or rail user)\\
 Travel time & Travel purpose \\
 Cost &  First class traveler\\ 
 Headway & Ticket type (one-way, two-way, annual pass...) \\
 &  Payer (self, employer...) \\
 &  Number of  Luggage \\
 & Age \\
 & Income brackets \\
 & Annual season ticket \\
 & Travel origin \\
 & Travel destination \\
\cmidrule[1pt](lr){1-2}
\end{tabular}
}
\end{table}
For more information, we  refer the reader to \cite{bierlaire2018swissmetro}. The original data has 10,728 observations\footnote{\url{https://biogeme.epfl.ch/data.html}} and has been used recurrently to test different choice modeling approach \citep{sifringer2020enhancing,han2020neural}. We preprocess the data by removing observations with unknown choice \citep{han2020neural}. We retain 10,719 observations, which we randomly split into training, validation and test sets with 7,505, 1,607 and 1,607 observations, respectively.

\subsection{Case study 2: Expedia dataset} \label{app:expedia}

We next evaluate RUMnets on a dataset of hotel searches on Expedia made publicly available through the competition ``Personalize Expedia Hotel Searches'' hosted by ICDM in 2013\footnote{\url{https://www.kaggle.com/c/expedia-personalized-sort}}. Each hotel search instance $(\bs{x}_t,\bs{z}_t,A_t)$ consists of the following types of information:
\begin{itemize}
\item Customer attributes $\bs{z}_t$: These attributes comprise user and search query features such as the visitor's country and search destination, the number of rooms, the duration of stay, whether there are children, how many days in advance of their trip the search is made.
\item Assortment $A_t$: The assortment includes all hotels displayed to the user on the search result page. Each alternative has product attributes that include average user ratings, current price, average historical price, location scores, display position, among others.
\item Choice $\bs{x}_t \in A_t$: In response to the displayed assortment, each user either booked a hotel $\bs{x}_t \in A_t$ or left without making any booking. As explained in the data pre-processing steps, we focus on the former type of events, meaning that we do not include the no-purchase option in the assortment.
\end{itemize}

The dataset is pre-processed as follows.\footnote{A preliminary version of the paper used different data processing, excluding the outside option (no-bookings) or with fewer product and customer features.} The initial dataset comprises 399,344 search instances. We create a one-hot encoding of the following categorical features:  \texttt{site\_id},  \texttt{visitor\_location\_country\_id}, \texttt{prop\_country\_id},  \texttt{srch\_destination\_id}, whereby all categories with fewer than 1,000 occurrences are lumped into a single binary indicator `-1'. The features \texttt{price\_usd} and \texttt{srch\_booking\_window} both exhibit unrealistic values for a few choice events. We filter searches with hotel a hotel price between \$10 and \$1,000. Additionally, we filter any search query made more than 1 year in advance or with a length of stay greater than 14 days. Consequently, we apply a log-transformations to the price and booking window features ($\texttt{price}\mapsto \log(\texttt{price})$ and $\texttt{search\_booking\_window} \mapsto \log(1+\texttt{search\_booking\_window})$). We create an outside option whose product features are all equal to zero, except the dummy indicator \texttt{is\_no\_purchase}$=1$. Finally, all missing observations are marked with the value `-1'.

Random Forests require a fixed assortment size as input. Hence, as part of our data pre-processing, we let $\kappa$ be the largest number of hotel options, counting the no-purchase alternative, and ``fill'' the assortment corresponding to each search query with dummy products. We create an \texttt{availability} binary feature, which indicates which items are real items on offer. By convention, dummy products have all their features equal to zero, except the log-price which is set as a large value $10^4$. Intuitively, this design enables distinguishing (unavailable) dummy products from products on offer, across multiple features. We also post-process the output of Random Forests, by re-normalizing choice probabilities, to ensure that unavailable items are picked with probability zero. We note that these operations are not relevant to the neural network-based models, as we can implement a `mask' in the training process to only consider available products. 

Following these transformations, the dataset counts 397,618 search queries, 46 hotel features, 85 customer and search features, and up to 39 alternatives, including the no-booking option.

\newpage 

\subsection{Predictive performance of individual models tested} \label{app:full_table}

Table~\ref{tab:numerics_detailed} gives detailed performance for the individual models tested in Section~\ref{sec:numerical}.
\begin{table}[H]\centering
\caption{Average predictive performance of the fitted choice models on the data splits} \label{tab:numerics_detailed}
\setlength\extrarowheight{2pt} 
\resizebox{16cm}{!}{
\begin{tabular}{cccccccccccccccc}
& &  &  & \multicolumn{6}{c}{Swissmetro} & \multicolumn{6}{c}{Expedia }    \\
 \cmidrule[1pt](lr){5-10} \cmidrule[1pt](lr){11-16}
 & \multicolumn{3}{c}{Model}  & \multicolumn{3}{c}{Log-likelihood loss} & \multicolumn{3}{c}{Accuracy} & \multicolumn{3}{c}{Log-likelihood loss} & \multicolumn{3}{c}{Accuracy}   \\
 \cmidrule[1pt](lr){2-4} \cmidrule[1pt](lr){5-7} \cmidrule[1pt](lr){8-10} \cmidrule[1pt](lr){11-13} \cmidrule[1pt](lr){14-16}
& Type & $(\ell,w)$ & $K$  & Train & Val & Test & Train & Val & Test &Train & Val & Test  &Train & Val & Test \\
\cmidrule[1pt](lr){2-16}
\multirow[c]{10}{*}{\rotatebox{90}{{{\sf \tiny MNL and extensions}}}} & MNL & - & - & 0.841 & 0.841 & 0.842 & 0.624 & 0.625 & 0.623 &2.482 & 2.482 & 2.482 & 0.307 & 0.307 & 0.307 \\
\cmidrule(lr){2-16} 
& \multirow{3}{*}{Latent class MNL} & - &  5 & 0.777 & 0.778 & 0.778 & 0.637 & 0.637 & 0.635 & 2.439 & 2.439 & 2.439 & 0.316 & 0.316 & 0.316  \\
&& -  &10  & 0.764 & 0.766 & 0.766 & 0.649 & 0.650 & 0.651 & 2.435 & 2.435 & 2.435 & 0.317 & 0.317 & 0.317 \\
&& - & 20 & 0.762 & 0.763 & 0.764 & 0.651 & 0.652 & 0.651  & 2.434 & 2.435 & 2.435 & 0.317 & 0.317 & 0.317 \\
\cmidrule(lr){2-16} 
& \multirow{3}{*}{TasteNet} & (3,10) &  - &  0.458 & 0.571 & 0.589 & 0.817 & 0.777 & 0.772 &2.111 &2.091 & 2.095  & 0.413 & 0.412 & 0.413 \\
&& (5,20)  &-   & 0.388 & 0.545 & 0.570 & 0.847 & 0.793 & 0.779 &2.087 & 2.097 & 2.097 & 0.415 & 0.413 & 0.413 \\
&& (10,30)  & - & 0.382 & 0.540 & 0.565 & 0.847 & 0.790 & 0.784 &2.079 & 2.094 & 2.094 & 0.416 & 0.412 & 0.412 \\
\cmidrule(lr){2-16} 
& \multirow{3}{*}{DeepMNL} 
& (3,10)& -  & 0.485 & 0.587 & 0.600 & 0.804 & 0.763 & 0.758 &2.069 & 2.074 & 2.073 & 0.413 & 0.412 & 0.412 \\
& &  (5,20) & -  &0.401 & 0.560 & 0.585 & 0.840 & 0.779 & 0.772  &2.065 & 2.077 & 2.077 & 0.413 & 0.410 & 0.410 \\
& & (10,30) & -  &  0.406 & 0.563 & 0.585 & 0.836 & 0.775 & 0.773 &2.066 & 2.079 & 2.079 & 0.413 & 0.410 & 0.410 \\
 \cmidrule[1pt](lr){2-16} 
 &\multirow{4}{*}{RUMnet} 
& (3,10) & 5  &   0.416 & 0.570 & 0.586 & 0.833 & 0.771 & 0.769&2.004 & 2.021 & 2.021 & 0.428 & 0.425 & 0.425 \\
& & (3,10) & 10   &0.375 & 0.541 & 0.566 & 0.852 & 0.781 & 0.777 &2.005 & 2.023 & 2.023 & 0.428 & 0.425 & 0.424 \\
\cmidrule(lr){3-16} 
&  &  (5,20) & 5  &0.353 & 0.552 & 0.589 & 0.862 & 0.788 & 0.783 &2.001 & 2.018 & 2.019 & 0.429 & 0.426 & 0.426 \\
& & (5,20) & 10   & 0.326 & 0.528 & 0.558 & 0.876 & 0.792 & 0.794  &-&-&-&-&-&-\\
 \cmidrule[1pt](lr){2-16} 
\multirow[c]{16}*{\rotatebox{90}{{{\sf \tiny Model-free ML}}}} & \multirow{3}{*}{Vanilla Neural Network} & (3,10)  &-   &  0.519 & 0.609 & 0.625 & 0.784 & 0.748 & 0.739 &2.447 & 2.484 & 2.483 & 0.311 & 0.308 & 0.308\\
& &  (5,20) & -  &   0.471 & 0.591 & 0.610 & 0.806 & 0.754 & 0.748&2.336 & 2.417 & 2.417 & 0.344 & 0.328 & 0.328 \\
&    &  (10,30) & -  &0.466 & 0.599 & 0.611 & 0.805 & 0.751 & 0.748 &2.334 & 2.407 & 2.407 & 0.346 & 0.328 & 0.328 \\
 \cmidrule(lr){2-16} 
& \multirow{14}{*}{Random Forest} &- & - &0.516 & 0.610 & 0.611 & 0.805 & 0.735 & 0.734  &2.732 & 2.852 & 2.852 & 0.309 & 0.307 & 0.307 \\
& & - & - &  0.517 & 0.610 & 0.611 & 0.804 & 0.737 & 0.734&2.727 & 2.849 & 2.849 & 0.307 & 0.307 & 0.307 \\
& & - & - &  0.518 & 0.611 & 0.611 & 0.804 & 0.735 & 0.735&2.725 & 2.848 & 2.848 & 0.307 & 0.307 & 0.307 \\
& & - & - &  0.286 & 0.542 & inf & 0.963 & 0.771 & 0.767  &1.985 & 2.780 & 2.779 & 0.661 & 0.308 & 0.308 \\
& & - & - &  0.286 & 0.540 & 0.540 & 0.965 & 0.771 & 0.770&1.974 & 2.772 & 2.772 & 0.666 & 0.308 & 0.308 \\
& & - & - &  0.288 & 0.540 & 0.540 & 0.964 & 0.772 & 0.769&1.966 & 2.767 & 2.767 & 0.677 & 0.307 & 0.308 \\
& & - & - & 0.168 & inf & inf & 0.999 & 0.774 & 0.775&1.289 & 2.739 & 2.739 & 0.992 & 0.310 & 0.310 \\
& & - & - &   0.168 & 0.522 & inf & 0.999 & 0.774 & 0.778&1.282 & 2.722 & 2.722 & 0.997 & 0.309 & 0.310 \\
& & - & - &   0.169 & 0.520 & inf & 0.999 & 0.776 & 0.777&1.275 & 2.713 & 2.712 & 0.999 & 0.309 & 0.309 \\
& & - & - &    &  &  &  &  &  &2.960 & 2.964 & 2.964 & 0.306 & 0.306 & 0.306 \\
& & - & - &    &  &  &  &  &   &2.959 & 2.964 & 2.964 & 0.307 & 0.307 & 0.307 \\
& & - & - &    &  &  &  &  &  &2.958 & 2.962 & 2.962 & 0.307 & 0.307 & 0.307 \\
& & - & - &    &  &  &  &  &  &0.333 & inf & inf & 1.000 & 0.311 & 0.312 \\
& & - & - &    &  &  &  &  &  &0.333 & inf & inf & 1.000 & 0.311 & 0.311 \\
 \cmidrule[1pt](lr){2-16} 
\end{tabular}
}
\end{table} 

\newpage 

\subsection{Additional materials} \label{app:add_viz} 

\subsubsection{Choice probabilities under random forest model.} 

In Figure~\ref{fig:monotonicityRF}, we plot the choice probabilities predicted by the trained random forests as a function of the cost of the Swissmetro alternative. We observe that the variations of the choice probabilities are not monotone in contrast with RUMnets; see Figure~\ref{fig:types} in the main paper.
\begin{figure}[H]
\centering
  \begin{tikzpicture}
    \newcommand{\probaTrain}{
( 0.00000 , 0.16950 )
( 0.00303 , 0.16950 )
( 0.00606 , 0.15688 )
( 0.00909 , 0.15065 )
( 0.01212 , 0.17377 )
( 0.01515 , 0.18984 )
( 0.01818 , 0.18829 )
( 0.02121 , 0.18894 )
( 0.02424 , 0.18261 )
( 0.02727 , 0.18230 )
( 0.03030 , 0.18036 )
( 0.03333 , 0.18230 )
( 0.03636 , 0.18671 )
( 0.03939 , 0.19793 )
( 0.04242 , 0.19987 )
( 0.04545 , 0.20160 )
( 0.04848 , 0.20192 )
( 0.05152 , 0.19862 )
( 0.05455 , 0.19668 )
( 0.05758 , 0.19473 )
( 0.06061 , 0.19473 )
( 0.06364 , 0.21208 )
( 0.06667 , 0.21947 )
( 0.06970 , 0.22838 )
( 0.07273 , 0.23614 )
( 0.07576 , 0.24391 )
( 0.07879 , 0.24585 )
( 0.08182 , 0.26294 )
( 0.08485 , 0.26100 )
( 0.08788 , 0.26488 )
( 0.09091 , 0.26488 )
( 0.09394 , 0.26262 )
( 0.09697 , 0.26262 )
( 0.10000 , 0.26445 )
( 0.10303 , 0.26445 )
( 0.10606 , 0.27532 )
( 0.10909 , 0.27532 )
( 0.11212 , 0.27953 )
( 0.11515 , 0.27953 )
( 0.11818 , 0.27759 )
( 0.12121 , 0.27759 )
( 0.12424 , 0.27759 )
( 0.12727 , 0.27759 )
( 0.13030 , 0.27759 )
( 0.13333 , 0.27759 )
( 0.13636 , 0.27759 )
( 0.13939 , 0.27953 )
( 0.14242 , 0.27953 )
( 0.14545 , 0.28147 )
( 0.14848 , 0.28147 )
( 0.15152 , 0.28147 )
( 0.15455 , 0.28147 )
( 0.15758 , 0.28147 )
( 0.16061 , 0.28147 )
( 0.16364 , 0.28147 )
( 0.16667 , 0.28147 )
( 0.16970 , 0.28147 )
( 0.17273 , 0.28147 )
( 0.17576 , 0.28147 )
( 0.17879 , 0.28147 )
( 0.18182 , 0.28147 )
( 0.18485 , 0.28147 )
( 0.18788 , 0.28147 )
( 0.19091 , 0.28147 )
( 0.19394 , 0.28147 )
( 0.19697 , 0.28147 )
( 0.20000 , 0.28147 )
( 0.20303 , 0.28147 )
( 0.20606 , 0.28147 )
( 0.20909 , 0.28147 )
( 0.21212 , 0.28147 )
( 0.21515 , 0.27953 )
( 0.21818 , 0.27953 )
( 0.22121 , 0.27953 )
( 0.22424 , 0.27953 )
( 0.22727 , 0.27953 )
( 0.23030 , 0.27953 )
( 0.23333 , 0.27953 )
( 0.23636 , 0.27953 )
( 0.23939 , 0.27953 )
( 0.24242 , 0.27953 )
( 0.24545 , 0.27953 )
( 0.24848 , 0.27953 )
( 0.25152 , 0.27953 )
( 0.25455 , 0.27953 )
( 0.25758 , 0.27953 )
( 0.26061 , 0.27953 )
( 0.26364 , 0.27953 )
( 0.26667 , 0.27953 )
( 0.26970 , 0.27953 )
( 0.27273 , 0.27953 )
( 0.27576 , 0.27953 )
( 0.27879 , 0.27953 )
( 0.28182 , 0.27953 )
( 0.28485 , 0.27953 )
( 0.28788 , 0.27953 )
( 0.29091 , 0.27953 )
( 0.29394 , 0.27953 )
( 0.29697 , 0.27953 )
( 0.30000 , 0.27953 )
}
\newcommand{\probaSM}{
( 0.00000 , 0.72200 )
( 0.00303 , 0.72200 )
( 0.00606 , 0.71956 )
( 0.00909 , 0.71997 )
( 0.01212 , 0.67949 )
( 0.01515 , 0.66600 )
( 0.01818 , 0.66709 )
( 0.02121 , 0.65494 )
( 0.02424 , 0.65370 )
( 0.02727 , 0.65246 )
( 0.03030 , 0.65246 )
( 0.03333 , 0.64663 )
( 0.03636 , 0.64222 )
( 0.03939 , 0.61945 )
( 0.04242 , 0.60586 )
( 0.04545 , 0.59938 )
( 0.04848 , 0.59574 )
( 0.05152 , 0.60331 )
( 0.05455 , 0.60849 )
( 0.05758 , 0.61431 )
( 0.06061 , 0.61625 )
( 0.06364 , 0.59891 )
( 0.06667 , 0.59346 )
( 0.06970 , 0.58455 )
( 0.07273 , 0.58067 )
( 0.07576 , 0.57290 )
( 0.07879 , 0.57484 )
( 0.08182 , 0.56844 )
( 0.08485 , 0.57038 )
( 0.08788 , 0.56649 )
( 0.09091 , 0.56649 )
( 0.09394 , 0.56798 )
( 0.09697 , 0.56798 )
( 0.10000 , 0.56698 )
( 0.10303 , 0.56698 )
( 0.10606 , 0.56020 )
( 0.10909 , 0.56020 )
( 0.11212 , 0.55793 )
( 0.11515 , 0.55793 )
( 0.11818 , 0.55987 )
( 0.12121 , 0.55987 )
( 0.12424 , 0.55987 )
( 0.12727 , 0.55987 )
( 0.13030 , 0.55987 )
( 0.13333 , 0.55987 )
( 0.13636 , 0.55987 )
( 0.13939 , 0.55793 )
( 0.14242 , 0.55793 )
( 0.14545 , 0.55599 )
( 0.14848 , 0.55599 )
( 0.15152 , 0.55599 )
( 0.15455 , 0.55599 )
( 0.15758 , 0.55599 )
( 0.16061 , 0.55599 )
( 0.16364 , 0.55599 )
( 0.16667 , 0.55599 )
( 0.16970 , 0.55599 )
( 0.17273 , 0.55599 )
( 0.17576 , 0.55599 )
( 0.17879 , 0.55599 )
( 0.18182 , 0.55599 )
( 0.18485 , 0.55599 )
( 0.18788 , 0.55599 )
( 0.19091 , 0.55599 )
( 0.19394 , 0.55599 )
( 0.19697 , 0.55599 )
( 0.20000 , 0.55599 )
( 0.20303 , 0.55599 )
( 0.20606 , 0.55599 )
( 0.20909 , 0.55599 )
( 0.21212 , 0.55599 )
( 0.21515 , 0.55793 )
( 0.21818 , 0.55793 )
( 0.22121 , 0.55793 )
( 0.22424 , 0.55793 )
( 0.22727 , 0.55793 )
( 0.23030 , 0.55793 )
( 0.23333 , 0.55793 )
( 0.23636 , 0.55793 )
( 0.23939 , 0.55793 )
( 0.24242 , 0.55793 )
( 0.24545 , 0.55793 )
( 0.24848 , 0.55987 )
( 0.25152 , 0.55987 )
( 0.25455 , 0.55987 )
( 0.25758 , 0.55987 )
( 0.26061 , 0.55987 )
( 0.26364 , 0.55987 )
( 0.26667 , 0.55987 )
( 0.26970 , 0.55987 )
( 0.27273 , 0.55987 )
( 0.27576 , 0.55987 )
( 0.27879 , 0.55987 )
( 0.28182 , 0.55987 )
( 0.28485 , 0.55987 )
( 0.28788 , 0.55987 )
( 0.29091 , 0.55987 )
( 0.29394 , 0.55987 )
( 0.29697 , 0.55987 )
( 0.30000 , 0.55987 )
}
\newcommand{\probaCar}{
( 0.00000 , 0.10850 )
( 0.00303 , 0.10850 )
( 0.00606 , 0.12356 )
( 0.00909 , 0.12939 )
( 0.01212 , 0.14674 )
( 0.01515 , 0.14416 )
( 0.01818 , 0.14461 )
( 0.02121 , 0.15613 )
( 0.02424 , 0.16369 )
( 0.02727 , 0.16524 )
( 0.03030 , 0.16718 )
( 0.03333 , 0.17107 )
( 0.03636 , 0.17107 )
( 0.03939 , 0.18262 )
( 0.04242 , 0.19427 )
( 0.04545 , 0.19902 )
( 0.04848 , 0.20235 )
( 0.05152 , 0.19807 )
( 0.05455 , 0.19484 )
( 0.05758 , 0.19095 )
( 0.06061 , 0.18901 )
( 0.06364 , 0.18901 )
( 0.06667 , 0.18707 )
( 0.06970 , 0.18707 )
( 0.07273 , 0.18319 )
( 0.07576 , 0.18319 )
( 0.07879 , 0.17930 )
( 0.08182 , 0.16862 )
( 0.08485 , 0.16862 )
( 0.08788 , 0.16862 )
( 0.09091 , 0.16862 )
( 0.09394 , 0.16940 )
( 0.09697 , 0.16940 )
( 0.10000 , 0.16857 )
( 0.10303 , 0.16857 )
( 0.10606 , 0.16448 )
( 0.10909 , 0.16448 )
( 0.11212 , 0.16254 )
( 0.11515 , 0.16254 )
( 0.11818 , 0.16254 )
( 0.12121 , 0.16254 )
( 0.12424 , 0.16254 )
( 0.12727 , 0.16254 )
( 0.13030 , 0.16254 )
( 0.13333 , 0.16254 )
( 0.13636 , 0.16254 )
( 0.13939 , 0.16254 )
( 0.14242 , 0.16254 )
( 0.14545 , 0.16254 )
( 0.14848 , 0.16254 )
( 0.15152 , 0.16254 )
( 0.15455 , 0.16254 )
( 0.15758 , 0.16254 )
( 0.16061 , 0.16254 )
( 0.16364 , 0.16254 )
( 0.16667 , 0.16254 )
( 0.16970 , 0.16254 )
( 0.17273 , 0.16254 )
( 0.17576 , 0.16254 )
( 0.17879 , 0.16254 )
( 0.18182 , 0.16254 )
( 0.18485 , 0.16254 )
( 0.18788 , 0.16254 )
( 0.19091 , 0.16254 )
( 0.19394 , 0.16254 )
( 0.19697 , 0.16254 )
( 0.20000 , 0.16254 )
( 0.20303 , 0.16254 )
( 0.20606 , 0.16254 )
( 0.20909 , 0.16254 )
( 0.21212 , 0.16254 )
( 0.21515 , 0.16254 )
( 0.21818 , 0.16254 )
( 0.22121 , 0.16254 )
( 0.22424 , 0.16254 )
( 0.22727 , 0.16254 )
( 0.23030 , 0.16254 )
( 0.23333 , 0.16254 )
( 0.23636 , 0.16254 )
( 0.23939 , 0.16254 )
( 0.24242 , 0.16254 )
( 0.24545 , 0.16254 )
( 0.24848 , 0.16060 )
( 0.25152 , 0.16060 )
( 0.25455 , 0.16060 )
( 0.25758 , 0.16060 )
( 0.26061 , 0.16060 )
( 0.26364 , 0.16060 )
( 0.26667 , 0.16060 )
( 0.26970 , 0.16060 )
( 0.27273 , 0.16060 )
( 0.27576 , 0.16060 )
( 0.27879 , 0.16060 )
( 0.28182 , 0.16060 )
( 0.28485 , 0.16060 )
( 0.28788 , 0.16060 )
( 0.29091 , 0.16060 )
( 0.29394 , 0.16060 )
( 0.29697 , 0.16060 )
( 0.30000 , 0.16060 )
}

     \begin{axis}[xlabel={Cost of Swissmetro}, ylabel={Choice probability}, axis y line=left, axis x line=bottom, xmin=0, xmax=0.3, ymin=0, ymax=1,legend pos=north east,
        xtick=\empty,
	    xtick pos=bottom,
	    extra x ticks={0.1,0.2,0.3},
		extra x tick style={grid=none},
	    extra x tick labels={0.1,0.2,0.3},] 
        \addplot[smooth,blue,very thick] coordinates {\probaTrain};
         \addlegendentry{Train}
       \addplot[smooth,orange,very thick] coordinates {\probaSM};
        \addlegendentry{SwissMetro}
        \addplot[smooth,black!40!green,very thick] coordinates {\probaCar};
        \addlegendentry{Car}
        \end{axis}
 \end{tikzpicture}
\caption{Predicted choice probabilities as a function of Swissmetro cost under the random forest approach.}
\label{fig:monotonicityRF}
\end{figure}

\subsubsection{Choice probabilities under various RUMnet models.}

Next, we explore how increasing the complexity of the RUMnet architecture affects the model structure and its resulting predictions. Two dimensions can be varied: (i) the complexity $(\ell,w)$ of each feed-forward neural network building block, controlling the non-linearity of the utility function, and (ii) the number of samples $K$, controlling the latent heterogeneity of customer and product attributes.

Figure~\ref{fig:complexity} explores the first dimension and illustrates how the predictions of RUMnet change when the complexity of each feed-forward neural network building block is increased. In particular, Figure~\ref{fig:complexity} shows that, for Customer 2, more complex neural networks (from left to  right) capture a ``sharper'' substitution between Swissmetro and Train; the choice probabilities are close to either 0 or 1 and a transition occurs at the cost level that makes the customer indifferent between these alternatives. We interpret this phenomenon as follows:  a more complex neural network better segments (shatters) the different types of customers, making the behavior of the resulting segments more predictable.
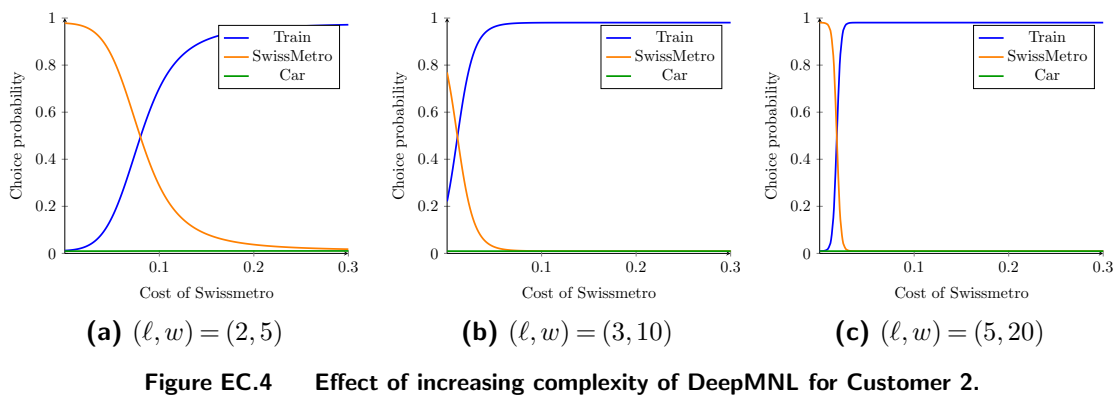
\begin{figure}[H]
\centering
\begin{subfigure}{.3\textwidth}
  \centering

  
  \caption{$(\ell,w)=(2,5)$}
\end{subfigure}
\begin{subfigure}{.3\textwidth}
  \centering
  %
  \caption{$(\ell,w)=(3,10)$}
\end{subfigure}%
\begin{subfigure}{.3\textwidth}
  \centering
   %
  \caption{$(\ell,w)=(5,20)$}
\end{subfigure}
\caption{Effect of increasing complexity of DeepMNL for Customer 2.}
\label{fig:complexity}
\end{figure}
 Figure~\ref{fig:hetero} reveals the  effect of increasing $K$. Latent heterogeneity implies that the choice probabilities are obtained as a mixture of different customer types. This is mirrored by the ``wavelets'' on the plots to the right. 
\begin{figure}[H]
\centering
\begin{subfigure}{.3\textwidth}
  \centering
  %
  \caption{DeepMNL}
  \label{fig:ec6a}
\end{subfigure}
\begin{subfigure}{.3\textwidth}
  \centering
  %
  \caption{RUMnet with $K=5$}
  \label{fig:ec6b}
\end{subfigure}%
\begin{subfigure}{.3\textwidth}
  \centering
   %
  \caption{RUMnet with $K=10$}
  \label{fig:ec6c}
\end{subfigure}
\begin{subfigure}{.3\textwidth}
  \centering
  %
  \caption{DeepMNL}
  \label{fig:ec6d} 
\end{subfigure}
\begin{subfigure}{.3\textwidth}
  \centering
  %
  \caption{RUMnet with $K=5$}
  \label{fig:ec6e}
\end{subfigure}%
\begin{subfigure}{.3\textwidth}
  \centering
   %
  \caption{RUMnet with $K=10$}
  \label{fig:ec6f}
\end{subfigure}
\caption{Effect of increasing heterogeneity. For all models, we have $\boldsymbol{(\ell,w)=(3,10)}$ in the first row (Figures~\ref{fig:ec6a},\ref{fig:ec6b} and \ref{fig:ec6c}) and $\boldsymbol{(\ell,w)=(5,20)}$ in the second row (Figures~\ref{fig:ec6d}, \ref{fig:ec6e} and \ref{fig:ec6f}).}
\label{fig:hetero}
\end{figure}

\putbib[ref]
\end{bibunit}
\end{APPENDICES}
\end{document}